\documentclass[12pt]{article}
\usepackage{helvet}

\usepackage[resetlabels]{multibib}
\newcites{checkworthy}{References}
\newcites{adversarial}{References}
\newcites{msda}{References}
\newcites{citeworth}{References}
\newcites{generating}{References}
\newcites{exaggeration}{References}

\usepackage{KUstyle}
\usepackage[margin=0.8in]{geometry} %
\usepackage{setspace} %
\usepackage{tabularx} %
\usepackage{hang}
\usepackage{amsmath}
\usepackage{times}
\usepackage{latexsym}
\usepackage{amssymb}
\usepackage{breqn}
\usepackage{graphicx}
\usepackage{enumitem}
\usepackage{anyfontsize}
\usepackage{booktabs}  
\usepackage{scalerel,xparse}

\usepackage{microtype}
\usepackage{arydshln}
\usepackage{titlesec}
\usepackage{euscript}
\usepackage{latexsym}
\usepackage{enumitem}

\usepackage{tabularx}
\usepackage{graphicx}       %
\usepackage{cleveref}       %
\usepackage{booktabs}       %
\usepackage{amsfonts}       %
\usepackage{nicefrac}       %
\usepackage{xcolor}
\usepackage{microtype}
\usepackage{arydshln}
\usepackage{times}
\usepackage{latexsym}
\usepackage{todonotes}
\usepackage{enumitem}
\usepackage{lipsum}
\usepackage{booktabs}
\usepackage{array}
\usepackage{multirow}
\usepackage{graphicx}
\usepackage[normalem]{ulem}

\usepackage{etoolbox}
\usepackage{todonotes}
\usepackage{amsmath}
\usepackage{color}
\usepackage{tcolorbox}
\usepackage{adjustbox}
\usepackage{caption}
\usepackage{subcaption}
\usepackage{bbm}
\usepackage{verbatim}
\usepackage{CJK}
\usepackage[T1]{fontenc}

\usepackage[utf8]{inputenc}

\usepackage{microtype}

\usepackage{inconsolata}

\usepackage{xspace}

\usepackage{multirow, makecell}

\usepackage{caption}
\usepackage{subcaption}
\usepackage{algorithm}
\usepackage{algpseudocode}
\usepackage{float}
\usepackage[normalem]{ulem}
\usepackage{xspace}
\usepackage{soul}
\usepackage{adjustbox}

\newcolumntype{L}[1]{>{\raggedright\let\newline\\\arraybackslash\hspace{0pt}}p{#1}}

\DeclareMathOperator*{\argmin}{arg\,min}

\usepackage[T1]{fontenc}

\usepackage{lipsum}
\usepackage{array}
\usepackage{multirow}
\usepackage[normalem]{ulem}

\usepackage{etoolbox}
\usepackage{color}
\usepackage{tcolorbox}
\usepackage{adjustbox}
\tcbset{width=0.9\textwidth,boxrule=0pt,colback=red,arc=0pt,auto outer arc,left=0pt,right=0pt,boxsep=5pt}
\usepackage[font=small,skip=-5pt]{subcaption}
\usepackage{diagbox}

\definecolor{citeworthy}{HTML}{32A852}

\makeatletter
\def\thickhline{%
  \noalign{\ifnum0=`}\fi\hrule \@height \thickarrayrulewidth \futurelet
   \reserved@a\@xthickhline}
\def\@xthickhline{\ifx\reserved@a\thickhline
               \vskip\doublerulesep
               \vskip-\thickarrayrulewidth
             \fi
      \ifnum0=`{\fi}}
\makeatother

\newlength{\thickarrayrulewidth}
\newcommand\METHOD{\textsc{MT-PET}}

\newcommand\NTEST{\textsc{563}}

\newcommand\DATASET{\textsc{\DATASET{}}}

\newcommand\SCORE{\textsc{IMS}}
\newcommand\SCOREFULL{\textsc{Information Matching Score}}

\newcommand{\Sref}[1]{\S\ref{#1}}

\newcommand{\cameraready}[1]{}

\newcommand{\citep}{\cite}
\newcommand{\citet}{\cite}
\newcommand{\citealt}{\cite}

\counterwithin{figure}{section}
\counterwithin{table}{section}
\counterwithin{algorithm}{section}
\counterwithin{equation}{section}

\newtheorem{definition}{Definition}[section]

\titleclass{\subsubsubsection}{straight}[\subsection]

\newcounter{subsubsubsection}[subsubsection]
\renewcommand\thesubsubsubsection{\thesubsubsection.\arabic{subsubsubsection}}

\titleformat{\subsubsubsection}
  {\normalfont\normalsize\bfseries}{\thesubsubsubsection}{1em}{}
\titlespacing*{\subsubsubsection}
{0pt}{3.25ex plus 1ex minus .2ex}{1.5ex plus .2ex}

\makeatletter
\renewcommand\paragraph{\@startsection{paragraph}{5}{\z@}%
  {3.25ex \@plus1ex \@minus.2ex}%
  {-1em}%
  {\normalfont\normalsize\bfseries}}
\renewcommand\subparagraph{\@startsection{subparagraph}{6}{\parindent}%
  {3.25ex \@plus1ex \@minus .2ex}%
  {-1em}%
  {\normalfont\normalsize\bfseries}}
\def\toclevel@subsubsubsection{4}
\def\toclevel@paragraph{5}
\def\toclevel@paragraph{6}
\def\l@subsubsubsection{\@dottedtocline{4}{7em}{4em}}
\def\l@paragraph{\@dottedtocline{5}{10em}{5em}}
\def\l@subparagraph{\@dottedtocline{6}{14em}{6em}}
\makeatother

\makeatletter
\newcounter{savesection}
\renewcommand\appendix{\par
  \setcounter{savesection}{\value{section}}%
  \setcounter{section}{0}%
  \setcounter{subsection}{0}%
  \gdef\thesection{\@Alph\c@section}}
\newcommand\unappendix{\par
  \setcounter{section}{\value{savesection}}%
  \setcounter{subsection}{0}%
  \gdef\thesection{\@arabic\c@section}}
\makeatother

\ptype{Ph.D. Thesis}
\author{Dustin Wright}
\title{Machine Understanding of Scientific Language}
\subtitle{This thesis has been accepted by the PhD School of The Faculty of Science, University of Copenhagen}
\date{Date: {December 2022}}
\advisor{Advisor: {Isabelle Augenstein}}
\fpimage{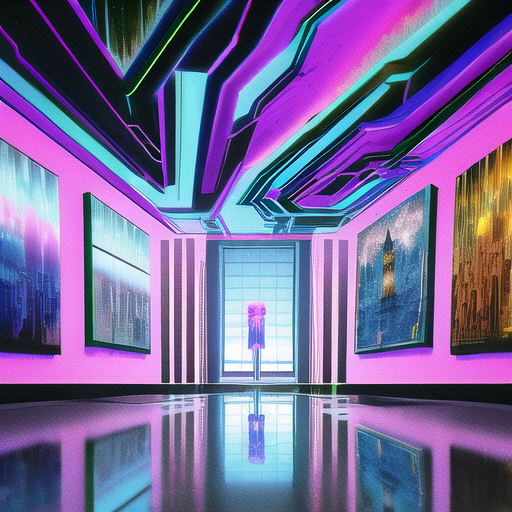} %

\newcommand\citeworthmodelname{\textsc{CiteBert}}
\newcommand\citeworthdataset{\textsc{CiteWorth}}

\begin{document}

\maketitle

\onehalfspacing
\pagenumbering{roman}

\begin{table}[h]
\def\arraystretch{1.5}
\begin{tabularx}{\textwidth}{l X}
Department & {Department of Computer Science} \\
Author(s): & {Dustin Wright} \\
Title and subtitle: & {Machine Understanding of Scientific Language} \\
Advisor: & {Isabelle Augenstein} \\
Date: & 20 12 2022
\end{tabularx}
\end{table}
\newpage

\section*{Abstract}
Scientific information expresses human understanding of nature. This knowledge is largely disseminated in different forms of text, including scientific papers, news articles, and discourse among people on social media. While important for accelerating our pursuit of knowledge, not all scientific text is faithful to the underlying science. As the volume of this text has burgeoned online in recent years, it has become a problem of societal importance to be able to identify the faithfulness of a given piece of scientific text automatically. This thesis is concerned with the cultivation of datasets, methods, and tools for machine understanding of scientific language, in order to analyze and understand science communication at scale. To arrive at this, I present several contributions in three areas of natural language processing and machine learning: automatic fact checking, learning with limited data, and scientific text processing. These contributions include new methods and resources for identifying check-worthy claims, adversarial claim generation, multi-source domain adaptation, learning from crowd-sourced labels, cite-worthiness detection, zero-shot scientific fact checking, detecting exaggerated scientific claims, and modeling degrees of information change in science communication. Critically, I demonstrate how the research outputs of this thesis are useful for effectively learning from limited amounts of scientific text in order to identify misinformative scientific statements and generate new insights into the science communication process.

\newpage

\section*{Resume}
Videnskabelig information udtrykker menneskets forståelse af naturen. Denne viden formidles i høj grad i forskellige former for tekst, herunder videnskabelige artikler, nyhedsartikler og diskurs blandt mennesker på sociale medier. Selvom det er vigtigt for at accelerere vores søgen efter viden, er ikke al videnskabelig tekst tro mod den underliggende videnskab. Som mængden af online tekst er vokset i de senere år, er det blevet en udfordring af samfundsmæssig betydning at kunne identificere troværdigheden af videnskabelig tekst automatisk. Dette speciale beskæftiger sig med skabelsen af datasæt, metoder og værktøjer til maskinforståelse af videnskabeligt sprog med henblik på at analysere og forstå videnskabelig kommunikation i større skala. For at nå frem til dette præsenterer jeg flere bidrag inden for tre områder af naturlig sprogbehandling og machine learning: automatisk faktatjek, læring med begrænset data og videnskabelig tekstbehandling. Disse bidrag omfatter nye metoder og ressourcer til at identificere checkværdige påstande, generering af modstridende krav, tilpasning af flere kilder til domæne, cite-worthiness detektion, nul-shot videnskabelig faktakontrol, opdagelse af overdrevne videnskabelige påstande og modellering af grader af informationsændringer i videnskabelig kommunikation . Slutteligt demonstrerer jeg, hvordan forskningsresultaterne fra denne afhandling er nyttige til effektivt at lære fra begrænsede mængder videnskabelig tekst for at identificere misinformative videnskabelige udsagn og generere ny indsigt i videnskabskommunikationsprocessen.

\newpage

\section*{Acknowledgements}
This work would not have been possible without the excellent mentorship of my advisor Isabelle Augenstein, who's insight, advice, and fine-tuned ability to foster quality scholarship enabled me to produce the work I did during my PhD. I would also like to acknowledge my collaborators who contributed immensely to the research described in this thesis. Additionally, I would like to thank my friends and colleagues who helped provide the balance needed to push through difficult periods and celebrate the good moments of my PhD. And finally I'd like to greatly thank my family, who though they are an ocean and a continent away, are always profoundly supportive of me.

\newpage

\tableofcontents
\newpage
\pagenumbering{arabic}

\section{Executive Summary}

Scientific knowledge is ubiquitous online, where people have access to text describing scientific knowledge in all of its forms. These forms are heterogeneous, including scientific papers intended for an expert audience, technical reports intended for science enthusiasts, and news articles and social media posts intended for the general public. Science communication is the pipeline of translation through which scientific information is disseminated at these different levels: from higher complexity (i.e. scientific papers) to lower complexity (e.g. news and social media posts). At the same time, the current media climate is rife with misinformation (content which is false or inaccurate) and disinformation  (content which is \textit{intentionally} false or inaccurate). Mis- and dis-informative scientific content online is equally widespread due to the misrepresentation of scientific information through the science communication process~\cite{Salita2015WritingFL,kua2004science,yavchitz2012misrepresentation,pellechia1997trends,walsh2018one,condit2004science,moore2006bad,sumner2014association,bratton2019association,bode2018see}, which has downstream consequences on people's behavior~\cite{kuru2021effects,Fischhoff2012CommunicatingUF,hart2016impact} e.g. how vaccines are talked about in the media has an effect on vaccine uptake~\cite{kuru2021effects} and discussions around climate change alter perceptions of efficacy around methods to address it~\cite{hart2016impact}. It is therefore a problem of societal importance to be able to combat the spread of scientific misinformation and improve science literacy among the public.

Towards this goal and due to the sheer volume of scientific text online, the automatic processing of scientific text using methods in natural language processing (NLP) and machine learning is an attractive option for assisting with organizing, understanding, and analyzing it at scale.
New deep learning algorithms such as transformers~\cite{vaswani2017attention}, self-supervised learning techniques for text such as masked language modeling~\cite{devlin2019bert}, large repositories of scientific text such as the Semantic Scholar Open Research Corpus (S2ORC)~\cite{lo2020s2orc}, and the availability of reliable implementations of state of the art models~\cite{DBLP:conf/emnlp/WolfDSCDMCRLFDS20} have enabled widespread and trackable progress on a range of important tasks in scientific language understanding. At the same time, given the need to acquire labels in order to perform supervised training, many of the tasks for which models perform acceptably well are relatively low complexity (e.g. determining which words and phrases refer to a limited set of drugs and chemicals, determine the intent of a citation in a scientific paper, which section of a paper a sentence belongs to, etc.), and limited to small set of scientific fields. As such, there is a large gap in the field of scientific NLP addressing machine understanding of scientific language for combating scientific misinformation.

This thesis seeks to fill this gap via the cultivation of methods, tools, and resources for machine processing and understanding of scientific texts. In particular, I'm concerned with how machines can be used to \textit{ensure information quality in science communication}; in other words, how to automatically detect and measure how accurate a piece of scientific information is. Prior to the work presented in this thesis, the area of information quality in science communication was sparsely studied in NLP, as most research in scientific NLP (including my own~\cite{conf/akbc/WrightKMH19,wright2022bioact,badal2019challenges}) had focused on tasks related to extracting information from scientific text such as relationships between biomedical entities~\cite{DBLP:conf/akbc/NadkarniWBSHH21} and discourse strategies in scientific writing~\cite{cohan2019structural}. Here I will present my work on defining tasks and building new datasets and algorithms within the area of scientific NLP for ensuring information quality.

To arrive at this, I first contribute new resources and methods on several tasks which serve as the bridge towards machine understanding of scientific language. As the goal is to build tools for ensuring information quality, I first look at tasks in automated fact checking and create new methods for check-worthiness detection and adversarial claim generation. Next, as data availability is an issue with scientific text, I work on methods for learning with limited data and present new models for domain adaptation, learning from crowd-sourced labels, dataset generation, and prompt-based learning in both the general domain and science. Finally, I apply the knowledge gained from these studies of fact checking and learning with limited data to construct new tasks, datasets, and methods for automating the study of science communication. As such, it will help to provide some background on each of these areas.

\subsection{Automated Fact Checking}
Automated fact checking is a process which can generally be broken down into three steps: 
\begin{enumerate}
    \item Claim check-worthiness detection
    \item Evidence retrieval
    \item Veracity prediction
\end{enumerate}
Additionally, a fourth step is added in the case of \textit{explainable} fact checking, namely producing a justification~\cite{DBLP:conf/acl/AtanasovaSLA20}. Here I will provide an overview of each of these steps and datasets available for each.

\paragraph{Check-Worthiness Detection}
Claim check-worthiness detection is concerned with identifying claims which are worthy of being fact checked. Definitions can vary across datasets, but range from more subjective definitions such as claims for which there is a general public interest~\cite{hassan2017claimbuster} to more objective definitions such as a claim which makes an assertion about the world that is checkable~\cite{konstantinovskiy2018towards}. This is presented in a number of different ways, such as identifying statements in political debates which should be checked~\cite{barron2018overview,elsayed2019overview} and identifying statements on Twitter which require verification~\cite{zubiaga2016analysing,zubiaga2017exploiting,zubiaga2018detection}. The task is usually studied as a first step in fact checking separate from the rest of the process.

\paragraph{Evidence Retrieval}
Once check-worthy claims have been identified, one must then select existing evidence which can be used to determine the veracity of those claims. This is framed as a retrieval task from some trustworthy source, such as Wikipedia or a set of scientific documents. Evidence documents are assumed to be factual, thus why they must be trustworthy. Additionally, evidence documents are necessary in order to assist in producing justifications for veracity predictions.

\paragraph{Veracity Prediction}
The next stage is to make a prediction of the veracity of the claim given the retrieved evidence. As such, a claim is determined to either be supported or refuted by the evidence, or that there is not enough evidence to predict either way. The veracity prediction task can also be performed in a late-fusion setup where evidence sentences are re-ranked during inference in order to improve performance~\cite{DBLP:conf/acl/MaGJW19,DBLP:conf/acl/SchlichtkrullKO20}.

Veracity prediction and evidence retrieval are tightly bound, so fact checking datasets tend contain data for both. Popular early datasets for general domain fact checking include the Liar dataset~\cite{DBLP:conf/acl/Wang17}, which contains real world-claims from Politifact, and the FEVER dataset~\cite{DBLP:conf/naacl/ThorneVCM18}, which is a large scale collection of manually written claims paired with evidence from Wikipedia. Fact checking datasets are evolving constantly, such as MultiFC which is largely multi-domain~\cite{DBLP:conf/emnlp/AugensteinLWLHH19}, FEVROUS which involves retrieving evidence and verifying claims over structured data~\cite{DBLP:conf/nips/AlyGST00CM21}, and multiple datasets for scientific fact checking such as SciFact~\cite{DBLP:conf/emnlp/WaddenLLWZCH20}, COVID-Fact~\cite{DBLP:conf/acl/SaakyanCM20}, and CoVERT~\cite{DBLP:journals/corr/abs-2204-12164}. For a comprehensive overview of methods and datasets for automatic fact checking, see the survey from~\cite{DBLP:journals/tacl/GuoSV22}.

\paragraph{Justification}
Finally, an emerging last step in the fact checking pipeline is to produce a justification which explains the prediction. This is required in order to ensure that the prediction is trustworthy and convince the user of the correctness of the prediction. One of the earliest works trains a joint extractive summarization and veracity prediction model to produce justifications for the predictions~\cite{DBLP:conf/acl/AtanasovaSLA20}. This was later followed up by work on using post-editing to improve the fluency and coherence of these justification~\cite{jolly2022generating}. The survey in~\cite{DBLP:journals/tacl/GuoSV22} contains a wide overview of methods and datasets for explainable fact checking and justification generation.

\subsubsection{A New View of Fact Checking for Science}
One of the contributions of this work is to rethink the fact checking pipeline in the context of scientific knowledge. The existing fact checking setup is designed to predict categorical truth and falsehood for a given claim. With scientific claims, I argue that veracity does not fully capture the types of subtle changes in information which are common in science journalism~\cite{sumner2014association,bratton2019association,Fischhoff2012CommunicatingUF,kua2004science,yavchitz2012misrepresentation,pellechia1997trends,walsh2018one,condit2004science,moore2006bad,bode2018see}. As I will demonstrate in Chapter \ref{paper:exaggeration} and Chapter \ref{paper:modeling}, we can reframe the problem with at least the following two steps:
\begin{enumerate}
    \item Identify texts which discuss the same information
    \item Measure how that information changes between those two texts
\end{enumerate}
Step 1 is similar to the evidence retrieval component of automatic fact checking, and step 2 is similar to the veracity prediction stage, however they are generalized and decoupled from the notion of veracity. This opens the door to defining different types of information changes one is interested in measuring (e.g. one sentence exaggerating another). The other two stages of the fact checking pipeline, namely check-worthiness detection and justification, can also be included in this framework. 

Check-worthiness can be included as is, but redefined towards predicting which scientific statements should be compared to the scientific literature. This is equally as subjective as the general domain version of the task, and is likely something that should be performed in tandem with humans in order to select the most salient scientific claims. The final step, justification, is a critical step in explainability which should be explored in future work. Namely, once one identifies statements describing the same information and measures how the statements differ, it is critical for a system to justify and explain exactly how those differences appear in text. In fact, an ideal system would be able to match scientific statements describing the same information, and simply explain in natural language how those statements differ, while at the same time binning statements into different classes of interest for the sake of triaging different types disinformation strategies e.g. exaggeration, cherry-picking data, changes in degree of certainty, etc. This would then enable new technologies where the average user browsing on the internet could have scientific misinformation automatically flagged for them, with an explanation of how the information they read online differs from what the source scientific literature says. Additionally, this could help improve science journalism by providing journalists with a tool to proof-read their articles and identify critical changes in their messaging which deviates from the scientific literature.

\subsection{Learning with Limited Data}
\label{sec:lld}

NLP has been seeing rapid progress in terms of the generalizability of models in recent years. Up until a few years ago, the dominant paradigm in NLP was to annotate large corpora for individual tasks and train a specialized model (generally an LSTM~\cite{hochreiter1997long} or a transformer~\cite{vaswani2017attention}) on that specific task or a set of highly related tasks. With the advent of ever-inflating transformers pre-trained in a self-supervised fashion with different flavors of language modeling~\cite{devlin2019bert,radford2019language}, models have become more general purpose and able to learn complex tasks with significantly less labeled data than previously. Here, I will focus on three areas which I make contributions to in this work for learning from limited labeled data: transfer learning, domain adaptation, and few-shot learning.

\paragraph{Transfer Learning} Transfer learning is ubiquitous in the field of NLP currently. The major shift to transfer learning approaches in NLP started with the works of Elmo~\cite{DBLP:conf/naacl/PetersNIGCLZ18}, ULM-FiT~\cite{DBLP:conf/acl/RuderH18}, and BERT~\cite{devlin2019bert}. The basic idea is simple: starting with a suitably large network, train the weights of this network on a massive corpus of general purpose text in a self-supervised fashion. Then, fine-tune this model on tasks of interest using labeled data. The main self-supervised learning techniques used are auto-regressive language modeling, where a model is trained to predict the next token in a piece of text given all of the previous tokens, and masked language modeling (MLM), where, given a piece of text, mask out some percentage of tokens in that text and try to predict the missing tokens. The large pre-trained language models resulting from this type of training can be used to achieve state-of-the-art performance on a large array of tasks with less labeled data than previous methods~\cite{devlin2019bert,DBLP:journals/corr/abs-1907-11692}. I make extensive use of large pre-trained language models in this thesis to help alleviate the need to employ massive amounts of labeled data for the scientific tasks I work on while still achieving reasonable results. Additionally, they form the foundation of the methods and models I develop throughout this thesis.

\paragraph{Domain Adaptation} Domain adaptation is a form of transfer learning where the goal is to generalize to data from distributions lying outside that of the training data. Domain differences in text can occur in various ways, from differences in the subject or topic of different texts to texts that are in completely different languages. Approaches generally fall into three categories: \textit{supervised} approaches (e.g. \citet{daumeiii:2007:ACLMain,finkel-manning-2009-hierarchical-fixed,conf/cvpr/KulisSD11}), where both labels for the source and the target domain are available; \textit{semi-supervised} approaches (e.g. \citet{conf/cvpr/DonahueHRSD13,conf/cvpr/YaoPNLM15}), where labels for the source and a small set of labels for the target domain are provided; and lastly \textit{unsupervised} approaches (e.g. \citet{blitzer-etal-2006-domain-fixed,ganin2015unsupervised,conf/aaai/SunFS16,conf/icml/LiptonWS18}), where only labels for the source domain are given. Additionally, different approaches exist in the single-source setting, where data in only one source domain is available, versus the multi-source setting, where data in multiple source domains is given~\cite{DBLP:journals/pieee/ZhuangQDXZZXH21}. Common methods in NLP for domain adaptation include domain adversarial training~\cite{ganin2015unsupervised}, pivot-based methods~\cite{DBLP:conf/naacl/ZiserR18}, progressive language model fine-tuning~\cite{DBLP:conf/acl/GururanganMSLBD20}, and mixture-of-experts~\cite{guo2018multi}.

For scientific text, domain adaptation is especially difficult given the stark differences in the language used between scientific domains. The definition of a domain is also particularly tricky -- even within the same academic field, say medicine, researchers working on different topics employ vastly different language, oftentimes even for the same concepts~\cite{DBLP:journals/nar/Bodenreider04}. Additionally, researchers in different fields have vastly different needs. For example, a common task in biomedical NLP is relation extraction for the construction of knowledge bases~\cite{DBLP:journals/biodb/WeiPLDMLWL16}, but each sub-discipline has its own particular set of entities and relations that they care about~\cite{badal2019challenges}. At the same time, annotating scientific text is highly expensive and time-consuming given the need for domain expertise and the technical nature of science. As such, it is highly beneficial to be able to generalize across domains. Part of this thesis will focus on domain adaptation, in particular multi-source domain adaptation with large pre-trained transformers and building structural scaffolding datasets with scientific text to improve transfer learning to new tasks.

These approaches are promising for two reasons. First, labeled data for scientific text tasks tend to be available for a few popular research areas, including medicine, biology, and computer science. It makes sense to investigate multi-source domain adaptation in order to leverage all available data to generalize to sparsely annotated scientific fields. Second, structural scaffolding tasks are generally much easier to acquire data for as they are acquired automatically. Scaffold tasks are weakly-supervised tasks where data is acquired via the structure of scientific documents and which help improve the generalization of models on tasks involving scientific texts~\cite{cohan2019structural}. Examples of scaffolding tasks include predicting the section of a scientific document where a piece of text resides and predicting whether or not a sentence should have a citation. As the training label is embedded in the structure of the document, massive corpora can be built cheaply and lead to gains in performance across tasks, as we will show in Chapter \ref{paper:citeworth}. 

\paragraph{Learning From Crowd-Sourced Labels} One method commonly used to acquire a large amount of annotations for a problem relatively cheaply is to employ crowd annotators on platforms such as Amazon Mechanical Turk to provide such annotations. While this is useful on extremely well-defined tasks, it is very difficult to acquire many high quality labels from crowd-workers for science. As such, one generally makes a tradeoff between annotation abundance and cost in order to acquire high quality labels from experts in science~\cite{DBLP:journals/corr/abs-1809-00537,dougan2014ncbi,DBLP:journals/biodb/WeiPLDMLWL16}.

In order to gain the most from these crowd annotations, recent work has looked into how to learn from them directly without selecting one single ground-truth label for a given sample~\cite{DBLP:journals/jair/UmaFHPPP21}. The work of \citet{DBLP:conf/iccv/PetersonBGR19} was one of the first, which demonstrated that learning directly from crowd annotations treated as soft-labels using the softmax function leads to better out of distribution performance in computer vision. This line of work has been followed by~\citet{DBLP:conf/hcomp/UmaFHPPP20} and \citet{DBLP:conf/naacl/FornaciariUPPHP21} in NLP, looking at the use of the KL divergence as an effective loss on the soft labels. The survey of \citet{DBLP:journals/jair/UmaFHPPP21} provides an extensive set of experiments on different methods for learning from crowd labels on a vast array of datasets. This is a potentially attractive option for scientific text in order to improve generalization from a limited set of crowd-annotations, which are expensive to acquire. In Chapter \ref{paper:aggregation}, I develop new methods for learning from crowd annotations, and demonstrate their efficacy on a number of tasks and datasets for out-of-domain performance, including with scientific text.

\paragraph{Prompt-Based Learning} In some cases, labeled data is either completely unavailable or it is prohibitively expensive to acquire a large enough set of labeled data to effectively train a model directly. Few-shot learning is an area of study in machine learning where the goal is to develop methods which generalize well from a highly limited set of labeled data. The area of few-shot learning investigated in this thesis for scientific text is prompt-based learning.

Prompt-based learning seeks to leverage language model pre-training for generalization. As such, the two widely used flavors of prompt-based learning are prefix prompting and cloze prompting~\cite{DBLP:journals/corr/abs-2107-13586}, reflecting the two major forms of language model pretraining used today. The core idea behind both methods for classification is the same: instead of solving a classification problem $P(y|x)$, where $x$ is the input and $y$ is one of a discrete set of labels, solve the problem $P(V(y)|p(x))$, where $p(x)$ is a function which transforms the input $x$ by inserting one or more tokens into $x$, with at least one token being a masked out token, and $V(y)$ maps the label $y$ to one or more tokens in the language model's vocabulary. Then, instead of learning a completely new classifier over a new label space, one can use the classifier which was trained over the model's vocabulary during language model pretraining and transform the classification task into a language modeling task, where $P(V(y)|p(x))$ is a prediction over the tokens $V(y)$ in the masked positions of $p(x)$. In this, the model has presumably learned useful patterns from the much larger pretraining corpus which can be transferred to the downstream classification task, provided the functions $p(x)$ and $V(y)$ are suitably reflective of the classification problem, particularly with respect to the language model being used.

As a concrete example, consider the problem of sentiment analysis of movie reviews. Given input $x = $ ``I was on the edge of my seat!'' and label $y = $ ``[POSITIVE]'', $y \in \{$ [POSITIVE] $, $ [NEGATIVE] $\}$, we can define $p(x)$ to be a function which takes input $x$ and produces output ``[x]. The movie is [MASK]'' and $V(y)$ to be a function which takes our discrete label $y$ and selects a token $V(y) \in \{$``good'', ``bad''$\}$ which we will train the model to predict in the position of the ``[MASK]'' token. In this case, we would hope that the model would predict ``good'' in the position of the ``[MASK]'', and that language model pretraining would provide a suitable initialization such that this can be fine-tuned with much fewer examples than training a classifier from scratch over the original input (in a large enough pretraining corpus, the language model has potentially seen similar examples associating ``being on the edge of one's seat'' as ``good'' in the context of film or other media). This style of prompt-based learning was popularized in the works of~\cite{schick2020exploiting,DBLP:conf/coling/SchickSS20}. Recent directions for prompt-based learning include automated prompt searching~\cite{DBLP:conf/emnlp/ShinRLWS20}, automated verbalizer searching~\cite{DBLP:conf/coling/SchickSS20}, learning soft prompts~\cite{DBLP:journals/corr/abs-2204-01172}, and learning from multiple prompts~\cite{DBLP:journals/tacl/JiangXAN20}.

Prompt-based learning presents a promising direction for scientific text understanding, as it both alleviates the need for labeling large corpora and allows for the injection of expert knowledge into the classification problem. One of the core challenges is determining useful prompts and verbalizers for a problem; models can be highly sensitive to the selection of patterns and label verbalizations~\cite{DBLP:conf/coling/SchickSS20}. The choice of model can also vastly change the optimal prompts for a given problem. As such, there is a tradeoff between the expense of annotating corpora for scientific text tasks and the time needed to train an appropriate language model and engineer or learn prompts which would reduce this expense while still providing acceptable performance. I will demonstrate in Chapter \ref{paper:exaggeration} that this is possible on the scientific text task of predicting exaggeration in science communication with as few as 200 training samples. However, I would argue that more work is needed in the direction of prompting for scientific text, particularly for determining which corpora are suitable for pretraining on which tasks, and how domain knowledge can be appropriately applied for a given task.

\subsection{NLP for Science}
\label{sec:science-nlp}
While significant progress has been made in NLP in recent years, much of the progress is restricted to a narrow domain of general text which does not require deep understanding of complex topics or jargon typical to much of science. Scientific text processing requires specialized datasets and resources for performing the various tasks one would wish to perform on scientific text. Additionally, specific scientific fields often require individualized resources, as each field has its own idiosyncrasies and jargon which aren't represented in or representative of other fields~\cite{DBLP:journals/nar/Bodenreider04}.

Much work on scientific NLP focuses on tasks relevant to the research community but not necessarily the general public. These tasks include named entity recognition (NER) and entity linking of biomedical concepts such as diseases and chemicals~\cite{dougan2014ncbi,conf/akbc/WrightKMH19}, as well as relation extraction to extract associations between these concepts~\cite{DBLP:journals/biodb/WeiPLDMLWL16}. Tools to perform these tasks are indeed useful: the construction of knowledge bases using automated tools can help researchers quickly organize a field of literature, offsetting the cognitive load required to perform their research and assist in the discovery of e.g. drug treatments for a novel disease. Similarly, a popular NLP task for scientific text is the summarization of scientific papers~\cite{DBLP:conf/aaai/YasunagaKZFLFR19}. The goal is to produce a concise summary of a paper which is easily consumable by a researcher in that field. The task is learned by training a generative model e.g. BART~\cite{DBLP:conf/acl/LewisLGGMLSZ20} to produce the abstract of a paper given the paper full text. More recent work has sought to build datasets and models for generating meta-summaries which ingest multiple documents on a particular topic and produce a summary of the major findings on that topic~\cite{DBLP:journals/corr/abs-2104-06486}. Finally, a variety of classification tasks over scientific texts exist, including citation intent classification ~\cite{jurgens2018measuring, cohan2019structural} and paper field prediction~\cite{beltagy2019scibert}. However, in this work I am concerned with machine understanding of scientific language more broadly than solely academic papers.
\begin{figure}[t]
  \centering
    \includegraphics[width=0.65\linewidth]{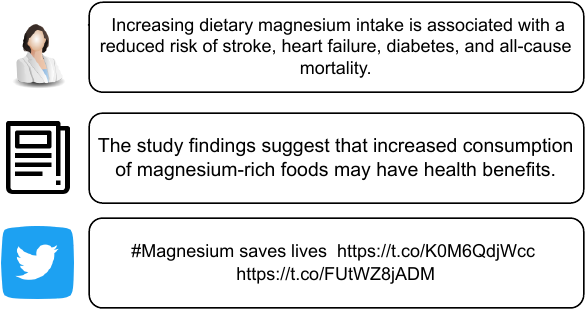}
    \caption{We are interested in measuring the information similarity of statements about scientific findings between different sources, including scientific papers, news, and tweets, shown here with real examples. The finding in this figure comes from~\citet{fang2016dietary} and the news quote is from ~\cite{rappaport2016dietary}.} 
    \label{fig:intro_claim_generation}
\end{figure}

For the interface between scientific literature and lay text, recent work has begun to investigate tasks such as lay summarization~\cite{DBLP:conf/emnlp/ChandrasekaranF20a} and scientific fact checking~\cite{DBLP:conf/emnlp/WaddenLLWZCH20,DBLP:conf/acl/SaakyanCM20,DBLP:journals/corr/abs-2204-12164}. The goal of lay summarization is to simplify complicated scientific literature and generate summaries which a lay person can understand. This is both a difficult linguistic task and also important for making the findings of science accessible to the general public. Another task at the interface of papers and the public, scientific fact checking is focused on the veracity of scientific information. The problem has been studied both on synthetic data derived from scientific abstracts~\cite{DBLP:conf/emnlp/WaddenLLWZCH20} and on real-world claims sourced from various social media websites~\cite{DBLP:conf/acl/SaakyanCM20,DBLP:journals/corr/abs-2204-12164}. The task is difficult, as it requires models for both information retrieval and entailment prediction, the difficulty if which is exacerbated by the complexity of scientific language.

This thesis addresses three gaps in the literature around scientific text understanding. The first is that information quality in science goes beyond veracity. While categorical falsehoods do exist in science communication, and it is important both as a semantics problem as well as for public well being, categorical falsehood only addresses one type of scientific misinformation. In practice, more subtle distortions such as exaggeration and hedging permeate science communication, and this has an impact on people's behavior~\cite{sumner2014association,bratton2019association,woloshin2009press,woloshin2002press,gustafson2019effects,Fischhoff2012CommunicatingUF,kuru2021effects}. Even well intentioned science communicators are prone to these distortions, as the quotes in \autoref{fig:intro_claim_generation} demonstrate. Here, I present a new paradigm with which to view the scientific misinformation problem in NLP: as one where we wish to identify the degree to which scientific statements are different and what those differences are. 

The second gap I address is that the majority of works on scientific language understanding, particularly for ensuring information quality, ignore most of the text of scientific papers. Most works will use paper abstracts as sources of claims and evidence to perform fact-checks against~\cite{DBLP:conf/emnlp/WaddenLLWZCH20,DBLP:conf/acl/SaakyanCM20,DBLP:journals/corr/abs-2204-12164}. While the abstract provides presumably the most salient information in a scientific article, tasks which focus solely on paper abstracts will fail to capture the more subtle pieces of information in an article such as caveats to findings and limitations, a well documented phenomenon for journalists as well~\cite{Fischhoff2012CommunicatingUF}. In Chapter \ref{paper:modeling} I will obviate the need to ingest the full texts of papers for fully understanding scientific language and the science communication pipeline. 

The third gap I address in this thesis is that the majority of existing work around scientific information quality is narrow in scope. Popular datasets for scientific fact checking and summarization, including~\cite{DBLP:conf/emnlp/WaddenLLWZCH20}, focus largely on biology and medicine. Additionally, they restrict themselves to just scientific papers or just scientific papers and one other domain (e.g. Reddit~\cite{DBLP:conf/acl/SaakyanCM20}, Twitter~\cite{DBLP:journals/corr/abs-2204-12164}, etc.). I present work in this thesis (particularly, Chapter \ref{paper:modeling}) which broadly covers three stages of the science communication pipeline (papers, news, and Twitter) and four scientific fields (medicine, biology, computer science, and psychology). Additionally, I argue that this is necessary in the context of ensuring information quality in science and machine understanding of scientific language in general, as it is important to build tools which are robust across fields of research and level of complexity for the sake of increased public good.

\subsection{Machine Understanding of Scientific Language}

The core problem explored in this thesis is how to enable machine understanding of scientific language. In particular, I'm concerned with how machines can understand what information is expressed in a scientific sentence, and how to determine the degree to which two scientific sentences express the same information. As a preliminary, it is important to explicitly define what I mean by \textit{scientific information}. Scientific information is expressed through scientific \textit{findings}, where a scientific finding has the following definition:
\begin{definition}
\label{def:findings}
A scientific finding is a statement that describes a particular research output of a scientific study, which could be a result, conclusion, product, etc.
\end{definition}
This general definition holds across fields; for example, many findings from medicine and psychology report on effects on some dependent variable via manipulation of an independent variable, while in computer science many findings are related to new systems, algorithms, or methods. The goal is to be able to build systems which can help analyze and improve science communication through automatic processing of scientific findings, a critically important topic which has large impact on both people's behavior and public policy~\cite{national2017communicating,kuru2021effects}. 

\begin{table*}[t]
\small

\newcommand{\tabincell}[2]{\begin{tabular}{@{}#1@{}}#2\end{tabular}}
\resizebox{0.99\textwidth}{!}{
\begin{tabular}{p{60mm}p{60mm}}%
\toprule
\textbf{Sentence 1} & \textbf{Sentence 2}\\
\midrule
The polar bear is sliding on the snow. & A polar bear is sliding across the snow.
\\
\midrule
A plane is taking off & An air plane is taking off\\
\midrule
A dog rides a skateboard & A dog is riding a skateboard\\
\midrule
A man is playing the drums & A man plays the drum\\
\bottomrule
\end{tabular}
}
\caption{Samples of sentence pairs in STSB which have a similarity score of 5 }
\label{tab:stsb-5-examples-main}
\end{table*}
\begin{table*}[t]
\small

\newcommand{\tabincell}[2]{\begin{tabular}{@{}#1@{}}#2\end{tabular}}
\resizebox{0.99\textwidth}{!}{
\begin{tabular}{p{60mm}p{60mm}}%
\toprule
\textbf{Sentence 1} & \textbf{Sentence 2}\\
\midrule
Higher-income professionals had less tolerance for smartphone use in business meetings. & We are intrigued by the result that professionals with higher incomes are less accepting of mobile phone use in meetings.
\\
\midrule
If we allow people to retract recently posted comments, then we may be able to minimize regret from posting in the heat of the moment. & Allowing users to retract recently posted comments may help minimize regret .\\
\midrule
Papers with shorter titles get more citations \#science \#metascience \#sciencemetrics & Our analysis suggests that papers with shorter titles do receive greater numbers of citations.\\
\midrule
Low levels of self-esteem and poor emotional processing skills were significantly correlated with gang involvement, as were low levels of parental monitoring, poor parental communication and housing instability. & Major findings also indicated that low levels of parental monitoring, poor parental communication and housing instability were significantly associated with gang involvement.\\
\bottomrule
\end{tabular}
}
\caption{Samples of sentence pairs in \textsc{Spiced}  (Chapter \ref{paper:modeling}) which have a matching score of 5.}
\label{tab:intro-dataset-5-examples}
\end{table*}

There are several core challenges on the path to this goal. First, what tasks are necessary in order to achieve it? As discussed in Section \ref{sec:science-nlp}, there are several existing tasks in scientific NLP. Information extraction can potentially help, but the information measured in these tasks are explicit entities and relations, which would potentially require defining and building labeled datasets for each type of relation which expresses a scientific finding we would be interested in. The task of automatic fact checking~\cite{DBLP:conf/naacl/ThorneVCM18} (and indeed the scientific version of it~\cite{DBLP:conf/emnlp/WaddenLLWZCH20}) is perhaps a better place to start, though it is concerned with a very specific type of information change: veracity. While it is important to be able to measure when one scientific statement is contradicted or supported by another, in practice the types of information changing between different utterances of the same finding tend to be more nuanced and not necessarily categorical falsehood~\cite{sumner2014association,bratton2019association,woloshin2009press,woloshin2002press}. It would seem that a more broad notion of information change is needed in order to achieve the goal of this work.

A promising line of work to emulate is semantic textual similarity (STS)~\cite{DBLP:journals/corr/abs-1708-00055,DBLP:conf/naacl/GanitkevitchDC13}. Here, the goal is to measure how similar are the meanings of two pieces of text, measured as a scalar from 1-5. Some examples of what would be considered highly similar sentences in a typical STS task (from STSB~\cite{DBLP:journals/corr/abs-1708-00055}) are given in \autoref{tab:stsb-5-examples-main}. Here we see a very strict notion of similarity: for a pair to be highly similar, the entire meaning of the sentence must be preserved from one sentence to the other. While closer to the type of information change concerned with in this work, this definition of similarity is too restrictive to be useful in the context of scientific information. As described in Definition \ref{def:findings}, a finding is an expression of a research output. In this, the salient information relates to what is said about the research output, so some information in a piece of text may change the semantics of the text but not what is meant by the finding. Consider the following sentences:
\begin{quote}
    \textbf{Sentence 1}: The study showed that increased dietary sugar led to weight gain in humans.
    
    \textbf{Sentence 2}: ``If a person eats more sugar, they'll gain more weight,'' said the researchers.
\end{quote}
The meaning of these two sentences is slightly different, but the information in the findings is equivalent. This is further demonstrated in real examples from the dataset I present in Chapter \ref{paper:modeling}, shown in Table \ref{tab:intro-dataset-5-examples}. Given this, STS is a good starting point if we can modulate the task to focus solely on the information in the scientific findings.

This thesis will build up to and ultimately define the task of measuring scientific information change, as well as develop and evaluate different ways of modeling and learning it. Prior to this work, this framing of the problem of scientific misinformation was not defined, and the most related work came in the forms of automatic fact checking~\cite{DBLP:conf/naacl/ThorneVCM18} and causal claim strength prediction of scientific statements~\cite{yu2019detecting,yu2020measuring,DBLP:conf/emnlp/LiZY17}. Because of this, I will ask several questions in this thesis: what tasks are relevant for understanding scientific language and how do we define them? How do we collect data for these tasks? How do we evaluate them? How do we model and learn them? I tackle these questions systematically in the following way, initially using existing datasets and tasks and eventually building new methods and datasets to achieve the goal of scientific language understanding for ensuring information quality: 

\paragraph{General Domain Fact Checking} I first present novel solutions to problems in general domain fact checking (Chapter \ref{paper:check-worthiness} and \ref{paper:adversarial}). This includes predicting when statements should be fact checked, as well as generating adversarial inputs for fact checking models in order to evaluate their robustness.

\paragraph{Modeling and Dataset Creation for Scientific Text Tasks} I next investigate several problems related to dataset creation and modeling for scientific tasks. This is predicated on the fact that dataset creation with scientific text is both expensive and time consuming~\cite{DBLP:conf/emnlp/WaddenLLWZCH20,sumner2014association,bratton2019association}, and as such, datasets for scientific tasks tend to be small and/or difficult to acquire. Additionally, one generally must acquire data for each scientific field and target task of interest~\cite{dougan2014ncbi,conf/akbc/WrightKMH19,DBLP:conf/emnlp/WaddenLLWZCH20,DBLP:conf/acl/SaakyanCM20,DBLP:journals/corr/abs-2204-12164,DBLP:journals/biodb/WeiPLDMLWL16,kringelum2016chemprot,DBLP:conf/aaai/YasunagaKZFLFR19,DBLP:journals/corr/abs-2104-06486,DBLP:conf/emnlp/ChandrasekaranF20a}. To alleviate some of these problems, with an eye to building up new datasets for the tasks required for scientific language understanding and measuring information change, I present contributions to domain adaptation (Chapter \ref{paper:msda}), learning from noisy crowd-sourced labels (Chapter \ref{paper:aggregation}), automatic dataset generation (Chapter \ref{paper:generating}), and few-shot learning (Chapter \ref{paper:exaggeration}).

\paragraph{Towards Scientific Language Understanding} Finally, I define, model, evaluate, and analyze several tasks in scientific language understanding, in particular with respect to measuring information change. For this I look into existing tasks such as cite-worthiness detection (Chapter \ref{paper:citeworth}) and scientific fact checking (Chapter \ref{paper:generating}), curate better evaluation data and develop models for the task of detecting exaggerated scientific statements (Chapter \ref{paper:exaggeration}), and define and build a comprehensive dataset for the new task of measuring information change in science communication (Chapter \ref{paper:modeling}). In addition, I demonstrate how the dataset and models built in Chapter \ref{paper:modeling} can be used to help both with other tasks in scientific language understanding and with analyzing science communication broadly.

In the following sections, I will break down and summarize each of these components and how they contribute to the goal of machine understanding of scientific language.

\subsubsection{General Domain Fact Checking}
As a part of automatically ensuring information quality in science, I develop new methods for ensuring information quality in general domain texts. For this, I work on two important issues in fact checking: detecting when a claim should be fact checked (Chapter \ref{paper:check-worthiness}) and fact checking model robustness against adversarial attacks (Chapter \ref{paper:adversarial}).

As the first step in automatic fact checking, check-worthiness detection involves determining if a statement ``makes an assertion about the world that is checkable''~\cite{konstantinovskiy2018towards}. This step is useful both for further processing by machine learning models to determine veracity as well as for notifying fact checkers of information worthy of a fact check. In this work, I use the observation that this is a highly subjective task~\cite{konstantinovskiy2018towards} to hypothesize that while samples labeled as positive are likely true positives, not all negative samples are true negatives. As such, I experiment with positive unlabeled (PU) learning for the check-worthiness detection task on three datasets: Wikipedia citation needed detection, rumor detection on Twitter, and political speech check-worthiness detection. I find that while PU learning is helpful for Wikipedia and Twitter, it is detrimental to performance in the political domain, noting some inconsistencies in the labeling of that dataset. This work and the observations made become some of the basis for the work I perform on check-worthiness in science in the form of cite-worthiness detection: the task of identifying scientific sentences which require a citation.

The second general domain fact checking task I investigate is adversarial claim generation. Adversarial claims are deceptive model inputs designed to mislead an ML system into making the wrong prediction. Its important to reveal such system vulnerabilities in order to correct them, especially for fact checking systems where an adversarial claim can trick a model into predicting a false claim is true. In this work I explore universal adversarial triggers -- single tokens which can be prepended to a wide range of inputs to force a particular model prediction to be changed in a certain direction (e.g. ``SUPPORTS'' to ``REFUTES'' in a fact checking model)~\cite{gao2019universal}. The primary issue with these types of attacks is that they tend to truly flip the input label (e.g. prepending a negation word such as ``None'' in order to flip a supported claim to a refuted claim) and make the input nonsensical. To address this, I introduce a secondary objective in the adversarial trigger search which optimizes the semantic textual similarity of the original claim and adversarial claim. To ensure the coherence of the adversarial claim, I additionally introduce a generation component using GPT-2~\cite{radford2019language} which is trained to include the trigger token in the output claim. Combining these two modules for adversarial claim generation results in more robust adversarial claims which are coherent and don't trivially flip the original label of the claim.

\subsubsection{Learning from Limited Data}
The next part of this thesis presents several contributions in the area of building and utilizing datasets for scientific language understanding tasks in the presence of limited data. The methods presented are general and applicable to a wide range of machine learning and NLP tasks, so I evaluate them on both general domain and scientific text. The methods I build are in the following areas of machine learning:
\begin{itemize}
    \item Domain adaptation
    \item Learning from crowd-sourced data
    \item Generation
    \item Few-shot learning
\end{itemize}

\paragraph{Domain adaptation} 
\begin{figure}[t]
  
  \centering
    \includegraphics[width=0.65\columnwidth]{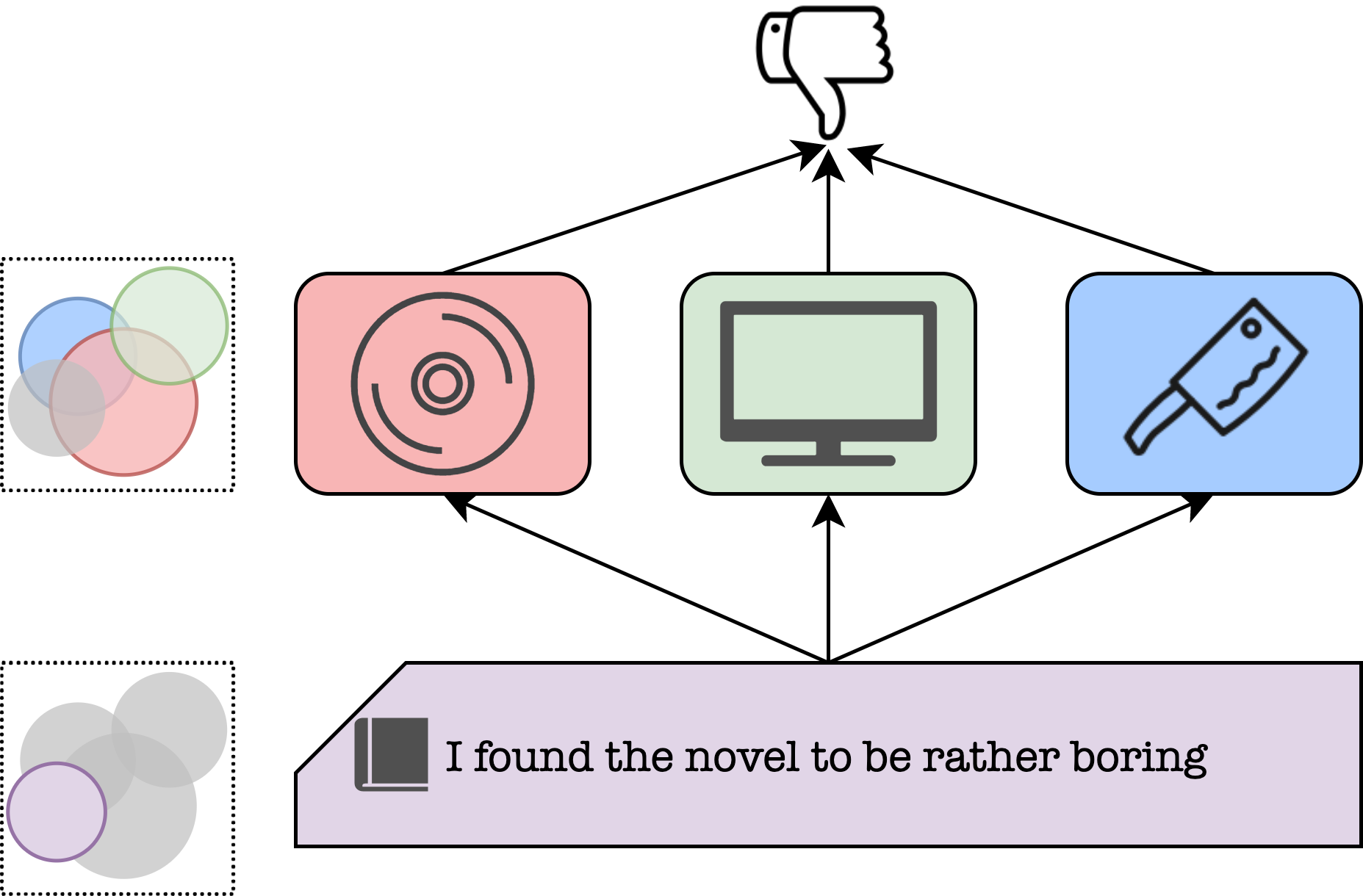}
    \caption{In multi-source domain adaptation, a model is trained on data drawn from multiple parts of the underlying distribution. At test time, the model must make predictions on data from a potentially non-overlapping part of the distribution.}
    \label{fig:msda-intro}
\end{figure}
In Chapter \ref{paper:msda} I present work on using large pre-trained transformer models to perform multi-source domain adaptation (MSDA). The main idea behind MSDA is to leverage data for a particular task but drawn from disparate modes of the underlying distribution in order to perform inference on a target mode of data for which no training labels are available (see \autoref{fig:msda-intro}). Examples of these different modes are different types of products in the case of reviews on Amazon or different fields of study in the case of scientific text. Domain adaptation is relevant to scientific language understanding due to the cost of obtaining high quality human annotated data in science. If we can make better use of less data and already existing data, we will have made progress towards improving NLP for science.

The particular methods I explore in this work are mixture-of-experts techniques and domain adversarial training~\cite{guo2018multi,ganin2015unsupervised}. The idea behind mixture-of-experts is to train individual models on particular domains and subsequently learn how to mix their predictions for the target domain. This is based on the hypothesis that some domains are more relevant than others for the target e.g. the language used in medicine is more similar to the language used in biology than in computer science, therefore a model trained on biology texts will be more relevant than one trained on computer science texts. Domain adversarial training on the other hand aims to induce a more uniform internal representation of data across domains such that the representations of data in the target domain are similar to the representations of data in the source domains. The net effect of this is that the classifier trained on source domain data generalizes better to target domain data, as the target domain data appears to lie within the distribution of data that the model was trained on.

I examine how mixture-of-experts and domain adversarial training can be effectively utilized with the current dominant large pretrained transformer models in NLP. I do so with several different types of mixing strategies, from simple ensembling to a learned attention mechanism, as well as including or excluding domain adversarial training. I find in this work that while simple ensembling provides some gains in performance across tasks, more complex mixing strategies provide no gain in performance. An analysis of the predictions of each individual domain expert reveals that these large transformer models learn highly homogeneous classifiers for a particular task despite being trained on \textit{completely different} data, helping to explain the result that complex mixing functions provide no gain in performance. Additionally, I find that while domain adversarial training does indeed induce a more uniform representation in a given model, this does not translate into improved generalization to target data.

\paragraph{Learning from Crowd-Sourced Data}
In Chapter \ref{paper:aggregation}, I propose new methods for learning from crowd annotations treated as soft-targets that confer more robust performance in the out-of-domain setting across a number of tasks. This is based on the fact that selecting an effective training signal for tasks in natural language processing is difficult: collecting expert annotations is expensive, and crowd-sourced annotations may not be reliable. Recent work in machine learning has demonstrated that learning from soft-labels acquired from crowd annotations can be effective~\cite{DBLP:conf/hcomp/UmaFHPPP20,DBLP:conf/naacl/FornaciariUPPHP21,DBLP:journals/jair/UmaFHPPP21,DBLP:conf/iccv/PetersonBGR19} especially when there is distribution shift in the test set~\cite{DBLP:conf/iccv/PetersonBGR19}. However, the best method for acquiring these soft labels is inconsistent across tasks. 

To address this, I propose new methods for acquiring soft-labels from crowd-annotations by aggregating the distributions produced by existing methods. In particular, I propose to find a distribution over classes by learning from multiple-views of crowd annotations via temperature scaling and finding the Jensen-Shannon centroid of their distributions. I demonstrate that using these aggregation methods leads to best or near-best performance across four NLP tasks on out-of-domain test sets, mitigating fluctuations in performance when using the constituent methods on their own. Additionally, these methods result in best or near-best uncertainty estimation across tasks. I argue that aggregating different views of crowd-annotations as soft-labels is an effective way to ensure performance which is as good or better than the best individual view, which is useful given the inconsistency in performance of the individual methods.

\paragraph{Generation} I next propose novel methods for dataset generation in the context of scientific fact checking in Chapter \ref{paper:generating}. Again due to the cost of annotation for scientific text tasks, one attractive option is to leverage existing data in order to automatically create new data with which to train models. One of the primary datasets for scientific fact checking, namely SciFact~\cite{DBLP:conf/emnlp/WaddenLLWZCH20}, is one such dataset in which human experts were required to manually write claims and pair them with ground truth statements from scientific abstracts which either support or refute those claims. In this work, I explore how the existing data in SciFact can be used to automatically generate new training data both in an unsupervised and supervised fashion. I find that both methods are effective, as training data generated using both can be used to train a model to within 90\% of the performance of a model trained on manually written claims on the veracity prediction task of scientific fact checking. 

\paragraph{Few-shot Learning} 
\begin{figure}[t]
  
  \centering
    \includegraphics[width=0.65\linewidth]{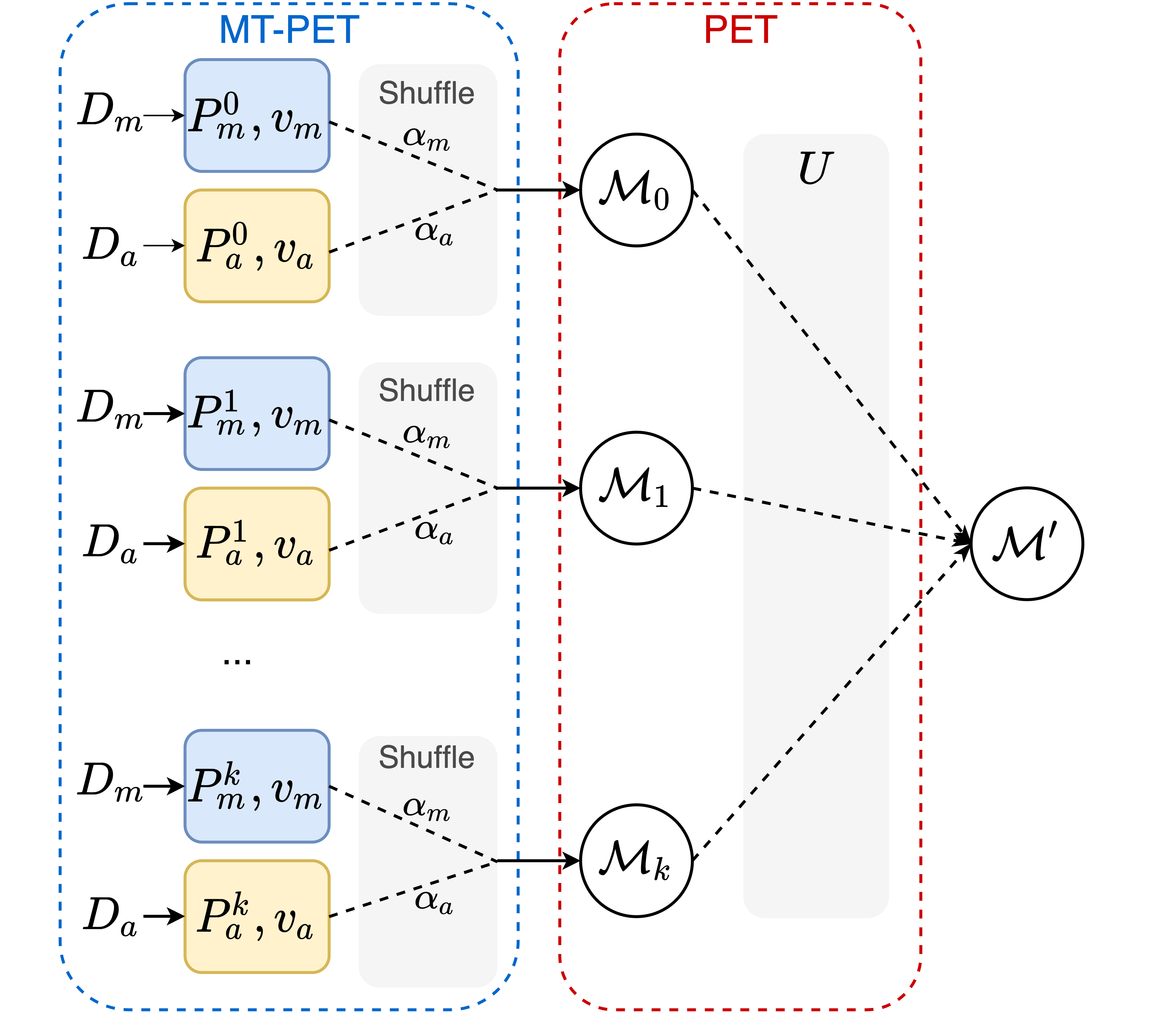}
    \caption{MT-PET design. We define pairs of complementary pattern-verbalizer pairs for a main task and auxiliary task. These PVPs are then used to train PET on data from both tasks.}
    \label{fig:mt-pet-intro}
\end{figure}
Next, I develop methods for few-shot learning evaluated on the nascent task of scientific exaggeration detection (Chapter \ref{paper:exaggeration}). Few-shot learning is an area of study which aims to achieve as much generalization as possible from as little data as possible. The particular area of few-shot learning explored in this work is prompt-based learning with large pretrained language models. For this we develop multi-task pattern exploiting training (MT-PET), a multi-task version of pattern exploiting training (PET)~\cite{schick2020exploiting,schick2020small}. 

As discussed in Section \ref{sec:lld}, the core idea behind PET is to transform a classic supervised learning task, in which the goal is to learn a classifier which produces a probability distribution over $K$ classes for a given input, to a cloze-style question answering problem which can make effective use of masked langauge model (MLM) pretraining. In this, one engineers a ``prompt'' for their input data with one or more tokens in this prompt masked, and the classification task is to predict the appropriate token in the language model's vocabulary which would fill the mask token in the prompt. These tokens are explicit verbalizations (a.k.a \textit{verbalizers}) of the classes which the model should be trained to predict. 

With PET, one defines prompts and verbalizers for the single task one wishes to solve. In Chapter \ref{paper:exaggeration} I develop \textbf{MT-PET} (see \autoref{fig:mt-pet-intro}), a multi-task version of this which can leverage prompts and verbalizers from complementary tasks to the main task one wishes to perform. The hypothesis is that some transfer learning may occur between similar tasks (e.g. semantic textual similarity and natural language inference), and thus having complementary patterns and verbalizers when training PET and using all of the training data from both tasks should help with few-shot learning. Indeed I find that MT-PET does help for the task of scientific exaggeration detection when using the complementary tasks of detecting exaggerated statements and detecting the causal claim strength of a statement with as few as 200 samples from each task.

\subsubsection{Tasks in Scientific Language Understanding}
Finally, I contribute to a number of tasks in scientific language understanding, culminating in a new task and dataset on measuring information change in science across different media. The tasks explored in this thesis are the following:
\begin{itemize}
    \item Cite-worthiness detection
    \item Scientific fact checking
    \item Scientific exaggeration detection
    \item Modeling information change in science
\end{itemize}

\paragraph{Cite-worthiness detection} I first present new data and models for cite-worthiness detection (Chapter \ref{paper:citeworth}). The task of cite-worthiness detection is: given a statement from a scientific paper, predict if that statement should have a citation i.e. that it requires external evidence in order to be validated. This task is similar to the check-worthiness detection tasks examined in Chapter \ref{paper:check-worthiness}. Additionally, as a structural scaffold it is easy to acquire large amounts of data for this task.

For this work, I observe that existing datasets for cite-worthiness are limited in size, limited in the number of domains studied, have high class imbalance, and are low-quality in terms of dataset cleanliness~\cite{sugiyama2010identifying,bird2008acl,farber2018cite,farber2018high}. In response to this, I develop CiteWorth, a large, rigorously curated, and high quality dataset for cite-worthiness detection across 10 scientific domains. CiteWorth contains over 1.1M sentences, of which 300K are cite-worthy and 800K are non-cite-worthy. Additionally, I develop a strict set of rules for curating and cleaning cite-worthy sentences such that the vast majority of trivial markers of possible citations are removed. The dataset is also \textit{contextualized} -- data is collected at the paragraph level, such that all surrounding context within a paragraph is available for each sentence.

I perform several baseline experiments on CiteWorth, finding that including context sentences can improve cite-worthiness detection by 5 points in F1 score. I additionally perform a domain analysis to show that CiteWorth is useful in the study of domain adaptation for scientific text as there exists strong differences in representations and cross-domain performance for different fields. Finally, I show that pre-training on the cite-worthiness detection task provides gains on several downstream tasks in scientific text understanding tasks, providing further evidence for the usefulness of scaffolding tasks in scientific NLP~\cite{cohan2019structural}.

\paragraph{Scientific fact checking} The next task I look into is scientific fact checking (Chapter \ref{paper:generating}). Scientific fact checking consists of the following: given a scientific claim $c$ and a corpus of scientific abstracts $D$, retrieve evidence abstracts from $D$ and predict if $c$ is either \textit{supported} or \textit{refuted} by those documents, or if there is \textit{not enough information} to make a prediction. Several datasets have been introduced for this task (e.g. \cite{DBLP:conf/emnlp/WaddenLLWZCH20,DBLP:conf/acl/SaakyanCM20,DBLP:journals/corr/abs-2204-12164}). I focus here on the SciFact dataset~\cite{DBLP:conf/emnlp/WaddenLLWZCH20}, which is manually created. 

The fact checking task is relevant in the context of combating scientific misinformation, but data is difficult to acquire. The SciFact dataset is built by having domain experts manually write scientific claims based off of findings described in scientific abstracts. These claims are then paired with source abstracts and sentences which support the claim. Negative instances are also created manually, where annotators manually rewrite claims to be contradicted by the source abstract. This is an expensive and time-consuming process, resulting in a somewhat small dataset ($\sim$1,400 claims). Given this, I build and test new methods for scientific fact checking dataset generation, achieving competetive performance on veracity prediction with no manually labeled samples.

Scientific fact checking addresses veracity, which is an important type of information change to model. As such, datasets such as SciFact are good starting points for building tools to combat scientific misinformation, but they are limited in scope both in terms of covering the types of misinformation that appear in science and in covering scientific language beyond academic papers. The next works I present are attempts to go beyond veracity and beyond solely scientific literature, to propose a different paradigm with which to think about and examine the problem of automating the process of ensuring information quality in science.

\paragraph{Scientific exaggeration detection} As a first step, I present work on scientific exaggeration detection in Chapter \ref{paper:exaggeration}. Similar to scientific fact checking, one of the goals of performing exaggeration detection is to combat scientific misinformation online and flag particular types of information change between statements made in source scientific literature and popular media. Exaggeration is one of the well documented issues in science communication~\cite{sumner2014association,bratton2019association}. 

While an important issue, little data was available for training machine learning models on this task prior to our study, and the problem had mostly been studied in NLP as one of detecting the causal strength of scientific claims as opposed to directly measuring differences in this claim strength~\cite{yu2019detecting,yu2020measuring,DBLP:conf/emnlp/LiZY17}. One of the goals of this work was to present a study which used real world data one would find in the wild, as opposed to artificially created data. Therefore, as a first step, I curate existing data from various studies on exaggeration in science communication into a comprehensive test set and small training set, necessary for measuring model performance and progress on the task.

The dataset comes from the studies in \cite{sumner2014association} and \cite{bratton2019association}, where domain experts manually label the primary findings as described in scientific papers and press releases along with their causal claim strength. Overall I curate 100 pairs of findings from papers and press releases for training and 553 pairs for evaluation. As the training dataset is small, I develop methods for prompt-based learning for this task, and demonstrate that one can achieve moderate levels of performance with only the 100 training instances in the data.

\paragraph{Modeling information change} Finally, I address the problem of modeling general information change in scientific findings between different media. This task is inspired by the fact that no comprehensive dataset had existed for the basic task of pairing together sentences which describe the same scientific findings. This is a necessary step if one wishes to make comparisons between how scientists and the media describe scientific findings, in order to analyze this communication and provide indications of where such communications fail. 

To address this gap, I build a dataset of paired scientific findings labeled with the degree to which the two findings describe \textbf{the same} findings using a 5-point scale which I call the Information Matching Score (IMS). Some examples of paired findings with a matching score of 5 are given in \autoref{tab:intro-dataset-5-examples}. The dataset, namely the \textsc{Scientific Paraphrase and Information ChangE Dataset} (\textsc{Spiced}), is built by first pairing together potential scientific findings as presented in scientific papers, news media, and Twitter using SentenceBERT (SBERT)~\cite{reimers-2019-sentence-bert}, then presenting the potential pairs to human annotators. The Prolific platform\footnote{\url{https://www.prolific.co/}} is used in order to hire domain experts in the scientific fields represented in the data: medicine, biology, psychology, and computer science. After constructing the dataset and ensuring the data is high quality, I train several baseline models and benchmark their performance, finding that SBERT models fine-tuned on \textsc{Spiced} are best suited to the task. 

Next, I show how models trained on \textsc{Spiced} are beneficial for multiple downstream applications. First, I show how models trained on \textsc{Spiced} perform significantly better on the task of evidence retrieval for scientific fact checking, despite differences in the domain and source of scientific claims. Then, I perform several large scale analyses of science communication using models trained on \textsc{Spiced} as well as the exaggeration detection dataset from Chapter \ref{paper:exaggeration}. I make three primary observations in this analysis:
\begin{enumerate}
    \item General news outlets systematically express higher information change than press releases and science and technology news outlets.
    \item Verified users and users with more followers express higher information change on average than organizational accounts.
    \item Findings as expressed in the limitations sections of papers tend to be exaggerated more in the media.
\end{enumerate}
Importantly, I show that models trained on \textsc{Spiced} can be used to reveal large scale trends in science communication, making the dataset and models useful for answering new research questions about how the message of science changes across media. These results also show that one shouldn't ignore the full-text of a paper when analyzing science communication, as stark differences exist between different sections of a paper in terms of how the message can change. These results and resources represent a new way to think about and study the problem of scientific misinformation.

\subsection{Towards Scientific Language Understanding}
The components of this thesis culminate into the first study on information change in science communication within natural language processing. As an entry point to building tools for combating scientific misinformation, I first develop new methods for general domain fact checking and scientific fact checking given the availability of data. I then contribute tools for modeling and dataset creation in order to assist with building new resources for new tasks in scientific language understanding. I gradually build upon various tasks in scientific language understanding, ultimately defining information change as an important and useful task for automatically analyzing scientific texts at all stages of the science communication pipeline. At the same time, there is still much work to be done in order to improve datasets, methods, and problem formulations for ensuring information quality in science communication.

\subsubsection{Datasets} The datasets developed in this thesis, while demonstrated to be useful, are still rather limited in size. The dataset presented in Chapter \ref{paper:exaggeration} consists of only 653 samples, and the dataset in Chapter \ref{paper:modeling} only 6,000. Additionally, they have limited scope, covering only the most popular scientific disciplines. As such, more resources should be invested in building larger and more comprehensive datasets for these tasks; in particular, most of those resources should be invested in developing difficult \textit{test sets}. In my view, the main need for more data is in order to track progress on these tasks as opposed to developing more accurate models. We should follow the current trend in NLP and machine learning to build models and methods which require less training data in order to mitigate the expense of annotation and collection of data. It therefore makes more sense to invest more resources into making difficult testing data covering a broad range of fields, topics, and tasks.

\subsubsection{Methods} In line with the need for larger testing data, new methods should be created for working with limited scientific training data. Massive language models the likes of GPT-3~\cite{DBLP:conf/nips/BrownMRSKDNSSAA20} are capable of impressive few- and zero-shot performance on general domain text. Given the training sets available in popular scientific domains, one avenue of research could be to first determine how to insert appropriate domain knowledge in a prompt-based fashion to perform well on those domains, as well as to explore how to develop prompting methods which transfer across domains. Those practices could then be applied to new scientific domains and tasks, where the main expense would come from hiring experts to develop small sets of prompts as opposed to hand annotating large training sets. 

\subsubsection{Problem Formulations} The problem formulation presented in Chapter \ref{paper:modeling} poses measuring information change between scientific sentences very generally. This is useful for revealing trends in science communication with very broad strokes. For example, we can ask research questions such as ``to what degree do different organizations change the message of science?'' and ``how do different social factors affect degree of information change?'', which are answerable with sufficiently large sets of unlabeled data.

Narrowing down the specific types of information change and strategies used by organizations is a different story. In the current formulation, one can narrow down the types of information change in a pipeline fashion, first by matching findings (considering a matched pair to be one where the IMS exceeds a certain threshold) and then performing a second analysis (human or machine) to identify what information changes between the pair. This is the setup used in Section \ref{sec:section_analysis} to determine what sections of a scientific paper tend to be overstated. The problem in this setup is: what types of information change do we care about measuring? Certain types of information change have been identified as being prevalent in science communication~\cite{sumner2014association,bratton2019association,Fischhoff2012CommunicatingUF}, but as of now there isn't a central resource defining all of them. I would argue that an important next step in building tools for measuring information change in science is to define a \textbf{taxonomy of information change in science}. Such changes should be of societal relevance and prevalent in science communication. Examples of changes that could be included in such a taxonomy are veracity, exaggeration, certainty, and cherry-picking.

Once such a taxonomy is defined, the next step is to determine how to automatically identify the specific changes listed in that taxonomy. Training a model for each type of information change would be cumbersome, potentially requiring separate training data for each label of interest. As such, new methods in the areas of learning from limited data explored in this thesis could be useful for overcoming the need to develop specialized training sets for every type of information change. For example, methods in multi-task learning, domain adaptation, and prompt-based learning could prove useful, given the proper injection of domain expert knowledge into the model. Large generative models (e.g. GPT-3) could also be one avenue to explore given their impressive zero-shot ability. Additionally, these models have an even more promising feature of potentially being able to explain their predictions in natural language. An ideal model would be able to both mark the types of changes occurring between two scientific sentences, as well as explain exactly how those changes appear in text.

One limitation of the current problem formulation is that I consider the matching problem to be 1-to-1. In practice, one may wish to compare a scientific sentence to multiple sources, and it may require integrating several different scientific findings to determine how a piece of science communication gets the message right and wrong. While it is possible in the current setup to simply rank multiple sentences and select any sentences above a certain threshold as the body for comparison, how to consolidate all of that information and select appropriate thresholds is something to explore in future work.

A final consideration is: what do we consider to be ``truth''? This is a problem in fact checking as well, where one must decide what source of information is considered the ground truth state of the world e.g. Wikipedia~\cite{DBLP:conf/naacl/ThorneVCM18}. In this thesis, I have assumed that scientific documents represent truth; in the real world, this isn't always the case~\cite{ioannidis2005most,taylor2015medical,simmons2016false,button2013power}. In fact, it is in the nature of science to change, and what is considered truth at one point in time will likely be disproved, replaced, or amended at a later point. It is then a critical next step to conjure new ways of predicting the trustworthiness and accuracy of scientific papers. This is a difficult task, since without a set ground truth, how does one know whether a scientific article is accurate? One could consider social factors such as the number of citations a paper has or the track record of the authors, but this route is fraught with potential for inadvertent biases and missteps. It is therefore, in my opinion, less a question to be answered solely by computer scientists, but an important question to be engaged with in an interdisciplinary conversation between social scientists, science of science researchers, ethicists, and the public.

The following chapters are prints of the various peer-reviewed papers which constitute this work, and hopefully represent a strong contribution in the area of natural language processing for scientific language understanding.

\newpage

\section*{References for Presented Papers}

\begin{hangingpar}
\textbf{Wright, D.}, \& Augenstein, I. (2020). Claim check-worthiness detection as positive unlabelled learning. In \textit{Findings of EMNLP}. Association for Computational Linguistics.
\\
\end{hangingpar}

\begin{hangingpar}
Atanasova, P.*, \textbf{Wright, D.}*, \& Augenstein, I. (2020). Generating label cohesive and well-formed adversarial claims. In \textit{EMNLP 2020}. Association for Computational Linguistics. \\ \begin{footnotesize}* denotes equal contribution\end{footnotesize}
\\
\end{hangingpar}

\begin{hangingpar}
\textbf{Wright, D.}, \& Augenstein, I. (2020). Transformer based multi-source domain adaptation. In \textit{EMNLP 2020}. Association for Computational Linguistics.
\\
\end{hangingpar}

\begin{hangingpar}
\textbf{Wright, D.}, \& Augenstein, I. (2022). Multi-View Knowledge Distillation from Crowd Annotations for Out-of-Domain Generalization. \textit{arXiv preprint arXiv:}.
\\
\end{hangingpar}

\begin{hangingpar}
\textbf{Wright, D.}, \& Augenstein, I. (2021). CiteWorth: Cite-Worthiness Detection for Improved Scientific Document Understanding. In \textit{Findings of ACL 2021}. Association for Computational Linguistics.
\\
\end{hangingpar}

\begin{hangingpar}
\textbf{Wright, D.}, \& Augenstein, I. (2021). Semi-Supervised Exaggeration Detection of Health Science Press Releases. In \textit{EMNLP 2021}. Association for Computational Linguistics.
\\
\end{hangingpar}

\begin{hangingpar}
\textbf{Wright, D.}, Wadden, D., Lo, K., Kuehl, B., Cohan, A., Augenstein, I., \& Wang, L. L. (2022). Generating Scientific Claims for Zero-Shot Scientific Fact Checking. In \textit{ACL 2022}. Association for Computational Linguistics.
\\
\end{hangingpar}

\begin{hangingpar}
\textbf{Wright, D.}*, Pei J.*, Jurgens D., \& Augenstein, I. (2022). Modeling Information Change in Science Communication with Semantically Matched Paraphrases. In \textit{EMNLP 2022}. Association for Computational Linguistics. \\ \begin{footnotesize}* denotes equal contribution\end{footnotesize}
\\
\end{hangingpar}

\newpage

\setlength\dashlinedash{1.0pt}
\setlength\dashlinegap{1.0pt}
\setlength\arrayrulewidth{0.3pt}

\section{Claim Check-Worthiness Detection as Positive Unlabelled Learning}
\label{paper:check-worthiness}

\subsection{Introduction}
Misinformation is being spread online at ever increasing rates \cite{del2016spreading} and has been identified as one of society's most pressing issues by the World Economic Forum~\cite{howell2013digital}. In response, there has been a large increase in the number of organizations performing fact checking \cite{graves2016rise}. However, the rate at which misinformation is introduced and spread vastly outpaces the ability of any organization to perform fact checking, so only the most salient claims are checked. This obviates the need for being able to automatically find check-worthy content online and verify it.

\begin{figure}[t]
  
  \centering
    \includegraphics[width=0.5\textwidth]{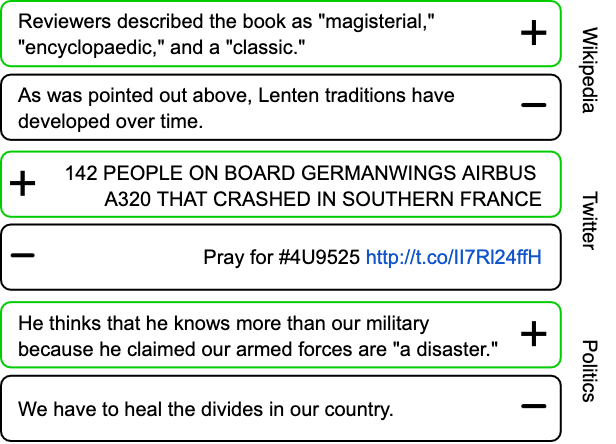}
    \caption{Examples of check-worthy and non check-worthy statements from three different domains. Check-worthy statements are those which were judged to require evidence or a fact check.}
    \label{fig:check-worthy-examples}
\end{figure}
The natural language processing and machine learning communities have recently begun to address the problem of automatic fact checking~\cite{vlachos2014fact,hassan2017claimbuster, thorne-vlachos-2018-automated-fixed,DBLP:conf/emnlp/AugensteinLWLHH19,DBLP:conf/acl/AtanasovaSLA20,atanasova-etal-2020-generating,ostrowski2020multihop,allein2020timeaware}. The first step of automatic fact checking is claim check-worthiness detection, a text classification problem where, given a statement, one must predict if the content of that statement makes ``an assertion about the world that is checkable'' \cite{konstantinovskiy2018towards}.
There are multiple isolated lines of research which have studied variations of this problem. \autoref{fig:check-worthy-examples} provides examples from three tasks which are studied in this work: rumour detection on Twitter~\cite{zubiaga2016analysing,journals/ipm/ZubiagaKLPLBCA18}, check-worthiness ranking in political debates and speeches~\cite{atanasova2018overview,elsayed2019overview,barron2020checkthat}, and citation needed detection on Wikipedia~\cite{redi2019citation}. Each task is concerned with a shared underlying problem: detecting claims which warrant further verification. However, no work has been done to compare all three tasks to understand shared challenges in order to derive shared solutions, which could enable improving claim check-worthiness detection systems across multiple domains.

Therefore, we ask the following main research question in this work: are these all variants of the same task, and if so, is it possible to have a unified approach to all of them? We answer this question by investigating the problem of annotator subjectivity, where annotator background and expertise causes their judgement of what is check-worthy to differ, leading to false negatives in the data~\cite{konstantinovskiy2018towards}. Our proposed solution is \textit{Positive Unlabelled Conversion (PUC)}, an extension of Positive Unlabelled (PU) learning, which converts negative instances into positive ones based on the estimated prior probability of an example being positive. We demonstrate that a model trained using \textit{PUC} improves performance on English \textit{citation needed detection} and %
\textit{Twitter rumour detection}. We also show that by pretraining a model on citation needed detection, one can further improve results on Twitter rumour detection over a model trained solely on rumours, highlighting that a unified approach to these problems is achievable. Additionally, we show that one attains better results on %
\textit{political speeches}  check-worthiness ranking without using any form of PU learning, arguing through a dataset analysis that the labels are much more subjective than the other two tasks.

The \textbf{contributions} of this work are as follows:
\begin{enumerate}[noitemsep]
    \item The first thorough comparison of multiple claim check-worthiness detection tasks.
    \item \textit{Positive Unlabelled Conversion (PUC)}, a novel extension of PU learning to support check-worthiness detection across domains.
    \item Results demonstrating that a unified approach to check-worthiness detection is achievable for 2 out of 3 tasks, improving over the state-of-the-art for those tasks.
\end{enumerate}
\subsection{Related Work}

\subsubsection{Claim Check-Worthiness Detection}
As the first step in automatic fact checking, claim check-worthiness detection is a binary classification problem which involves determining if a piece of text makes ``an assertion about the world which can be checked''~\cite{konstantinovskiy2018towards}. We adopt this broad definition as it allows us to perform a structured comparison of many publicly available datasets. The wide applicability of the definition also allows us to study if and how a unified cross-domain approach could be developed. 

Claim check-worthiness detection can be subdivided into three distinct domains: rumour detection on Twitter, check-worthiness ranking in political speeches and debates, and citation needed detection on Wikipedia. A few studies have been done which attempt to create full systems for mining check-worthy statements, including the works of \citet{konstantinovskiy2018towards}, ClaimRank \cite{jaradat2018claimrank}, and ClaimBuster \cite{hassan2017claimbuster}. They develop full software systems consisting of relevant source material retrieval, check-worthiness classification, and dissemination to the public via end-user applications. These works are focused solely on the political domain, using data from political TV shows, speeches, and debates. In contrast, in this work we study the claim check-worthiness detection problem across three domains which have publicly available data: Twitter~\cite{zubiaga2017exploiting}, political speeches~\cite{atanasova2018overview}, and Wikipedia~\cite{redi2019citation}.

\paragraph{Rumour Detection on Twitter}
Rumour detection on Twitter is primarily studied using the PHEME dataset~\cite{zubiaga2016analysing}, a set of tweets and associated threads from breaking news events which are either rumourous or not. Published systems which perform well on this task include contextual models (e.g. conditional random fields) acting on a tweet's thread~\cite{zubiaga2017exploiting,journals/ipm/ZubiagaKLPLBCA18}, identifying salient rumour-related words~\cite{abulaish2019graph}, and using a GAN to generate misinformation in order to improve a downstream discriminator~\cite{ma2019detect}.

\paragraph{Political Speeches}
For political speeches, the most studied datasets come from the Clef CheckThat! shared tasks ~\cite{atanasova2018overview, elsayed2019overview,barron2020checkthat} and ClaimRank~\cite{jaradat2018claimrank}. The data consist of transcripts of political debates and speeches where each sentence has been annotated by an independent news or fact-checking organization for whether or not the statement should be checked for veracity. The most recent and best performing system on the data considered in this paper consists of a two-layer bidirectional GRU network which acts on both word embeddings and syntactic parse tags~\cite{hansen2019neural}. In addition, they augment the native dataset with weak supervision from unlabelled political speeches.

\paragraph{Citation Needed Detection}
Wikipedia citation needed detection has been investigated recently in~\cite{redi2019citation}. The authors present a dataset of sentences from Wikipedia labelled for whether or not they have a citation attached to them. They also released a set of sentences which have been flagged as not having a citation but needing one (i.e. \textit{unverified}). In contrast to other check-worthiness detection domains, there are much more training data available on Wikipedia. However, the rules for what requires a citation do not necessarily capture all ``checkable'' statements, as ``all material in Wikipedia articles must be verifiable''~\cite{redi2019citation}. 
Given this, we view Wikipedia citation data as a set of positive and unlabelled data: statements which have attached citations are positive samples of check-worthy statements, and within the set of statements without citations there exist some positive samples (those needing a citation) and some negative samples. 
Based on this, this domain constitutes the most general formulation of check-worthiness among the domains we consider. Therefore, we experiment with using data from this domain as a source for transfer learning, training variants of PU learning models on it, then applying them to target data from other domains.

\subsubsection{Positive Unlabelled Learning}
PU learning methods attempt to learn good binary classifiers given only positive labelled and unlabelled data. Recent applications where PU learning has been shown to be beneficial include detecting deceptive reviews online~\cite{li2014spotting,ren2014positive}, keyphrase extraction~\cite{sterckx2016supervised} and named entity recognition~\cite{peng2019distantly}. For a survey on PU learning, see~\cite{bekker2018learning}, and for a formal definition of PU learning, see \S\ref{sec:pu_learning}.

Methods for learning positive-negative (PN) classifiers from PU data have a long history~\cite{denis1998pac,de1999positive,letouzey2000learning}, with one of the most seminal papers being from~\citet{elkan2008learning}. In this work, the authors show that by assuming the labelled samples are a random subset of all positive samples, one can utilize a classifier trained on PU data in order to train a different classifier to predict if a sample is positive or negative. The process involves training a PN classifier with positive samples being shown to the classifier once and \textit{unlabelled} samples shown as \textit{both} a positive sample and a negative sample. The loss for the duplicated samples is weighted by the confidence of a PU classifier that the sample is positive.

Building on this, du Plessis et al.~\citep{du2014analysis} propose an unbiased estimator which improves the estimator introduced in~\cite{elkan2008learning} by balancing the loss for positive and negative classes. The work of Kiryo et al.~\citep{kiryo2017positive} extends this method to improve the performance of deep networks on PU learning. Our work builds on the method of Elkan and Noto~\citep{elkan2008learning} by relabelling samples which are highly confidently positive.

\subsection{Methods}
\begin{figure*}[t]
  \centering
    \includegraphics[width=0.95\textwidth]{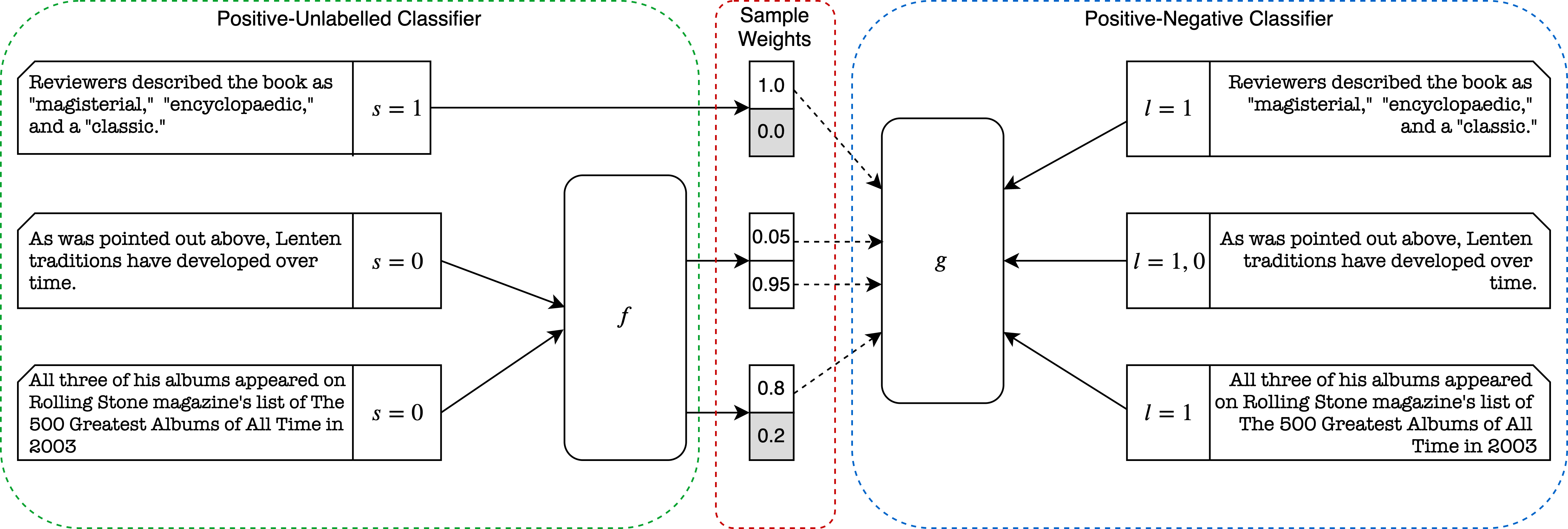}
    \caption{High level view of \textit{PUC}. A PU classifier ($f$, green box) is first learned using PU data (with $s$ indicating if the sample is positive or unlabelled). From this the prior probability of a sample being positive is estimated. Unlabelled samples are then ranked by $f$ (red box) and the most positive samples are converted into positives until the dataset is balanced according to the estimated prior. The model $g$ is then trained using the duplication and weighting method of \citet{elkan2008learning} as described in \S\ref{sec:pu_learning} with labels $l$ (blue box). Greyed out boxes are negative weights which are ignored when training the classifier $g$, as those examples are only trained as positives.}
      \label{fig:puc}
\end{figure*}
The task considered in this paper is to predict if a statement makes ``an assertion about the world that is checkable'' \cite{konstantinovskiy2018towards}. %
As the subjectivity of annotations for existing data on claim check-worthiness detection is a known problem~\cite{konstantinovskiy2018towards}, we view the data as a set of positive and unlabelled (PU) data. In addition, we unify our approach to each of them by viewing Wikipedia data as an abundant source corpus.
Models are then trained on this source corpus using variants of PU learning and transferred via fine-tuning to the other claim check-worthiness detection datasets, which are subsequently trained on as PU data. On top of vanilla PU learning, we introduce \textit{Positive Unlabelled Conversion (PUC)} which relabels examples that are most confidently positive in the unlabelled data. A formal task definition, description of PU learning, and explanation of the \textit{PUC} extension are given in the following sections.

\subsubsection{Task Definition}
The fundamental task is binary text classification. In the case of positive-negative (PN) data, we have a labelled dataset $\mathcal{D}: \{(x, y)\}$ with input features $x \in \mathbb{R}^{d}$ and labels $y \in \{0, 1\}$. The goal is to learn a classifier $g: x \rightarrow (0,1)$ indicating the probability that the input belongs to the positive class. With PU data, the dataset $\mathcal{D}$ instead consists of samples $\{(x, s)\}$, where the value $s \in \{0,1\}$ indicates if a sample is labelled or not. The primary difference from the PN case is that, unlike for the labels $y$, a value of $s = 0$ does not denote the sample is negative, but that the label is unknown. The goal is then to learn a PN classifier $g$ using a PU classifier $f: x \rightarrow (0,1)$ which predicts whether or not a sample is labelled~\cite{elkan2008learning}. 

\subsubsection{PU Learning}
\label{sec:pu_learning}
Our overall approach is depicted in~\autoref{fig:puc}. We begin with an explanation of the PU learning algorithm described in~\cite{elkan2008learning}. Assume that we have a dataset randomly drawn from some probability distribution $p(x,y,s)$, where samples are of the form $(x,s), ~s \in \{0,1\}$ and $s=1$ indicates that the sample is labelled. The variable $y$ is unknown, but we make two assumptions which allow us to derive an estimator for probabilities involving $y$. The first is that:
\begin{equation}
    p(y = 0 | s = 1) = 0
\end{equation}
In other words, if we know that a sample is labelled, then that label cannot be 0. The second assumption is that labelled samples are Selected Completely At Random from the underlying distribution (also known as the SCAR assumption). Check-worthiness data can be seen as an instance of SCAR PU data; annotators tend to only label those instances which are very clearly check-worthy in \textit{their} opinion~\cite{konstantinovskiy2018towards}. When combined across several annotators, we assume this leads to a random sample from the total set of check-worthy statements.

Given this, a classifier $f : x \rightarrow (0,1)$ is trained to predict $p(s=1|x)$ from the PU data. It is then employed to train a classifier $g$ to predict $p(y=1|x)$ by first estimating $c = p(s=1|y=1)$ on a set of validation data. Considering a validation set $V$ where $P \subset V$ is the set of positive samples in $V$, $c$ is estimated as:
\begin{equation}
\label{eq:c_estimate}
    c \approx \frac{1}{|P|}\sum_{x \in P}f(x)
\end{equation}
This says our estimate of $p(s=1|y=1)$ is the average confidence of our classifier on known positive samples. Next, we can estimate $E_{p(x,y,s)}[h(x,y)]$ for any arbitrary function $h$ empirically from a dataset of $k$ samples as follows:
\begin{dmath}
    E[h] = \frac{1}{k}(\sum_{(x,s=1)}h(x,1) + \sum_{(x,s=0)}w(x)h(x,1) + (1-w(x))h(x,0))
\end{dmath}
\begin{equation}
    w(x) = p(y=1|x,s=0) = \frac{1-c}{c}\frac{p(s=1|x)}{1-p(s=1|x)}
\end{equation}
In this case, $c$ is estimated using~\autoref{eq:c_estimate} and $p(s=1|x)$ is estimated using the classifier $f$. The derivations for these equations can be found in~\cite{elkan2008learning}.

To estimate $p(y=1|x)$ empirically, the unlabelled samples in the training data are duplicated, with one copy negatively labelled and one copy positively labelled. Each copy is trained on with a weighted loss $w(x)$ when the label is positive and $1 - w(x)$ when the label is negative. Labelled samples are trained on normally (i.e. a single copy with unit weight).

\subsubsection{Positive Unlabelled Conversion}
For \textit{PUC}, the motivation is to relabel those samples from the unlabelled data which are very clear cut positive. To accomplish this, we start with the fact that one can also estimate the prior probability of a sample having a positive label using $f$. If instead of $h$ we want to estimate $E[y] = p(y=1)$, the following is obtained:
\begin{equation}
    p(y=1) \approx \frac{1}{k}(\sum_{x,s=1}1 + \sum_{x,s=0}w(x))
\end{equation}
This estimate is then utilized to convert the most confident unlabelled samples into positives. First, all of the unlabelled samples are ranked according to their calculated weight $w(x)$. The ranked samples are then iterated through and converted into positive-only samples until the distribution of positive samples is greater than or equal to the estimate of $p(y=1)$. Unlike in vanilla PU learning, these samples are discretized to have a positive weight of 1, and trained on by the classifier $g$ once per epoch as positive samples along with the labelled samples. The remaining unlabelled data are trained on in the same way as in vanilla PU learning. %

\subsubsection{Implementation}
In order to create a unified approach to check-worthiness detection, transfer learning from Wikipedia citation needed detection is employed. 
To accomplish this, we start with a training dataset $\mathcal{D}^{s}$ of statements from Wikipedia featured articles that are either labelled as containing a citation (positive) or unlabelled. We train a classifier $f^{s}$ on this dataset and obtain a classifier $g^{s}$ via \textit{PUC}. For comparison, we also train models with vanilla PU learning and PN learning as baselines. The network architecture for both $f^{s}$ and $g^{s}$ is BERT~\cite{devlin2019bert}, a large pretrained transformer-based~\cite{vaswani2017attention} language model. We use the HuggingFace transformers implementation of the 12-layer 768 dimensional variation of BERT~\cite{DBLP:conf/emnlp/WolfDSCDMCRLFDS20}. The classifier in this implementation is a two layer neural network acting on the \texttt{[CLS]} token.

From $g^{s}$, we train a classifier $g^{t}$ using downstream check-worthiness detection dataset $D^{t}$ by initializing $g^{t}$ with the base BERT network from $g^{s}$ and using a new randomly initialized final layer. In addition, we train a model $f^{t}$ on the target dataset, and train $g^{t}$ with \textit{PUC} from this model to obtain the final classifier. As a baseline, we also experiment with training on just the dataset $D^{t}$ without any pretraining. In the case of citation needed detection, since the data comes from the same domain we simply test on the test split of statements labelled as ``citation needed'' using the classifier $g^{s}$. We compare our models to the published state of the art baselines on each dataset.

For all of our models ($f^s$, $g^s$, $f^t$, $g^t$) we train for two epochs, saving the weights with the best F1 score on validation data as the final model. Training is performed with a max learning rate of 3e-5 and a triangular learning rate schedule~\cite{DBLP:conf/acl/RuderH18} that linearly warms up for 200 training steps, then linearly decays to 0 for the rest of training. For regularization we add L2 loss with a coefficient of 0.01, and dropout with a rate of 0.1. Finally, we split the training sets into 80\% train and 20\% validation, and train with a batch size of 8. 
The code to reproduce our experiments can be found here.\footnote{\url{https://github.com/copenlu/check-worthiness-pu-learning}} 

\subsection{Experimental Results}
To what degree is claim check-worthiness detection a PU learning problem, and does this enable a unified approach to check-worthiness detection? In our experiments, we progressively answer this question by answering the following: 1) is PU learning beneficial for the tasks considered? 2) Does PU citation needed detection transfer to rumour detection? 3) Does PU citation needed detection transfer to political speeches? To investigate how well the data in each domain reflects the definition of a check-worthy statement as one which ``makes an assertion about the world which is checkable'' and thus understand subjectivity in the annotations, we perform a dataset analysis comparing the provided labels of the top ranked check-worthy claims from the \textit{PUC} model with the labels given by two human annotators. In all experiments, we report the mean performance of our models and standard deviation across 15 different random seeds. Additionally, we report the performance of each model ensembled across the 15 runs through majority vote on each sample.

\subsubsection{Datasets}
\begin{table*}[t]
    \centering
    \fontsize{10}{10}\selectfont
    \begin{tabular}{l c c c : c c c}
    \toprule %
    Method & P & R & \multicolumn{1}{c}{F1} & eP & eR & eF1\\
    \midrule %
       \citealt{redi2019citation}  & 75.3& 70.9& 73.0 [76.0]*& - & - & -\\
    \hdashline
       \rule{0pt}{2ex}BERT  & \underline{78.8 $\pm$ 1.3}& 83.7 $\pm$ 4.5& 81.0 $\pm$ 1.5 & 79.0 & 85.3 & 82.0 \\
       BERT + PU  & \textbf{78.8 $\pm$ 0.9}& \underline{84.3 $\pm$ 3.0}& \underline{81.4 $\pm$ 1.0} & 79.0 & \underline{85.6} & \underline{82.2}\\
       BERT + \textit{PUC}  & 78.4 $\pm$ 0.9& \textbf{85.6 $\pm$ 3.2}& \textbf{81.8 $\pm$ 1.0} & 78.6 & \textbf{87.1} & \textbf{82.6}\\
    \bottomrule %

    \end{tabular}
    \caption{F1 and ensembled F1 score for citation needed detection training on the FA split and testing on the LQN split of \cite{redi2019citation}. The FA split contains statements with citations from featured articles and the LQN split consists of statements which were flagged as not having a citation but needing one. Listed are the mean, standard deviation, and ensembled results across 15 seeds (eP, eR, and eF1). \textbf{Bold} indicates best performance, \underline{underline} indicates second best. *The reported value is from rerunning their released model on the test dataset. The value in brackets is the value reported in the original paper.}
    \label{tab:citation_detection_results}
\end{table*}
\paragraph{Wikipedia Citations}
We use the dataset from \citet{redi2019citation} for citation needed detection. The dataset is split into three sets: one coming from featured articles (deemed `high quality', 10k positive and 10k negative statments), one of statements which have no citation but have been flagged as needing one (10k positive, 10k negative), and one of statements from random articles which have citations (50k positive, 50k negative). In our experiments the models were trained on the high quality statements from featured articles and tested on the statements which were flagged as `citation needed'. The key differentiating features of this dataset from the other two datasets are: 1) the domain of text is Wikipedia and 2) annotations are based on the decisions of Wikipedia editors following Wikipedia guidelines for citing sources\footnote{\url{https://en.wikipedia.org/wiki/Wikipedia:Citing_sources}}.

\paragraph{Twitter Rumours}
The PHEME dataset of rumours is employed for Twitter claim check-worthiness detection~\cite{zubiaga2016analysing}. The data consists of 5,802 annotated tweets from 5 different events, where each tweet is labelled as rumourous or non-rumourous (1,972 rumours, 3,830 non-rumours). We followed the leave-one-out evaluation scheme of~\cite{zubiaga2017exploiting}, namely, we performed a 5-fold cross-validation for all methods, training on 4 events and testing on 1. The key differentiating features of this dataset from the other two datasets are: 1) the domain of data is tweets and 2) annotations are collected from professional journalists specifically for building a dataset to train machine learning models.

\paragraph{Political Speeches}
The dataset we adopted in the political speeches domain is the same as in~\citet{hansen2019neural}, consisting of 4 political speeches from the 2018 Clef CheckThat! competition~\cite{atanasova2018overview} and 3 political speeches from ClaimRank~\cite{jaradat2018claimrank} (2,602 statements total). We performed a 7-fold cross-validation, using 6 splits as training data and 1 as test in our experimental setup. The data from ClaimRank is annotated using the judgements from 9 fact checking organizations, and the data from Clef 2018 is annotated by factcheck.org. The key differentiating features of this dataset from the other two datasets are: 1) the domain of data is transcribed spoken utterances from political speeches and 2) annotations are taken from 9 fact checking organizations gathered independently.

\subsubsection{Is PU Learning Beneficial for Citation Needed Detection?}

Our results for citation needed detection are given in \autoref{tab:citation_detection_results}. The vanilla BERT model already significantly outperforms the state of the art model from Redi et al.~\citep{redi2019citation} (a GRU network with global attention) by 6 F1 points. We see further gains in performance with PU learning, as well as when using \textit{PUC}. Additionally, the models using PU learning have lower variance, indicating more consistent performance across runs. The best performing model we see is the one trained using \textit{PUC} with an F1 score of 82.6. We find that this confirms our hypothesis that citation data is better seen as a set of positive and unlabelled data when used for check-worthiness detection. In addition, it gives some indication that PU learning improves the generalization power of the model, which could make it better suited for downstream tasks.

\subsubsection{Does PU Citation Needed Detection Transfer to Rumour Detection?}
\begin{table*}[t]
    \centering
    \fontsize{10}{10}\selectfont
    \begin{tabular}{l c c c : c c c}
    \toprule
    Method & $\mu$P & $\mu$R & \multicolumn{1}{c}{$\mu$F1}& eP & eR & eF1\\
    \midrule
       \citealt{zubiaga2017exploiting}  & 66.7& 55.6& 60.7 & - & - & -\\
       BiLSTM & 62.3 &	56.4 &	59.0 & - & - & -\\
    \hdashline
       \rule{0pt}{2ex}BERT  & \underline{69.9 $\pm$ 1.7}& 60.8 $\pm$ 2.6& 65.0 $\pm$ 1.3 & 71.3 & 61.9	& 66.3\\
       BERT + Wiki  & 69.3 $\pm$ 1.6& 61.4 $\pm$ 2.6& 65.1 $\pm$ 1.2 & 70.7 & 62.2 & 66.2\\
       BERT + WikiPU  & \underline{69.9 $\pm$ 1.3}& 62.5 $\pm$ 1.6& 66.0 $\pm$ 1.1 & \textbf{72.2} & 64.6 & 68.2\\
       BERT + Wiki\textit{PUC}  & \textbf{70.1 $\pm$ 1.1}& 61.8 $\pm$ 1.8& 65.7 $\pm$ 1.0 & \underline{71.5} & 62.7 & 66.8\\
       BERT + PU  & 68.7 $\pm$ 1.2& 64.7 $\pm$ 1.8& 66.6 $\pm$ 0.9 & 69.9 & 65.2 & 67.5\\
       BERT + \textit{PUC}  & 68.1 $\pm$ 1.5& 65.3 $\pm$ 1.6& 66.6 $\pm$ 0.9 & 69.1 & 66.3 & 67.7\\
       BERT + PU + WikiPU  & 68.4 $\pm$ 1.2& \textbf{66.1 $\pm$ 1.2}& \textbf{67.2 $\pm$ 0.6} & 69.3 & \underline{67.2} & \underline{68.3}\\
       BERT + \textit{PUC} + WikiPUC  & 68.0 $\pm$ 1.4& \underline{66.0 $\pm$ 2.0}& \underline{67.0 $\pm$ 1.3} & 69.4 & \textbf{67.5} & \textbf{68.5}\\
    \bottomrule

    \end{tabular}
    \caption{micro-F1 ($\mu$F1) and ensembled F1 (eF1) performance of each system on the PHEME dataset. Performance is averaged across the five splits of~\cite{zubiaga2017exploiting}. Results show the mean, standard deviation, and ensembled score across 15 seeds. \textbf{Bold} indicates best performance, \underline{underline} indicates second best.}
    \label{tab:pheme_results}
\end{table*}
\subsubsubsection{Baselines}
The best published method that we compare to is the CRF from~\cite{zubiaga2017exploiting}. which utilizes a combination of content and social features. Content features include word vectors, part-of-speech tags, and various lexical features, and social features include tweet count, listed count, follow ratio, age, and whether or not a user is verified. The CRF acts on a timeline of tweets, making it contextual. In addition, we include results from a 2-layer BiLSTM with FastText embeddings~\cite{bojanowski2017enriching}. There exist other deep learning models which have been developed for this task, including \cite{ma2019detect} and \cite{abulaish2019graph}, but they do not publish results on the standard splits of the data and we were unable to recreate their results, and thus are omitted.%

\subsubsubsection{Results}
The results for the tested systems are given in \autoref{tab:pheme_results}. Again we see large gains from BERT based models over the baseline from \cite{zubiaga2017exploiting} and the 2-layer BiLSTM. Compared to training solely on PHEME, fine tuning from basic citation needed detection sees little improvement (0.1 F1 points). However, fine tuning a model trained using PU learning leads to an increase of 1 F1 point over the non-PU learning model, indicating that PU learning enables the Wikipedia data to be useful for transferring to rumour detection i.e. the improvement is not only from a better semantic representation learned from Wikipedia data. For \textit{PUC}, we see an improvement of 0.7 F1 points over the baseline and lower overall variance than vanilla PU learning, meaning that the results with \textit{PUC} are more consistent across runs. The best performing models also use PU learning on in-domain data, with the best average performance being from the models trained using PU/\textit{PUC} on in domain data and initialized with weights from a Wikipedia model trained using PU/\textit{PUC}. When models are ensembled, pretraining with vanilla PU learning improves over no pretraining by almost 2 F1 points, and the best performing models which are also trained using PU learning on in domain data improve over the baseline by over 2 F1 points. We conclude that framing rumour detection on Twitter as a PU learning problem leads to improved performance.

Based on these results, we are able to confirm two of our hypotheses. The first is that Wikipedia citation needed detection and rumour detection on Twitter are indeed similar tasks, and a unified approach for both of them is possible. Pretraining a model on Wikipedia provides a clear downstream benefit when fine-tuning on Twitter data, \textit{precisely when PU/PUC is used}. Additionally, training using \textit{PUC} on in domain Twitter data provides further benefit. This shows that \textit{PUC} constitutes a unified approach to these two tasks.

The second hypothesis we confirm is that both Twitter and Wikipedia data are better seen as positive and unlabelled for claim check-worthiness detection. When pretraining with the data as a traditional PN dataset there is no performance gain and in fact a performance loss when the models are ensembled. PU learning allows the model to learn better representations for general claim check-worthiness detection.

To explain why this method performs better, \autoref{tab:citation_detection_results} and \autoref{tab:pheme_results} show that \textit{PUC} improves model recall at very little cost to precision. The aim of this is to mitigate the issue of subjectivity in the annotations of check-worthiness detection datasets noted in previous work \cite{konstantinovskiy2018towards}. Some of the effects of this are illustrated in \autoref{tab:pheme_pos_better} and \autoref{tab:pheme_neg_better} in \S\ref{sec:puc_examples}. The \textit{PUC} models are better at distinguishing rumours which involve claims of fact about people i.e. things that people said or did, or qualities about people. For non-rumours, the \textit{PUC} pretrained model is better at recognizing statements which describe qualitative information surrounding the events and information that is self-evident e.g. a tweet showing the map where the Charlie Hebdo attack took place.

\subsubsection{Does PU Citation Needed Detection Transfer to Political Speeches?}
\subsubsubsection{Baselines}
The baselines we compare to are the state of the art models from ~\citet{hansen2019neural} and \citet{konstantinovskiy2018towards}. The model from \citet{konstantinovskiy2018towards} consists of InferSent embeddings~\cite{conneau2017supervised} concatenated with POS tag and NER features passed through a logistic regression classifier. The model from \citet{hansen2019neural} is a bidirectional GRU network acting on syntatic parse features concatenated with word embeddings as the input representation.

\subsubsubsection{Results}
The results for political speech check-worthiness detection are given in \autoref{tab:clef_results}. We find that the BERT model initialized with weights from a model trained on plain Wikipedia citation needed statements performs the best of all models. As we add transfer learning and PU learning, the performance steadily drops. We perform a dataset analysis to gain some insight into this effect in \S\ref{sec:dataset_analysis}.
\begin{table}
    \centering
    \fontsize{10}{10}\selectfont
    \begin{tabular}{l c}
    \toprule
    Method & MAP\\
    \midrule
        \citealt{konstantinovskiy2018towards} & 26.7\\
       \citealt{hansen2019neural}  & 30.2\\
    \hdashline
       \rule{0pt}{2ex}BERT  &33.0 $\pm$ 1.8\\
       BERT + Wiki  & \textbf{34.4 $\pm$ 2.7}\\
       BERT + WikiPU  & \underline{33.2 $\pm$ 1.7}\\
       BERT + Wiki\textit{PUC}  & 31.7 $\pm$ 1.8\\
       BERT + PU  & 18.8 $\pm$ 3.7\\
       BERT + \textit{PUC}  & 26.7 $\pm$ 2.8\\
       BERT + PU + WikiPU  & 16.8 $\pm$ 3.5\\
       BERT + \textit{PUC} + Wiki\textit{PUC}  & 27.8 $\pm$ 2.7\\
    \bottomrule

    \end{tabular}
    \caption{Mean average precision (MAP) of models on political speeches. \textbf{Bold} indicates best performance, \underline{underline} indicates second best.}
    \label{tab:clef_results}
\end{table}

\subsubsection{Dataset Analysis}
\label{sec:dataset_analysis}
In order to understand our results in the context of the selected datasets, we perform an analysis to learn to what extent the positive samples in each dataset reflect the definition of a check-worthy claim as ``an assertion about the world that is checkable''. We ranked all of the statements based on the predictions of 15 \textit{PUC} models trained with different seeds, where more positive class predictions means a higher rank (thus more check-worthy), and had two experts manually relabel the top 100 statements. The experts were informed to label the statements based on the definition of check-worthy given above. We then compared the manual annotation to the original labels using F1 score. Higher F1 score indicates the dataset better reflects the definition of check-worthy we adopt in this work. Our results are given in \autoref{tab:relabel_results}.
\begin{table}
    \centering
    \fontsize{10}{10}\selectfont
    \begin{tabular}{l c c c}
    \toprule
    Dataset & P & R & F1\\
    \midrule
                  &  81.7 &87.0 &84.3\\
       Wikipedia  & 84.8 & 87.0 & 85.9\\
                  & \textit{83.3}& \textit{87.0}& \textit{85.1}\\
       \hdashline
        \rule{0pt}{2ex}  & 87.5& 82.4& 84.8 \\
       Twitter  & 86.3 & 81.2 & 83.6\\
                & \textit{86.9}& \textit{81.8}& \textit{84.2} \\
       \hdashline
        \rule{0pt}{2ex} &33.8& 89.3& 49.0\\
       Politics  &31.1&  100.0&  47.5\\
                &\textit{32.5} &\textit{94.7}& \textit{48.3}\\
    \bottomrule

    \end{tabular}
    \caption{F1 score comparing manual relabelling of the top 100 predictions by \textit{PUC} model with the original labels in each dataset by two different annotators. \textit{Italics} are average value between the two annotators.}
    \label{tab:relabel_results}
\end{table}

We find that the Wikipedia and Twitter datasets contain labels which are more general, evidenced by similar high F1 scores from both annotators ($>$ 80.0). For political speeches, we observe that the human annotators both found many more examples to be check-worthy than were labelled in the dataset. This is evidenced by examples such as \textit{It's why our unemployment rate is the lowest it's been in so many decades} being labelled as not check-worthy and \textit{New unemployment claims are near the lowest we've seen in almost half a century} being labelled as check-worthy in the same document in the dataset's original annotations. This characteristic has been noted for political debates data previously~\cite{konstantinovskiy2018towards}, which was also collected using the judgements of independent fact checking organizations~\cite{gencheva-etal-2017-context-fixed}. Labels for this dataset were collected from various news outlets and fact checking organizations, which may only be interested in certain types of claims such as those most likely to be false. This makes it difficult to train supervised machine learning models for general check-worthiness detection based solely on text content and document context due to labelling inconsistencies. 

\subsection{Discussion and Conclusion}
In this work, we approached claim check-worthiness detection by examining how to unify three distinct lines of work. We found that check-worthiness detection is challenging in any domain as there exist stark differences in how annotators judge what is check-worthy. We showed that one can correct for this and improve check-worthiness detection across multiple domains by using positive unlabelled learning. Our method enabled us to perform a structured comparison of datasets in different domains, developing a unified approach which outperforms state of the art in 2 of 3 domains and illuminating to what extent these datasets reflect a general definition of check-worthy. 

Future work could explore different neural base architectures. Further, 
it could potentially benefit all tasks to consider the greater context in which statements are made. We would also like to acknowledge again that all experiments have only focused on English language datasets; developing models for other, especially low-resource languages, would likely result in additional challenges. We hope that this work will inspire future research on check-worthiness detection, which we see as an under-studied problem, with a focus on developing resources and models across many domains such as Twitter, news media, and spoken rhetoric.

\section*{Acknowledgements}
$\begin{array}{l}\includegraphics[width=1cm]{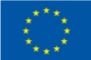} \end{array}$ This project has received funding from the European Union's Horizon 2020 research and innovation programme under the Marie Sk\l{}odowska-Curie grant agreement No 801199.

\newpage

\section{Generating Label Cohesive and Well-Formed Adversarial Claims}
\label{paper:adversarial}

\subsection{Introduction}
\begin{figure}[t]
\centering
\includegraphics[width=0.65 \columnwidth]{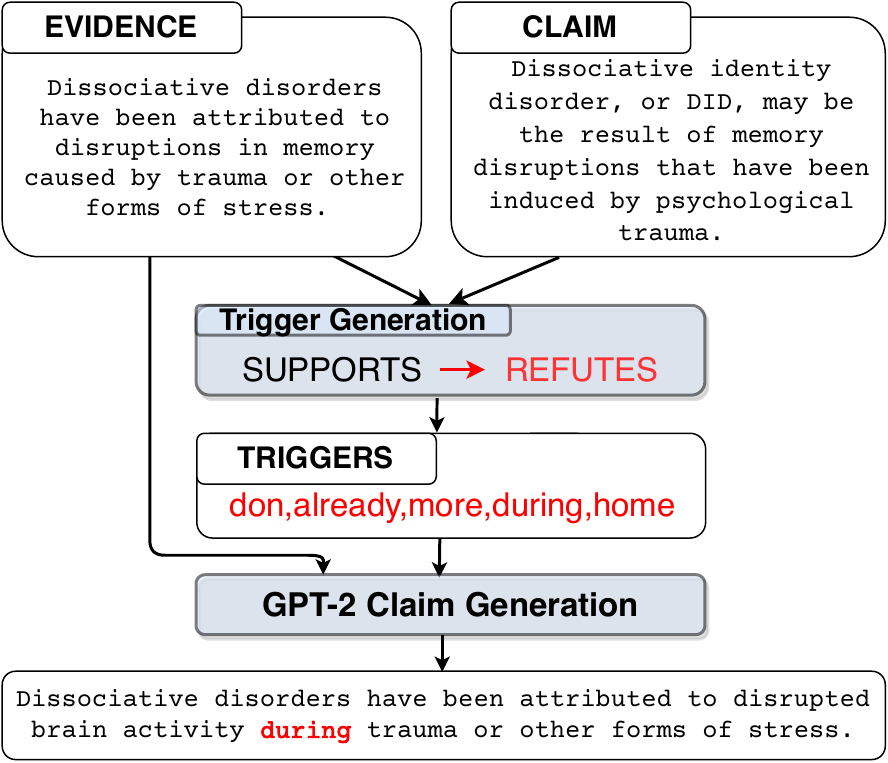}
\caption{High level overview of our method. First, universal triggers are discovered for flipping a source to a target label (e.g. SUPPORTS $\rightarrow$ REFUTES). These triggers are then used to condition the GPT-2 language model to generate novel claims with the original label, including at least one of the found triggers.}
\label{fig:adversarial-architecture}
\end{figure}
Adversarial examples~\cite{goodfellow2015explaining, szegedy2013intriguing} are deceptive model inputs designed to mislead an ML system into making the wrong prediction. They expose regions of the input space that are outside the training data distribution where the model is unstable. 
It is important to reveal such vulnerabilities and correct for them, especially for tasks such as fact checking (FC). %

In this paper, we explore the vulnerabilities of FC models trained on the FEVER dataset~\cite{DBLP:conf/naacl/ThorneVCM18}, where the inference between a claim and evidence text is predicted. We particularly construct \textit{universal adversarial triggers}~\cite{wallace2019universal} -- single n-grams appended to the input text that can shift the prediction of a model from a source class to a target one. Such adversarial examples are of particular concern, as they can apply to a large number of input instances. 

However, we find that the triggers also change the meaning of the claim such that the true label is in fact the target class. For example, when attacking a claim-evidence pair with a `SUPPORTS' label, a common unigram found to be a universal trigger when switching the label to `REFUTES' is `none'. Prepending this token to the claim drastically changes the meaning of the claim such that the new claim is in fact a valid 
`REFUTES' claim as opposed to an adversarial `SUPPORTS' claim. %
Furthermore, we find adversarial examples constructed in this way to be nonsensical, as a new token is simply being attached to an existing claim. 

Our \textbf{contributions} are as follows. We \textit{preserve the meaning} of the source text and \textit{improve the semantic validity} of universal adversarial triggers to automatically construct more potent adversarial examples. This is accomplished via: 1) a \textit{novel extension to the HotFlip attack}~\cite{ebrahimi2018hotflip}, where we jointly minimize the target class loss of a FC model and the attacked class loss of a natural language inference model; 2) a \textit{conditional language model} trained using GPT-2~\cite{radford2019language}, which takes %
trigger tokens and a piece of evidence, and generates a semantically coherent new claim containing at least one trigger. %
The resulting triggers maintain potency against a FC model while preserving the original claim label. Moreover, the conditional language model produces semantically coherent adversarial examples containing triggers, on which a FC model performs 23.8\% worse than with the original FEVER claims. The code for the paper is publicly available.\footnote{https://github.com/copenlu/fever-adversarial-attacks}

\subsection{Related Work}
\subsubsection{Adversarial Examples}
Adversarial examples for %
NLP systems can be constructed as automatically generated text~\cite{ren2019generating} or perturbations of existing input instances~\cite{jintextfool,ebrahimi2018hotflip}. For a %
detailed literature overview, see~\citet{zhang2019adversarial}.

One potent type of adversarial techniques are universal adversarial attacks~\cite{gao2019universal, wallace2019universal} -- single perturbation changes that can be applied to a large number of input instances and that cause significant performance decreases of the model under attack. 
~\citet{wallace2019universal} find universal adversarial triggers that can change the prediction of the model using the HotFlip algorithm~\cite{ebrahimi2018hotflip}. 

However, for NLI tasks, they also change the meaning of the instance they are appended to, and the prediction of the model remains correct. ~\citet{michel2019evaluation} %
address this by exploring only perturbed instances in the neighborhood of the original one.
Their approach is for instance-dependent attacks, whereas we suggest finding \textit{universal} adversarial triggers that also preserve the original meaning of input instances. 
Another approach to this 
are rule-based perturbations of the input~\cite{ribeiro2018semantically} or imposing adversarial constraints on the produced perturbations~\cite{dia2019semantics}. %
By contrast, we extend the HotFlip method by including an auxiliary Semantic Textual Similarity (STS) objective. We additionally use the extracted universal adversarial triggers to generate adversarial examples with low perplexity.

\subsubsection{Fact Checking}

Fact checking systems consist of components to identify check-worthy claims \cite{atanasova2018overview,hansen2019neural,Wright2020ClaimCD}, retrieve and rank evidence documents \cite{conf/emnlp/0001R18,allein2020timeaware}, determine the relationship between claims and evidence documents \cite{DBLP:conf/emnlp/BowmanAPM15,conf/emnlp/AugensteinRVB16,conf/naacl/BalyMGMMN18}, and finally predict the claims' veracity \cite{DBLP:conf/naacl/ThorneVCM18,DBLP:conf/emnlp/AugensteinLWLHH19}.
As this is a relatively involved task, models easily overfit to shallow textual patterns, necessitating the need for adversarial examples to evaluate the limits of their performance.

\citet{thorne2019evaluating} are the first to propose hand-crafted adversarial attacks. %
They follow up on this with the FEVER 2.0 %
task~\cite{thorne-etal-2019-fever2}, where participants design adversarial attacks for existing FC systems. The first two winning systems~\cite{niewinski-etal-2019-gem, hidey-etal-2020-deseption} produce claims requiring multi-hop reasoning, which has been shown to be challenging for fact checking models \cite{ostrowski2020multihop}. The other remaining system~\cite{kim-allan-2019-fever} generates adversarial attacks manually. We instead find universal adversarial attacks that can be applied to most existing inputs while markedly decreasing fact checking performance.
\citet{niewinski-etal-2019-gem} additionally feed a pre-trained GPT-2 model with the target label of the instance along with the text for conditional adversarial claim generation. Conditional language generation has also been employed by \citet{keskar2019ctrl} to control the style, content, and the task-specific behavior of a Transformer.

\subsection{Methods}

\subsubsection{Models}
We take a RoBERTa~\cite{DBLP:conf/acl/GururanganMSLBD20} model pretrained with a LM objective and fine-tune it to classify claim-evidence pairs from the FEVER dataset as SUPPORTS, REFUTES, and NOT ENOUGH INFO (NEI). The evidence used is the gold evidence, available for the SUPPORTS and REFUTES classes. For NEI claims, we use the system of \citet{malon2018team} to retrieve evidence sentences. 
To measure the semantic similarity between the claim before and after prepending a trigger, we use a large RoBERTa model fine-tuned on the Semantic Textual Similarity Task.\footnote{https://huggingface.co/SparkBeyond/roberta-large-sts-b} For further details, we refer the reader to \S\ref{sec:appendixA}.

\subsubsection{Universal Adversarial Triggers Method}
The Universal Adversarial Triggers method is developed to find n-gram trigger tokens $\mathbf{t_{\alpha}}$, which, appended to the original input $x$,  $f(x) = y$, cause the model to predict a target class $\widetilde{y}$ : $f(t_{\alpha}, x) = \widetilde{y}$. In our work, we generate unigram triggers, as generating longer triggers would require additional objectives to later produce well-formed adversarial claims. We start by initializing the triggers with the token `a'. Then, we update the embeddings of the initial trigger tokens $\mathbf{e}_{\alpha}$ with embeddings $\mathbf{e}_{w_i}$ of candidate adversarial trigger tokens $w_i$ that minimize the loss $\mathcal{L}$ for the target class $\widetilde{y}$. Following the HotFlip algorithm, we reduce the brute-force optimization problem using a first-order Taylor approximation around the initial trigger embeddings:
\begin{equation}
\underset{\mathbf{w}_{i} \in \mathcal{V}}{\arg \min }\left[\mathbf{e}_{w_i}-\mathbf{e}_{\alpha}\right]^{\top} \nabla_{\mathbf{e}_{\alpha}} \mathcal{L}
\end{equation}
where $\mathcal{V}$ is the vocabulary of the RoBERTa model and $\nabla_{\mathbf{e}_{\alpha}} \mathcal{L}$ is the average gradient of the task loss accumulated for all batches. This approximation allows for a $\mathcal{O}(|\mathcal{V}|)$ space complexity of the brute-force candidate trigger search.

While HotFlip %
finds universal adversarial triggers that successfully fool the model for many instances, we find that the most potent triggers are often negation words, e.g., `not', `neither', `nowhere'. Such triggers change the meaning of the text, making the prediction of the target class correct. Ideally, adversarial triggers would preserve the original label of the claim. To this end, we propose to include an auxiliary STS model objective when searching for candidate triggers. The additional objective is used to minimize the loss $\mathcal{L'}$ for the maximum similarity score (5 out of 0) between the original claim and the claim with the prepended trigger. Thus, we arrive at the combined optimization problem:
\begin{equation}
\small
\underset{\mathbf{w}_{i} \in \mathcal{V}}{\arg \min }([\mathbf{e}_{w_i}-\mathbf{e}_{\alpha}]^{\top} \nabla_{\mathbf{e}_{\alpha}} \mathcal{L} + [\mathbf{o}_{w_i}-\mathbf{o}_{\alpha}]^{\top} \nabla_{\mathbf{o}_{\alpha}} \mathcal{L'})
\end{equation}
where $\mathbf{o}_w$ is the STS model embedding of word $w$. For the initial trigger token, we use ``[MASK]'' as STS selects candidates from the neighborhood of the initial token.

\subsubsection{Claim Generation}
\label{sec:claim_generation}
In addition to finding highly potent adversarial triggers, it is also of interest to generate coherent statements containing the triggers. To accomplish this, we use the HuggingFace implementation of the GPT-2 language model~\cite{radford2019language,DBLP:conf/emnlp/WolfDSCDMCRLFDS20}, a large transformer-based language model trained on 40GB of text. 
The objective is to generate a coherent claim, which either entails, refutes, or is unrelated a given piece of evidence, while also including trigger words.

The language model is first fine tuned on the FEVER FC corpus with a specific input format. FEVER consists of claims and evidence with the labels \texttt{SUPPORTS}, \texttt{REFUTES}, or \texttt{NOT ENOUGH INFO} (NEI). We first concatenate evidence and claims with a special token. %
Next, to encourage generation of claims with certain tokens, a sequence of tokens separated by commas is prepended to the input. For training, the sequence consists of a single token randomly selected from the original claim, and four random tokens from the vocabulary. %
This encourages the model to only select the one token most likely to form a coherent and correct claim. The final input format is \texttt{[trigger tokens]}\textbar\textbar\texttt{[evidence]}\textbar\textbar\texttt{[claim]}.
Adversarial claims are then generated by providing an initial input of a series of five comma-separated trigger tokens plus evidence, and progressively generating the rest of the sequence. Subsequently, the set of generated claims is pruned to include only those which contain a trigger token, %
and constitute the desired label. The latter is ensured by passing both evidence and claim through an external NLI model trained on SNLI \cite{DBLP:conf/emnlp/BowmanAPM15}. %

\subsection{Results}
We present results for universal adversarial trigger generation and coherent claim generation. 
Results are measured using the original FC model on claims with added triggers and generated claims (macro F1). We also measure how well the added triggers maintain the claim's original label (semantic similarity score), the perplexity (PPL) of the claims with prepended triggers, and the semantic quality of generated claims (manual annotation). PPL is measured with a pretrained RoBERTa LM.

\subsubsection{Adversarial Triggers}
\begin{table}[t]
\small
\centering
\begin{tabular}{l@{\hspace{1.2\tabcolsep}}l@{\hspace{1.2\tabcolsep}}l@{\hspace{1.2\tabcolsep}}l}
\toprule
\textbf{Class} & \textbf{F1} & \textbf{STS} & \textbf{PPL}\\ \midrule
\multicolumn{4}{c}{\bf No Triggers} \\
All & .866 & 5.139 & 11.92 ($\pm$45.92) \\
S & .938 & 5.130 & 12.22 ($\pm$40.34) \\
R & .846 & 5.139 &  12.14 ($\pm$37.70) \\
NEI & .817 & 5.147 & 14.29 ($\pm$84.45) \\
\midrule
\multicolumn{4}{c}{\bf FC Objective} \\
All & .602 ($\pm$.289) & 4.586 ($\pm$.328) & 12.96 ($\pm$55.37) \\
S$\rightarrow$R & .060 ($\pm$.034) & 4.270 ($\pm$.295) & 12.44 ($\pm$41.74) \\
S$\rightarrow$NEI & .611 ($\pm$.360) & 4.502 ($\pm$.473) & 12.75 ($\pm$40.50) \\
R$\rightarrow$S & .749 ($\pm$.027) & 4.738 ($\pm$.052) & 11.91 ($\pm$36.53) \\
R$\rightarrow$NEI & .715 ($\pm$.026) & 4.795 ($\pm$.094) & 11.77 ($\pm$36.98) \\
NEI$\rightarrow$R & .685 ($\pm$.030) & 4.378 ($\pm$.232) & 14.20 ($\pm$83.32) \\
NEI$\rightarrow$S & .793 ($\pm$.054) & 4.832 ($\pm$.146) & 14.72 ($\pm$93.15) \\
\midrule
\multicolumn{4}{c}{\bf FC+STS Objectives} \\
All & .763 ($\pm$.123) & 4.786 ($\pm$.156) & 12.97 ($\pm$58.30) \\
S$\rightarrow$R & .702 ($\pm$.237) & 4.629 ($\pm$.186) & 12.62 ($\pm$41.91) \\
S$\rightarrow$NEI & .717 ($\pm$.161) & 4.722 ($\pm$.152) & 12.41 ($\pm$39.66) \\
R$\rightarrow$S & .778 ($\pm$.010) & 4.814 ($\pm$.141) & 11.93 ($\pm$37.04) \\
R$\rightarrow$NEI & .779 ($\pm$.009) & 4.855 ($\pm$.098) & 12.20 ($\pm$37.67) \\
NEI$\rightarrow$R & .780 ($\pm$.078) & 4.894 ($\pm$.115) & 15.27 ($\pm$111.2) \\
NEI$\rightarrow$S & .821 ($\pm$.008) & 4.800 ($\pm$.085) & 13.42 ($\pm$82.30) \\
\bottomrule
\end{tabular}
\caption{Universal Adversarial Trigger method performance. Triggers are generated given claims from a source class to fool the classifier to predict a target class (column \textit{Class}, with SUPPORTS (S), REFUTES (R), NEI). %
The results are averaged over the top 10 triggers.}
\label{tab:eval}
\end{table}
Table~\ref{tab:eval} presents the results of applying universal adversarial triggers to claims from the source class.
The top-performing triggers for each direction are found in \S\ref{sec:appendixC}. 
The adversarial method with a single FC objective successfully deteriorates model performance by a margin of 0.264 F1 score overall. The biggest performance decrease is when the adversarial triggers are constructed to flip the predicted class from SUPPORTS to REFUTES. We also find that 8 out of 18 triggers from the top-3 triggers for each direction, are negation words such as  `nothing', `nobody', `neither', `nowhere' (see Table~\ref{tab:evalonetrig} in the appendix). The first of these triggers decreases the performance of the model to 0.014 in F1. While this is a significant performance drop, these triggers also flip the meaning of the text. The latter is again indicated by the decrease of the semantic similarity between the claim before and after prepending a trigger token, which is the largest for the SUPPORTS to REFUTES direction. We hypothesise that the success of the best performing triggers is partly due to the meaning of the text being flipped.

Including the auxiliary STS objective increases the similarity between the claim before and after prepending the trigger for five out of six directions. Moreover, we find that now only one out of the 18 top-3 triggers for each direction are negation words. Intuitively, these adversarial triggers are worse at fooling the FC model as they also have to preserve the label of the original claim. Notably, for the SUPPORTS to REFUTES direction the trigger performance is decreased with a margin of 0.642 compared to the single FC objective.
We conclude that including the STS objective for generating Universal Adversarial triggers helps to preserve semantic similarity with the original claim, but also makes it harder to both find triggers preserving the label of the claim while substantially decreasing the performance of the model.

\begin{table*}[t]%
\fontsize{9}{9}\selectfont
\renewcommand{\arraystretch}{1.4}
\centering
\begin{tabular}{p{6cm} p{4.2cm} p{4.6cm}}
\toprule
\textbf{Evidence} & \textbf{Triggers} & \textbf{Generated Claim}\\ 
\midrule

\multicolumn{3}{c}{\bf SUPPORTS Claims} \\
Since the 19th century, some Romani have also migrated to the Americas. & don,already,more,\textcolor{red}{during},home & Romani have moved to the Americas \textcolor{red}{during} the 19th century. \\

Cyprus is a major tourist destination in the Mediterranean. & foreign,biggest,\textcolor{red}{major},every, friends & Cyprus is a \textcolor{red}{major} tourist destination. \\

The first Nobel Prize in Chemistry was awarded in 1901 to Jacobus Henricus van't Hoff, of the Netherlands, ``for his discovery of the laws of chemical dynamics and osmotic pressure in solutions.'' & later,\textcolor{red}{already},quite,altern, whereas & Henricus Van't Hoff was \textcolor{red}{already} awarded the Nobel Prize. \\
\midrule
\multicolumn{3}{c}{\bf REFUTES Claims} \\

California Attorney General Kamala Harris defeated Sanchez , 61.6\% to 38.4\%. & phys,incarn,\textcolor{red}{not},occasionally,
something & Kamala Harris did \textcolor{red}{not} defeat Sanchez, 61.6\% to 38.4\%. \\

Uganda is in the African Great Lakes region. & unless,endorsed,picks,pref, \textcolor{red}{against} & Uganda is \textcolor{red}{against} the African Great Lakes region. \\

Times Higher Education World University Rankings is an annual publication of university rankings by Times Higher Education (THE) magazine. & interested,reward,visit, \textcolor{red}{consumer},conclusion & Times Higher Education World University Rankings is a \textcolor{red}{consumer} magazine. \\

\midrule
\multicolumn{3}{c}{\bf NOT ENOUGH INFO Claims} \\
The KGB was a military service and was governed by army laws and regulations, similar to the Soviet Army or MVD Internal Troops. & nowhere,\textcolor{red}{only},none,no,nothing & The KGB was \textcolor{red}{only} controlled by a military service. \\

The series revolves around Frank Castle, who uses lethal methods to fight crime as the vigilante ``the Punisher'', with Jon Bernthal reprising the role from Daredevil. & says,said,\textcolor{red}{take},say,is & \textcolor{red}{Take} Me High is about Frank Castle's use of lethal techniques to fight crime. \\

The Suite Life of Zack \& Cody is an American sitcom created by Danny Kallis and Jim Geoghan. & whilst,interest,applic,\textcolor{red}{someone}, nevertheless & The Suite Life of Zack \& Cody was created by \textcolor{red}{someone} who never had the chance to work in television. \\
\bottomrule
\end{tabular}
\caption{Examples of generated adversarial claims. These are all claims which the FC model incorrectly classified.}
\label{tab:generation_examples}
\end{table*}

\subsubsection{Generation}
We use the method described in \S\ref{sec:claim_generation} to generate 156 claims using triggers found with the additional STS objective, and 156 claims without. 52 claims are generated for each class (26 flipping to one class, 26 flipping to the other). A different GPT-2 model is trained to generate claims for each specific class, with triggers specific to attacking that class used as input. The generated claims are annotated manually (see \S\ref{app:B3} for the procedure). The overall average claim quality is 4.48, indicating that most generated statements are highly semantically coherent. The macro F1 of the generative model w.r.t. the intended label is 58.9 overall. For the model without the STS objective, the macro F1 is 56.6, and for the model with the STS objective, it is 60.7, meaning that using triggers found with the STS objective helps the generated claims to retain their intended label.

We measure the performance of the original FC model on generated claims (\autoref{tab:generation_eval}). We compare between using triggers that are generated with the STS objective (Ex2) and without (Ex1). In both cases, the adversarial claims effectively fool the FC model, which performs 38.4\% worse and 23.8\% worse on Ex1 and Ex2, respectively.  Additionally, the overall sentence quality increases when the triggers are found with the STS objective (Ex2). The FC model's performance is higher on claims using triggers generated with the STS objective but still significantly worse than on the original claims. We provide examples of generated claims with their evidence in \autoref{tab:generation_examples}.
\begin{table}%
\fontsize{10}{10}\selectfont
\centering
\begin{tabular}{lccc}
\toprule
\textbf{Target} & \textbf{F1} & \textbf{Avg Quality} & \textbf{\# Examples}\\ \midrule
\multicolumn{4}{c}{\bf FC Objective} \\
Overall& 0.534& 4.33&156\\
SUPPORTS& 0.486& 4.79& 39\\
REFUTES& 0.494& 4.70&32\\
NEI& 0.621& 3.98 &85\\
\midrule
\multicolumn{4}{c}{\bf FC+STS Objectives} \\
Overall& 0.635& 4.63&156\\
SUPPORTS& 0.617& 4.77&67\\
REFUTES& 0.642& 4.68&28\\
NEI& 0.647& 4.44&61\\
\bottomrule
\end{tabular}
\caption{FC performance for generated claims.}
\label{tab:generation_eval}
\end{table}

Comparing FC performance with our generated claims vs. those from the development set of adversarial claims from the FEVER shared task %
, we see similar drops in performance (0.600 and 0.644 macro F1, respectively). While the adversarial triggers from FEVER cause a larger performance drop, they were manually selected to meet the label coherence and grammatical correctness requirements. Conversely, we automatically generate claims that meet these requirements.

\subsection{Conclusion}
We present a method for automatically generating highly potent, well-formed, label cohesive claims for FC. 
We improve upon previous work on universal adversarial triggers by determining how to construct valid claims containing a trigger word. 
Our method is fully automatic, whereas previous work on generating claims for fact checking is generally rule-based or requires manual intervention. As FC is only one test bed for adversarial attacks, it would be interesting to test this method on other NLP tasks requiring semantic understanding such as question answering %
to better understand shortcomings of models. %

\section*{Acknowledgements}
$\begin{array}{l}\includegraphics[width=1cm]{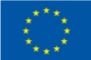} \end{array}$ This project has received funding from the European Union's Horizon 2020 research and innovation programme under the Marie Sk\l{}odowska-Curie grant agreement No 801199.

\newpage

\section{Transformer Based Multi-Source Domain Adaptation}
\label{paper:msda}

\subsection{Introduction}
\begin{figure}[t]
  
  \centering
    \includegraphics[width=0.75\columnwidth]{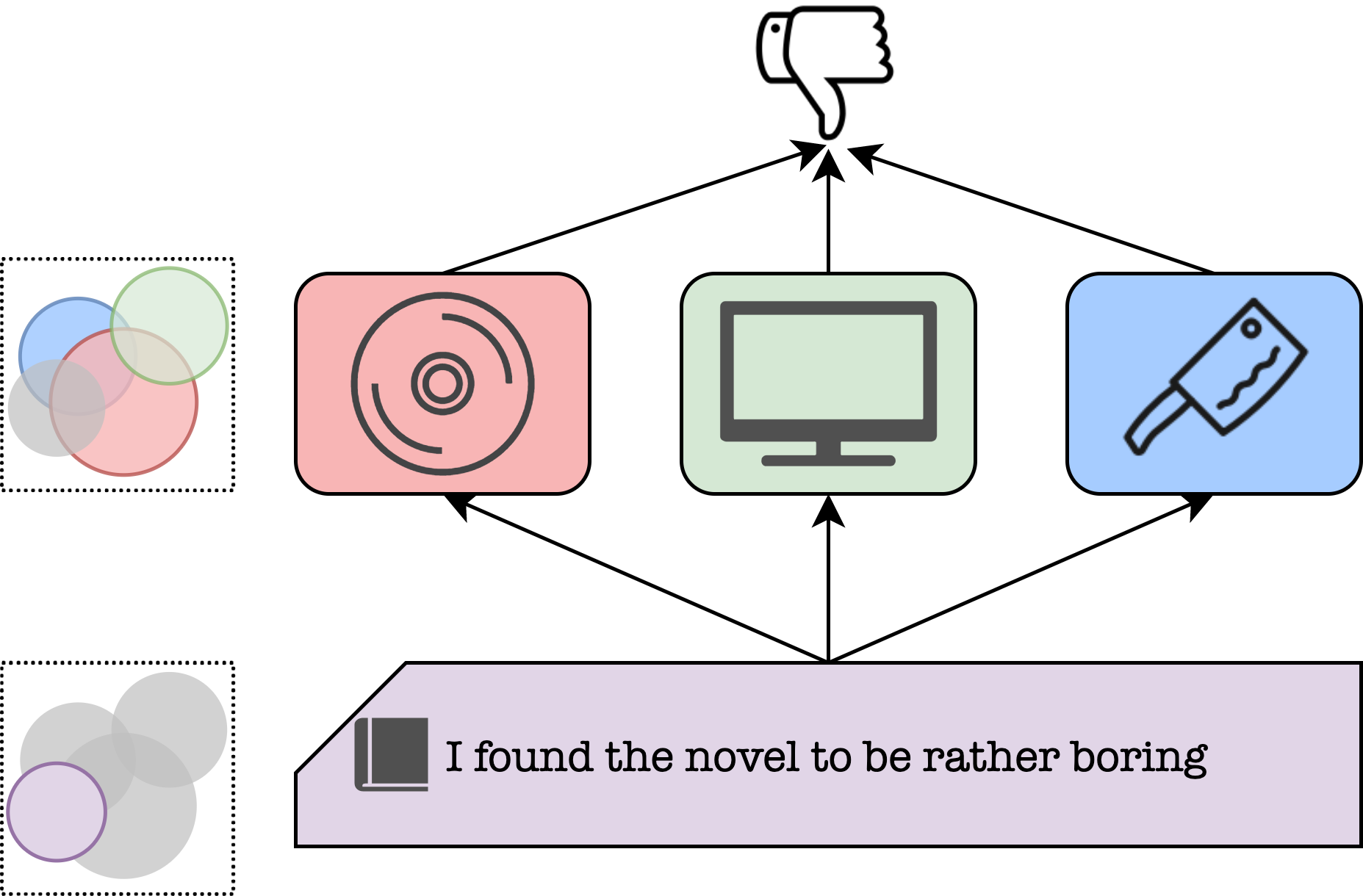}
    \caption{In multi-source domain adaptation, a model is trained on data drawn from multiple parts of the underlying distribution. At test time, the model must make predictions on data from a potentially non-overlapping part of the distribution.}
    \label{fig:msda}
\end{figure}

Machine learning practitioners are often faced with the problem of evolving test data, leading to mismatches in training and test set distributions.
As such, the problem of \textit{domain adaptation} is of particular interest to the natural language processing community in order to build models which are robust this shift in distribution. For example, a model may be trained to predict the sentiment of product reviews for DVDs, electronics, and kitchen goods, and must utilize this learned knowledge to predict the sentiment of a review about a book (\autoref{fig:msda}). This paper is concerned with this setting, namely \textit{unsupervised multi-source domain adaptation}.

Multi-source domain adaptation is a well studied problem in deep learning for natural language processing. Prominent techniques are generally based on data selection strategies and representation learning. For example, a popular representation learning method is to induce domain invariant representations using unsupervised target data and domain adversarial learning~\cite{ganin2015unsupervised}. Adding to this, mixture of experts techniques attempt to learn both domain specific and global shared representations and combine their predictions~\cite{guo2018multi,li2018s,ma2019domain}. These methods have been primarily studied using convolutional nets (CNNs) and recurrent nets (RNNs) trained from scratch, while the NLP community has recently begun to rely more and more on large pretrained transformer (LPX) models e.g. BERT~\cite{devlin2019bert}. %
To date there has been some preliminary investigation of how LPX models perform under domain shift in the single source-single target setting~\cite{ma2019domain,han2019unsupervised,rietzler2019adapt,DBLP:conf/acl/GururanganMSLBD20}. What is lacking is a study into the effects of and best ways to apply classic multi-source domain adaptation techniques with LPX models, which can give insight into possible avenues for improved application of these models in settings where there is domain shift.

Given this, we present a study into unsupervised multi-source domain adaptation techniques for large pretrained transformer models. %
Our main research question is: do mixture of experts and domain adversarial training offer any benefit when using LPX models? The answer to this is not immediately obvious, as such models have been shown to generalize quite well across domains and tasks while still learning representations which are not domain invariant. Therefore, we experiment with four mixture of experts models, including one novel technique based on attending to different domain experts; as well as domain adversarial training with gradient reversal. %
Surprisingly, we find that, while domain adversarial training helps the model learn more domain invariant representations, this does not always result in increased target task performance.
When using mixture of experts, we see significant gains on out of domain rumour detection, and some gains on out of domain sentiment analysis. Further analysis reveals that the classifiers learned by domain expert models are highly homogeneous, making it challenging to learn a better mixing function than simple averaging.

\subsection{Related Work}
Our primary focus is multi-source domain adaptation with LPX models. We first review domain adaptation in general, followed by studies into domain adaptation with LPX models.

\subsubsection{Domain Adaptation}
Domain adaptation approaches generally fall into three categories: \textit{supervised} approaches (e.g. \citet{daumeiii:2007:ACLMain,finkel-manning-2009-hierarchical-fixed,conf/cvpr/KulisSD11}), where both labels for the source and the target domain are available; \textit{semi-supervised} approaches (e.g. \citet{conf/cvpr/DonahueHRSD13,conf/cvpr/YaoPNLM15}), where labels for the source and a small set of labels for the target domain are provided; and lastly \textit{unsupervised} approaches (e.g. \citet{blitzer-etal-2006-domain-fixed,ganin2015unsupervised,conf/aaai/SunFS16,conf/icml/LiptonWS18}), where only labels for the source domain are given. Since the focus of this paper is the latter, we restrict our discussion to unsupervised approaches. A more complete recent review of unsupervised domain adaptation approaches is given in \citet{kouw2019review}.

A popular approach to unsupervised domain adaptation is to induce representations which are invariant to the shift in distribution between source and target data. For deep networks, this can be accomplished via domain adversarial training using a simple gradient reversal trick~\cite{ganin2015unsupervised}. This has been shown to work in the multi-source domain adaptation setting too~\cite{li2018s}. Other popular representation learning methods include minimizing the covariance between source and target features~\cite{conf/aaai/SunFS16} and using maximum-mean discrepancy between the marginal distribution of source and target features as an adversarial objective~\cite{guo2018multi}.

Mixture of experts has also been shown to be effective for multi-source domain adaptation. \citet{kim2017domain} use %
attention to combine the predictions of domain experts. \citet{guo2018multi} propose learning a mixture of experts using a point to set metric, which combines the posteriors of models trained on individual domains. Our work attempts to build on this to study how multi-source domain adaptation can be improved with LPX models.

\subsubsection{Transformer Based Domain Adaptation}
There are a handful of studies which investigate how LPX models can be improved in the presence of domain shift. These methods tend to focus on the data and training objectives for single-source single-target unsupervised domain adaptation. The work of \citet{ma2019domain} shows that curriculum learning based on the similarity of target data to source data improves the performance of BERT on out of domain natural language inference. Additionally, \citet{han2019unsupervised} demonstrate that domain adaptive fine-tuning with the masked language modeling objective of BERT leads to improved performance on domain adaptation for sequence labelling. \citet{rietzler2019adapt} offer similar evidence for task adaptive fine-tuning on aspect based sentiment analysis. \citet{DBLP:conf/acl/GururanganMSLBD20} take this further, showing that significant gains in performance are yielded when progressively fine-tuning on in domain data, followed by task data, using the masked language modeling objective of RobERTa. Finally, \citet{lin2020does} explore whether domain adversarial training with BERT would improve performance for clinical negation detection, finding that the best performing method is a plain BERT model, giving some evidence that perhaps well-studied domain adaptation methods may not be applicable to LPX models.

What has not been studied, to the best of our knowledge, is the impact of domain adversarial training via gradient reversal on LPX models on natural language processing tasks, as well as if mixture of experts techniques can be beneficial. As these methods have historically benefited deep %
models for domain adaptation, we explore their effect when applied to LPX models in this work. %

\subsection{Methods}
\begin{figure}[t]
  
  \centering
    \includegraphics[width=0.5\columnwidth]{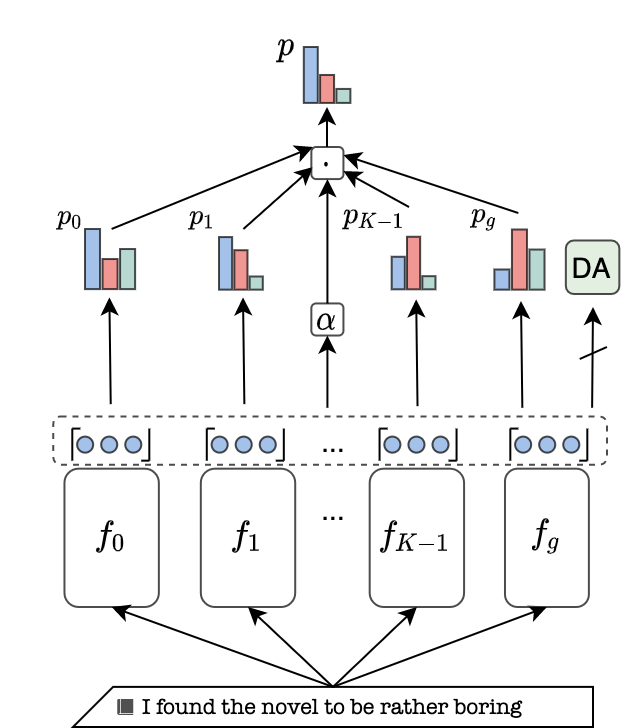}
    \caption{The overall approach tested in this work. A sample is input to a set of expert and one shared LPX model as described in \S\ref{sec:modeling}. The output probabilities of these models are then combined using an attention parameter alpha (\S\ref{sec:avg}, \S\ref{sec:fta}, \S\ref{sec:dc}, \S\ref{sec:attention}). In addition, a global model $f_g$ learns domain invariant representations via a classifier \texttt{DA} with gradient reversal (indicated by the slash, see \S\ref{sec:da_method}).}
    \label{fig:model-architecture}
\end{figure}
This work is motivated by previous research on domain adversarial training and mixture of domain experts for domain adaptation. In this, the data consists of %
$K$ source domains $\mathcal{S}$ and a target domain $\mathcal{T}$. The source domains consist of labelled datasets $D_{s}, s \in \{1,...,K\}$ and the target domain consists only of unlabelled data $U_{t}$. The goal is to learn a classifier $f$, which generalizes well to $\mathcal{T}$ using only the labelled data from $\mathcal{S}$ and optionally unlabelled data from $\mathcal{T}$. We consider a base network $f_{z}, z \in \mathcal{S} \cup \{g\}$ corresponding to either a domain specific network or a global shared network. These $f_{z}$ networks are initialized using LPX models, in particular DistilBert~\cite{sanh2019distilbert}. %

\subsubsection{Mixture of Experts Techniques}\label{sec:modeling}
We study four different mixture of expert techniques: simple averaging, fine-tuned averaging, attention with a domain classifier, and a novel sample-wise attention mechanism based on transformer attention~\cite{vaswani2017attention}. Prior work reports that utilizing mixtures of domain experts and shared classifiers leads to improved performance when having access to multiple source domains~\cite{guo2018multi,li2018s}. Given this, we investigate if mixture of experts can have any benefit when using LPX models. %

Formally, for a setting with $K$ domains, we have set of $K$ different LPX models $f_{k}, k \in \{0...K-1\}$ corresponding to each domain. There is also an additional LPX model $f_{g}$ corresponding to a global shared model. The output predictions of these models are $p_{k}, k \in \{0...K-1\}$ and $p_{g}$, respectively. Since the problems we are concerned with are binary classification, these are single values in the range $(0,1)$. The final output probability is calculated as a weighted combination of a set of domain expert probabilities $\bar{\mathcal{K}} \subseteq \mathcal{S}$ and the probability from the global shared model. Four methods are used for calculating the weighting. 

\subsubsubsection{Averaging}
\label{sec:avg}
The first method is a simple averaging of the predictions of domain specific and shared classifiers. The final output of the model is
\begin{equation}
    p_A(x,\bar{\mathcal{K}}) = \frac{1}{|\bar{\mathcal{K}}|+1}\sum_{k \in \bar{\mathcal{K}}}p_{k}(x) + p_{g}(x)
\end{equation}

\subsubsubsection{Fine Tuned Averaging}
\label{sec:fta}
As an extension to simple averaging, we fine tune the weight given to each of the domain experts and global shared model. This is performed via randomized grid search evaluated on validation data, after the models have been trained. A random integer between zero and ten is generated for each of the models, which is then normalized to a set of probabilities $\alpha_{F}$. The final output probability is then given as follows.

\begin{equation}
    p_F(x) = \sum_{k \in \bar{\mathcal{K}}}p_k(x) * \alpha_{F}^{(k)}(x) + p_g(x) * \alpha_{F}^{(g)}(x)
\end{equation}

\subsubsubsection{Domain Classifier}
\label{sec:dc}
It was recently shown that curriculum learning using a domain classifier can lead to improved performance for single-source domain adaptation~\cite{ma2019domain} when using LPX models. Inspired by this, we experiment with using a domain classifier as a way to attend to the predictions of domain expert models. First, a domain classifier $f_C$ is trained to predict the domain of an input sample $x$ given $\mathbf{r}_{g} \in \mathbb{R}^{d}$, the representation of the \texttt{[CLS]} token at the output of a LPX model. From the classifier, a vector $\alpha_{C}$ is produced with the probabilities that a sample belongs to each source domain. 
\begin{equation}
    \alpha_{C} = f_{C}(x) = \text{softmax}(\mathbf{W}_C\mathbf{r}_{g} + b_{C})
\end{equation}
where $\mathbf{W}_{C} \in \mathbb{R}^{d \times K}$ and $b_{C} \in \mathbb{R}^{K}$. The domain classifier is trained before the end-task network and is held static throughout training on the end-task. For this, a set of domain experts $f_{k}$ are trained and their predictions combined through a weighted sum of the attention vector $\alpha_{C}$.
\begin{equation}
    p_C(x) = \sum_{k \in S}p_k(x) * \alpha_C^{(k)}(x)
\end{equation}
where the superscript $(k)$ indexes into the $\alpha_C$ vector. Note that in this case we only use domain experts and not a global shared model. In addition, the probability is always calculated with respect to each source domain.

\subsubsubsection{Attention Model}
\label{sec:attention}
Finally, a novel parameterized attention model is learned which attends to different domains based on the input sample. The attention method is based on the scaled dot product attention applied in transformer models~\cite{vaswani2017attention}, where a global shared model acts as a query network attending to each of the expert and shared models. As such, a shared model $f_{g}$ produces a vector $\mathbf{r}_{g} \in \mathbb{R}^{d}$, and each domain expert produces a vector $\mathbf{r}_{k} \in \mathbb{R}^{d}$. First, for an input sample $x$, a probability for the end task is obtained from the classifier of each model yielding probabilities $p_{g}$ and $p_{k}, k \in {0...K-1}$.
An attention vector $\alpha_{X}$ is then obtained via the following transformations.
\begin{equation}
    \mathbf{q} = \mathbf{g}\mathbf{Q}^{T}
\end{equation}
\begin{equation}
    \mathbf{k} = \begin{bmatrix}
           \mathbf{r}_{1} \\
           \vdots \\
           \mathbf{r}_{K} \\
           \mathbf{r}_{g}
         \end{bmatrix} \mathbf{K}^{T}
\end{equation}
\begin{equation}
    \alpha_{X} = \text{softmax}(\mathbf{q}\mathbf{k}^{T})
\end{equation}
where $\mathbf{Q} \in \mathbb{R}^{d \times d}$ and $\mathbf{K} \in \mathbb{R}^{d \times d}$. The attention vector $\alpha_{X}$ then attends to the individual predictions of each domain expert and the global shared model. 
\begin{equation}
    p_X(x,\bar{\mathcal{K}}) = \sum_{k \in \bar{\mathcal{K}}}p_k(x) * \alpha_{X}^{(k)}(x) + p_g(x) * \alpha_{X}^{(g)}(x)
\end{equation}

To ensure that each model is trained as a domain specific expert, a similar training procedure to that of \citealt{guo2018multi} is utilized, described in \S\ref{sec:training}.

\subsubsection{Domain Adversarial Training}
\label{sec:da_method}
The method of domain adversarial adaptation we investigate here is the well-studied technique described in~\citet{ganin2015unsupervised}. It has been shown to benefit both convolutional nets and recurrent nets on NLP problems~\cite{li2018s,gui2017part}, so is a prime candidate to study in the context of LPX models. Additionally, some preliminary evidence indicates that adversarial training might improve LPX generalizability for single-source domain adaptation~\cite{ma2019domain}. 

To learn domain invariant representations, we train a model such that the learned representations maximally confuse a domain classifier $f_d$. This is accomplished through a min-max objective between the domain classifier parameters $\theta_{D}$ and the parameters $\theta_{G}$ of an encoder $f_g$. The objective can then be described as follows.
\begin{equation}
    \mathcal{L}_D = \max_{\theta_{D}}\min_{\theta_{G}}-d\log  f_{d}(f_{g}(x))
\end{equation}
where $d$ is the domain of input sample $x$. The effect of this is to improve the ability of the classifier to determine the domain of an instance, while encouraging the model to generate maximally confusing representations via minimizing the negative loss. In practice, this is accomplished by training the model using standard cross entropy loss, but reversing the gradients of the loss with respect to the model parameters $\theta_{G}$.

\subsubsection{Training}
\label{sec:training}
Our training procedure follows a multi-task learning setup in which the data from a single batch comes from a single domain. Domains are thus shuffled on each round of training and the model is optimized for a particular domain on each batch. 

For the attention based (\S\ref{sec:attention}) and averaging (\S\ref{sec:avg}) models we adopt a similar training algorithm to~\citet{guo2018multi}. For each batch of training, a meta-target $t$ is selected from among the source domains, with the rest of the domains treated as meta-sources $\mathcal{S}' \in \mathcal{S} \setminus \{t\}$. Two losses are then calculated. The first is with respect to all of the meta-sources, where the attention vector is calculated for only those domains. For target labels $y_{i}$ and a batch of size $N$ with samples from a single domain, this is given as follows.
\begin{equation}
    \mathcal{L}_{s} = -\frac{1}{N}\sum_{i}y_{i}\log p_{X}(x, \mathcal{S}')
\end{equation}
The same procedure is followed for the averaging model $p_{A}$. The purpose is to encourage the model to learn attention vectors for out of domain data, thus why the meta-target is excluded from the calculation. 

The second loss is with respect to the meta-target, where the cross-entropy loss is calculated directly for the domain expert network of the meta-target.
\begin{equation}
    \mathcal{L}_{t} = -\frac{1}{N}\sum_{i}y_{i}\log p_t(x)
\end{equation}
This allows each model to become a domain expert through strong supervision. The final loss of the network is a combination of the three losses described previously, with $\lambda$ and $\gamma$  hyperparameters controlling the weight of each loss.
\begin{equation}
    \mathcal{L} = \lambda \mathcal{L}_s + (1 - \lambda) \mathcal{L}_t + \gamma \mathcal{L}_D
\end{equation}

For the domain classifier (\S\ref{sec:dc}) and fine-tuned averaging (\S\ref{sec:fta}), the individual LPX models are optimized directly with no auxiliary mixture of experts objective. In addition, we experiment with training the simple averaging model directly.

\subsection{Experiments and Results}

We focus our experiments on text classification problems with data from multiple domains. To this end, we experiment with sentiment analysis from Amazon product reviews and rumour detection from tweets. For both tasks, we perform cross-validation on each domain, holding out a single domain for testing and training on the remaining domains, allowing a comparison of each method on how well they perform under domain shift. The code to reproduce all of the experiments in this paper can be found here\footnote{\url{https://github.com/copenlu/xformer-multi-source-domain-adaptation}}.

\paragraph{Sentiment Analysis Data} The data used for sentiment analysis come from the legacy Amazon Product Review dataset~\cite{blitzer2007biographies}. This dataset consists of 8,000 total tweets from four product categories: books, DVDs, electronics, and kitchen and housewares. Each domain contains 1,000 positive and 1,000 negative reviews. In addition, each domain has associated unlabelled data. Following previous work we focus on the transductive setting~\cite{guo2018multi,ziser2017neural} where we use the same 2,000 out of domain tweets as unlabelled data for training the domain adversarial models.  This data has been well studied in the context of domain adaptation, making for easy comparison with previous work.

\paragraph{Rumour Detection Data} The data used for rumour detection come from the PHEME dataset of rumourous tweets~\cite{zubiaga2016analysing}. There are a total of 5,802 annotated tweets from 5 different events labelled as rumourous or non-rumourous (1,972 rumours, 3,830 non-rumours). Methods which have been shown to work well on this data include context-aware classifiers \cite{zubiaga2017exploiting} and positive-unlabelled learning \cite{Wright2020ClaimCD}. Again, we use this data in the transductive setting when testing domain adversarial training.

\subsubsection{Baselines}

\paragraph{What's in a Domain?}
We use the model from~\citet{li2018s} as a baseline for sentiment analysis. This model consists of a set of domain experts and one general CNN, and is trained with a domain adversarial auxiliary objective.

\paragraph{Mixture of Experts}
Additionally, we present the results from~\citet{guo2018multi} representing the most recent state of the art on the Amazon reviews dataset. Their method consists of domain expert classifiers trained on top of a shared encoder, with predictions being combined via a novel metric which considers the distance between the mean representations of target data and source data. %

\begin{table*}[t!]
    \centering
    \fontsize{10}{10}\selectfont
    \begin{tabular}{l c c c c c | c c c c c c c }
    \toprule %
     Method &\multicolumn{5}{c}{Sentiment Analysis (Accuracy)}&\multicolumn{6}{c}{Rumour Detection (F1)}\\
    \cmidrule(lr){2-6}
    \cmidrule(lr){7-12}
     & D & E & K & B & macroA & CH & F & GW & OS & S & $\mu$F1\\
    \midrule %
         \citealt{li2018s} &77.9 &80.9 &80.9 &77.1 & 79.2 & - & - & - & - & - & -\\
      \citealt{guo2018multi} &87.7 &89.5 & 90.5& 87.9 &88.9 & - & - & - & - & - & -\\
      \citealt{zubiaga2017exploiting}&- &- &- &- & - & 63.6& \textbf{46.5}& 70.4& 69.0& 61.2& 60.7\\
    \midrule
        Basic & 89.1& 89.8& 90.1& 89.3& 89.5& 66.1& 44.7& 71.9& 61.0& 63.3& 62.3\\
        \midrule
        Adv-6 & 88.3& 89.7& 90.0& 89.0& 89.3 & 65.8& 42.0& 66.6& 61.7& 63.2& 61.4\\
        Adv-3 & 89.0& 89.9& 90.3& 89.0& 89.6& 65.7& 43.2& 72.3& 60.4& 62.1& 61.7\\
        \midrule
        Independent-Avg & 88.9& \textbf{90.6}& 90.4& \textbf{90.0}& \textbf{90.0}& 66.1& 45.6& 71.7& 59.4& 63.5& 62.2\\
        Independent-Ft & 88.9& 90.3& \textbf{90.8}& \textbf{90.0}& \textbf{90.0}& 65.9& 45.7& 72.2& 59.3& 62.4& 61.9\\
        MoE-Avg & \textbf{89.3}& 89.9& 90.5& 89.9& 89.9& \textbf{67.9}& 45.4& \textbf{74.5}& 62.6& \textbf{64.7}& \textbf{64.1}\\
        MoE-Att & 88.6& 90.0& 90.4& 89.6& 89.6& 65.9& 42.3& 72.5& 61.2& 63.3& 62.2\\
        MoE-Att-Adv-6 & 87.8& 89.0 & 90.5& 88.3& 88.9& 66.0& 40.7& 69.0& 63.8& 63.7& 61.8\\
        MoE-Att-Adv-3 & 88.6& 89.1& 90.4& 88.9& 89.2& 65.6& 42.7& 73.4& 60.9& 61.0& 61.8\\
        MoE-DC & 87.8& 89.2& 90.2& 87.9& 88.8& 66.5& 40.6& 70.5& \textbf{70.8}& 62.8& 63.8\\
    \bottomrule %

    \end{tabular}
    \caption{Experiments for sentiment analysis in (D)VD, (E)lectronics, (K)itchen and housewares, and (B)ooks domains and rumour detection for different events ((C)harlie(H)ebdo, (F)erguson, (G)erman(W)ings, (O)ttawa(S)hooting, and (S)ydneySiege) using leave-one-out cross validation. Results are averaged across 5 random seeds. The results for sentiments analysis are in terms of accuracy and the results for rumour detection are in terms of F1.}
    \label{tab:sentiment_results}
\end{table*}

\paragraph{\citealt{zubiaga2017exploiting}} Though not a domain adaptation technique, we include the results from \citealt{zubiaga2017exploiting} on rumour detection to show the current state of the art performance on this task. The model is a CRF, which utilizes a combination of content and social features acting on a timeline of tweets.

\subsubsection{Model Variants}
A variety of models are tested in this work. Therefore, each model is referred to by the following.

\paragraph{Basic} Basic DistilBert with a single classification layer at the output. 

\paragraph{Adv-$X$} DistilBert with domain adversarial supervision applied at the $X$'th layer (\S\ref{sec:da_method}).

\paragraph{Independent-Avg} DistilBert mixture of experts averaged but trained individually (not with the algorithm described in \S\ref{sec:training}).

\paragraph{Independent-FT} DistilBert mixture of experts averaged with mixing attention fine tuned after training (\S\ref{sec:fta}), trained individually.

\paragraph{MoE-Avg} DistilBert mixture of experts using averaging (\S\ref{sec:avg}).

\paragraph{MoE-Att} DistilBert mixture of experts using our novel attention based technique (\S\ref{sec:attention}).

\paragraph{MoE-Att-Adv-$X$} DistilBert mixture of experts using attention and domain adversarial supervision applied at the $X$'th layer.

\paragraph{MoE-DC} DistilBert mixture of experts using a domain classifier for attention (\S\ref{sec:dc}).

\subsubsection{Results}

Our results are given in \autoref{tab:sentiment_results}. Similar to the findings of \citet{lin2020does} on clinical negation, we see little overall difference in performance from both the individual model and the mixture of experts model when using domain adversarial training on sentiment analysis. For the base model, there is a slight improvement when domain adversarial supervision is applied at a lower layer of the model, but a drop when applied at a higher level. Additionally, mixture of experts provides some benefit, especially using the simpler methods such as averaging.

For rumour detection, again we see little performance change from using domain adversarial training, with a slight drop when supervision is applied at either layer. The mixture of experts methods overall perform better than single model methods, suggesting that mixing domain experts is still effective when using large pretrained transformer models. In this case, the best mixture of experts methods are simple averaging and static grid search for mixing weights, indicating the difficulty in learning an effective way to mix the predictions of domain experts. We elaborate on our findings further in \S\ref{sec:discussion}. Additional experiments on domain adversarial training using Bert can be found in \autoref{tab:bert_appendix_results} in \S\ref{sec:bert_appendix}, where we similarly find that domain adversarial training leads to a drop in performance on both datasets.

\begin{figure*}[t]
  
  \centering
    \includegraphics[width=0.85\textwidth]{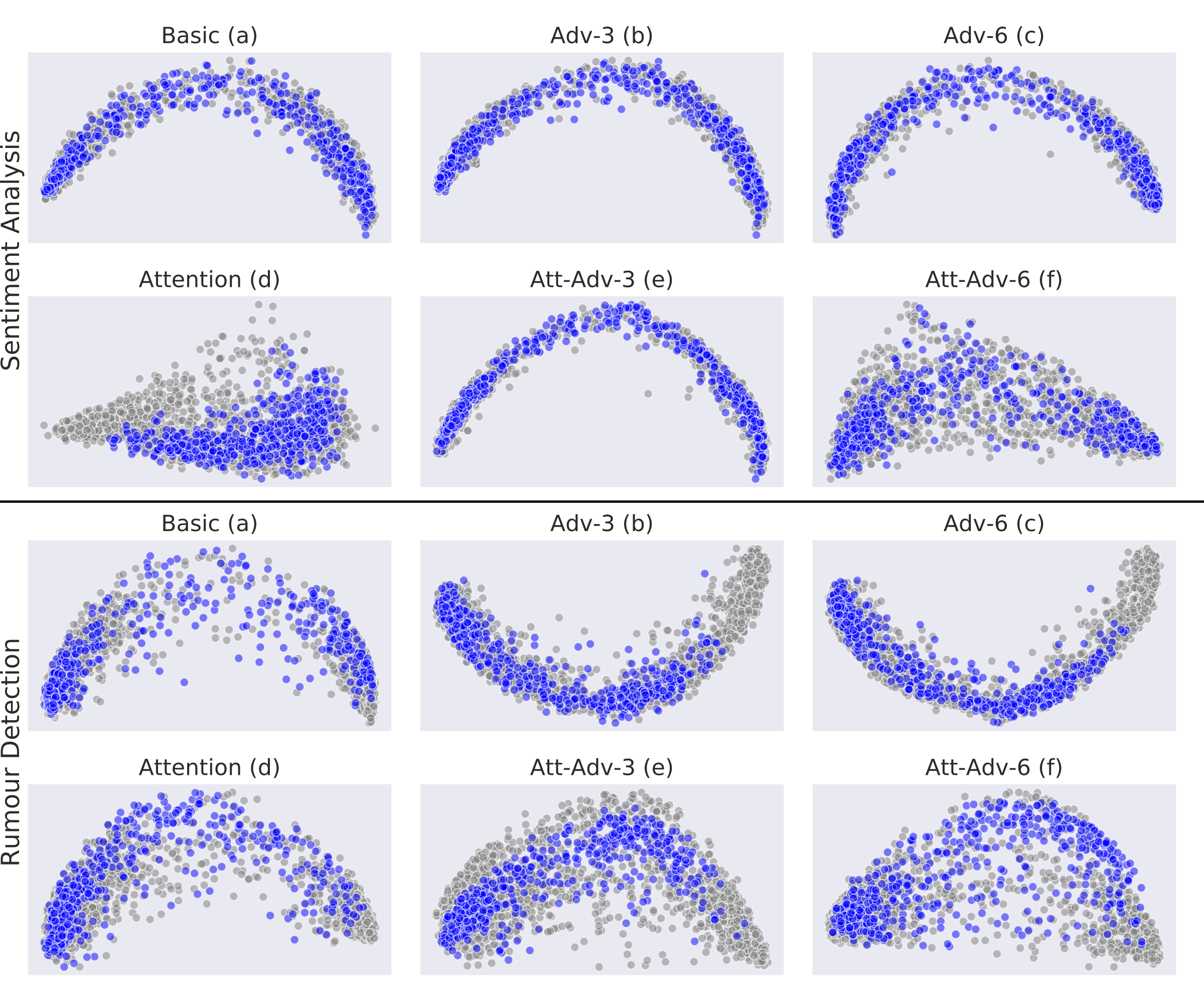}
    \caption{Final layer DistilBert embeddings for 500 randomly selected examples from each split for each tested model for sentiment data (top two rows) and rumour detection (bottom two rows). The blue points are out of domain data (in this case from Kitchen and Housewares for sentiment analysis and Sydney Siege for rumour detection) and the gray points are in domain data. }
    \label{fig:all_reps}
\end{figure*}

\subsection{Discussion}
\label{sec:discussion}
We now discuss our initial research questions in light of the results we obtained, and provide explanations for the observed behavior.
  
\subsubsection{What is the Effect of Domain Adversarial Training?}
We present PCA plots of the representations learned by different models in \autoref{fig:all_reps}. These are the final layer representations of 500 randomly sampled points for each split of the data. In the ideal case, the representations for out of domain samples would be indistinguishable from the representations for in domain data.

In the case of basic DistilBert, we see a slight change in the learned representations of the domain adversarial models versus the basic model (\autoref{fig:all_reps} top half, a-c) for sentiment analysis. When the attention based mixture of experts model is used, the representations of out of domain data cluster in one region of the representation space (d). With the application of adversarial supervision, the model learns representations which are more domain agnostic. Supervision applied at layer 6 of DistilBert (plot f) yields a representation space similar to the version without domain adversarial supervision. Interestingly, the representation space of the model with supervision at layer 3 (plot e) yields representations similar to the basic classifier. This gives some potential explanation as to the similar performance of this model to the basic classifier on this split (kitchen and housewares). Overall, domain adversarial supervision has some effect on performance, leading to gains in both the basic classifier and the mixture of experts model for this split. Additionally, there are minor improvements overall for the basic case, and a minor drop in performance with the mixture of experts model.

The effect of domain adversarial training is more pronounced on the rumour detection data for the basic model (\autoref{fig:all_reps} bottom half, a), where the representations exhibit somewhat less variance when domain adversarial supervision is applied. Surprisingly, this leads to a slight drop in performance for the split of the data depicted here (Sydney Siege). For the attention based model, the variant without domain adversarial supervision (d) already learns a somewhat domain agnostic representation. The model with domain adversarial supervision at layer 6 (f) furthers this, and the classifier learned from these representations perform better on this split of the data. Ultimately, the best performing models for rumour detection do not use domain supervision, and the effect on performance on the individual splits are mixed, suggesting that domain adversarial supervision can potentially help, but not in all cases.

\subsubsection{Is Mixture of Experts Useful with LPX Models?}
\label{sec:moe_discussion}
\begin{figure}
  
  \centering
    \includegraphics[width=0.6\columnwidth]{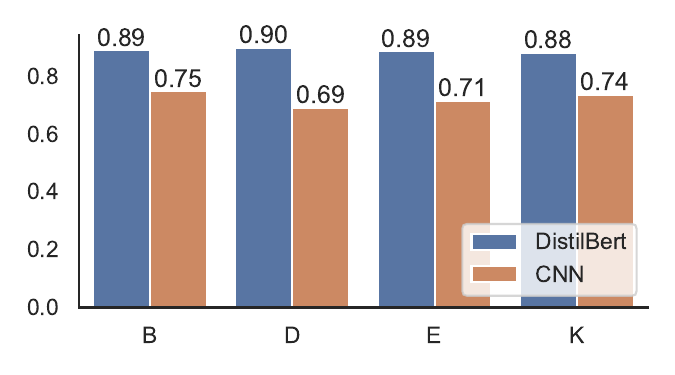}
    \caption{Comparison of agreement (Krippendorff's alpha) between domain expert models when the models are either DistilBert or a CNN. Predictions are made on unseen test data by each domain expert, and agreement is measured between their predictions ((B)ooks, (D)VD, (E)lectronics, and (K)itchen). The overall agreement between the DistilBert experts is greater than the CNNs, suggesting that the learned classifiers are much more homogenous.}
    \label{fig:krippendorff}
\end{figure}
We performed experiments with several variants of mixture of experts, finding that overall, it can help, but determining the optimal way to mix LPX domain experts remains challenging. Simple averaging of domain experts (\S\ref{sec:avg}) gives better performance on both sentiment analysis and rumour detection over the single model baseline. Learned attention (\S\ref{sec:attention}) has a net positive effect on performance for sentiment analysis and a negative effect for rumour detection compared to the single model baseline. Additionally, simple averaging of domain experts consistently outperforms a learned sample by sample attention. This highlights the difficulty in utilizing large pretrained transformer models to learn to attend to the predictions of domain experts.

\paragraph{Comparing agreement} To provide some potential explanation for why it is difficult to learn to attend to domain experts, we compare the agreement on the predictions of domain experts of one of our models based on DistilBert, versus a model based on CNNs (\autoref{fig:krippendorff}). CNN models are chosen in order to compare the agreement using our approach with an approach which has been shown to work well with mixture of experts on this data~\cite{guo2018multi}. Each CNN consists of an embedding layer initialized with 300 dimensional FastText embeddings~\cite{bojanowski2017enriching}, a series of 100 dimensional convolutional layers with widths 2, 4, and 5, and a classifier. The end performance is on par with previous work using CNNs~\cite{li2018s} (78.8 macro averaged accuracy, validation accuracies of the individual models are between 80.0 and 87.0). Agreement is measured using Krippendorff's alpha~\cite{krippendorff2011computing} between the predictions of domain experts on test data. 

We observe that the agreement between DistilBert domain experts on test data is significantly higher than that of CNN domain experts, indicating that the learned classifiers of each expert are much more similar in the case of DistilBert. Therefore, it will potentially be more difficult for a mixing function on top of DistilBert domain experts to gain much beyond simple averaging, while with CNN domain experts, there is more to be gained from mixing their predictions. This effect may arise because each DistilBert model is highly pre-trained already, hence there is little change in the final representations, and therefore similar classifiers are learned between each domain expert.

\subsection{Conclusion}
In this work, we investigated the problem of multi-source domain adaptation with large pretrained transformer models. Both domain adversarial training and mixture of experts techniques were explored. While domain adversarial training could effectively induce more domain agnostic representations, it had a mixed effect on model performance. Additionally, we demonstrated that while techniques for mixing domain experts can lead to improved performance for both sentiment analysis and rumour detection, determining a beneficial mixing of such experts is challenging. The best method we tested was a simple averaging of the domain experts, and we provided some evidence as to why this effect was observed. We find that LPX models may be better suited for data-driven techniques such as that of~\citet{DBLP:conf/acl/GururanganMSLBD20}, which focus on inducing a better prior into the model through pretraining, as opposed to techniques which focus on learning a better posterior with architectural enhancements. We hope that this work can help inform researchers of considerations to make when using LPX models in the presence of domain shift.

\section*{Acknowledgements}
$\begin{array}{l}\includegraphics[width=1cm]{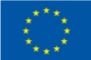} \end{array}$ This project has received funding from the European Union's Horizon 2020 research and innovation programme under the Marie Sk\l{}odowska-Curie grant agreement No 801199.

\newpage

\section{Multi-View Knowledge Distillation from Crowd Annotations for Out-of-Domain Generalization}
\label{paper:aggregation}

\subsection{Introduction}
One of the primary concerns in supervised machine learning is how to define, collect, and use labels as training data for a given task. There are a multitude of tradeoffs associated with this decision, including the cost, the number of labels to collect, the time to collect those labels, the accuracy of those labels with respect to the task under consideration, and how well those labels enable model generalization. These tradeoffs are made based on how the labels are collected (e.g. crowdsourcing, expert labeling, distant supervision) and how they are trained on in practice, for example as one-hot categorical labels (hard labeling) or as a distribution over possible classes (soft labeling).

A large body of literature exists which examines all facets of this question~\cite{DBLP:journals/jair/UmaFHPPP21}. Recent work has focused on utilizing soft-labeling schemes for classification tasks as a method for improving both model accuracy and uncertainty estimation~\cite{DBLP:conf/iccv/PetersonBGR19,DBLP:conf/hcomp/UmaFHPPP20,DBLP:conf/naacl/FornaciariUPPHP21}. When using soft-labels, models are trained to minimize the divergence between their predictive distribution and a distribution over the labels obtained from crowd annotations~\cite{DBLP:conf/hcomp/UmaFHPPP20}. While this has been shown to potentially improve model generalization for vision tasks~\cite{DBLP:conf/iccv/PetersonBGR19}, little work has systematically compared how different soft-labeling schemes affect out-of-distribution performance and uncertainty estimation in NLP. We seek to fill this gap in this work, providing an in-depth study into soft-labeling techniques and best practices for improving model generalization and uncertainty estimation across eight methods, 4 NLP tasks, and and 7 datasets.

Soft-labeling methods have been compared in both \citet{DBLP:conf/naacl/FornaciariUPPHP21} and \citet{DBLP:journals/jair/UmaFHPPP21} for an in-domain testing setting. These studies are primarily focused on identifying the best methods for learning from these soft distributions within a particular domain without going in great depth about which methods for obtaining soft-labels lead to best performance. As such, no clear best method emerges when comparing across soft-labeling approaches in the in-domain setting, making it difficult to decide which aggregation technique to use for a given task. Additionally, these studies do not examine the out of domain test setting, where the benefits of soft-labeling have been made clear in the computer vision literature~\cite{DBLP:conf/iccv/PetersonBGR19}. Here, we demonstrate that soft-labels which are aggregated across multiple soft-labeling techniques mitigate changes in performance for these different techniques across different tasks for out of domain data.
To do this, we propose four multi-view aggregation methods to generate aggregated soft-labels, including three novel methods based on the Jensen-Shannon centroid and temperature scaling. 

In sum, this work makes the following contributions:
\begin{enumerate}[noitemsep]
    \item A comprehensive comparison of soft-labeling techniques for learning from crowd annotations for 4 NLP tasks across 7 datasets in the out-of-domain test setting, including text classification (recognizing textual entailment, medical relation extraction, and toxicity detection) and sequence tagging (part-of-speech tagging);
    \item Novel methods for aggregating different views of soft-labels derived from crowd-annotations;
    \item Insights and suggestions into best practices and tradeoffs for different soft-labeling methods in terms of performance and uncertainty estimation.
\end{enumerate}

\subsection{Related Work}

\paragraph{Learning from Crowd-Sourced Labels}
An efficient way to collect training data for a new task is to ask crowd annotators on platforms such as Amazon Mechanical Turk to manually annotate training data. How to select an appropriate training signal from these noisy crowd labels has a rich set of literature (e.g. see the survey from \citealt{DBLP:journals/tacl/PaunCCHKP18}). There exist a multitude of studies on selecting the best label from a set of crowd annotations, many focusing on Bayesian methods to learn a latent distribution over the true class for each sample influenced by factors such as annotator behavior~\cite{DBLP:conf/naacl/HovyBVH13,dawid1979maximum} and item difficulty~\cite{carpenter2008multilevel}. While selecting a single true label to use as training is the dominant paradigm in machine learning, this discards potentially useful information regarding uncertainty over classes inherent in many tasks, for example where items can be especially difficult or ambiguous~\cite{DBLP:conf/chi/GordonZPHB21}. Given this, recent work has looked into how to learn directly from crowd-annotations~\cite{DBLP:journals/jair/UmaFHPPP21}. The work of \citet{DBLP:conf/iccv/PetersonBGR19} was one of the first, which demonstrated that learning directly from crowd annotations treated as soft-labels using the softmax function leads to better out of distribution performance in computer vision. This line of work has been followed by~\citet{DBLP:conf/hcomp/UmaFHPPP20} and \citet{DBLP:conf/naacl/FornaciariUPPHP21} in NLP, looking at the use of the KL divergence as an effective loss on the soft labels. The survey of \citet{DBLP:journals/jair/UmaFHPPP21} provides an extensive set of experiments on different methods for learning from crowd labels on a vast array of datasets. What has not been done is a systematic comparison of different soft-labeling methods in the out of domain setting, where we reveal that the selection of the best method is not obvious. We fill this gap in this work, and propose new methods for aggregating soft-labels which yield more consistent and robust performance without requiring new annotations or learning methods. %

\paragraph{Knowledge Distillation} Knowledge distillation seeks to build compact but robust models by training them on the probability distribution learned by a much larger teacher network~\cite{DBLP:conf/nips/BaC14,DBLP:journals/corr/HintonVD15}. The goal is to impart the ``dark knowledge'' contained in the distribution learned by the larger network, which can indicate the degree to which a particular example resembles each class if the soft-targets from the classifier are well calibrated. Oftentimes the distribution of the larger network will be smoothed via some temperature in order to accomplish this~\cite{DBLP:journals/corr/HintonVD15}, as has also been done in learning from crowd labels~\cite{DBLP:journals/frai/UmaAP22}, or ensembled with multiple large classifiers~\cite{DBLP:journals/corr/HintonVD15,DBLP:journals/corr/abs-2012-09816}. \citet{DBLP:journals/corr/abs-2012-09816} demonstrate that the data used to train an ensemble should constitute a multi-view structure (i.e. multiple different features in the data are predictive of a particular class) in order for ensembling to improve the test set performance of the final distilled model. Inspired by this, we develop several methods for aggregating multiple views of crowd-sourced labels in order to obtain a distribution that can induce robust classifiers in the out-of-domain setting. As such, ``multi-view'' in this work is defined as multiple posterior distributions over true classes from crowd annotations that are explained by different factors e.g. annotator behavior or raw number of votes per class.

\subsection{Methods}
\begin{figure*}[t]
  \centering
    \includegraphics[width=0.9\linewidth]{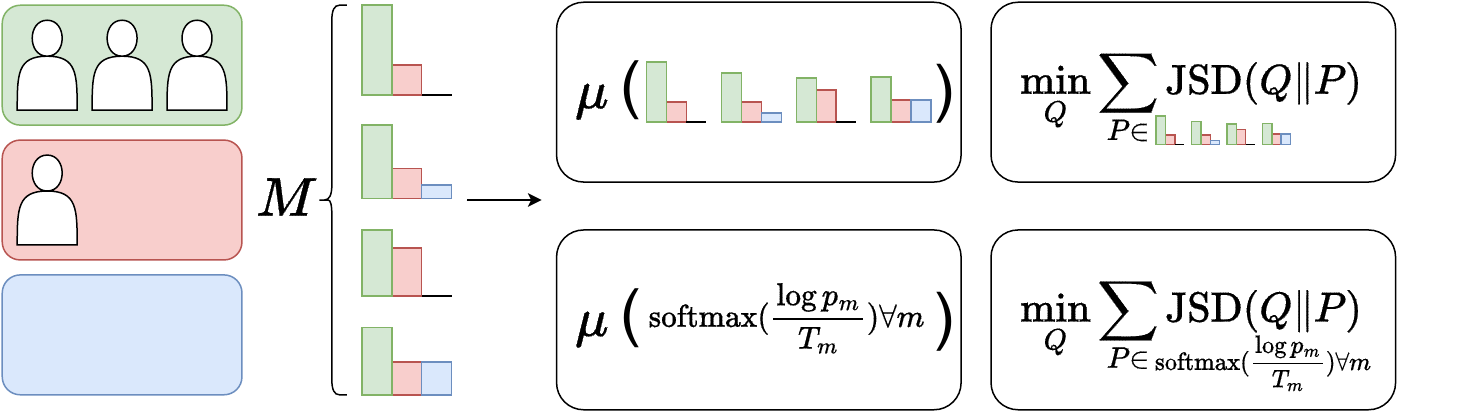}
    \caption{We experiment with four methods for aggregating soft labels in this work: Averaging, the Jensen-Shannon centroid, temperature scaling, and a hybrid approach. } 
    \label{fig:claim_generation}
\end{figure*}
We build upon a rich literature around the topic of learning from crowd annotations. As opposed to hard labels, soft-labeling schemes give us a distribution over the possible classes in a given dataset. In this, more difficult or ambiguous classes can have their probability mass distributed over multiple classes, which can help regularize a downstream classifier. This may also reflect potential ``dark knowledge''~\cite{DBLP:journals/corr/HintonVD15}, or the relative similarity of an input sample to different classes, learnable from the crowd annotations. 
Multiple soft-labeling schemes have been demonstrated to provide good training signals on different NLP tasks, but none of these methods are consistently best across tasks~\cite{DBLP:journals/jair/UmaFHPPP21}. Given this, we start with several well-studied methods for learning from crowd-labels, described in \autoref{sec:soft_labels}~\cite{DBLP:conf/hcomp/UmaFHPPP20,DBLP:conf/naacl/FornaciariUPPHP21,DBLP:conf/naacl/HovyBVH13,dawid1979maximum}, adding to this literature by providing a systematic analysis of their performance and best practices when considering generalization to out of domain data, which is under-explored in NLP. Then, we propose several methods for aggregating the distributions over class labels produced by each of these methods in \autoref{sec:soft_label_methods}, which we will demonstrate produce robust soft-labels across tasks.

\subsubsection{Soft Labeling Methods}
\label{sec:soft_labels}
We experiment with two widely used normalization schemes for obtaining soft labels, as well as two methods based on Bayesian models of annotation. 

\paragraph{Standard Normalization} The standard normalization scheme presented in~\cite{DBLP:conf/hcomp/UmaFHPPP20} obtains soft-labels for a given sample by transforming a set of crowd-sourced labels into a probability distribution. This is done by normalizing the number of votes given to each label by the total number of annotations for a given sample, as described in \autoref{eq:standard_norm}.
\begin{equation}
\label{eq:standard_norm}
    p_{stand}(i,y) = \frac{c_{i,y}}{\sum_{\hat{y}}c_{i,\hat{y}}}
\end{equation}
where $c_{i,y}$ is the number of votes label $y$ received for item $i$.

\paragraph{Softmax Normalization} The standard normalization scheme does not distribute probability mass to any label which receives no votes from any annotator. The works of~\citet{DBLP:conf/iccv/PetersonBGR19,DBLP:conf/naacl/FornaciariUPPHP21} propose to use the softmax function directly from label vote counts as a way to obtain soft labels for a given sample, as in \autoref{eq:softmax_norm}. 
\begin{equation}
\label{eq:softmax_norm}
    p_{soft}(i,y) = \frac{e^{c_{i,y}}}{\sum_{\hat{y}}e^{c_{i,\hat{y}}}}
\end{equation}
This can potentially help to further regularize a model.

\paragraph{Dawid \& Skene} A common method for aggregating crowd-sourced labels into a single ground-truth label is to treat the true label as a latent variable to be learned from annotations. Several models have been proposed in the literature to accomplish this~\cite{dawid1979maximum,DBLP:conf/naacl/HovyBVH13,carpenter2008multilevel}, often accounting for other aspects of the annotation problem such as annotator competence and item difficulty. One such method is the Dawid and Skene model~\cite{dawid1979maximum}, a highly popular method across fields for aggregating labels from crowd-annotations, which focuses in particular on modeling the true class based on each annotator's ability to correctly identify true instances of a given class. 
In other words, the model is designed to explain away inconsistencies of individual annotators, which may be desirable for use as a training signal when gold labels are unavailable. To obtain a soft label for a given sample $i$ from this model, we use the posterior distribution of the latent variable $c_{i}$ which models the true class for a given instance.\footnote{Implementation: \url{https://github.com/sukrutrao/Fast-Dawid-Skene}}

\paragraph{MACE} Multi-Annotator Competence Estimation (MACE, \citealt{DBLP:conf/naacl/HovyBVH13})\footnote{Implementation: \url{https://github.com/dirkhovy/MACE}} is another Bayesian method popular in NLP which focuses specifically on explaining away poor performing annotators. It does this by learning to differentiate between annotators which likely follow the global labeling strategy of selecting the true underlying label from those which follow a labeling strategy which deviates from this e.g. spamming a single label for every example. To do this, it learns a distribution over the true label for each sample, as well as the likelihood that each annotator is faithfully labeling each sample. For extensive details on both the Dawid and Skene and MACE models, as well as several other Bayesian annotation models, see the survey by \citet{DBLP:journals/tacl/PaunCCHKP18}.

\subsubsection{Combining Soft Labels}
\label{sec:soft_label_methods}
In order to acquire labels which capture the multiple views of the annotations learned by individual soft-labeling methods,  we develop and investigate novel methods for combining the soft-labels obtained from these different views. This is inexpensive, requiring zero additional annotations, and we will demonstrate that it is robust across tasks.

The goal for a single example $x_{i}$ is as follows: given a set of categorical distributions $p_{m}(y_{i}|x_{i})$ with $m \in \{1...M\}$ for $M$ different soft-labeling methods, produce a categorical distribution $p(y_{i}|x_{i}) = f(p_{1:M}(y_{i}|x_{i}))$ which will serve as a soft target for example $x_{i}$. The function $f$ should be selected to capture the different aspects of the problem which each distribution $p_{m}$ models. For example, MACE produces a posterior probability over the true class which explains away bad annotator behavior, while the model of Dawid and Skene produces a posterior which best explains each annotator's individual annotation biases. Our hypothesis is that combining several different models (i.e. different \textbf{views} of the crowd-sourced annotations) will yield labels that can induce more robust classifiers. 

\paragraph{Averaging} The most basic model to acquire an ensembled probability distribution is to take an average of the individual probabilities $p_{1:M}$. More formally, the averaging function $f_a$ takes the following form:
\begin{equation}
    f_{a}(p_{1:M}(y_{i}|x_{i})) = \frac{1}{M}\sum_{m}p_{m}(y_{i}|x_{i})
\end{equation}
This effectively yields a distribution which is the center of mass of the given distributions $p_{1:M}$.

\paragraph{Jensen-Shannon Centroid} The Jensen-Shannon centroid (JSC) is the minimizer of the sum of the Jensen-Shannon divergences (JSD) between a proposed distribution $Q$ and a set of probability distributions $p_{1:M}$. It is defined as:
\begin{equation}
\label{eq:js_centroid}
    f_{c}(p_{1:M}(y_{i}|x_{i})) = \argmin_{Q}\sum_{m}\text{JS}(p_{m}\|Q)
\end{equation}
where $\text{JS}(P\|Q)$ is the JSD, a symmetric version of the Kullback-Leibler divergence (KLD), defined as follows for discrete probability distributions:
\begin{equation}
\label{eq:jsd}
    \text{JS}(P\|Q) = \frac{1}{2}\text{KLD}(P\|S) + \frac{1}{2}\text{KLD}(Q\|S)
\end{equation}
\begin{equation*}
    S = \frac{1}{2}(P + Q)
\end{equation*}
\begin{equation}
    \text{KLD}(P\|Q) = \sum_{j}P^{(j)}\log\frac{P^{(j)}}{Q^{(j)}}
\end{equation}
In other words, the JSC is a probability distribution which has the least average total divergence to the average between itself and each probability distribution in a set of probability distributions. 

Finding the JSC can be done efficiently using methods from convex optimization. In particular, we use the ConCave-Convex procedure (CCCP, \citealt{DBLP:conf/nips/YuilleR01}) developed in \citet{DBLP:journals/entropy/Nielsen20}. The full derivation and definition of the method can be found in \citet{DBLP:journals/entropy/Nielsen20} Equations 94-104 and Algorithm 1, but at a high level, we can define a categorical distribution $p$ with $K$ classes using the natural parameter $\theta$ consisting of $K-1$ components as:
\begin{equation*}
    p = \{\theta_{1:(K-1)}, 1 - \sum_{k=1}^{K-1}\theta_{k}\}
\end{equation*}
The negative entropy of this distribution is then calculated in terms of $\theta$ as follows:
\begin{equation}
\label{eq:negentropy}
    F(\theta) = -H(\theta) = \sum_{k=1}^{K-1}\theta_{k}\log\theta_{k} \\ + (1 - \sum_{k=1}^{K-1}\theta_{k})\log(1 - \sum_{k=1}^{K-1}\theta_{k})
\end{equation}
which has partial derivatives and inverse gradient:
\begin{equation}
\label{eq:partial}
    \frac{\partial}{\partial \theta_{k}} = \log \frac{\theta_{k}}{1 - \sum_{k=1}^{K-1}\theta_{k}}
\end{equation}
\begin{equation}
\label{eq:inverse}
    \theta_{k} = (\nabla F^{-1}(\eta))_{k} = \frac{e^{\eta_{k}}}{1 + \sum_{k=1}^{K-1}e^{\eta_{k}}}
\end{equation}
The JSD between two categorical distributions $p_{1}$ and $p_{2}$ under this view can then be calculated in terms of the negative entropy $F$ defined in \autoref{eq:negentropy}:
\begin{equation*}
    \text{JS}(\theta_{1}\|\theta_{2}) = \frac{F(\theta_{1}) + F(\theta_{2})}{2} - F(\frac{\theta_{1} + \theta_{2}}{2})
\end{equation*}
Finally, the hyperparameterless update rule used to find the locally optimum JSC of a set of probability distributions $p_{1:M}$ using their natural parameters $\theta_{1:M}$ is defined in terms of \autoref{eq:partial} and \autoref{eq:inverse}:
\begin{equation}
    \theta^{(t+1)} = (\nabla F)^{-1}(\frac{1}{M}\sum_{m}F(\frac{\theta_{m} + \theta^{(t)}}{2}))
\end{equation}
where $\theta^{(0)} = [f_{a}(p_{1:M})]_{1:K-1}$. This optimization procedure is efficient and improves linearly to a local minimum.

\paragraph{Temperature Scaling} One approach in knowledge distillation is to scale the softmax output of the larger teacher network prior to using it to produce soft labels to teach the smaller student network. Here, we develop a method for optimizing a temperature parameter for each distribution in our ensemble based on the JSD between each distribution.

For each soft-labeling method $p_{m}$, we optimize a temperature parameter $T_{m}, m \in \{1...M\}$ which softens each distribution produced by that method. The optimization procedure for a given parameter $T_{m}$ is then given in \autoref{eq:temp_scale} and \autoref{eq:temp_scale_loss}
\begin{equation}
\label{eq:temp_scale}
    \tilde{p}_{k}(y_{i}|x_{i}) = \text{softmax}(\frac{l_{k}(y_{i}|x_{i})}{T_{k}})
\end{equation}
\begin{equation}
\label{eq:temp_scale_loss}
    \mathcal{L}(T_{m}) = \frac{1}{M - 1}\sum_{k != m}\text{JSD}(\tilde{p}_{m}\|\tilde{p}_{k})
\end{equation}
where $l_{k}$ are the log-probabilities for a given sample and $\text{JSD}$ is calculated as in \autoref{eq:jsd}. In practice, since the JSD is symmetric, we only need to calculate the loss for the $\frac{M(M-1)}{2}$ combinations of distributions in the ensemble. Since optimizing this loss directly will encourage the temperature to scale to infinity, as the loss will be 0 when a large enough temperature drives all distributions to be uniform, we add a regularization loss on the temperature parameters in order to discourage them from being exceedingly large. The final loss (assuming averaging the JSD over a batch of samples) is given in \autoref{eq:temp_scale_loss_full}.
\begin{equation}
\label{eq:temp_scale_loss_full}
    \mathcal{L} = \frac{1}{Z}\sum_{j=1}^{M}\sum_{k=j+1}^{M}\text{JSD}(\tilde{p}_{j}\|\tilde{p}_{k}) + \lambda T_{j}^{2}
\end{equation}
where $\lambda$ is a regularization constant and $Z = \frac{M(M-1)}{2}$. Finally, after optimizing for the temperature parameters $T_{m}$, we aggregate the distributions by averaging over the temperature scaled ensemble.
\begin{equation}
    f_{t}(p_{1:M}) = f_{a}(\tilde{p}_{1:M})
\end{equation}

\paragraph{Hybrid} Finally, we develop a hybrid approach where we first temperature scale the distributions in the ensemble via \autoref{eq:temp_scale_loss_full}, followed by finding the JSC as in \autoref{eq:js_centroid}.
\begin{equation}
    f_{h}(p_{1:M}) = f_{c}(\tilde{p}_{1:M})
\end{equation}

\subsubsection{Learning From Soft labels}
Learning from soft labels requires methods which minimize the divergence between the probability distribution produced by a classifier and the distribution over labels. 
To do this, we adopt the strategies used in knowledge distillation and learning from crowd-sourced labels,  using the Kullback-Leibler divergence as a loss function between the probability distribution produced by a given classifier and the soft labels produced by the methods in \autoref{sec:soft_labels} and \autoref{sec:soft_label_methods}.

\subsection{Experimental Setup}
\label{sec:experimental_setup}
We focus our experiments on the following research questions:
\begin{itemize}[noitemsep]
    \item \textbf{RQ1}: Which methods for learning from crowd-sourced labels are most robust in out-of-domain settings?
    \item \textbf{RQ2}: Does aggregating multiple views of crowd annotations lead to more robust out-of-domain performance?
    \item \textbf{RQ3}: Which soft-labeling methods lead to better uncertainty estimation?
\end{itemize}
Our experiments focus on the out-of-domain setting, where there is distribution shift between training data and test data. We use pairs of datasets which capture the same high level tasks such that the training data has both gold and crowd-annotations available while the testing data has only gold annotations. Thus, to perform well on the test set, a model must be able to generalize across these two factors: input data sourced from different corpora and/or labels acquired from different sources. Additionally, two of our experiments employ limited datasets for training, giving a very difficult out of domain setting. We choose this setup in order to understand the impact of crowd-sourced soft targets on model generalization, whereas in the in-domain setting where train and test data use gold labels obtained from the same source, performance is dominated by the use of expert labels. In other words, when does the knowledge contained in soft-targets confer benefits over expert annotations in NLP?

For all experiments we use RoBERTa as our base network~\cite{DBLP:journals/corr/abs-1907-11692} with the same training hyperparameters in order to provide a stable comparison across different soft-labeling techniques. Additionally, this allows us to observe how the same soft-labeling techniques on the same network perform on different tasks. For the soft-labeling experiments (labeled ``KLD'') we only use soft labels obtained using one of the crowd-labeling methods described in \autoref{sec:soft_labels} and \autoref{sec:soft_label_methods} and trained using the KL divergence as the loss. Additionally, we experiment with the multi-task learning setup used in \citet{DBLP:conf/naacl/FornaciariUPPHP21,DBLP:journals/jair/UmaFHPPP21}, where the model is trained on both gold labels and soft crowd-sourced labels (labeled ``Gold + KLD''). This allows us to differentiate performance between when gold annotations are available vs. not, which is clearly beneficial in the in-domain test setting where the same method of acquiring gold labels is used for test data~\cite{DBLP:conf/naacl/FornaciariUPPHP21}, but not necessarily in the out-of-domain setting~\cite{DBLP:conf/iccv/PetersonBGR19}. The tasks and datasets used in our experiments are described in the following paragraphs.

\paragraph{Recognizing Textual Entailment (RTE)} The first task we consider is recognizing textual entailment (RTE). In the RTE task, a model must predict whether a hypothesis is entailed (i.e. supported) by a given premise. For training, we use the Pascal RTE-1 dataset~\cite{DBLP:conf/mlcw/DaganGM05} with crowd-sourced labels from~\citet{DBLP:conf/emnlp/SnowOJN08}. The dataset consists of 800 premise-hypothesis pairs annotated by 164 different annotators with 10 annotations per pair. As an out-of-domain test set, we use the Stanford Natural Langauge Inference dataset (SNLI)~\cite{DBLP:conf/emnlp/BowmanAPM15}, where we transform the task into binary classification by collapsing the ``neutral'' and ``contradiction'' classes into a single class.

\paragraph{Medical Relation Extraction (MRE)} Medical relation extraction (MRE) seeks to predict what relations hold between different biomedical entities in sentences extracted from biomedical papers. The MRE dataset used for training in this work is the crowd-sourced dataset from \cite{DBLP:journals/corr/abs-1809-00537}, which collected crowd annotations from 3,984 sentences from PubMed abstracts~\cite{DBLP:conf/acl/0001F14} annotated by at least 15 annotators for 14 different UMLS~\cite{DBLP:journals/nar/Bodenreider04} relations. Here we focus on the 975 sentence subset which also received expert annotations, specifically for the ``cause'' relationship. As such, we follow previous work~\cite{DBLP:journals/jair/UmaFHPPP21} and frame the task as a binary classification problem, where a positive label indicates the ``cause'' relation exists. For testing, we use the causal claim-strength dataset curated from~\cite{wright2021semi}, which contains 1,126 sentences from news articles and scientific papers related to health science labeled for causal claim strength (statement of no relation, correlational, conditional causal, and causal). We convert the dataset to a binary classification problem by combining the ``conditional causal'' and ``causal'' classes into the positive class and the ``correlational'' and ``no relation'' classes into the negative class.

\paragraph{Part-of-Speech Tagging (POS)} The POS tagging task is a sequence tagging task, where the goal is to predict the correct part-of-speech for each token in a sentence. For training data, we use the Gimpel dataset from~\citet{DBLP:conf/acl/GimpelSODMEHYFS11} with the crowd-sourced labels provided by~\citet{DBLP:conf/acl/HovyPS14} mapped to the universal POS tag set in~\citet{DBLP:conf/acl/PlankHS14}. The dataset consists of 1000 tweets (17,503 tokens) labeled with Universal POS tags and annotated by 177 annotators. Each token received at least 5 annotations. We use the publicly available sample of the Penn Treebank POS dataset~\cite{DBLP:journals/coling/MarcusSM94} accessed from NLTK~\cite{DBLP:conf/acl/Bird06} as our out-of-domain test set, which consists of 3,914 sentences from Wall Street Journal articles (100,676 tokens). 

\paragraph{Toxicity Detection} Finally, to measure performance on a highly subjective task, we use the toxicity detection dataset created as a part of the Google Jigsaw unintended bias in toxicity classification competition\footnote{\url{https://www.kaggle.com/competitions/jigsaw-unintended-bias-in-toxicity-classification}}. The dataset we use comes from~\citet{DBLP:journals/corr/abs-2205-00501}, which annotated 25,500 comments from the original Civil Comments dataset. The pool of annotators is specifically selected and split into multiple rating pools based on self-indicated identity group membership (African American and LGBTQ). This study revealed that one's identity group correlates with shifts in perception of which comments are toxic, thus demonstrating the subjectivity of the task. We randomly split the dataset into training and test, and for the test data we use the annotations in the original crowd-sourcing task; in other words, using a completely separate annotator pool that isn't selected based on identity groups. 

\subsection{Results and Discussion}

We evaluate the performance of each soft-labeling method across two metrics: F1 score and calibrated log-likelihood (CLL, \citealt{DBLP:conf/iclr/AshukhaLMV20}). F1 score provides an indication of the predictive ability for the classifiers trained on the soft-labels, while CLL describes how well calibrated each classifier is in terms of its predictive uncertainty. We use the CLL in order to obtain a fair comparison between methods, as it first performs temperature scaling on a held-out portion of the test set and measures the temperature-scaled negative log-likelihood on the rest of the test set, averaging results over 5 splits. Formal definitions of each metric can be found in \autoref{sec:eval_metrics}. Additionally, we show the average performance of the individual methods ($\mu$) to illustrate how the aggregation methods compare with the average individual performance, as well as results using only expert annotations (Gold) and only majority vote (Silver). We will first discuss general observations from our results, and based on this will provide answers for the research questions proposed in  \autoref{sec:experimental_setup}.

\subsubsection{Raw Performance} 

\begin{figure}[t]
         \centering
     \begin{subfigure}[b]{0.4\textwidth}
         \centering
         \includegraphics[width=\textwidth]{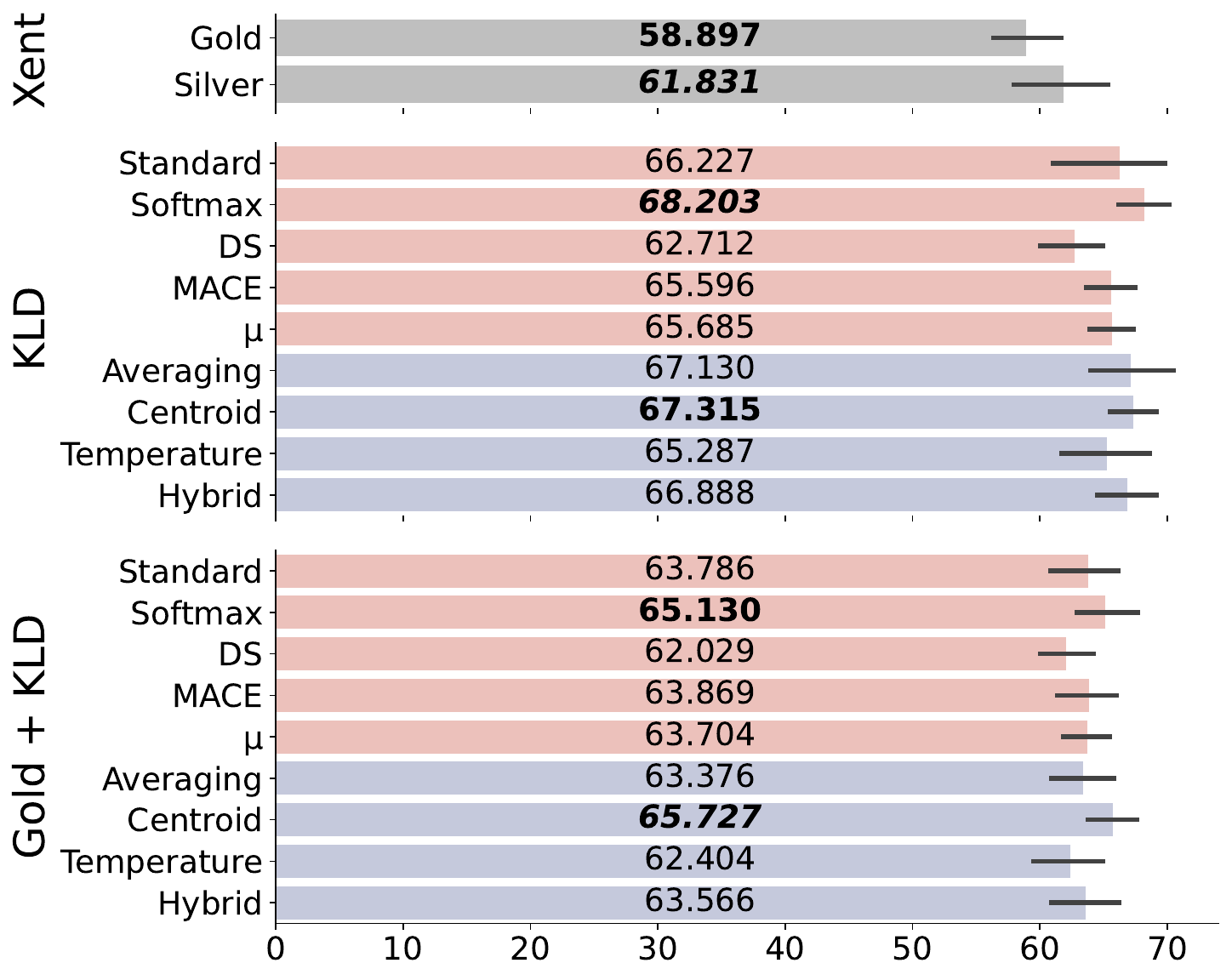}
         \caption{RTE}
         \label{fig:rte-f1}
     \end{subfigure}
     \hfill
     \begin{subfigure}[b]{0.4\textwidth}
         \centering
         \includegraphics[width=\textwidth]{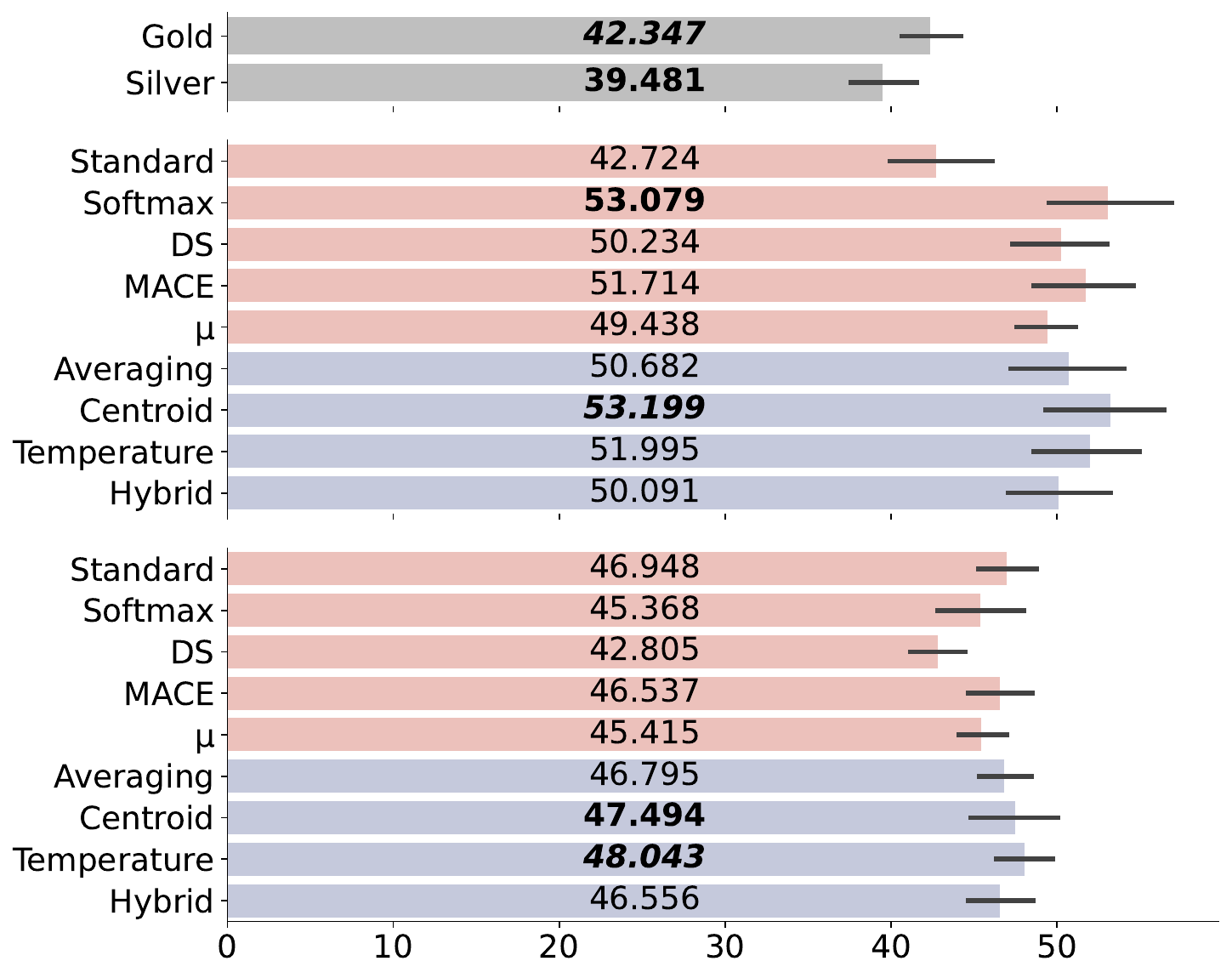}
         \caption{MRE}
         \label{fig:mre-f1}
     \end{subfigure} 
     \\
     \begin{subfigure}[b]{0.4\textwidth}
         \centering
         \includegraphics[width=\textwidth]{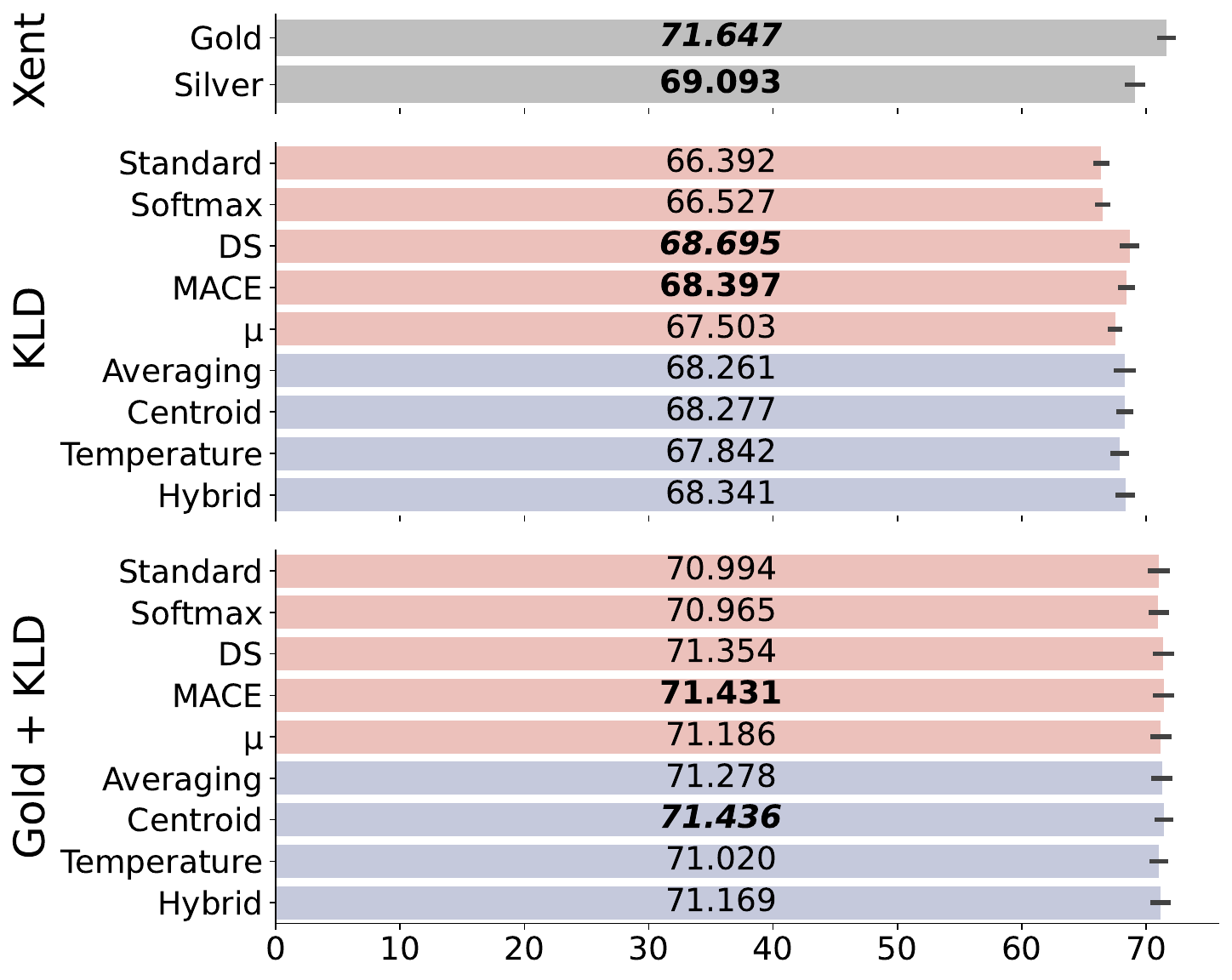}
         \caption{POS}
         \label{fig:pos-f1}
     \end{subfigure}
     \hfill
     \begin{subfigure}[b]{0.4\textwidth}
         \centering
         \includegraphics[width=\textwidth]{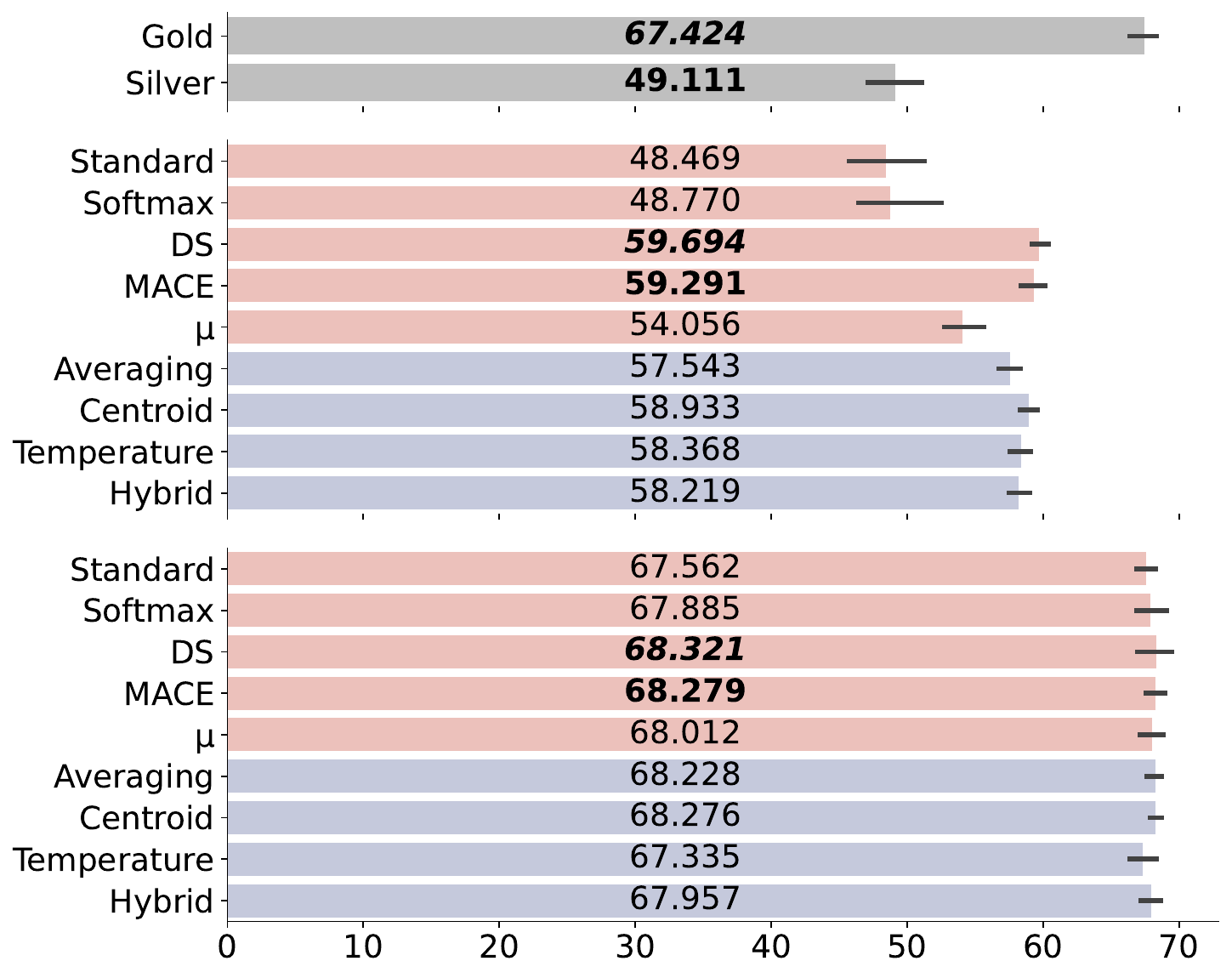}
         \caption{Toxicity}
         \label{fig:Jigsaw-F1}
     \end{subfigure} 
 
    \caption{F1 scores for the (a) RTE, (b) MRE, (c) POS, and (d) Toxicity datasets on out of domain test sets for each given method. Results are averaged across 20 random seeds (5 seeds for Toxicity detection). Grey are hard labels only, red are individual methods and blue are aggregation methods. Best results within each setting (Xent, KLD, Gold + KLD) are given in \textbf{\textit{bold italics}}, second best results in \textbf{bold}.}
    \label{fig:F1}
\end{figure}

\begin{figure*}[th]
         \centering
     \begin{subfigure}[b]{0.4\textwidth}
         \centering
         \includegraphics[width=\textwidth]{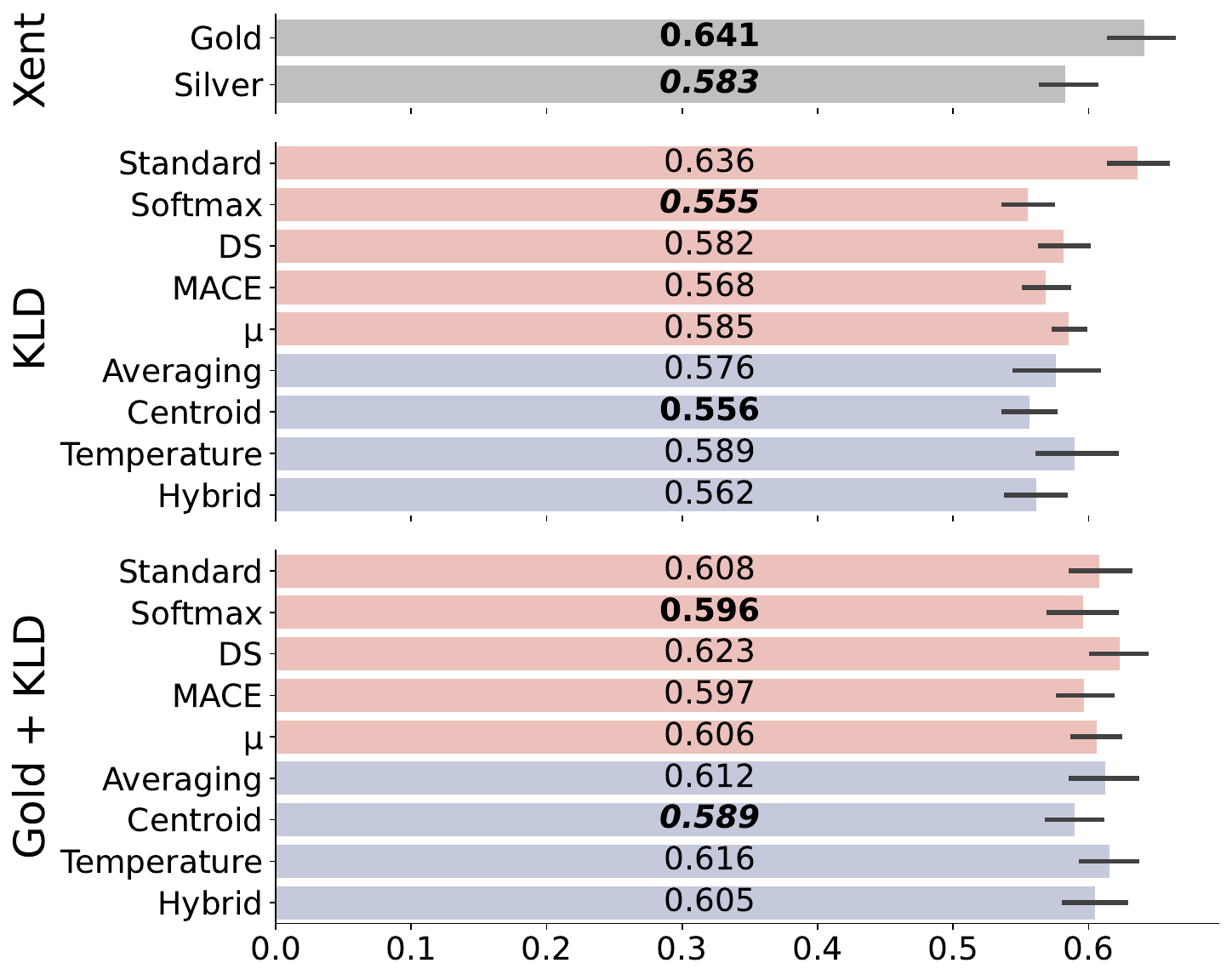}
         \caption{RTE}
         \label{fig:rte-nll}
     \end{subfigure}
     \hfill
     \begin{subfigure}[b]{0.4\textwidth}
         \centering
         \includegraphics[width=\textwidth]{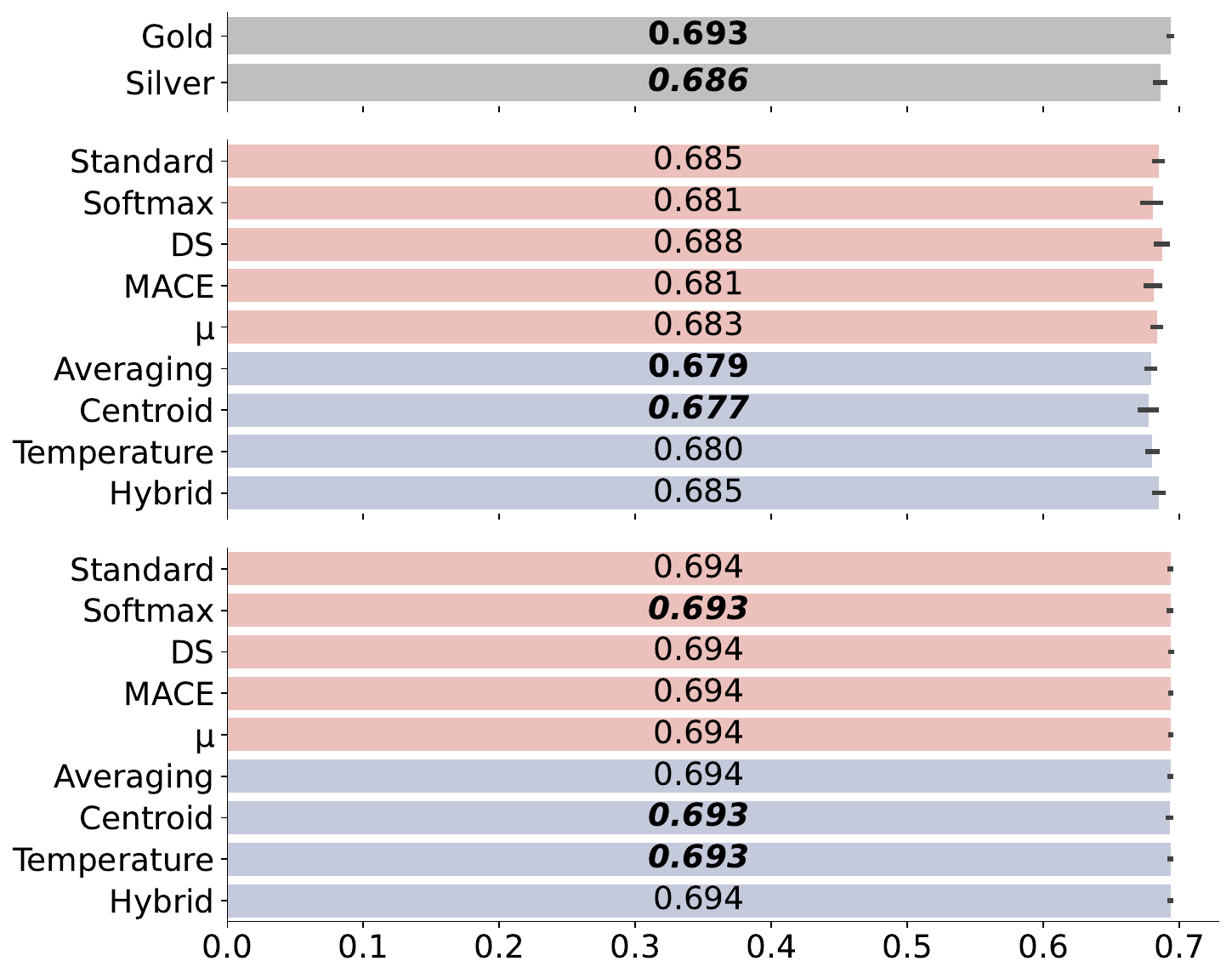}
         \caption{MRE}
         \label{fig:mre-nll}
     \end{subfigure} 
     \\
     \begin{subfigure}[b]{0.4\textwidth}
         \centering
         \includegraphics[width=\textwidth]{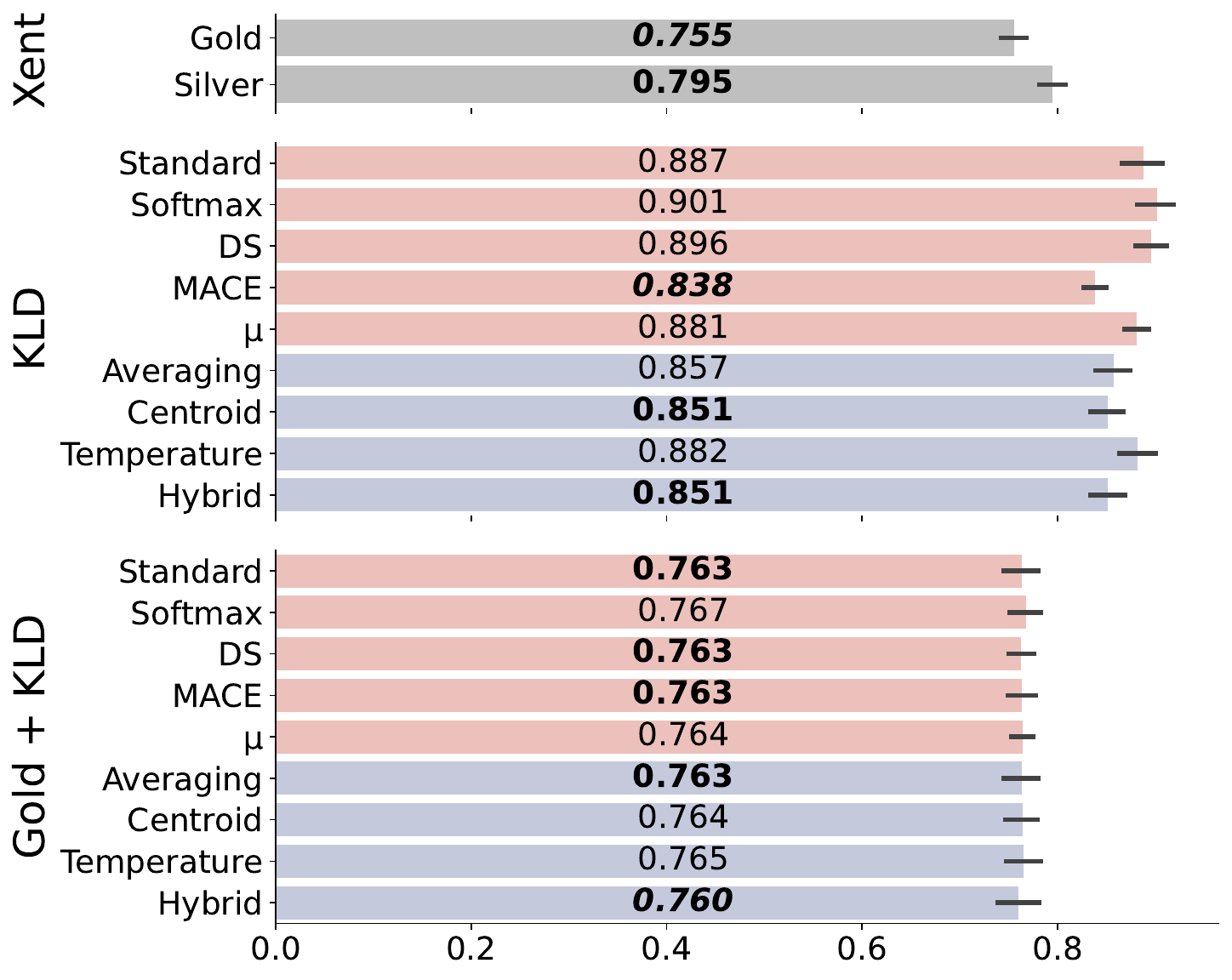}
         \caption{POS}
         \label{fig:pos-nll}
     \end{subfigure}
     \hfill
     \begin{subfigure}[b]{0.4\textwidth}
         \centering
         \includegraphics[width=\textwidth]{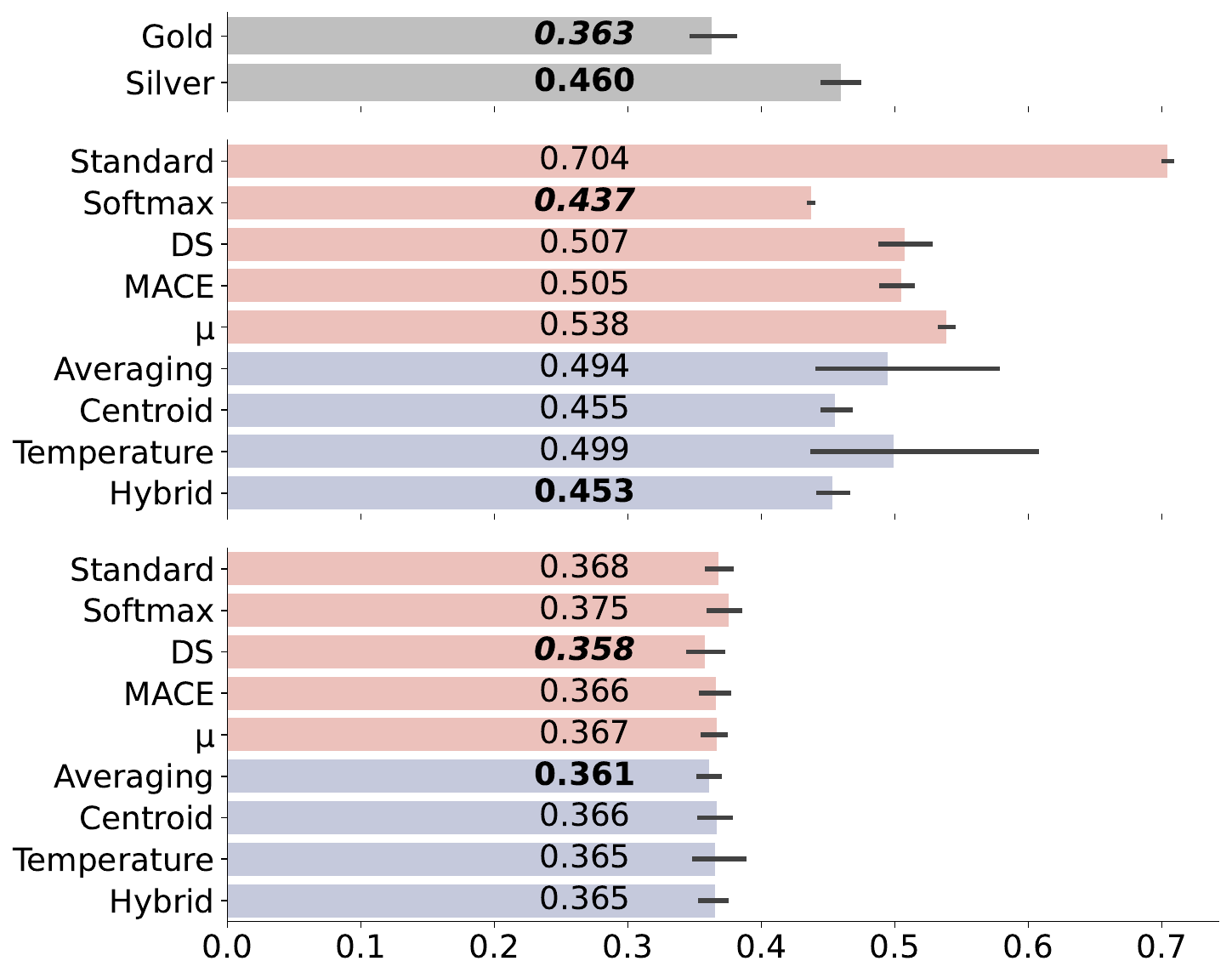}
         \caption{Toxicity}
         \label{fig:Jigsaw-NLL}
     \end{subfigure} 
 
    \caption{Calibrated log-likelihood scores (lower is better) for the (a) RTE, (b) MRE, (c) POS, and (d) Toxicity datasets on out of domain test sets for each given method. Results are averaged across 20 random seeds (5 seeds for Toxicity detection). Grey are hard labels only, red are individual methods and blue are aggregation methods. Best results within each setting (Xent, KLD, Gold + KLD) are given in \textbf{\textit{bold italics}}, second best results in \textbf{bold}.}
    \label{fig:NLL}
\end{figure*}

\paragraph{Overall} In general, we see that the RTE and MRE datasets are much more difficult to generalize from than the POS and Jigsaw tasks, as reflected in the high variance of the results. Additionally, gold labels in these two settings yield worse performance than simply training on soft labels, as opposed to the in-domain setting reported in \citet{DBLP:journals/jair/UmaFHPPP21} where gold labels are needed for high performance, indicating that it may be more beneficial to use only crowd-sourced labels in the out-of-domain setting. POS tagging sees the best performance when using only gold labels, contrasting with results reported in~\cite{DBLP:journals/jair/UmaFHPPP21} which show that adding soft labels with gold labels improves performance in the in-domain setting. The Toxicity task on the other hand benefits from both gold and crowd labels. This may be due to the input data coming from the same distribution, even though the gold labels come from disjoint sets of annotators. Additionally, all experiments using only soft-labels, with the exception of POS tagging, perform better than using hard labels obtained with majority vote. 

\paragraph{Soft Labels} Looking towards which soft-labeling method provides the best performance in the absence of gold labels, it is inconsistent across tasks. This was also seen in the survey by ~\citet{DBLP:journals/jair/UmaFHPPP21}. Where the aggregation methods provide the most benefit is in their \textbf{consistency} of performance. In particular, the aggregation method using the JSC (Centroid in \autoref{fig:F1}) yields best or near-best performance across datasets, while the hybrid method works slightly better on POS tagging. This is despite fluctuations in performance for the unaggregated methods across tasks. For example, the softmax method works well for RTE and MRE, but worse for POS tagging and much worse for the highly subjective Toxicity detection task. The Bayesian methods show the opposite behavior, working well for POS tagging and Toxicity detection (potentially due to there being far more annotations from which to learn), but much worse for RTE and MRE. Looking at the average performance of the individual soft-labeling methods vs. the aggregation methods, we see that aggregation consistently performs better than average. Aggregation is also resistant to low-performing constituent distributions, as can be seen in the toxicity experiment where both the standard and softmax distributed labels produce significantly worse classifiers than those trained on labels from either Bayesian method, while each aggregation method remains close to the best performers. Finally, we also see that temperature scaling does not benefit performance in this setting, and robust performance is achieved with the JSC alone.

\paragraph{Gold + KLD} Adding gold labels for the RTE and MRE tasks leads to worse performance, potentially due to the limited amount of labeled data. This adds further evidence to the literature that soft labels can provide benefits over expert annotations for out-of-domain performance~\cite{DBLP:conf/iccv/PetersonBGR19}. In terms of raw performance, gold labels are sufficient to obtain best performance for POS tagging, with soft labels not conferring benefits in the out-of-domain setting. This may be explained by the observation that the gold annotations for the POS dataset~\cite{DBLP:conf/acl/GimpelSODMEHYFS11} were collected by researchers correcting labels for tweets pre-tagged by a tagger trained on Wall Street Journal articles (as in PTB), while the crowd-sourced annotations we use from \cite{DBLP:conf/acl/HovyPS14} are annotated from scratch with minimal context, only seeing three words at a time. As such, while there is a significant difference between the source of input data between train and test, there may be less difference in terms of gold labels. For the toxicity detection task, all methods perform within reasonable ranges of each other, with the Bayesian methods and basic averaging conferring slightly better performance.

\subsubsection{Uncertainty Estimation}

\paragraph{Overall} We see that uncertainty estimation as measured using calibrated log-likelihood can be improved with the addition of soft-labels in all cases except for POS tagging. The benefits are again more pronounced for the RTE and MRE tasks, where training data is limited. We also see inconsistency from the individual soft labeling methods across tasks, while the aggregation methods (and particularly the JSC) offer much more consistent uncertainty estimation which is better or approximately equal to the performance of the best performing individual method.

\paragraph{Soft Labels} When looking at soft-labels only, the JSC aggregation method provides the most consistent results across tasks, with either the best or second best performance. The hybrid method also offers good uncertainty estimation, especially in the large-data regime of POS tagging and Toxicity detection, though less so for MRE. 

\paragraph{Gold + KLD} As with the raw performance results, including gold labels in a multi-task setup yields better uncertainty estimation when labeled data is abundant; otherwise using only soft-labels yields better uncertainty estimation.

\subsubsection{Research Questions}
\paragraph{RQ1: Best methods for OOD performance.} In the out-of-domain setting, we find that among individual soft-labeling techniques, no consistent and clear best performer arises. Aggregating the soft-labels appears to mitigate these fluctuations in performance; in particular, aggregating using the JSC of the individual distributions, which leads to consistently best or near-best performance on all tasks.

\paragraph{RQ2: Does aggregation help?} We find that aggregating multiple views of crowd-labels sometimes leads to better performance in the out of distribution setting, but will generally be at least approximately as good as the best performing individual methods regardless of poor performance induced by some individual methods. This is illustrated by the observation that on all tasks in both the multi-task and single-task settings, at least one individual soft labeling method leads to noticeably poorer performance than the best individual methods, while aggregating across the soft-labeling methods using the JSC is consistently high performing.

\paragraph{RQ3: Uncertainty estimation from soft-labeling.} We find that in the absence of hard-labels, different individual soft-labeling methods are inconsistent in their uncertainty estimation across tasks. Again, aggregating these different views of the crowd-sourced labels mitigates these fluctuations. As with raw performance, we find that the Jenson-Shannon centroid is a sensible and consistent choice across tasks in the out-of-distribution setting,

\subsubsection{Analysis} 

\begin{table}
    \def\arraystretch{1.2}
    \centering
    \begin{tabular}{l c c c c}
    \toprule 
    Dataset & Avg & Centroid & Temp & Hybrid \\
    \midrule 
RTE & -  & 0.993** & - & - \\
MRE & -  & 0.998** & - & - \\
POS & -  & 0.987** & - & - \\
Jigsaw & -  & 0.996** & - & - \\

    \bottomrule 

    \end{tabular}
    \caption{Pearon correlation between the JS divergence of different aggregation methods and individual methods correlated with the individual method's average JS divergence to each other individual method. We only report correlation scores that are significant at p < 0.05 (- indicates no significance).} 
    \label{tab:correlation_results}
\end{table}

We briefly analyze the difference in JSD (\autoref{eq:jsd}) between the aggregation methods and each individual method on each dataset. We first ask if the performance of an aggregation method correlates with its average JSD to the distribution generated by the best performing individual method. Looking at the Pearson correlation on each dataset, we find that a statistically significant correlation appears for the RTE dataset (RTE: 0.962, p < 0.05), while the scores obtained for other datasets were not significant. As such, it suggests the possibility that performance is correlated with distance to the best performing distribution, though more experiments would be needed for statistically significant scores on the other datasets.

We next explore how the JSD of each aggregation method to each individual method relates to how close the distribution produced by that individual method is to all other individual methods. This is to understand if there is a relationship between how close the individual methods are and the distribution obtained from each aggregation method. In other words, we correlate the following values:
\begin{equation*}
    \text{JSD}(Q\|p_{m}), \frac{1}{M-1}\sum_{k != m}\text{JSD}(p_{m}\|p_{k})
\end{equation*}
where $Q$ is the distribution produced by one of the aggregation methods and $p_{m}$ is the distribution of individual method $m$. Our hypothesis is that the JSC aggregation method produces a distribution which is closer to the distributions in the ensemble which are more similar to each other, as opposed to simple averaging which may be influenced by more divergent distributions. Our hypothesis for the JSC is confirmed by the results in \autoref{tab:correlation_results}, where the average JSD of the distributions produced by the JSC aggregation method to those of a given individual method is highly correlated with the average JSD of that individual method to all other individual methods. This suggests that aggregating using the JSC will lead to distributions closer to the hubs of an ensemble, where many of the individual distributions are similar. This may be desirable if those different views are representative of the problem one is modeling; the downside is the potential to ignore divergent views of the data which could be informative. We leave further exploration of this tradeoff to future work.

\subsection{Conclusion} In this work we present a systematic comparison of soft-labeling techniques from crowd-sourced labels and demonstrate their utility on out-of-domain performance for several text-classification tasks. The out-of-domain setting allows us to observe how well learning from crowd-sourced soft-labels enables generalization to unseen domains of data, potentially reflecting the ``dark knowledge'' imparted by these labels. Given than no consistent best performing model appears, we propose four novel methods for aggregating multiple views of crowd-sourced labels into a combined distribution, demonstrating that doing so leads to consistently robust performance across tasks despite fluctuations in performance shown by the constituent views. Concretely, we show that using the JSC between the constituent distributions yields consistently high raw performance and good uncertainty estimation, and is resilient to fluctuations in performance of the individual methods. This constitutes a low-cost solution to acquiring reliable soft-labels from crowd-annotations which oftentimes outperform expert labels on out-of-domain data.

\section*{Acknowledgements}
$\begin{array}{l}\includegraphics[width=1cm]{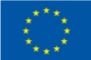} \end{array}$ This project has received funding from the European Union's Horizon 2020 research and innovation programme under the Marie Sk\l{}odowska-Curie grant agreement No 801199.

\newpage

\section{\citeworthdataset{}: Cite-Worthiness Detection for Improved Scientific Document Understanding}
\label{paper:citeworth}

\subsection{Introduction}
Building effective NLP systems from scientific text is challenging due to the highly domain-specific and diverse nature of scientific language, and a lack of abundant sources of labelled data to capture this. While large scale repositories of extracted, structured, and unlabelled plain-text scientific documents have recently been introduced~\cite{lo2020s2orc}, most datasets for downstream tasks such as named entity recognition~\cite{li2016biocreative} and citation intent classification~\cite{cohan2019structural} remain limited in size and highly domain specific. This begs the question: what useful training signals can be automatically extracted from massive unlabelled scientific text corpora to help improve systems for scientific document processing?

Scientific documents contain much inherent structure (sections, tables, equations, citations, etc.), which can facilitate creating large labelled datasets. Some recent examples %
include using paper field~\cite{beltagy2019scibert}, the section to which a sentence belongs~\cite{cohan2019structural}, and the cite-worthiness of a sentence~\cite{cohan2019structural,sugiyama2010identifying} as a training signal.

Cite-worthiness detection is the task of identifying \textit{citing sentences}, i.e. sentences which contain a reference to an external source. It has useful applications, such as in assistive document editing, and as a first step in citation recommendation~\cite{farber2018cite}. In addition, cite-worthiness has been shown to be useful in helping to improve the ability of models to learn other tasks~\cite{cohan2019structural}. %
We also hypothesize that there is a strong domain shift between how different fields use citations, and that such a dataset is useful for studying domain adaptation problems with scientific text.

However, constructing such a dataset to be of high quality is surprisingly non-trivial. Building a dataset for cite-worthiness detection involves extracting sentences from a scientific document, labelling whether each sentence contains a citation, and removing all citation markers. As a form of distant supervision, this naturally comes with the hazard of adding spurious correlations, such as poorly removed citation text causing ungrammatical sentences and hanging punctuation, which can trivially indicate a cite-worthy or non-cite-worthy sentence. Additionally, the task itself is quite difficult to learn, as different fields employ citations differently, and whether or not a sentence contains a citation depends on factors such as the context in which it appears. Given this, we present \citeworthdataset{}, a %
rigorously curated dataset for cite-worthiness detection in English. \citeworthdataset{}~contains rich metadata, such as authors and links to cited papers, and all data is provided in \textit{full paragraphs}: every sentence in a paragraph is labelled in order to provide sentence \textit{context}. We offer the dataset to the research community to facilitate further research on cite-worthiness detection and related scientific document processing tasks.

Using \citeworthdataset{}, we ask the following primary research questions:
\begin{quote}
\textbf{RQ1}: How can a dataset for cite-worthiness detection be automatically curated with low noise (\S\ref{sec:dataset-construction})?

\textbf{RQ2}: What methods are most effective for automatically detecting cite-worthy sentences (\S\ref{sec:benchmarks})?

\textbf{RQ3}: How does domain affect learning cite-worthiness detection (\S\ref{sec:domain-adaptation})?

\textbf{RQ4}: Can large scale cite-worthiness data be used to perform transfer learning to downstream scientific text tasks (\S\ref{sec:transfer-learning})?
\end{quote}
We demonstrate that \citeworthdataset{}~is of high quality through a manual evaluation, that there are large differences in how models generalize to data from different fields, and that sentence context leads to significant performance improvements on cite-worthiness detection. Additionally, we find that cite-worthiness is a useful task for transferring to downstream scientific text tasks, in particular citation intent classification, for which we offer performance improvements over the current state-of-the-art model SciBERT~\cite{beltagy2019scibert}.

In sum, our \textbf{contributions} are as follows:
\begin{itemize}[noitemsep]
    \item \citeworthdataset{}, a dataset of 1.2M rigorously cleaned sentences from scientific papers labelled for cite-worthiness, balanced across 10 diverse scientific fields.
    \item A method for cite-worthiness detection which considers the entire paragraph a sentence resides in, improving by 5 F1 points over the state of the art model for scientific document processing, SciBERT~\cite{beltagy2019scibert}.
    \item A thorough analysis of the problem of cite-worthiness detection, including explanations of predictions and insight into how scientific domain affects performance.
    \item New state of the art on citation intent detection via transfer learning from joint citation detection and language model fine-tuning on \citeworthdataset{}, with improved performance over SciBERT on several other tasks.
\end{itemize}
\subsection{Related Work}
\subsubsection{Cite-Worthiness Detection}
Cite-worthiness detection is the task of identifying \textit{citing sentences}, i.e. sentences which contain a reference to an external source. The reasons for citing are varied, e.g. to give credit to existing ideas or to provide evidence for a claim being made. %
\citet{sugiyama2010identifying} perform cite-worthiness detection using SVMs with features such as unigrams, bigrams, presence of proper nouns, and the classification of previous and next sentences. They create a dataset from the ACL Anthology Reference corpus (ACL-ARC, \citet{bird2008acl}), using heuristics to remove citation markers. \citet{farber2018cite} document the performance of convolutional recurrent neural nets on a larger set of three datasets coming from ACL-ARC, arXiv CS~\cite{farber2018high}, and Scholarly Dataset 2.\footnote{\url{http://www.comp.nus.edu.sg/~sugiyama/SchPaperRecData.html}} Datasets from these studies suffer from high class imbalance, are limited to only one or a few domains, and little analysis of the datasets is performed to understand the quality of the data or what aspects of the problem are difficult or easy. Additionally, no study to date has considered how sentence context can affect learning to perform cite-worthiness detection.%

In addition to being a useful task in itself, %
cite-worthiness detection is useful for other tasks in scientific document understanding. In particular, it has been shown to help improve performance on the closely related task of citation intent classification~\cite{jurgens2018measuring} when used as an auxiliary task in a multi-task setup~\cite{cohan2019structural}. However, cite-worthiness detection has not been studied in a transfer learning setup as a pretraining task for multiple scientific text problems. In this work, we seek to understand to what extent cite-worthiness detection is a transferable task.

\paragraph{Scientific Document Understanding}
Numerous problems related to scientific document understanding have been studied previously. Popular tasks include named entity recognition~\cite{li2016biocreative,kim2004introduction,dougan2014ncbi,luan-etal-2018-multi} and linking~\cite{conf/akbc/WrightKMH19}, keyphrase extraction~\cite{augenstein2017semeval,augenstein-sogaard-2017-multi}, relation extraction~\cite{kringelum2016chemprot,luan-etal-2018-multi}, dependency parsing~\cite{kim2003genia}, citation prediction~\cite{holm2020longitudinal}, citation intent classification~\cite{jurgens2018measuring,cohan2019structural}, summarization~\cite{DBLP:conf/conll/CollinsAR17}, and fact checking~\cite{DBLP:conf/emnlp/WaddenLLWZCH20}.

Datasets for scientific document understanding tasks tend to be limited in size and restricted to only one or a few fields, making it difficult to build models with which one can study cross-domain performance and domain adaptation. Here, we curate a large dataset of cite-worthy sentences spanning 10 different fields, showing that such data is both useful for studying domain adaptation and for transferring to related downstream scientific document understanding tasks.

\subsection{RQ1: \citeworthdataset{}~Dataset Construction}
\label{sec:dataset-construction}
The first research question we ask is: How can a dataset for cite-worthiness detection be automatically curated with low noise? To answer this, we start with the S2ORC dataset of extracted plain-text scientific articles~\cite{lo2020s2orc}. It consists of data from 81.1M English scientific articles, with full structured text for 8.1M articles. S2ORC uses \textsc{ScienceParse}\footnote{\url{https://github.com/allenai/scienceparse}} to parse PDF documents and \textsc{Grobid}\footnote{\url{https://github.com/kermitt2/grobid}} to extract structured data from text. As such, the data also includes rich metadata, e.g. Microsoft Academic Graph (MAG) categories, linked citations, and linked figures and tables. Throughout this work, a ``citation span'' denotes a span containing citation text (e.g. ``[2]''), and a ``citation marker'' is any text that trivially indicates a citation, such as the phrase ``is shown in.'' A citation span is also a type of citation marker. It is important to remove all citation markers from the dataset to prevent the model learning to use these signals for prediction.

\subsubsection{Data Filtering}
\label{sec:data-filtering}
Given the size of S2ORC, we first reduce the candidate set of data to papers where all of the following are available. 
\begin{itemize}[noitemsep]
    \item Abstract
    \item Body text
    \item Bibliography
    \item Tables and figures
    \item Venue information
    \item Inbound citations
    \item Microsoft Academic Graph categories
\end{itemize}
Filtering based on these criteria results in 5,494,387 candidate papers from which to construct the dataset.
After filtering the candidate set of papers, we perform the following checks on the sentences in the body text.
\begin{enumerate}[noitemsep]
    \item Citation spans are parenthetical author-year or bracketed-numerical form.
    \item Citation spans are at the end of a sentence.
    \item All possible citation spans have been extracted by S2ORC. %
    \item No citation markers are left behind after removing citation spans from the text.
    \item Sentence starts with a capital letter, ends with `.', `!', or `?', and is at least 20 characters long.
\end{enumerate}
The detailed steps of extracting and labelling sentences based on these criteria are given in \S\ref{sec:extraction}. With the first two criteria, we restrict the scope of cite-worthy sentences to being only those whose citation span comes at the end of a sentence, and whose citation format is parenthetical author-year form or bracketed-numerical form. In other words, cite-worthy sentences in our data are constrained to those of the following forms.

\begin{quote}
    This result has been shown in previous work (Author1 et al., \#\#\#\#, ...).
    
    This result has been shown in previous work [\#-\#].
\end{quote}
In this, we ignore citation sentences which contain inline citations, such as ``The work of Authors et al. (\#\#\#\#) has shown this in previous work'', as well as any sentence with a citation format that does not match the two we have selected. 

Curating cite-worthy sentences as such helps prevent spurious correlations in the data. Removing citations in the middle of a sentence runs the risk of rendering the sentence ungrammatical (for example, the above sample would turn into ``The work \textit{of has} shown this in previous work''), providing a signal to machine learning models. While there are cases where inline citations could potentially be removed in their entirety and not destroy the sentence structure, this is beyond the scope of this paper and left to future work.

\subsubsection{Extracting Cite-Worthy Sentences in Context}
\label{sec:extraction}
As we are interested in using sentence context for prediction, we perform extraction at the \textit{paragraph level}, ensuring that all of the sentences in a given paragraph meet the checks given in \S\ref{sec:data-filtering}. As such, our dataset construction pipeline for a given paper begins by first extracting all paragraphs from the body text which belong to sections with titles coming from a constrained list of permissible titles (e.g. ``Introduction,'' ``Methods,'' ``Discussion'') . The full list is provided in \S\ref{sec:premissible-section-titles}. 

For a given paragraph, we first word and sentence tokenize the text with SciSpacy~\cite{DBLP:conf/bionlp/NeumannKBA19}. Each sentence is then checked for containing citations using the provided citation spans in the S2ORC dataset. In some cases, the sentence contains citations which were missed by S2ORC; these are checked using regular expressions (see \S\ref{sec:regexes}). If a match is found the paragraph is ignored, as we only consider paragraphs where all citations have been extracted by S2ORC. Otherwise, the location and format of the citation is checked, again using regular expressions (see \S\ref{sec:regexes}). If the citation is not at the end of the sentence, the paragraph is ignored. We then remove the citation text using the provided citation spans for all sentences which pass the above checks.
 
 \begin{table*}[t]
    \centering
    \fontsize{10}{10}\selectfont
    \begin{tabular}{p{15cm}}
    \toprule %
    \textbf{Biology} \\
    \midrule
    Wood Frogs (Rana sylvatica) are a charismatic species of frog common in much of North America. They breed in explosive choruses over a few nights in late winter to early spring. \textcolor{citeworthy}{\emph{The incidence in Wood Frogs was associated with a die-off of frogs during the breeding chorus in the Sylamore District of the Ozark National Forest in Arkansas\sout{ (Trauth et al., 2000)}.}} \\
    \bottomrule \\
    \toprule
    \textbf{Computer Science} \\
    \midrule
    \textcolor{citeworthy}{\emph{Land use or cover change is a direct reflection of human activity, such as land use, urban expansion, and architectural planning, on the earth's surface caused by urbanization\sout{ [1].}}} Remote sensing images are important data sources that can efficiently detect land changes. \textcolor{citeworthy}{\emph{Meanwhile, remote sensing image-based change detection is the change identification of surficial objects or        geographic phenomena through the remote observation of two or more different phases\sout{ [2]}.}} \\
    \bottomrule %

    \end{tabular}
    \caption{Excerpts from training samples in \citeworthdataset{}~from the Biology and Computer Science fields. Green sentences are cite-worthy sentences, from which citation markers are removed during dataset construction.}
    \label{tab:dataset_examples}
\end{table*}

\begin{table}[t]
    \centering
    \fontsize{10}{10}\selectfont
    \begin{tabular}{l r}
    \toprule %
    Metric & \# \\
    \midrule
    Total sentences & 1,181,793 \\
    Total number of tokens & 34,170,708 \\
    Train sentences & 945,426 \\
    Dev sentences & 118,182 \\
    Test sentences & 118,185 \\
    Total cite-worthy & 375,388 (31.76\%)\\
    Total non-cite-worthy & 806,405 (68.24\%)\\
    Min char length & 21 \\
    Max char length & 1,447 \\
    Average char length & 152 \\
    Median char length & 142 \\
    \bottomrule %

    \end{tabular}
    \caption{Various statistics of the \citeworthdataset{}~dataset.}
    \label{tab:dataset_statistics}
    \vspace{-4mm}
\end{table}

Simply removing the citation span runs the risk of leaving other types of citation markers, such as hanging punctuation and prepositional phrases e.g. ``This was shown by the work of \sout{Author et al. (\#\#\#\#)}.'' To mitigate this, we remove all hanging punctuation at the end of a sentence that is not a period, exclamation point, or question mark, and check for possible hanging citations using the regular expression provided in \S\ref{sec:regexes}. The regular expression checks for many common prepositional phrases and citation markers occurring as the last phrase of a sentence such as ``see,'' ``of,'' ``by,'' etc. 

To handle issues with sentence tokenization, we also ensure that the first character of each sentence is a capital letter, and that the sentence ends with a period, exclamation point, or question mark. If all criteria are met for all sentences in a paragraph, the paragraph is added to the dataset. Finally, we build a dataset which is diverse across domains by evenly sampling paragraphs from the following 10 MAG categories, ensuring that each paragraph belongs to exactly one category: Biology, Medicine, Engineering, Chemistry, Psychology, Computer Science, Materials Science, Economics, Mathematics, and Physics. Example excerpts from the dataset are presented in \autoref{tab:dataset_examples}, and the statistics for the final dataset are given  in~\autoref{tab:dataset_statistics}.\footnote{The full dataset can be downloaded from this repository: \url{https://github.com/copenlu/cite-worth}}

\subsubsection{Manual Evaluation}
In order to provide some measure of the general quality of \citeworthdataset{}, we perform a manual evaluation of a sample of the data. We annotate the data for whether or not citation markers are completely removed, and for whether or not the sentences are well-formed, containing no obvious extraction artifacts. We sample 500 cite-worthy sentences and 500 non-cite-worthy sentences randomly from the data. Additionally, we compare to a baseline where the only heuristic used is to remove citation spans based on the provided spans in the S2ORC dataset. We again sample 500 cite-worthy and 500 non-cite-worthy sentences for annotation. The two sets are shuffled together and given to an independent expert annotator with a PhD in computer science for labelling. The annotator is instructed to label if the sentences are complete and have no hanging punctuation or obvious extraction errors, and if there are any textual indicators that the sentences contain a citation. The results for the manual annotation can be seen in \autoref{tab:manual_annotation_results}.

\begin{table}[t]
    \centering
    \fontsize{10}{10}\selectfont
    \begin{tabular}{l c c}
    \toprule %
    Method & Extracted Correct & Markers Removed \\
    \midrule
    Baseline & 92.07 & 92.78 \\
    Ours & \textbf{98.90} & \textbf{98.10} \\
    \bottomrule %

    \end{tabular}
    \caption{Results of manually annotating 1000 random sentences (per method) from \citeworthdataset{}~and a naive baseline which only removes citations based on provided citation spans . ``Extracted Correct'' are results for correctly extracting the sentences (i.e. that sentences are tokenized correctly and are grammatical), and ``Markers Removed'' are results for successfully removing citation markers. The data curated using our method has ~6\% fewer errors in terms of extraction and removal of citation markers, and less than 2\% of the samples have some form of citation marker.}
    \label{tab:manual_annotation_results}
\end{table}

We see that the \citeworthdataset{}~data are of a much higher quality than removing citation markers based only on the citation spans. Overall, our heuristics improve on extraction quality by 6.83\% absolute and on removing markers of citations by 5.32\% absolute. This results in 1.1\% of the sample data containing sentence cleaning issues, and 1.9\% having trivial markers indicating a citation is present. 
We argue that this is a strong indicator of the quality of the data for supervised learning.

\subsection{RQ2: System Evaluation}
\label{sec:benchmarks}

Next, we ask: what methods are most effective for performing cite-worthiness detection? To answer this and characterize the difficulty of the problem, we run a variety of baseline models on \citeworthdataset{}. The hyperparameters selected for each model, as well as hyperparameter sweep information, are given in Appendix \ref{sec:hyperparams}.

\paragraph{Logistic Regression}
As a simple baseline, we use a logistic regression model with TF-IDF input features. 

\paragraph{\citet{farber2018cite}} The convolutional recurrent neural network (CRNN) model from \citet{farber2018cite}. They additionally use oversampling to deal with class imbalance.

\paragraph{Transformer} We additionally train a Transformer model from scratch~\cite{vaswani2017attention}, tuning the model hyperparameters on a subset of the training data via randomized grid search.

\paragraph{BERT} We use a pretrained BERT model~\cite{devlin2019bert} due to the strong performance of large pretrained Transformer models on downstream tasks.

\paragraph{SciBERT} SciBERT~\cite{beltagy2019scibert} is a BERT model pretrained on a large corpus of scientific text from Semantic Scholar~\cite{ammar2018construction}, and is therefore potentially better suited to fine-tuning on scientific cite-worthiness detection.

\paragraph{SciBERT + PU Learning} We experiment with SciBERT trained using positive-unlabelled (PU) learning~\cite{elkan2008learning} which has been shown to significantly improve performance on citation needed detection in Wikipedia and rumour detection on Twitter~\cite{Wright2020ClaimCD}. The intuition behind PU learning is to assume that cite-worthy data is labelled and non-cite-worthy data is unlabelled, containing some cite-worthy examples. This is to mitigate the subjectivity involved in adding citations to sentences. Technically, this involves training a classifier on the positive-unlabeled data which will predict the probability that a sample is labeled, and using this to estimate the probability that a sample is positive given that it is unlabeled. One then trains a second model where positive samples are trained on normally and unlabeled samples are duplicated and trained on twice, once as positive and once as negative data, weighed by the first model's estimate of the probability that the sample is positive.

\paragraph{Longformer-Ctx} Finally, we test our novel contextualized prediction model based on Longformer~\cite{DBLP:journals/corr/abs-2004-05150}. Longformer is a Transformer based language model which uses a sparse attention mechanism to scale better to longer documents. %
We process an entire paragraph at a time, separating each sentence with a \texttt{[SEP]} token. Each \texttt{[SEP]} token representation at the output of Longformer is then passed through a network with one hidden layer and a classifier. As a control, we also experiment with Longformer using only single sentences as input (Longformer-Solo).

Due to the imbalance in the distribution of classes, the loss for each of the models is weighted. For comparison, we include results for SciBERT without weighting the loss function. The results for our baseline models on the test set of the dataset are given in \autoref{tab:citeworth_baselines}.\footnote{The code for all experiments can be found here: \url{https://github.com/copenlu/cite-worth}}

\begin{table}[t]
    \setlength{\tabcolsep}{1.5pt}
    \def\arraystretch{1.2}
    \centering
    \fontsize{10}{10}\selectfont
    \begin{tabular}{l c c c}
    \toprule %
    Method & P & R & \multicolumn{1}{c}{F1} \\
    \midrule %
       Logistic Regression & $46.65_{0.00}$& $64.88_{0.00}$& $54.28_{0.00}$\\
       \citet{farber2018cite} & $49.57_{0.96}$& $65.56_{2.61}$& $56.41_{0.34}$\\
       Transformer & $47.92_{0.78}$& $71.59_{1.74}$& $57.39_{0.10}$\\
       BERT & $55.04_{0.66}$& $69.02_{1.33}$& $61.23_{0.21}$\\
       SciBERT-no-weight & $\mathbf{65.94}_{0.37}$& $51.62_{0.53}$& $57.91_{0.30}$\\
       SciBERT & $57.03_{0.50}$& $68.08_{1.03}$& $62.06_{0.15}$\\
       SciBERT + PU & $49.46_{0.83}$& $\mathbf{82.12}_{1.40}$& $61.73_{0.27}$\\
       Longformer-Solo & $57.21_{0.25}$& $68.00_{0.41}$& $62.14_{0.02}$\\
       Longformer-Ctx & $59.92_{0.28}$& $77.15_{0.49}$& $\mathbf{67.45}_{0.06}$\\
    \bottomrule %

    \end{tabular}
    \caption{F1 performance of baselines on the test set of \citeworthdataset{}. Results are averaged across 5 seeds, with standard deviations given in the subscripts.}
    \label{tab:citeworth_baselines}
\end{table}
Our results indicate that context is critical, resulting in the best F1 score of 67.45 (Longformer-Ctx) and a 5.31 point improvement over the next best model. Using class weighting is also highly important, resulting in another increase of over 4 F1 points. Compared to not using class weights, PU learning performs significantly better, and leads to the highest recall of all models under test. Additionally, language model pre-training is useful, as BERT, SciBERT, and Longformer all perform significantly better than a Transformer trained from scratch and the model from \citet{farber2018cite}. 

To gain some insight into what the model learns, we visualize the most salient features from SciBERT for selected easy and hard examples. We use the single-sentence model instead of the paragraph model for simplicity. ``Easy'' samples are defined as those which the model predicted correctly with high confidence, and ``hard'' examples are defined as those for which the model had low confidence in its prediction. We use the InputXGradient method~\cite{kindermans2016investigating}, specifically the variant using L2 normalization over neurons to get a pre-embedding score, as it has been recently shown to have the best overall agreement with human rationales versus several other explainability techniques~\cite{atanasova2020diagnostic}. The method works by calculating the gradient of the output with respect to the input, then multiplies this with the input. In the examples below ``C'' refers to an example whose gold label is cite-worthy, and ``N'' refers to an example whose gold label is non-cite-worthy.

The model is able to pick up on obvious markers of cite-worthy and non-cite-worthy sentences for the following correctly classified examples, such as that a sentence refers to a preprint or to different sections within the paper itself:
\begin{quote}
\centering
C: 
{\setlength{\fboxsep}{0pt}\colorbox{white!0}{\parbox{0.35\textwidth}{
\colorbox{red!12.105263157894738}{\strut [CLS]} \colorbox{red!4.7894736842105265}{\strut in} \colorbox{red!5.7894736842105265}{\strut this} \colorbox{red!13.157894736842106}{\strut note} \colorbox{red!3.736842105263158}{\strut ,} \colorbox{red!8.947368421052634}{\strut we} \colorbox{red!10.526315789473685}{\strut follow} \colorbox{red!5.052631578947369}{\strut the} \colorbox{red!9.473684210526317}{\strut approach} \colorbox{red!5.7894736842105265}{\strut to} \colorbox{red!5.105263157894737}{\strut the} \colorbox{red!8.421052631578949}{\strut en} \colorbox{red!20.526315789473685}{\strut \#\#och} \colorbox{red!7.894736842105264}{\strut \#\#s} \colorbox{red!23.157894736842106}{\strut conjecture} \colorbox{red!22.105263157894736}{\strut outlined} \colorbox{red!10.526315789473685}{\strut in} \colorbox{red!12.105263157894738}{\strut the} \colorbox{red!100.0}{\strut preprint} \colorbox{red!13.157894736842106}{\strut .} \colorbox{red!13.684210526315791}{\strut [SEP]} 
}}}

N: 
{\setlength{\fboxsep}{0pt}\colorbox{white!0}{\parbox{0.35\textwidth}{
\colorbox{red!33.92857142857142}{\strut [CLS]} \colorbox{red!85.71428571428571}{\strut conclusions} \colorbox{red!28.57142857142857}{\strut are} \colorbox{red!42.857142857142854}{\strut provided} \colorbox{red!23.21428571428571}{\strut in} \colorbox{red!100.0}{\strut section} \colorbox{red!37.49999999999999}{\strut 4} \colorbox{red!21.428571428571427}{\strut .} \colorbox{red!37.49999999999999}{\strut [SEP]} 
}}}
\end{quote}
We also see that the dataset contains many relatively difficult instances, as we show in the following incorrectly classified examples. E.g., the model observes ``briefly discussed'' as an indicator that an instance is non-cite-worthy when it is in fact cite-worthy, and that ``described earlier'' and ``previous work'' signal that a sentence is cite-worthy when it is in fact labelled as non-cite-worthy.
\begin{quote}
\centering
C:
{\setlength{\fboxsep}{0pt}\colorbox{white!0}{\parbox{0.35\textwidth}{
\colorbox{red!26.190476190476193}{\strut [CLS]} \colorbox{red!76.19047619047619}{\strut some} \colorbox{red!71.42857142857143}{\strut approaches} \colorbox{red!23.80952380952381}{\strut for} \colorbox{red!20.476190476190478}{\strut the} \colorbox{red!57.142857142857146}{\strut solution} \colorbox{red!16.666666666666668}{\strut as} \colorbox{red!29.523809523809526}{\strut well} \colorbox{red!15.714285714285715}{\strut as} \colorbox{red!30.0}{\strut their} \colorbox{red!66.66666666666667}{\strut limitations} \colorbox{red!32.38095238095239}{\strut are} \colorbox{red!100.0}{\strut briefly} \colorbox{red!61.904761904761905}{\strut discussed} \colorbox{red!26.66666666666667}{\strut .} \colorbox{red!30.0}{\strut [SEP]} 
}}}

N:
{\setlength{\fboxsep}{0pt}\colorbox{white!0}{\parbox{0.35\textwidth}{
\colorbox{red!28.358208955223876}{\strut [CLS]} \colorbox{red!34.32835820895522}{\strut this} \colorbox{red!43.28358208955223}{\strut simple} \colorbox{red!17.91044776119403}{\strut and} \colorbox{red!49.253731343283576}{\strut fast} \colorbox{red!49.253731343283576}{\strut technique} \colorbox{red!19.402985074626862}{\strut for} \colorbox{red!16.41791044776119}{\strut the} \colorbox{red!47.76119402985074}{\strut production} \colorbox{red!14.776119402985072}{\strut of} \colorbox{red!89.55223880597013}{\strut snps} \colorbox{red!40.298507462686565}{\strut was} \colorbox{red!68.65671641791045}{\strut described} \colorbox{red!99.99999999999999}{\strut earlier} \colorbox{red!41.7910447761194}{\strut in} \colorbox{red!70.14925373134326}{\strut our} \colorbox{red!86.56716417910447}{\strut previous} \colorbox{red!73.13432835820895}{\strut work} \colorbox{red!26.86567164179104}{\strut .} \colorbox{red!32.83582089552238}{\strut [SEP]} 
}}}
\end{quote}
We hypothesize that in such instances, context can help the most in disambiguating which sentences in a paragraph should be labelled as cite-worthy. Additionally, other information such as the section in which a sentence resides could help. E.g., to correctly label the fourth statement above as ``non-cite-worthy'', it may help to see that the last sentence of the paragraph is ``In our previously published work, it was reported that SNPs were joined together by the heat treatment, and this process led to increase in the sizes of SNPs which finally resulted in sharper XRD peaks'' which is a cite-worthy sentence. Additionally, it may help to know that it resides in the ``Discussion'' section of the paper.

\subsection{RQ3: Domain Evaluation}
\label{sec:domain-adaptation}
\begin{figure}[t]
  
  \centering
    \includegraphics[width=0.5\textwidth]{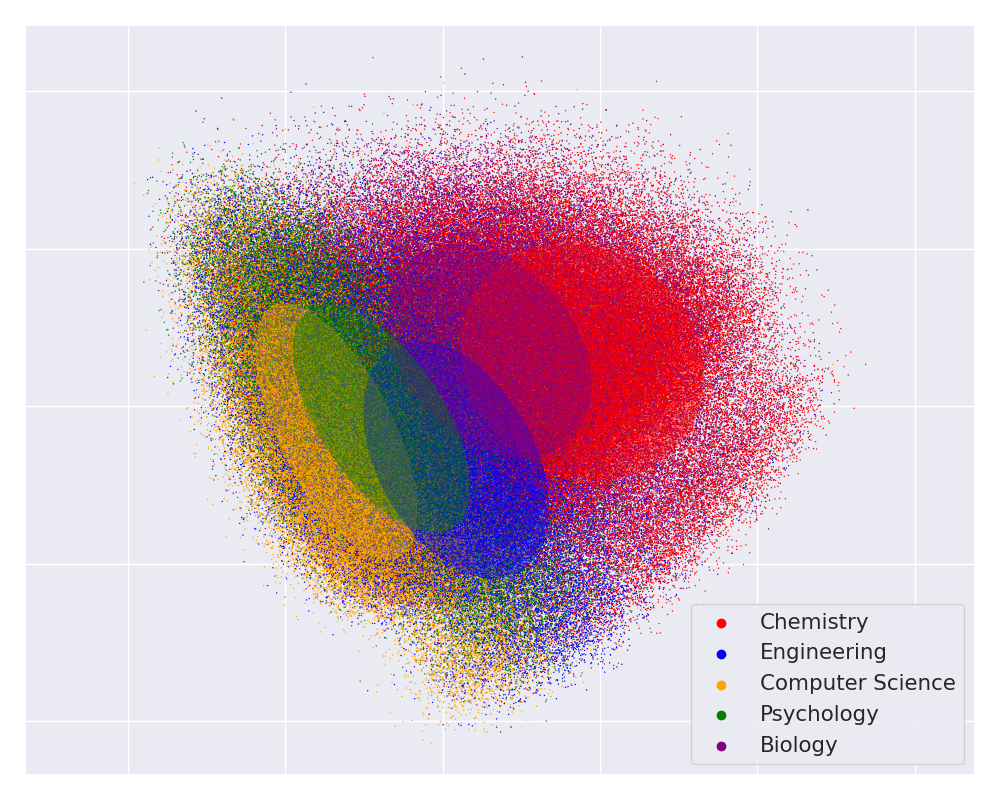}
    \caption{Visualizing the BERT embeddings for 5 of the 10 domains from \citeworthdataset{}~using the method by \citet{Aharoni2020UnsupervisedDC}. Clustering is performed using Gaussian Mixture Models.}
    \label{fig:domain-visualization}
\end{figure}

We next ask: how does domain affect learning to perform cite-worthiness? To answer this, we study the relationships between cite-worthiness data from different fields and how the Longformer-Ctx model performs in a cross-domain setup. For ease of analysis we limit the scope of fields to 5 of the 10 fields in the dataset: Chemistry, Engineering, Computer Science, Psychology, and Biology. 

First, we visualize the embedding space for data from each of these domains using the method of \citet{Aharoni2020UnsupervisedDC}. In this, the data is passed through BERT (specifically the base, uncased variant) and the output representations for each token in a sentence are average pooled. These representations are visualized in 2D space via PCA in \autoref{fig:domain-visualization}. It is clear that similar fields occupy closer space, with `engineering' and `computer science' sharing closer representations, as well as `biology' and `chemistry'. We perform clustering on this data using a Gaussian mixture model similarly to \citet{Aharoni2020UnsupervisedDC}, finding that domains form somewhat distinct clusters with a cluster purity of 57.61. This demonstrates that the data in different fields are drawn from different distributions, thus differences could exist in a model's ability to perform cite-worthiness detection on out of domain data.

To test this, we perform a cross-validation experiment using the 5 selected fields, training on one field and testing on another for all 25 combinations. The results for the 5x5 train/test setup using Longformer-Ctx are given in \autoref{tab:domain-adaptation-results}. 

Not surprisingly, the best performance for each split occurs when training on data from the same field. We also observe high variance in the maximum performance for each field ($\sigma$ = 3.32), and between different fields on the same test data, despite large pretrained Transformer models being relatively invariant across domains~\cite{wright2020transformer}. This suggests stark differences in how different fields employ citations. Additionally, we observe a strong (inverse) correlation between distance in the embedding space and performance on different domains, showing that using more similar data for training helps on out-of-domain performance~\cite{Aharoni2020UnsupervisedDC}.

\begin{table}[t]
\def\arraystretch{1.5}
    \centering
    \fontsize{10}{10}\selectfont
    \begin{tabular}{l|c|c|c|c|c|}
    \multicolumn{1}{l}{\diagbox[innerwidth=1cm]{Train}{Test}} & \multicolumn{1}{c}{Ch} & \multicolumn{1}{c}{E} & \multicolumn{1}{c}{CS} & \multicolumn{1}{c}{P} & \multicolumn{1}{c}{B}\\
    \cline{2-6}
    Ch & \textbf{67.58}& 58.41& 56.86& 62.35& 68.23\\
    \cline{2-6}
    E & 66.62& \textbf{60.25}& 60.11& 64.02& 68.07\\
    \cline{2-6}
    CS & 65.05& 59.36& \textbf{61.99}& 63.85& 66.72\\
    \cline{2-6}
    P & 65.49& 58.03& 56.69& \textbf{65.10}& 68.27\\
    \cline{2-6}
    B & 66.59& 58.80& 58.22& 64.54& \textbf{69.12}\\
    \cline{2-6}
    \multicolumn{1}{c}{} \\
    \cline{2-6}
    $\sigma$ & 0.90& 0.78& 2.02& 0.92& 0.77\\
    \cline{2-6}
    $\rho$ & 0.87& 0.86& 0.76& 0.67& 0.79\\
    \cline{2-6}
    \end{tabular}
    \caption{F1 performance on different domain adaptation settings for the fields (Ch)emistry, (E)ngineering, (C)omputer (S)cience, (P)sychology, and (B)iology. %
    Out-of-domain tests use the entire set of data from that field, while in domain tests use 80\% of data for training, 10\% for validation, and 10\% for test. $\sigma$ is the standard deviation of performance of different train domains on the given test domain, and $\rho$ is Pearson correlation between performance and Euclidean distance from the train domain cluster to the test domain cluster. }
    \label{tab:domain-adaptation-results}
\end{table}

\subsection{RQ4: Cite-Worthiness for Transfer Learning}
\label{sec:transfer-learning}
\begin{table*}[t]
    \centering
    \fontsize{10}{10}\selectfont
    \begin{tabular}{l l l c c c c}
    \toprule %
    Dataset & Reference & Task & \multicolumn{1}{l}{Base} & \multicolumn{1}{l}{LM} &
    \multicolumn{1}{l}{Cite}& \multicolumn{1}{l}{LMCite}\\
    \midrule %
       BC5CDR & \citet{li2016biocreative} & NER & $89.84_{0.18}$& $\mathbf{90.03_{0.11}}$& $89.73_{0.25}$& $90.02_{0.79}$ \\
       JNLPBA & \citet{kim2004introduction} & NER & $77.02_{0.36}$& $77.13_{0.53}$& $76.97_{0.44}$& $\mathbf{77.15_{0.58}}$ \\
       NCBI-Disease & \citet{dougan2014ncbi}& NER & $\mathbf{88.79_{0.35}}$& $88.53_{0.58}$& $88.66_{0.57}$& $88.31_{0.43}$\\
       SciERC & \citet{luan-etal-2018-multi}& NER & $67.08_{0.50}$& $66.64_{0.47}$& $67.12_{0.46}$& $\mathbf{67.48_{0.45}}$ \\
    \midrule
       EBM-NLP & \citet{nye2018corpus} & PICO & $76.61_{0.21}$& $\mathbf{76.69_{0.28}}$& $76.55_{0.88}$& $76.41_{0.32}$ \\
    \midrule
       ChemProt & \citet{kringelum2016chemprot}& REL & $83.17_{0.43}$& $\mathbf{83.26_{0.90}}$& $82.70_{1.06}$& $83.16_{0.63}$\\
       SciERC &\citet{luan-etal-2018-multi} & REL & $80.21_{0.81}$& $\mathbf{80.68_{1.04}}$& $80.00_{1.73}$& $80.58_{0.96}$ \\
    \midrule
       ACL-ARC & \citet{jurgens2018measuring} & CLS & $71.82_{2.93}$& $70.95_{2.25}$& $\mathbf{73.68_{2.75}}$& $72.92_{3.76}$ \\
       SciCite & \citet{cohan2019structural} & CLS & $84.83_{0.65}$& $85.18_{0.47}$& $85.32_{0.16}$& $\mathbf{85.35_{0.29}}$ \\
       PaperField & \citet{beltagy2019scibert} & CLS & $65.48_{0.18}$& $\mathbf{65.57_{0.27}}$& $65.46_{0.24}$& $65.42_{0.48}$ \\
    \midrule
       Average & & & $78.386$ & $78.466$& $78.619$ & $\mathbf{78.680}$ \\
    \bottomrule %

    \end{tabular}
    \caption{Performance on various downstream scientific document understanding tasks as presented by \citet{beltagy2019scibert}. The metrics used are the same as in their paper: NER is span-level F1, PICO is token level F1, relation extraction is macro-F1, and ChemProt is micro-F1. All runs are averaged across 5 seeds. Subscripts are the standard deviation for 5 runs.}
    \label{tab:downstream_task_results}
\end{table*}
The final question we ask is: to what extent is cite-worthiness detection transferable to downstream tasks in scientific document understanding? To answer this, we fine tune SciBERT on the task of cite-worthiness detection as well as masked language modeling (MLM) on \citeworthdataset{}, followed by fine-tuning on several document understanding tasks. We use SciBERT in order to have a direct comparison with previous work \cite{beltagy2019scibert}. The tasks we evaluate on come from~\citet{beltagy2019scibert} and are categorized as follows. 
\begin{itemize}[noitemsep]
    \item Named Entity Recognition (NER)/PICO: These tasks involve labelling the spans of different types of entities in a document.%
    \item Relation Extraction (REL): This task involves labelling a sequence for the relationship between two entities.%
    \item Text classification (CLS): Finally, we test on several text classification tasks (citation intent classification and paper field classification), where the goal is to classify a sentence into one or more categories. %
\end{itemize}
We compare five variants of pre-training and fine-tuning, given as follows.

\paragraph{Base} The original SciBERT model.

\paragraph{LM} SciBERT with MLM fine tuning on \citeworthdataset{}.

\paragraph{Cite} SciBERT fine-tuned for the task of cite-worthiness detection. The classifier is a pooling layer on top of the \texttt{[CLS]} representation of SciBERT, followed by a classification layer.

\paragraph{LMCite} SciBERT with MLM fine tuning and cite-worthiness detection. The two tasks are trained jointly i.e. on each batch of training, the model incurs a loss for both MLM and cite-worthiness detection which are summed together.

The results for all experiments are given in \autoref{tab:downstream_task_results}. Note that the reported results for SciBERT are on re-running the model locally for fair comparison. We first observe that incorporating our dataset into fine-tuning tends to improve model performance across all tasks to varying degrees, with the exception of NER on the NCBI-Disease corpus. The tasks where cite-worthiness as an objective has the most influence are the two citation intent classification tasks (ACL-ARC and SciCite). We see average improvements of 1.8 F1 points for the ACL-ARC dataset (including 2 points F1 improvement over the minumum and maximum model performance of SciBERT) and 0.5 F1 points on SciCite. %
The best average performance is from the model which incorporates both MLM and cite-worthiness as an objective, which we call \citeworthmodelname{}.\footnote{We release two \citeworthmodelname{}\ models available from the HuggingFace model hub: \texttt{copenlu/citebert} and \texttt{copenlu/citebert-cite-only}.}

For other tasks, fine-tuning the language model on \citeworthdataset{}~data tends to be sufficient for improving performance, though the margin of improvement tends to be minimal. This is in line with previous work reporting that language model fine-tuning on in-domain data leads to improvements on end-task fine-tuning \cite{DBLP:conf/acl/GururanganMSLBD20}. \citeworthdataset{}~is relatively small compared to the corpus on which SciBERT is originally trained (30.7M tokens for the train and dev splits on which we train versus 3.1B), so one could potentially see further improvements by incorporating more data or including cite-worthiness as an auxiliary task during language model pre-training. However, this is outside the scope of this work. 

\subsection{Conclusion}
In this work, we present an in-depth study into the problem of cite-worthiness detection in English. We rigorously curate \citeworthdataset{}, a high-quality dataset for cite-worthiness detection; present a paragraph-level contextualized model which improves by 5.31 F1 points on the task of cite-worthiness detection over the existing state-of-the-art; show that \citeworthdataset{}~is a good testbed for studying domain adaptation in scientific text; and show that in a transfer-learning setup one can achieve state of the art results on the task of citation intent classification using this data. In addition to studying cite-worthiness and transfer learning, \citeworthdataset{}~is suitable for use in downstream natural language understanding tasks. %
As we retain the S2ORC metadata with the data, one could potentially use the data to study joint cite-worthiness detection and citation recommendation. Additionally, one could explore other useful problems such as modeling different authors' writing styles and incorporating the author network as a signal. We hope that the data and accompanying fine-tuned models will be useful to the research community working on problems in the space of scientific language processing.

\section*{Acknowledgements}
$\begin{array}{l}\includegraphics[width=1cm]{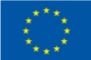} \end{array}$ The research documented in this paper has received funding from the European Union's Horizon 2020 research and innovation programme under the Marie Sk\l{}odowska-Curie grant agreement No 801199.

\newpage

\newcommand\methodfull{Knowledge Base Informed Negations\xspace}
\newcommand\method{\textsc{KBIN}\xspace}
\newcommand\cgentity{\textsc{ClaimGen-entity}\xspace}
\newcommand\cgbart{\textsc{ClaimGen-BART}\xspace}
\newcommand\nottt[1]{\bgroup\let\texttt\relax#1\egroup}

\section{Generating Scientific Claims for Zero-Shot Scientific Fact Checking}
\label{paper:generating}

\subsection{Introduction}
Scientific documents contain complex assertions about scientific processes, making it difficult to automate important tasks such as claim extraction and scientific fact checking. 
Additionally, the collection of manually annotated labels to train models on tasks with scientific data is time consuming and expensive due to the need for domain expertise~\cite{DBLP:conf/conll/CollinsAR17,augenstein-sogaard-2017-multi,lehman-etal-2019-inferring,DBLP:conf/emnlp/WaddenLLWZCH20,DBLP:journals/corr/abs-2104-06486}. As such, methods which require less manual annotation are especially useful in this domain. This work 
addresses this challenge by exploring how automatic generation of scientific claims can assist with dataset creation and zero-shot fact checking in the biomedical domain.

\begin{figure}[t]
  
  \centering
    \includegraphics[width=0.65\linewidth]{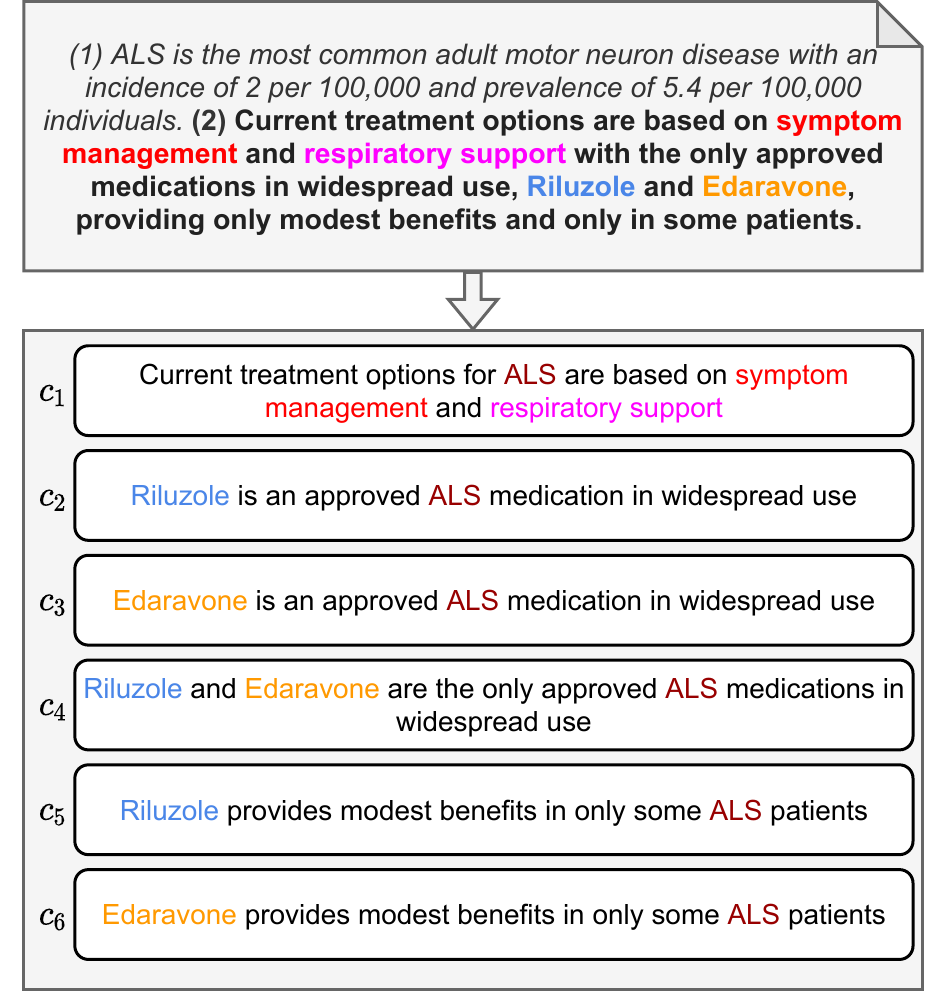}
    \caption{A complex excerpt from \citet{Mejzini2019ALSGM} (top) and the set of valid claims that can be generated from the bolded sentence (c1-c6).} %
    \label{fig:claim_generation_scifact}
\end{figure}

Being able to reduce scientific text to atomic assertions has numerous possible applications, and is known to be helpful for scientific communication and machine processing of scientific concepts~\cite{DBLP:conf/esws/KuhnBNK13}. 
Claim generation can enable zero-shot fact checking, reducing the need for expert-labeled data~\cite{DBLP:conf/acl/PanCXKW20},
and can be used to expand existing datasets such as \citet{DBLP:conf/emnlp/WaddenLLWZCH20} and \citet{DBLP:conf/acl/SaakyanCM20} without additional manual annotation. 
In this work we focus on the use of claim generation in scientific fact checking, demonstrating that claim generation enables zero-shot biomedical fact checking.

Generating scientific claims involves distilling a complex scientific sentence into one or more valid claims (see examples in \autoref{fig:claim_generation_scifact}).
As in previous work, we focus on biomedical claims as biomedical literature has long been a major focus in scientific natural language processing, as well as scientific fact checking~\cite{DBLP:conf/acl/SaakyanCM20,DBLP:conf/emnlp/WaddenLLWZCH20,kotonya-toni-2020-explainable}. While in \citet{DBLP:conf/emnlp/WaddenLLWZCH20}, claims were rewritten by domain experts from complex citation sentences (citances), we propose methods for automatically generating claims and claim negations from this source. 

Similar to other generation tasks, evaluating the quality of generated output requires multiple judgements beyond the fluency of the generated text, e.g., whether each claim is faithful to the source sentence, and is understandable on its own~\cite{DBLP:journals/corr/abs-2008-12009}. However, there are also other quality attributes that are important to assess specifically for scientific claims, such as whether each claim is atomic or check-worthy \citep{Wright2020ClaimCD}. Given this, we propose a set of manual evaluation criteria and annotation guidelines for evaluating claim generation (\S\ref{sec:quality_eval}).

Additionally, when generating claims to build datasets for tasks such as fact checking, a major challenge is creating refuted claims as negative training instances. 
Previous work has proposed automatic ways of generating refutations based on negating existing claims or creating claim variants via entity-replacement \citep{DBLP:conf/acl/PanCXKW20} and text-infilling using a pre-trained masked language model \citep{DBLP:conf/acl/SaakyanCM20}. We improve upon this by introducing \methodfull (\method), a principled method to generate refutations that performs entity-replacement using the relations and learned embeddings of entities in a domain-specific knowledge base. 

\paragraph{Contributions} In sum, our contributions are:
\begin{itemize}[noitemsep]
    \item The first study on scientific claim generation, comparing both unsupervised (\cgentity) and fully supervised (\cgbart) generation on biomedical text.
    \item \method, a novel method for generating refuted scientific claims which produces more convincing negations than previous work.
    \item Application of our claim generation methods on zero-shot scientific fact checking resulting in 90\% of the performance of a model trained on in-domain manually written claims. Additionally, a rigorous evaluation study showing that \cgbart and \method produce significantly higher quality claims and more convincing negations than previous work.
\end{itemize}

\subsection{Preliminaries}
\paragraph{Valid Claims} In this work, we define a \textit{valid claim} as one which is fluent, atomic, de-contextualized, and accurately reflects the meaning of the original sentence. Fluency is concerned with a claim being a generally well-formed English sentence, and atomicity with a claim being a ``verifiable statement expressing a finding about one aspect of a scientific entity or process,
which can be verified from a single source'' \cite{DBLP:conf/emnlp/WaddenLLWZCH20}. De-contextualilzation is concerned with a sentence being interpretable on its own, requiring none of the original surrounding text to resolve aspects of the sentence such as pronouns, abbreviations, etc., and can be handled by either directly de-contextualizing a sentence~\cite{DBLP:journals/tacl/ChoiPLKDC21} or by ensuring that all of the context sentences are available to a model~\cite{DBLP:journals/corr/abs-2112-01640}. Check-worthy claims in the wild may not be fluent, atomic, or de-contextualized, however it is useful to generate such claims as they have been shown to be useful for automated processing of science concepts~\cite{DBLP:conf/esws/KuhnBNK13} and scientific fact checking \citep{DBLP:conf/emnlp/WaddenLLWZCH20}.

\paragraph{Scientific Claim Generation} At a high level, scientific claim generation is the task of distilling one or more \textit{valid claims} from one or more sentences concerned with a scientific fact. More specifically, the task is defined as: given a scientific sentence $s$ and optionally additional context sentences $X$, generate one or more claims $c_{i} \in C$ which are valid and entailed by $s$ and $X$. In the context of fact checking, we must generate claims which are either \textit{supported} or \textit{refuted} by the literature, as well as those for which \textit{not enough information} is present to make a veracity judgement, in order that they may be paired with appropriate evidence documents to serve as training data for fact checking systems.
As such, we require methods which can take the claims in $C$ which are entailed by the source sentence and generate negations to acquire \textit{refuted} claims.

\subsection{Generating Supported Claims}
We experiment with two generation methods designed to produce claims which are \textit{supported} by the source sentence. The first method is an entity-centric unsupervised method adapted from \citet{DBLP:conf/acl/PanCXKW20} which requires no <sentence, claim> pairs (\cgentity). We also introduce a new method that uses BART \cite{DBLP:conf/acl/LewisLGGMLSZ20} trained on a small set of <sentence, claim> pairs to directly generate claims (\cgbart). 
For each sample $i$, we refer to the input source sentence as $s_{i}$, the context sentences as $x_{l}^{(i)} \in X_{i}$ and the output claims as $C_{i}$ consisting of $k$ claims $\{c_{1}^{(i)} \dots c_{k}^{(i)}\}$ 
Following \citet{DBLP:conf/emnlp/WaddenLLWZCH20}, we use citation sentences as unlabelled sentences for generation since these provide a natural link to an evidence document. Various components of our modeling pipelines take advantage of models pretrained on datasets for NER, NLI, QA, and fact-checking. We provide an overview of these datasets in \S\ref{sec:data_info}.

\subsubsection{\cgentity}
We adapt the entity-centric method presented in \citet{DBLP:conf/acl/PanCXKW20} as an unsupervised claim generation approach. This method has been tested on general domain fact checking, but has not been used for science claim generation and zero-shot scientific fact checking. In particular, we re-implement the base method used for generating supported claims and adapt it to the biomedical domain, substituting in a domain specific model for named-entity recognition. %
The method consists of the following steps for a given sample $i$:
\begin{enumerate}[noitemsep]
    \item Run named entity recognition (NER) on the input text to obtain a set of named entities $E_{i}$.
    \item For each named entity $e_{j}^{(i)}$, generate a question $q_{j}^{(i)}$ about that entity which can be answered from $s_{i}$.
    \item From $q_{j}^{(i)}$, generate the declarative form of the question to obtain claim $c_{j}^{(i)}$. 
\end{enumerate}

\paragraph{Named Entity Recognition} For NER, we employ scispaCy~\cite{DBLP:conf/bionlp/NeumannKBA19}, a spaCy\footnote{\href{https://spacy.io/}{https://spacy.io/}} pipeline for scientific NLP. The NER model is trained on the MedMentions dataset~\cite{DBLP:conf/akbc/MohanL19}, which consists of 4,392 PubMed abstracts exhaustively annotated for mentions of UMLS entities~\cite{DBLP:journals/nar/Bodenreider04}. 

\paragraph{Question Generation} For question generation, we use BART trained on questions from SQuAD~\cite{DBLP:conf/emnlp/RajpurkarZLL16}. As input for training, we encode a concatenation of the context and answer text from a given SQuAD question, and train the model to decode the question. During inference, we concatenate the source sentence $s_{i}$ and an entity $e_{j}^{(i)}$ and sample a question $q_{j}^{(i)}$ for this pair using beam search.

\paragraph{Question to Claim} Finally, as in \citet{DBLP:conf/acl/PanCXKW20}, we use a second BART model to generate declarative claims from questions. We train the model on the QA2D dataset~\cite{DBLP:journals/corr/abs-1809-02922}, which contains declarative full sentences paired with questions and their answer from SQuAD. The model is trained by encoding a concatenation of the question and answer, and decoding the full declarative sentence. At inference time, we concatenate and encode $q_{j}^{(i)}$ and $e_{j}^{(i)}$, and use beam search at the decoder to generate a claim $c_{j}^{(i)}$.

\subsubsection{\cgbart} 
We introduce a fully-supervised model for claim generation based on BART trained on <citance, claim> pairs. For this, we use the manual citance re-writes 
released by the SciFact authors,\footnote{\href{https://github.com/allenai/scifact/blob/master/doc/claims-with-citances.md}{https://github.com/allenai/scifact/blob/master/doc/claims-with-citances.md}}
which consist of citances from scientific papers rewritten as one or more atomic claims which are directly entailed by the citance. 

For training, we encode the citance, as well as the sentences immediately before and after the citance (the context), and train the decoder to generate claims directly. We choose to encode the context as well to help \textit{de-contextualize} generated claims. We concatenate the citance and context using a double pipe (i.e. $X_{i}$\texttt{||}$s_{i}$), and train the encoder to generate one claim at a time. 
We use top-$k$ sampling to generate multiple claims, with $k$ set to the number of noun chunks in the original source citance.\footnote{We use scispaCy to identify noun chunks}

\subsection{Knowledge Base Informed Negations}
\begin{figure*}[t]
  
  \centering
    \includegraphics[width=0.95\linewidth]{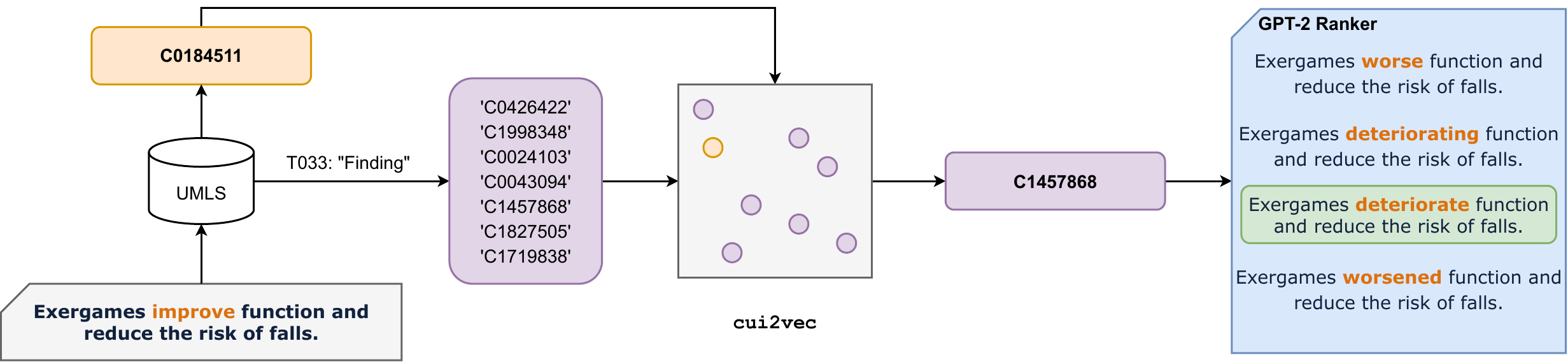}
    \caption{\method method. We start with NER and linking to UMLS using scispaCy. We then find the most similar concepts with the same type using \texttt{cui2vec}, replace the entity in the source sentence using the canonical name and aliases of similar entities, and rank them using GPT-2. Finally, from the highest ranked replacements, we select the claim which maximizes contradiction with the original claim using an external NLI model.}
    \label{fig:kbin_method}
\end{figure*}
\label{sec:kbin}
\begin{algorithm}[t]
\caption{KBIN algorithm}
\label{alg:kbin}
\begin{algorithmic}[1]
\Function{GetNegation}{$c, \text{KB}, V, N$}
\State $E \gets \text{NER}(c)$
\State $\bar{C} \gets []$
\For{$e_{j} \text{ in } E$}
    \State $u_{j} \gets \text{LINK}(e_{j})$
    \State $R \gets \text{KB.siblings}(u_{j})$
    \State $\text{filter}(R, \text{KB.type}(u_{j}))$
    \State $dist \gets \text{cosdist}(V[u_{j}], V[R])$
    \For{$r \text{ in argsort}(dist)[:N]$}
        \State $A \gets \text{KB.aliases}(R[r])$ 
        \State $T \gets \text{replace}(c, e_{j}, a) \text{ for } a \text{ in } A$
        \State $\bar{C}\text{.add}(\text{rank\_perplexity}(T)[0])$
    \EndFor
\EndFor
\State\Return $\text{rank\_contradiction}(c, \bar{C})[0]$ 
\EndFunction
\end{algorithmic}
\end{algorithm}

\cgentity and \cgbart only produce claims which are entailed by the source sentence. Additionally, we are interested in producing claim variants which are directly refuted by the original sentence, as 
these negations are needed when building fact checking datasets and for training fact checking models. Work in \citet{DBLP:conf/emnlp/WaddenLLWZCH20} created these negations manually, and some work has begun to explore automatically generating these negations for scientific claims~\cite{DBLP:conf/acl/SaakyanCM20}.
To this end, we leverage the availability of 
large curated biomedical knowledge bases
to develop a principled approach to claim variant generation. In particular, we use the UMLS metathesaurus~\cite{DBLP:journals/nar/Bodenreider04}, which unifies hundreds of different ontologies in biomedicine, 
as a source of term replacements for negations.

We provide an overview of the KBIN algorithm in Algorithm \autoref{alg:kbin} and \autoref{fig:kbin_method}. KBIN works by first performing NER on an input claim $c$, obtaining entities $\{e_1,\dots,e_{n}\} \in E$. For each entity $e_{j}$ in $E$,  we link the entity to its unique concept $u_{j}$ in UMLS using the scispaCy entity linker. If the entity is linked, we select all concepts which are siblings to $u_{j}$ in the concept hierarchy, and which have the same semantic type (e.g. ``Clinical Drug''). We rank all selected concepts by their cosine distance to the entity concept using pre-trained UMLS concept vectors, retaining the top 20 closest concepts. For this, we use \texttt{cui2vec}~\cite{DBLP:conf/psb/BeamKSFWPSCK20}, which contains pre-trained concept vectors for 108,477 concepts from UMLS trained on medical documents from diverse sources.

For each of the related concepts, we generate candidate claim variants by replacing the entity text in the original claim with the canonical name and aliases of the related concept from UMLS. We rank all replacement sentences by their perplexity using a pre-trained GPT-2 model~\cite{radford2019language},
keeping the sentence with least perplexity for each replacement.
Finally, from among these most fluent sentences, we select the replacement which maximizes the NLI prediction of \textit{contradiction} with the original claim.
For this, we use a RoBERTa model~\cite{DBLP:journals/corr/abs-1907-11692} pre-trained on MNLI~\cite{DBLP:conf/naacl/WilliamsNB18}. %

\subsection{Experiments}
We investigate three primary research questions:
\begin{itemize}[noitemsep,leftmargin=9.5mm]
    \item[\textbf{RQ1}]{Do automatically generated claims enable zero-shot scientific fact checking?}
    \item[\textbf{RQ2}]{What is the percentage of high-quality claims generated using our methods?}
    \item[\textbf{RQ3}]{How does KBIN compare with previous work for claim negation in terms of generating contradictions?}
\end{itemize}
For \textbf{RQ1}, we use \cgentity and \cgbart generated claims to train a fact checking model, evaluating on the SciFact dataset~\cite{DBLP:conf/emnlp/WaddenLLWZCH20} and comparing to relevant baselines. To answer \textbf{RQ2} and \textbf{RQ3}, we design annotation criteria and perform manual evaluations with a group of expert annotators (details in \S\ref{sec:quality_eval}). %

\subsubsection{RQ1: Fact Checking Performance}
\label{sec:factcheck}

\paragraph{SciFact Task} The SciFact fact verification task consists of: given a claim $c$ and a corpus of scientific abstracts $D$, retrieve evidence abstracts from $D$, predict if the claim is \textit{supported} or \textit{refuted} by those documents or if there is \textit{not enough information (NEI)} to make a prediction, and optionally determine what the rationale sentences are that explain the prediction. Here we focus on the oracle abstract setting of the task, in which gold abstracts are provided to the model and there is no retrieval component. This setup exists in the scientific fact checking literature~\cite{DBLP:conf/acl/SaakyanCM20}, and allows us to focus on one component of the fact checking pipeline for evaluating the impacts of claim generation.

\paragraph{Creating Training Data for the Zero-shot Setting} We require a set of claim-abstract pairs for training where the abstract either supports, refutes, or does not provide evidence for the given claim.
We exploit citation relationships to generate claims paired with potential evidence, using citances from the CiteWorth dataset~\cite{DBLP:conf/acl/WrightA21} as source citances for generation.
\textit{Supports} claims are produced
by directly pairing a generated claim with the abstracts of documents cited by the source citance. For \textit{refutes} claims, we negate a generated claim using \method 
and pair it with the same abstract. 
For claims labelled \textit{NEI}, we pair the generated claim or negated claim with the abstract of the source document of the citance; the source document is related to the claim but presumably does not directly support or refute the claim given the need for a citation.

\paragraph{Experimental Setup} In our experimental setup, we use LongChecker~\cite{DBLP:journals/corr/abs-2112-01640}, a Longformer~\cite{DBLP:journals/corr/abs-2004-05150} model adapted for scientific fact checking. 
The model forms its input by concatenating a claim with its evidence abstract, inserting separator tokens between sentences, and uses a classification head to predict the veracity label from the representation of the \texttt{[CLS]} token.

We explore several different setups for our training data. As a baseline, we experiment with pre-training only on FEVER claims~\cite{DBLP:conf/naacl/ThorneVCM18}, which are general domain fact checking data based on Wikipedia. We also include an experiment where we manually tune a threshold for the prediction of \textit{NEI} on the SciFact training data, as we saw that the model tends to overpredict this label without any fine-tuning on in-domain data. We also provide an upper bound on performance by fine-tuning on the in-domain train split of SciFact. Finally, we experiment with both \cgentity and \cgbart as sources of training data generated from CiteWorth citances, pairing both with \method for negations. We note that though \cgbart requires manually re-written claims as training data for generating \textit{supports} claims, it does not use any claims paired with evidence manually labelled for veracity, thus making it zero-shot for the SciFact fact-checking task. In all cases we test on the SciFact dev split.
Hyperparameter information, including number of training instances, is given in \S\ref{sec:hyperparams}, and code and data will be released upon paper acceptance. In all cases, results are reported as macro-F1. %

\paragraph{Results} Our results on SciFact are given in \autoref{tab:scifact_eval}. With an upper bound of 77.70 F1, we see that a model fine-tuned on automatically generated claims is able to achieve within 90\% of the performance of a model trained on in-domain manually written claims. This is also invariant to the method used to generate claims, as both \cgentity and \cgbart produce similar results. Additionally, both methods provide significant gains over pre-training on FEVER only, especially when no threshold on \textit{NEI} claims is used but also when re-calibrating the model to predict \textit{NEI} less often.

\begin{table}%
    \def\arraystretch{1.2}
    \centering
    \fontsize{10}{10}\selectfont
    \begin{tabular}{l c c c}
    \toprule %
    Method & P & R & F1 \\
    \midrule %
       FEVER only & $86.21$& $11.96$& $21.01$\\
       FEVER + thresh & $69.15$& $66.51$& $67.80$\\
       SciFact (Upper Bound) & $77.88$& $77.51$& $77.70$\\
       \midrule
       \cgentity & $\mathbf{72.86}$& $69.38$& $\mathbf{71.08}$ \\
       \cgbart & $64.09$& $\mathbf{79.43}$& $70.94$ \\
    \bottomrule %

    \end{tabular}
    \caption{Results for veracity prediction on the SciFact dataset using different sources of training data.} %
    \label{tab:scifact_eval}
\end{table}

\subsubsection{RQ2: Claim Quality Evaluation}
\label{sec:quality_eval}
\begin{table*}[t]
    \def\arraystretch{1.3}
    \centering
    \fontsize{10}{10}\selectfont
    \begin{tabular}{l p{12cm}}
    \toprule %
    Metric & Labels\\
    \midrule %
    \multirow{3}{*}{Fluency} & 3 - The claim contains no grammatical errors and its meaning can be understood \\ 
    & 2 - The claim contains some grammatical errors but is still understandable \\ 
    & 1- The claim contains many grammatical errors and cannot be understood\\
    \hline
    \multirow{3}{*}{De-Contextualized} & 1 - The claim is interpretable on its own and requires no context; the addition of the original context does not alter the meaning of the claim \\ 
    & 0 - The claim cannot be interpreted in a meaningful way without the original context\\
    \hline
    \multirow{2}{*}{Atomicity} & 1 - The claim is about a single entity/process (atomic) \\
    & 0 - The claim is non-atomic and can be broken down into multiple claims \\
    \hline
    \multirow{9}{*}{Faithfulness} & 5 - The claim is correct and fully supported and complete with respect to the original sentence and context\\ 
    & 4 - The claim is correct with respect to the original sentence and context but leaves out information from the original sentence and context\\ 
    & 3 - The claim is related to the original sentence and does not contain incorrect information but is not explicitly stated in the original sentence\\ 
    & 2 - The claim contains explicitly incorrect information relative to the original sentence and context\\ 
    & 1 - The claim has nothing to do with the original sentence\\
    \bottomrule %

    \end{tabular}
    \caption{Claim quality evaluation metrics and their possible values}
    \label{tab:quality_metrics}
\end{table*}
Next, we explore if there are differences between our methods in terms of claim quality and the percentage of valid claims. For this, we ask three expert annotators to manually assess generated claims along a number of quality criteria. One annotator has undergraduate training in the life sciences and graduate training in computer science; the other two annotators have undergraduate training in the life sciences and materials science respectively. We define a set of criteria for evaluation, given in \autoref{tab:quality_metrics}.
These criteria are inspired by the AIDA (Atomic, Independent, Declarative, and Absolute) framework for scientific claims introduced in \citet{DBLP:conf/esws/KuhnBNK13}. They are also based on similar human evaluation criteria used to assess generation quality for related tasks~\cite{DBLP:journals/corr/abs-2008-12009}. We develop an initial set of guidelines for the annotators and conduct two rounds of pilot annotations to improve instructions and increase agreement. 
For the final evaluation, we generate claims on
a set of 100 citances sampled from the CiteWorth dataset~\cite{DBLP:conf/acl/WrightA21}, which contains citations in context for over 1M citances spanning 10 domains.

\begin{table*}[t]
    \def\arraystretch{1.2}
    \centering
    \fontsize{10}{10}\selectfont
    \begin{tabular}{l c c c c c c c}
    \toprule %
    Method & Fluency & De-Con. (\%) & Atomic (\%) & Faithfulness & \# Gen & \# Accept & P\\
    \midrule %
       \cgentity & $2.51$& $55.63$& $\mathbf{85.28}$& $3.54$& $893$& $111$& $12.43$\\
       \cgbart & $\mathbf{2.74}$& $\mathbf{84.35}$& $80.65$& $\mathbf{4.15}$& $156$& $69$& $\mathbf{44.23}$ \\
       \midrule
       $\alpha$ (236 claims)& 82.74& 64.53& 58.71& 53.01& - & - & -\\
    \bottomrule %

    \end{tabular}
    \caption{Average annotation score, agreement, and claim yield for each category. De-contextualized is only annotated if fluency > 1; atomicity and faithfulness are only annotated if fluency > 1 and de-contextualized == 1. \# Gen are the total claims generated by the method, and \# Accept are the number of acceptable claims generated.}
    \label{tab:manual_eval}
\end{table*}

\begin{table*}[t]
    \def\arraystretch{1.3}
    \centering
    \fontsize{10}{10}\selectfont
    \begin{tabular}{p{8.5cm} p{4.9cm} c}
    \toprule %
    Citance & Generated & Fl,D,A,Fa\\
    \midrule %
    Due to its geographic position and geological history, the island of Sardinia is characterized by a remarkable richness of endemic species and represents one of the most prominent biodiversity hotspots in the Mediterranean basin.& The island of Sardinia is characterized by a remarkable richness of endemic species.& 3,1,1,5\\
    \midrule
    Frequently reported symptom-eliciting chemicals and environmental agents include fragranted products, motor-vehicle exhaust fumes, cleaning agents, freshly printed papers or magazines, and smoke from wood burners. &Frequently reported symptom-eliciting chemicals and environmental agents are fragranted products.& 3,1,1,5\\
    \midrule
    The herbicide inhibits EPSPS (5-enolpyruvylshikimate-3-phosphate synthase) in the shikimate pathway, which has a key role in the biosynthesis of aromatic amino acids and is required for survival of the plant.& The herbicide inhibits EPSPS in the shikimate pathway. & 3,1,1,5\\
    \midrule
    Experimental models of OA, such as the intra-articular injection of monosodium acetate (MIA), are associated with joint pathology and pain behaviour comparable to clinical OA. &OA is associated with joint pathology and pain behaviour comparable to clinical OA.& 3,1,0,4\\
    \bottomrule %

    \end{tabular}
    \caption{Sample generated claims with their ratings for (Fl)uency, (D)e-Contextualized, (A)tomicity, (Fa)ithfulness}
    \label{tab:claim_examples}
\end{table*}

We limit the citances to those from papers in biology and medicine to match the domain of 
SciFact.
Annotator agreement is measured as Krippendorff's $\alpha$ \cite{krippendorff2011computing} on 236 claims for each category except fluency, where we measure the percentage of claims where all annotators agree.\footnote{Fluency agreement is measured in terms of agreement percentage as most ratings are the same (3), thus any disagreements have an oversized influence on $\alpha$.} The annotators then assess 1,049 total claims (including the 236 shared claims). Each annotator rates all criteria for an individual claim, starting with fluency, then de-contextualized, then atomicity, then faithfulness. We are mainly interested in claim quality and yield, so annotators only annotate ``de-contextualized'' if the claim is legible (fluency > 1), and only annotate ``atomicity'' and ``faithfulness'' if the claim is also de-contextualized (so one is able to discern meaning from the claim). This results in the following rules for acceptable claims based on the definitions for the labels in each category: Fluency $>$ 1 \texttt{AND} De-Contextualized $=$ 1 \texttt{AND} Atomicity $=$ 1 \texttt{AND} Faithfulness $>$ 3. An acceptable claim is thus legible, meaningful, represents a single aspect of a scientific entity or process, and accurately reflects the information presented in the original citance.

The results of claim quality annotation are given in \autoref{tab:manual_eval}. Note that these are on claims generated by \cgentity and \cgbart (see examples in \autoref{tab:claim_examples}), and thus are only \textit{supports} claims.
We first note that inter-annotator agreement is very high for fluency and moderate across all other criteria. Generated claims are quite fluent across methods, with a small minority of instances being illegible. Unsurprisingly, \cgbart improves over \cgentity across all categories except for atomicity. This intuitively makes sense as \cgentity directly produces claims which are about a single entity. \cgentity yields a higher number of claims per citance as it generates one claim for every entity in the sentence, but the precision of acceptable claims is much lower than that of \cgbart. Thus, there is a tradeoff between the two methods between the number of claims generated and their acceptability. While higher yield could lead to higher coverage of claims in the original text, this study is left to future work.

Next, we examine the similarity between generated claims 
and manually written claims from SciFact. We generate claims for each source citance $s_{i}$ in the SciFact dev split, and calculate the ROUGE score~\cite{lin2004rouge} between each generated claim $c^{(i)}_{j}$ and each manually written claim $d^{(i)}_{k}$. From this, we take an average of the max ROUGE score for each generated claim. Formally, given $|C|$ claims we calculate:
\begin{equation*}
    score = \frac{1}{|C|}\sum_{i}\sum_{j} \max_{k}\text{ROUGE}(c^{(i)}_{j}, d^{(i)}_{k})
\end{equation*}
Our evaluation results are given in \autoref{tab:rouge_eval}.
\begin{table}%
    \def\arraystretch{1.2}
    \centering
    \fontsize{10}{10}\selectfont
    \begin{tabular}{l c c c}
    \toprule %
    Method & R-1 & R-2 & R-L \\
    \midrule %
       Entity & $47.12$& $27.63$& $42.30$ \\
       BART & $\mathbf{56.58}$ & $\mathbf{40.12}$ & $\mathbf{53.38}$ \\
    \bottomrule %

    \end{tabular}
    \caption{ROUGE score between generated and manually written reference claims in the SciFact dataset} %
    \label{tab:rouge_eval}
\end{table}
Both methods produce claims which have high overlap with the reference claims, though claims generated directly using BART are significantly closer to the reference claims than those generated using \cgentity. Finally, we note the these scores are in the range of state-of-the-art models used for paraphrase generation, establishing a solid baseline for this task~\cite{DBLP:conf/emnlp/ZhouB21}.

\subsubsection{RQ3: Negation Evaluation}
\label{sec:negeval}
\begin{table}[t]
    \def\arraystretch{1.3}
    \centering
    \fontsize{10}{10}\selectfont
    \begin{tabular}{p{5cm} p{0.1cm} p{3cm} p{6.2cm}}
    \toprule %
    Original Claim & & Method & Generated Negation\\
    \midrule
    Tonic signaling from the SCFV prevents constitutive stimulation.
        & & Entity replace & Tonic signaling from the SCFV {\color{red}under care of respiratory physician (finding)} constitutive stimulation. \\ \cline{3-4}
        & & \citet{DBLP:conf/acl/SaakyanCM20} & Tonic signaling from the {\color{red}inflammatory stimulation.} \\  \cline{3-4}
        & & \method & Tonic signaling from the SCFV {\color{red}accelerates} constitutive stimulation. \\
    \midrule
    Activation of the RAC1 homolog CED-10 kills viable cells in SRGP-1 mutant \textit{Caenorhabditis Elegans}.
        & & Entity replace & Activation of the {\color{red}LASS4} homolog CED-10 kills viable cells in SRGP-1 mutant \textit{Caenorhabditis Elegans}. \\ \cline{3-4}
        & & \citet{DBLP:conf/acl/SaakyanCM20} & Activation of the RAC1 homolog CED-10 kills viable cells in SRGP-1 {\color{red}\textit{Helicobacter Elegans}}. \\ \cline{3-4}
        & & \method & Activation of the RAC1 homolog CED-10 {\color{red}mediate} viable cells in SRGP-1 mutant \textit{Caenorhabditis Elegans}. \\
    \bottomrule %

    \end{tabular}
    \caption{Example negations generated using three methods. Span replacements are highlighted in {\color{red}{red}}. In addition to replacing noun phrases, \method also has the ability to replace verb phrases as shown in these examples.}
    \label{tab:claim_negations}
\end{table}
\begin{table}[t]
    \def\arraystretch{1.2}
    \centering
    \fontsize{10}{10}\selectfont
    \begin{tabular}{l c c c c}
    \toprule %
     & & \multicolumn{3}{c}{Entailment} \\
    Method & Fluency & 3 & 2 & 1 \\
    \midrule %
      Entity replace & 83& 1& 81& 1\\
      \citet{DBLP:conf/acl/SaakyanCM20} & 83& 10& 64& 9\\
      \method & \textbf{93}& \textbf{15}& 75& 3\\
    \bottomrule %

    \end{tabular}
    \caption{Results for manual annotation of claim negations on 100 negations for each method. Fluent claims received annotations other than ``SKIP''.} %
    \label{tab:negation_eval}
\end{table}
Finally, we perform a manual evaluation to compare \method against other methods of negation generation. Annotators evaluate negations based on Fluency and Entailment. 
We adopt the definitions used to annotate the SNLI corpus~\cite{DBLP:conf/emnlp/BowmanAPM15}, in which the annotator is given an original claim (premise) and a generated negation (hypothesis) and asked to select from among the following options, including a SKIP option for Fluency:
\begin{itemize}[noitemsep,leftmargin=10mm,itemindent=-15pt]
    \item[]\textbf{3}\quad The hypothesis is DEFINITELY FALSE given the premise
    \item[]\textbf{2}\quad The hypothesis MIGHT BE TRUE given the premise
    \item[]\textbf{1}\quad The hypothesis is DEFINITELY TRUE given the premise
    \item[]\textbf{SKIP}~ The hypothesis contains a lot of grammatical errors and cannot be understood
\end{itemize}

We compare \method to two baselines.
The first baseline replaces a single entity in the claim with a random entity of the same type, similar to the method in \citet{DBLP:conf/acl/PanCXKW20}. The second is the proposed negation generation method in \citet{DBLP:conf/acl/SaakyanCM20}. The method is based on extracting keywords using YAKE~\cite{DBLP:journals/isci/CamposMPJNJ20} (an unsupervised method based on statistical text features), replacing those keywords using text infilling with a pre-trained language model, and selecting the replacement with the highest contradiction score using a model pre-trained for NLI. We generate negations for 100 claims using all three methods. For annotation, generated negations from all three methods are aggregated and the order of negation method randomized for each of the 100 claims.

Example negations generated by all three methods are given in \autoref{tab:claim_negations} and annotation results for fluency and entailment are given in \autoref{tab:negation_eval}. First, \method produces more fluent claims than both baselines.
Additionally, \method produces more convincing negations on average than both baselines. We observe that the most common operation performed by all three methods is to replace a noun phrase. \method has the benefit of being able to replace many entity types 
corresponding to concepts found in UMLS, 
which also include verb phrases that encode relations. Finally, \method improves over the baseline from \citet{DBLP:conf/acl/SaakyanCM20} by producing fewer claims which are directly entailed by the source claim, i.e., that maintain the original meaning and do not negate the original claim.

\subsubsection{Further Analysis}
\begin{figure}[t]
  
  \centering
    \includegraphics[width=\linewidth]{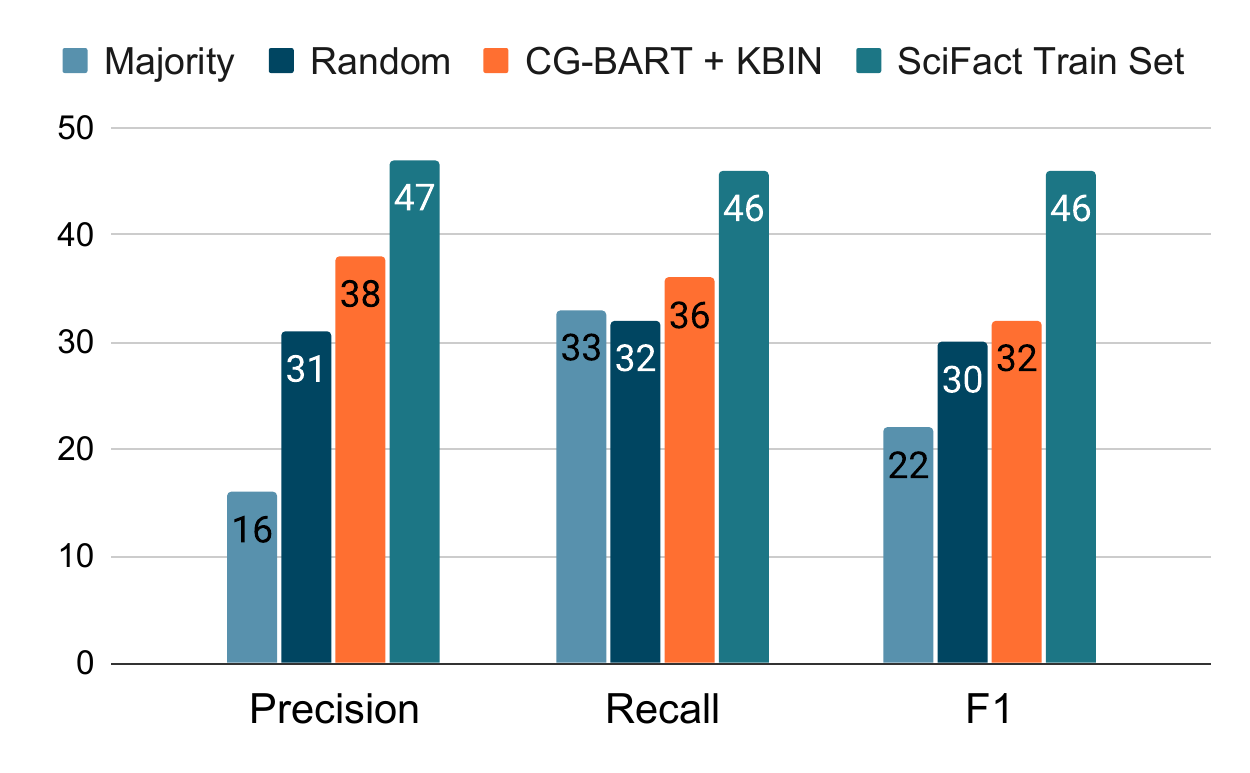}
    \caption{Fact checking performance of models trained only on claims (i.e. no evidence). Training on our generated claims result in performance closer to random (indicating fewer data artifacts) than training on the original SciFact claims.}
    \label{fig:claim_only_baseline}
\end{figure}

To give further insight into the quality of claims generated using our methods, we perform an experiment where we train and test models for scientific fact checking using claims only. This ``claim-only'' experiment helps us assess whether the negation process introduces data artifacts that can be leveraged by the model to predict veracity. We present results from training on claims generated using \cgbart and \method, compared against training on the original SciFact training data (which has manually written negations), along with random and majority baselines, in \autoref{fig:claim_only_baseline}. 

We observe that there are likely some dataset artifacts in the original SciFact claims that lead to model performance well above the majority  and random baselines.\footnote{It is difficult to fully separate the contributions of data artifacts and model performance in this setting, i.e., there is no situation which guarantees *no* undesirable data artifacts. Performance ought to be better than a random baseline in this theoretical setting, due to the pretrained language model likely having had some exposure to the content of the claims during pretraining.} This phenomenon has been observed in general domain natural language inference datasets as well~\cite{DBLP:conf/starsem/PoliakNHRD18}. Training on claims generated using our methods results in performance that is much more proximal to random performance on the SciFact dev set, indicating that the label-associated bias in the original training data is not present and a possible domain shift between the original SciFact claims and our generated claims. This can further explain some of the performance gap we observe between zero-shot fact-checking and the upper bound of training on manually labeled training data (\autoref{tab:scifact_eval}).

\subsection{Related Work}

\paragraph{Scientific Fact Checking} Our work follows a line of recent literature on scientific fact checking~\cite{DBLP:conf/emnlp/WaddenLLWZCH20}. The goal of this task is to determine the veracity of claims related to scientific topics by retrieving appropriate documents from scientific literature, finding evidentiary sentences from those documents, and determining whether claims are supported, refuted, or there is not enough evidence to make a judgement. The task closely resembles the task of general domain fact-checking~\cite{DBLP:conf/naacl/ThorneVCM18,DBLP:conf/emnlp/AugensteinLWLHH19}. Well-performing systems on this task use large language models to perform neural document retrieval~\cite{DBLP:journals/corr/abs-2010-11930} or multi-task learning of rationale prediction and stance prediction~\cite{DBLP:conf/aaai/LiBP21, DBLP:journals/corr/abs-2112-01640}. 
Recent work on general domain fact checking has also introduced methods for adversarial generation of claims which are particularly difficult to fact-check~\cite{thorne-etal-2019-fever2,atanasova-etal-2020-generating}, and for performing the task without any labeled data~\cite{DBLP:conf/acl/PanCXKW20}. Our proposed methods extend zero-shot fact checking to the scientific domain, demonstrating that one can achieve 90\% of the inference performance of state-of-the-art systems without domain-specific labeled data.

\paragraph{Generating Training Data} Our work is also related to methods for the automatic generation of training data.
Generation of synthetic data has been used for multiple tasks, for example question answering~\cite{DBLP:conf/emnlp/DuanTCZ17,DBLP:conf/emnlp/RiabiSKSSS21}, knowledge-base completion~\cite{DBLP:conf/emnlp/SafaviZK21}, and fact-checking~\cite{DBLP:conf/acl/PanCXKW20}. Most similar to our setting, the COVID-Fact dataset~\cite{DBLP:conf/acl/SaakyanCM20} contains claims related to COVID-19 crawled from Reddit, and is constructed semi-automatically. Claims which are supported by evidence are extracted from Reddit and verified by human annotators, while negations of these claims are generated automatically via masked language model infilling. \method improves upon the negation method proposed in this work by leveraging in-domain structured knowledge via UMLS.

\subsection{Conclusion}
In this work, we propose the task of scientific claim generation, presenting \cgbart, \cgentity, and \method to perform the task.
We demonstrate that generated claims can be used to train a model for zero-shot scientific fact checking and obtain within 90\% of the performance of a model trained on human-written claims. Through a rigorous user study we demonstrate that \cgbart produces higher quality claims than \cgentity, and that \method produces more fluent and more convincing negations than previous work.
Work remains to improve claim generation quality and assess the impacts of generated claims in other domains of science, as well as how generated claims can be used in the evidence retrieval component of fact checking systems.
We hope that our methods will be used to facilitate future work 
by enabling faster creation of training datasets and improving the performance of models on the timely and important task of scientific fact checking.

\section*{Acknowledgements}
$\begin{array}{l}\includegraphics[width=1cm]{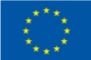} \end{array}$ This project is supported in part by the European Union's Horizon 2020 research and innovation programme under the Marie Sk\l{}odowska-Curie grant agreement No 801199, and by the United States National Science Foundation Grant OIA-2033558. 
We thank Doug Downey, Hannaneh Hajishirzi, the reviewers, and members of the Semantic Scholar research team for their valuable feedback.

\section*{Ethical Considerations}
Automated scientific fact checking has great potential value to the scientific community, as well as for addressing phenomenon such as the propagation of scientific misinformation. Our aim in releasing models for scientific claim generation is to improve the generalizability of science fact checking systems in domains with less training resources. When training our fact checking models with generated or synthetic data, there are questions regarding the veracity of the generated data and whether a model trained on inferred labels could produce trustworthy judgments. We hope that by introducing this task and models, we will enable the community to study such questions, while contributing to data curation in a domain in which such curation would normally require significant manual efforts and cost.

\newpage

\section{Semi-Supervised Exaggeration Detection of Health Science Press Releases}
\label{paper:exaggeration}

\subsection{Introduction}
\begin{figure}[t]
  
  \centering
    \includegraphics[width=\linewidth]{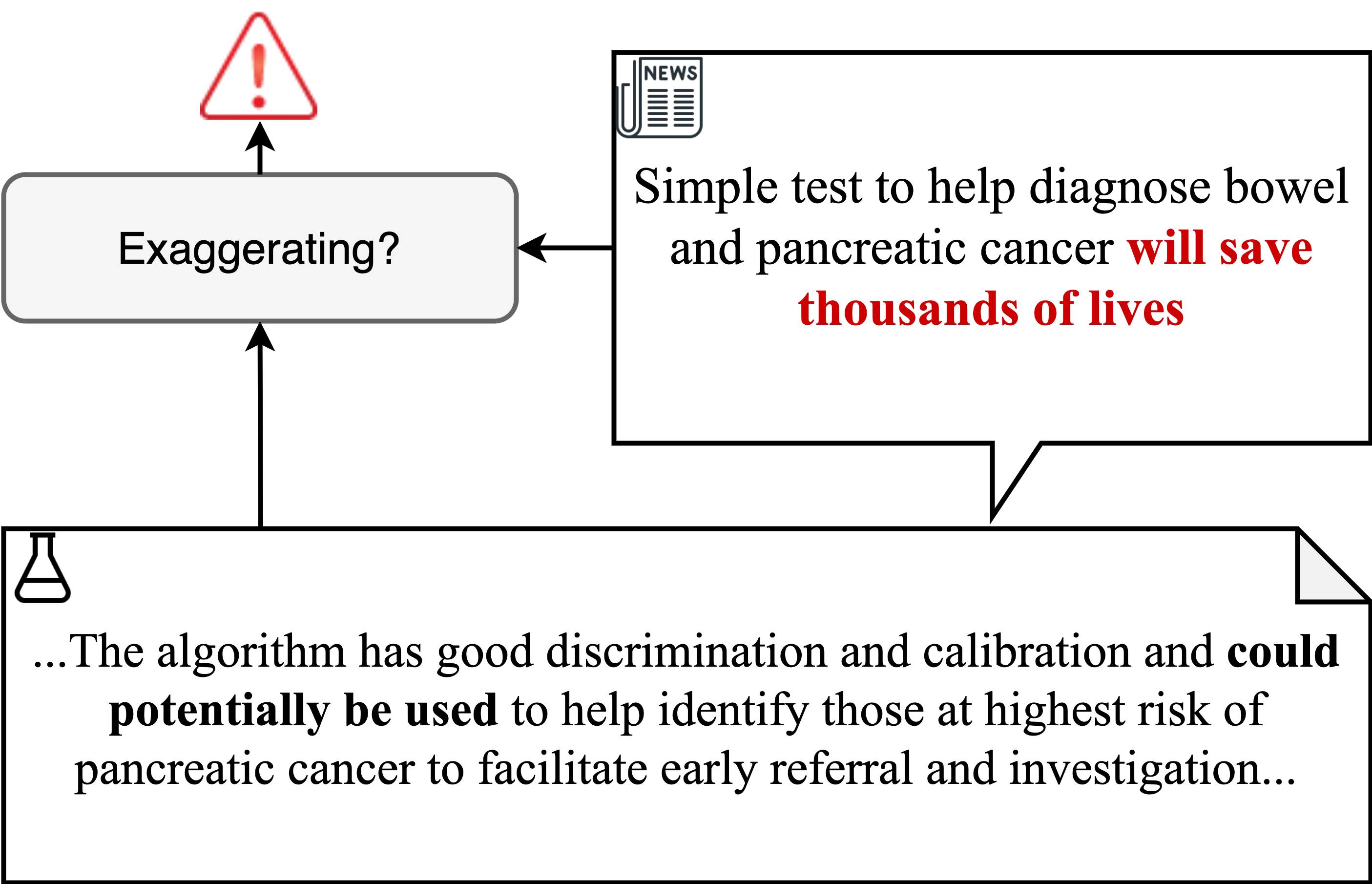}
    \caption{Scientific exaggeration detection is the problem of identifying when a news article reporting on a scientific finding has exaggerated the claims made in the original paper. In this work, we are concerned with predicting exaggeration of the main finding of a scientific abstract as reported by a press release.}
    \label{fig:exaggeration-detection}
\end{figure}
Factual and honest science communication is important for maintaining public trust in science~\cite{nelkin1987selling,moore2006bad}, and the ``dominant link between academia and the media'' are press releases about scientific articles~\cite{sumner2014association}. However, multiple studies have demonstrated that press releases have a significant tendency to sensationalize their associated scientific articles \cite{sumner2014association,bratton2019association,woloshin2009press,woloshin2002press}. 
In this paper, we explore how natural language processing can help identify exaggerations of scientific papers in press releases.

While \citet{sumner2014association} and \citet{bratton2019association} performed manual analyses to understand the prevalence of exaggeration in press releases of scientific papers from a variety of sources, recent work has attempted to expand this using methods from NLP \cite{yu2019detecting,yu2020measuring,DBLP:conf/emnlp/LiZY17}. These works focus on the problem of automatically detecting the difference in the strength of causal claims made in scientific articles and press releases. They accomplish this by first building datasets of main claims taken from PubMed abstracts and (unrelated) press releases from EurekAlert\footnote{\url{https://www.eurekalert.org/}} labeled for their strength. With this, they train machine learning models to predict claim strength, and analyze unlabelled data using these models. This marks an important first step toward the goal of automatically identifying exaggerated scientific claims in science reporting.

However, existing work has only partially attempted to address this task using NLP. Particularly, there exists no standard benchmark data for the exaggeration detection task with \textbf{paired} press releases and abstracts i.e. where the data consist of tuples of the form (press release, abstract) and the press release is written about the paired scientific paper. Collecting paired data labeled for exaggeration is critical for understanding how well any solution performs on the task, but is challenging and expensive as it requires domain expertise~\cite{sumner2014association}. The focus of this work is then to curate a standard set of benchmark data for the task of scientific exaggeration detection, provide a more realistic task formulation of the problem, and develop methods effective for solving it using limited labeled data. To this end, we present~\METHOD, a multi-task implementation of Pattern Exploiting Training (PET, \citet{schick2020exploiting,schick2020small}) for detecting exaggeration in health science press releases. We test our method by curating a benchmark test set of data from the expert annotated data of \citet{sumner2014association} and \citet{bratton2019association}, which we release to help advance research on scientific exaggeration detection.

\paragraph{Contributions}
In sum, we introduce:
\vspace{-0.3cm}
\begin{itemize}[noitemsep]
    \item A new, more realistic task formulation for scientific exaggeration detection.
    \item A curated set of benchmark data for testing methods for scientific exaggeration detection consisting of \NTEST~press release/abstract pairs.
    \item \METHOD, a multi-task extension of PET which beats strong baselines on scientific exaggeration detection.
\end{itemize}

\subsection{Problem Formulation}
We first provide a formal definition of the problem of scientific exaggeration detection, which guides the approach described in \S\ref{sec:approach}. We start with a set of document pairs $\{(t,s) \in \mathcal{D}\}$, where $s$ is a source document (e.g. a scientific paper abstract) and $t$ is a document written about the source document $s$ (e.g. a press release for the paper). The goal is to predict a label $l \in \{0,1,2\}$ for a given document pair $(t,s)$, where $0$ implies the target document \textit{undersells} source document, $1$ implies the target document accurately reflects the source document, and $2$ implies the target document \textit{exaggerates} the source document. 

Two realizations of this formulation are investigated in this work. The first (defined as \textbf{T1}) is an \textit{inference} task consisting of labeled document pairs used to learn to predict $l$ directly. In other words, we are given training data of the form $(t, s, l)$ and can directly train a model to predict $l$ from both $t$ and $s$. The second (defined as \textbf{T2}) is as a \textit{classification} task consisting of a training set of documents $d \in \mathcal{D}'$ from \textbf{both} the source and the target domain, and a classifier is trained to predict the \textit{claim strength} $l'$ of sentences from these documents. In other words, we don't require \textbf{paired} documents $(t,s)$ at train time. At test time, these classifiers are then applied to document pairs $(t,s)$ and the predicted claim strengths $(l'_{s}, l'_{t})$ are compared to get the final label $l$. Previous work has used this formulation to estimate the prevalence of \textit{correlation to causation} exaggeration in press releases~\cite{yu2020measuring}, but have not evaluated this on paired labeled instances.

 Following previous work~\cite{yu2020measuring}, we simplify the problem by focusing on detecting when the \textit{main finding} of a paper is exaggerated. The first step is then to identify the main finding from $s$, and the sentence describing the main finding in $s$ from $t$. In our semi-supervised approach, we do this as an intermediate step to acquire unlabeled data, but for all labeled training and test data, we assume the sentences are already identified and evaluate on the sentence-level exaggeration detection task.

\subsection{Approach}
\label{sec:approach}
\begin{figure}[t]
  
  \centering
    \includegraphics[width=0.85\linewidth]{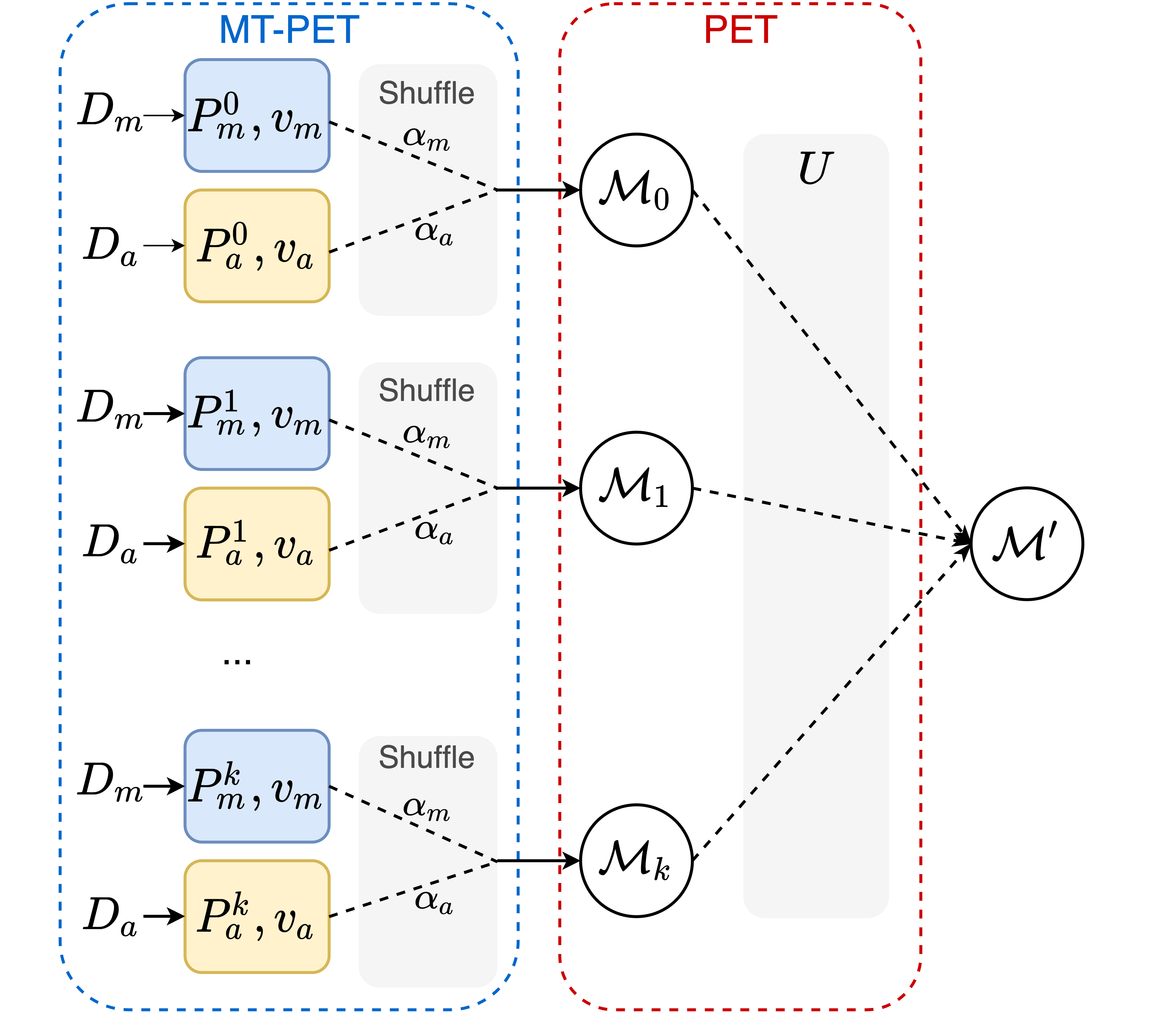}
    \caption{MT-PET design. We define pairs of complementary pattern-verbalizer pairs for a main task and auxiliary task. These PVPs are then used to train PET on data from both tasks.}
    \label{fig:mt-pet}
\end{figure}
One of the primary challenges for scientific exaggeration detection is a lack of labeled training data. Given this, we develop a semi-supervised approach for few-shot exaggeration detection based on pattern exploiting training (PET, \citet{schick2020exploiting,schick2020small}). Our method, multi-task PET (\METHOD, see \autoref{fig:mt-pet}), improves on PET by using multiple complementary cloze-style QA tasks derived from different source tasks during training. %
We first describe PET, followed by \METHOD.

\subsubsection{Pattern Exploiting Training (PET)}
PET~\cite{schick2020exploiting} uses the masked language modeling objective of pretrained language models to transform a task into one or more cloze-style question answering tasks. %
The two primary components of PET are \textit{patterns} and \textit{verbalizers}. \textit{Patterns} are cloze-style sentences which mask a single token e.g. in sentiment classification with the sentence ``We liked the dinner'' a possible pattern is: ``We liked the dinner. It was \texttt{[MASK]}.'' \textit{Verbalizers} are single tokens which capture the meaning of the task's labels in natural language, and which the model should predict to fill in the masked slots in the provided patterns (e.g. in the sentiment analysis example, the verbalizer could be \texttt{Good}). 

Given a set of \textit{pattern-verbalizer pairs (PVPs)}, an ensemble of models is trained on a small labeled seed dataset to predict the appropriate verbalizations of the labels in the masked slots. These models are then applied on unlabeled data, and the raw logits are combined as a weighted average to provide soft-labels for the unlabeled data. A final classifier is then trained on the soft labeled data using a distillation loss based on KL-divergence.

\subsubsection{Notation}
We adopt the notation in the original PET paper~\cite{schick2020exploiting} to describe \METHOD. In this, we have a masked language model $\mathcal{M}$ with a vocabulary $V$ and mask token \texttt{[MASK]} $\in V$. A pattern is defined as a function $P(x)$ which transforms a sequence of input sentences $\mathbf{x} = (s_{0},...,s_{k-1}), s_{i} \in V^{*}$ to a phrase or sentence which contains exactly one mask token. Verbalizers $v(x)$ map a label in the task's label space $\mathcal{L}$ to a set of tokens in the vocabulary $V$ which $\mathcal{M}$ is trained to predict.

For a given sample $\mathbf{z} \in V^{*}$ containing exactly one mask token and $w \in V$ corresponding to a word in the language model's vocabulary, $M(w|\mathbf{z})$ is defined as the unnormalized score that the language model gives to word $w$ at the masked position in $\mathbf{z}$. The score for a particular label as given in \citet{schick2020exploiting} is then
\begin{equation}
    s_{\mathbf{p}}(l|\mathbf{x}) = M(v(l) | P(\mathbf{x}))
\end{equation}
For a given sample, PET then assigns a score $s$ for each label based on all of the verbalizations of that label. When applied to unlabeled data, this produces soft labels from which a final model $\mathcal{M}'$ can be trained via distillation using KL-divergence.

\subsubsection{\METHOD}

In the original PET implementation, PVPs are defined for a single target task. \METHOD~extends this by allowing for auxiliary PVPs from related tasks, adding complementary cloze-style QA tasks during training. The motivation for the multi-task approach is two-fold: 1) complementary cloze-style tasks can potentially help the model to learn different aspects of the main task; in our case, the similar tasks of exaggeration detection and claim strength prediction; 2) data on related tasks can be utilized during training, which is important in situations where data for the main task is limited.

Concretely, we start with a main task $T_{m}$ with a small labeled dataset $(x_{m},y_{m}) \in D_{m}$, where $y_{m} \in \mathcal{L}_{m}$ is a label for the instance, as well as an auxiliary task $T_{a}$ with labeled data $(x_{a},y_{a}) \in D_{a}, y_{a} \in \mathcal{L}_{a}$. Each pattern $P_{m}^{i}(x)$ for the main task has a corresponding complementary pattern $P_{a}^{i}(x)$ for the auxiliary task. Additionally, the labels in $\mathcal{L}_{a}$ have their own verbalizers $v_{a}(x)$. Thus, with $k$ patterns, the full set of PVP tuples is given as 
\begin{equation*}
    \mathcal{P} = \{((P_{m}^{i}, v_{m}), (P_{a}^{i}, v_{a})) | 0 \le i < k\}
\end{equation*}
Finally, a large set of unlabeled data $U$ for the \textit{main task only} is available. \METHOD~then trains the ensemble of $k$ masked language models using the pairs defined for the main and auxiliary task. In other words, for each individual model both the main PVP $(P_{m},v_{m})$ and auxiliary PVP $(P_{a},v_{a})$ are used during training. 

For a given model $\mathcal{M}_{i}$ in the ensemble, on each batch we randomly select one task $T_{c}, c \in \{m,a\}$ on which to train. The PVP for that task is then selected as $(P_{c}^{i}, v_{c})$. Inputs $(x_{c}, y_{c})$ from that dataset are passed through the model, producing raw scores for each label in the task's label space.
\begin{equation}
    s_{\mathbf{p}_{c}^{i}}(\cdot|\mathbf{x}_{c}) = \{\mathcal{M}_{i}(v_{c}(l)|P_{c}^{i}(\mathbf{x}_{c})) | \forall~ l \in \mathcal{L}_{c}\}
\end{equation}
The loss is calculated as the cross-entropy between the task label $y_{c}$ and the softmax of the score $s$ normalized over the scores for all label verbalizations~\cite{schick2020exploiting}, weighted by a term $\alpha_{c}$.
\begin{equation}
    q_{\mathbf{p}_{c}^{i}} = \frac{e^{s_{\mathbf{p}_{c}^{i}}(\cdot|\mathbf{x}_{c})}}{\sum_{l\in\mathcal{L}_{c}}e^{s_{\mathbf{p}_{c}^{i}}(l|\mathbf{x}_{c})}}
\end{equation}
\begin{equation}
    L_{c} = \alpha_{c} * \frac{1}{N}\sum_{n} H(y_{c}^{(n)},q^{(n)}_{\mathbf{p}_{c}^{i}})
\end{equation}
where $N$ is the batch size, $n$ is a sample in the batch, $H$ is the cross-entropy, and $\alpha_{c}$ is a hyperparameter weight given to task $c$. 

\METHOD~then proceeds in the same fashion as standard PET. Different models are trained for each PVP tuple in $\mathcal{P}$, and each model produces raw scores $s_{\mathbf{p}_{m}^{i}}$ for all samples in the unlabeled data. The final score for a sample is then a weighted combination of the scores of individual models.
\begin{equation}
    s(l|\mathbf{x}_{u}^{j}) = \sum_{i}w_{i}*s_{\mathbf{p}_{m}^{i}}(l|\mathbf{x}_{u}^{j})
\end{equation}
where the weights $w_{i}$ are calculated as the accuracy of model $\mathcal{M}_{i}$ on the train set $D_{m}$ before training. The final classification model is then trained using KL-divergence between the predictions of the model and the scores $s$ as target logits.

\subsubsection{\METHOD~for Scientific Exaggeration}
\label{sec:mtpet_task_details}

\begin{table*}[t]
    \def\arraystretch{1.4}
        \centering
        \fontsize{10}{10}\selectfont
        \begin{tabular}{c c}
        \toprule %
        Name & Pattern\\
        \midrule %
    $P_{T_{1}}^{0}(x)$&Scientists claim $a$. || Reporters claim $b$.The reporters claims are \texttt{[MASK]} \\
    $P_{T_{1}}^{1}(x)$& Academic literature claims $a$. || Popular media claims $b$. The media claims are \texttt{[MASK]} \\
    \midrule
    $P_{T_{2}}^{0}(x)$& [Reporters|Scientists] say $a$. The claim strength is \texttt{[MASK]} \\
    $P_{T_{2}}^{1}(x)$&[Academic literature|Popular media] says $a$. The claim strength is \texttt{[MASK]}\\
    \bottomrule
    \end{tabular}
    \caption{Patterns for both \textbf{T1} (exaggeration detection) and \textbf{T2} (claim strength prediction)}
    \label{tab:patterns}
\end{table*}
\begin{table}[t]%
    \def\arraystretch{1.1}
    \centering
    \fontsize{10}{10}\selectfont
    \begin{tabular}{c c p{3.5cm}}
    \toprule %
    Pattern & Label & Verbalizers\\
    \midrule %
        & Downplays& preliminary, competing, uncertainties\\
       $P_{T_{1}}^{0}$ & Same& following, explicit\\
        & Exaggerates& mistaken, wrong, hollow, naive, false, lies\\
    \midrule
        & Downplays& hypothetical, theoretical, conditional\\
       $P_{T_{1}}^{1}$ & Same& identical\\
        & Exaggerates& mistaken, wrong, premature, fantasy, noisy, artifical\\
    \midrule
        \multirow{12}{*}{$P_{T_{2}}^{*}$}& NA& sufficient, enough, authentic, medium\\
       & Correlational& inferred, estimated, calculated, borderline, approximately, variable, roughly\\
        & Cond. Causal& cautious, premature, uncertain, conflicting, limited\\
        & Causal& touted, proven, replicated, promoted, distorted\\
    \bottomrule %

    \end{tabular}
    \caption{Verbalizers for PVPs from both \textbf{T1} and \textbf{T2}. Verbalizers are obtained using PETAL~\cite{DBLP:conf/coling/SchickSS20}, starting with the top 10 verbalizers per label and then manually filtering out words which do not make sense with the given labels.}
    \vspace{-4mm}
    \label{tab:verbalizers1}
\end{table}
We use \METHOD~to learn from data labeled for both of our formulations of the problem (\textbf{T1}, \textbf{T2}). In this, the first step is to define PVPs for exaggeration detection (\textbf{T1}) and claim strength prediction (\textbf{T2}).

To do this, we develop an initial set of PVPs and use PETAL~\cite{DBLP:conf/coling/SchickSS20} to automatically find verbalizers which adequately represent the labels for each task. We then update the patterns manually and re-run PETAL, iterating as such until we find a satisfactory combination of verbalizers and patterns which adequately reflect the task. Additionally, we ensure that the patterns between \textbf{T1} and \textbf{T2} are roughly equivalent. This yields 2 patterns for each task, provided in \autoref{tab:patterns}, and verbalizers given in \autoref{tab:verbalizers1}. The verbalizers found by PETAL capture multiple aspects of the task labels, selecting words such as ``mistaken,'' ``wrong,'' and ``artificial'' for exaggeration, ``preliminary'' and ``conditional'' for downplaying, and multiple levels of strength for strength detection such as ``estimated'' (correlational), ``cautious'' (conditional causal), and ``proven'' (direct causal).

For unlabeled data, we start with unlabeled pairs of full text press releases and abstracts. As we are concerned with detecting exaggeration in the primary conclusions, we first train a classifier based on single task PET for conclusion detection using a set of seed data. The patterns and verbalizers we use for conclusion detection are given in \autoref{tab:conc_patterns} and \autoref{tab:verbalizers2}. After training the conclusion detection model, we apply it to the press releases and abstracts, choosing the sentence from each with the maximum score $s_{\mathbf{p}}(1|\mathbf{x})$.
\begin{table}[t]
    \def\arraystretch{1.5}
        \centering
        \fontsize{10}{10}\selectfont
        \begin{tabular}{c c p{3.5cm}}
        \toprule %
        Name & Pattern\\
        \midrule %
    $P_{0}(x)$&\texttt{[MASK]}: $a$ \\
    $P_{1}(x)$&\texttt{[MASK]} - $a$\\
    $P_{2}(x)$&``\texttt{[MASK]}'' statement: $a$\\
    $P_{3}(x)$&$a$ (\texttt{[MASK]}) \\
    $P_{4}(x)$&(\texttt{[MASK]}) $a$\\
    $P_{5}(x)$&[Type: \texttt{[MASK]}] $a$ \\
\bottomrule
    \end{tabular}
    \caption{Patterns for conclusion detection.}
    \label{tab:conc_patterns}
\end{table}
\begin{table}[t]
    \def\arraystretch{1.2}
    \centering
    \fontsize{10}{10}\selectfont
    \begin{tabular}{c l}
    \toprule %
    Label & Verbalizers\\
    \midrule %
        0& Text\\
        1& Conclusion\\
    \bottomrule %

    \end{tabular}
    \caption{Verbalizers for PVPs for conclusion detection.}
    \label{tab:verbalizers2}
\end{table}

\subsection{Data Collection}
One of the main contributions of this work is a curated benchmark dataset for scientific exaggeration detection. Labeled datasets exist for the related task of claim strength detection in scientific abstracts and press releases~\cite{yu2020measuring,yu2019detecting}, but these 
data are from press releases and abstracts which are unrelated (i.e. the given press releases are not written about the given abstracts), making them unsuitable for benchmarking exaggeration detection. Given this, we curate a dataset of paired sentences from abstracts and associated press releases, labeled by experts for exaggeration based on their claim strength. We then collect a large set of unlabeled press release/abstract pairs useful for semi-supervised learning.

\subsubsection{Gold Data}
The gold test data used in this work are from~\citet{sumner2014association} and \citet{bratton2019association}, who annotate scientific papers, their abstracts, and associated press releases along several dimensions to characterize how press releases exaggerate papers. 
The original data consists of 823 pairs of abstracts and press releases. The 462 pairs from \citet{sumner2014association} have been used in previous work to test claim strength prediction~\cite{DBLP:conf/emnlp/LiZY17}, but the data, which contain press release and abstract conclusion sentences that are mostly paraphrases of the originals, are used as is.

We focus on the annotations provided for claim strength. The annotations consist of six labels which we map to the four labels defined in \citet{DBLP:conf/emnlp/LiZY17}. The labels and their meaning are given in \autoref{tab:all_labels}. %
This gives a claim strength label $l_{\rho}$ for the press release and $l_{\gamma}$ for the abstract. The final exaggeration label is then defined as follows:
\begin{equation*}
l_{e} = \begin{cases}
    0 & l_{\rho} < l_{\gamma}\\
    1 & l_{\rho} = l_{\gamma}\\
    2 & l_{\rho} > l_{\gamma}
    \end{cases}
\end{equation*}

As the original abstracts in the study are not provided, we automatically collect them using the Semantic Scholar API.\footnote{\url{https://api.semanticscholar.org/}}
We perform a manual inspection of abstracts to ensure the correct ones are collected, discarding missing and incorrect abstracts. 
Gold conclusion sentences are obtained by sentence tokenizing abstracts using SciSpaCy~\cite{DBLP:conf/bionlp/NeumannKBA19} and finding the best matching sentence to the provided paraphrase in the data using ROUGE score~\cite{lin2004rouge}. We then manually fix sentences which do not correspond to a single sentence from the abstract. Gold press release sentences are gathered in the same way from the provided press releases.

This results in a dataset of 663 press release/abstract pairs labeled for claim strength and exaggeration. The label distribution is given in \autoref{tab:gold_statistics}. %
We randomly sample 100 of these instances as training data for few shot learning (\textbf{T1}), leaving 553 instances for testing. Additionally, we create a small training set of 1,138 sentences labeled for whether or not they are the main conclusion sentence of the press release or abstract. This data is used in the first step of \METHOD~to identify conclusion sentences in the unlabeled pairs.

\begin{table*}[t]
    \def\arraystretch{1.3}
    \centering
    \fontsize{10}{10}\selectfont
    \begin{tabular}{c p{4cm} c p{4cm}}
    \toprule %
    \citet{sumner2014association} & Description & \citet{DBLP:conf/emnlp/LiZY17}& Description\\
    \midrule %
    0& No relationship mentioned& - & -\\
    \hline
    1& Statement of no relationship & 0& Statement of no relationship\\
    \hline
    2& Statements of correlation & \multirow{2}{*}{1}& \multirow{2}{*}{Statement of correlation}\\
    3& Ambiguous statement of relationship & &\\
    \hline
    4& Conditional statement of causation &\multirow{2}{*}{2}& \multirow{2}{4cm}{Conditional statement of causation}\\
    5& Statement of ``can'' & &\\
    \hline
    6& Statements of causation & 3& Statement of causation\\
    \bottomrule %

    \end{tabular}
    \caption{Claim strength labels and their meaning from the original data in \citet{sumner2014association} and \citet{bratton2019association} and the mappings to the labels from \citet{DBLP:conf/emnlp/LiZY17}. We use the labels from \citet{DBLP:conf/emnlp/LiZY17} in this study, including for deriving the exaggeration labels.}
    \label{tab:all_labels}
\end{table*}
\begin{table}
    \centering
    \fontsize{10}{10}\selectfont
    \begin{tabular}{l c}
    \toprule %
     Label & Count\\
    \midrule
        Downplays & 113\\
        Same & 406\\
        Exaggerates& 144\\
    \bottomrule %

    \end{tabular}
    \caption{Number of labels per class for benchmark exaggeration detection data.}
    \label{tab:gold_statistics}
\end{table}
For \textbf{T2} we use the data from~\citet{yu2020measuring,yu2019detecting}. \citet{yu2019detecting} create a dataset of 3,061 conclusion sentences labeled for claim strength from structured PubMed abstracts of health observational studies with conclusion sections of 3 sentences or less. \citet{yu2020measuring} then annotate statements from press releases from EurekAlert. The selected data are from the title and first two sentences of the press releases, as \citet{sumner2014association} note that most press releases contain their main conclusion statements in these sentences, following an inverted pyramid structure common in journalism~\cite{po2003news}. Both studies use the labeling scheme from \citet{DBLP:conf/emnlp/LiZY17} (see \autoref{tab:all_labels}). %
The final data contains 2,076 labeled conclusion statements. From these two datasets, we select a random stratified sample of 4,500 instances for training in our full-data experiments, and subsample 200 for few-shot learning (100 from abstracts and 100 from press releases). 

\subsubsection{Unlabeled Data}
We collect unlabeled data from ScienceDaily,\footnote{\url{https://www.sciencedaily.com/}} a science reporting website which aggregates and re-releases press releases from a variety of sources. To do this, we crawl press releases from ScienceDaily via the Internet Archive Wayback Machine\footnote{\url{https://archive.org/web/}} between January 1st 2016 and January 1st 2020 using Scrapy.\footnote{\url{https://scrapy.org/}} We discard press releases without paper DOIs and then pair each press release with a paper abstract by querying for each DOI using the Semantic Scholar API. This results in an unlabeled set of 7,741 press release/abstract pairs. Additionally, we use only the title, lead sentence, and first three sentences of each press release.%

\subsection{Experiments}
Our experiments are focused on the following primary research questions:
\begin{itemize}[noitemsep]
    \item \textbf{RQ1}: Does \METHOD~improve over PET for scientific exaggeration detection?
    \item \textbf{RQ2}: Which formulation of the problem leads to the best performance?
    \item \textbf{RQ3}: Does few-shot learning performance approach the performance of models trained with many instances?
    \item \textbf{RQ4}: What are the challenges of scientific exaggeration prediction?
\end{itemize}
We experiment with the following model variants:
\begin{itemize}[noitemsep]
    \item \textbf{Supervised}: A fully supervised setting where only labeled data is used.
    \item \textbf{PET}: Standard single-task PET.
    \item \textbf{\METHOD}: We run \METHOD~with data from one task formulation as the main task and the other formulation as the auxiliary task.
\end{itemize}

We perform two evaluations in this setup: one with \textbf{T1} as the main task and one with \textbf{T2}. For \textbf{T1}, we use the 100 expert annotated instances with paired press release and abstract sentences labeled for exaggeration (200 sentences total). For \textbf{T2}, we use 100 sentences from the press data from \citet{yu2020measuring} and 100 sentences from the abstract data in \citet{yu2019detecting} labeled for claim strength. We use RoBERTa base~\cite{DBLP:journals/corr/abs-1907-11692} from the HuggingFace Transformers library~\cite{DBLP:conf/emnlp/WolfDSCDMCRLFDS20} as the main model, and set $\alpha_{m}$ to be $1$, and $\alpha_{a} = \text{min}(2, \frac{|D_{m}|}{|D_{a}|})$. All methods are evaluated using macro-F1 score, and results are reported as the average performance over 5 random seeds.

\subsubsection{Performance Evaluation}
\begin{table}[t]
    \def\arraystretch{1.2}
    \centering
    \fontsize{10}{10}\selectfont
    \begin{tabular}{l c c c}
    \toprule %
    Method & P & R & \multicolumn{1}{c}{F1} \\
    \midrule %
       Supervised & $28.06$& $33.10$& $29.05$\\
       PET & $41.90$& $39.87$& $39.12$\\
       MT-PET & $\mathbf{47.80}$& $\mathbf{47.99}$& $\mathbf{47.35}$\\
    \bottomrule %

    \end{tabular}
    \caption{Results for exaggeration detection with paired conclusion sentences from abstracts and press releases (\textbf{T1}). MT-PET uses 200 sentences for strength classification, 100 each from press releases and abstracts.} %
    \label{tab:t1_base_results}
\end{table}
\begin{table*}[t]
    \def\arraystretch{1.2}
    \centering
    \fontsize{10}{10}\selectfont
    \begin{tabular}{l c c c c c c}
    \toprule %
    Method & $|$\textbf{T2}$|$,$|$\textbf{T1}$|$& P & R & \multicolumn{1}{c}{F1} & Press F1 & Abstract F1 \\
    \midrule %
       Supervised & 200,0 & $49.28$& $51.07$& $49.03$& $54.78$& $59.41$\\
       PET& 200,0 & $55.76$& $58.58$& $56.57$& $63.56$& $62.76$\\
       MT-PET& 200,100 & $\mathbf{56.68}$& $\mathbf{60.13}$& $\mathbf{57.44}$& $\mathbf{64.72}$& $\mathbf{63.27}$\\
     \midrule %
       Supervised& 4500,0 & $58.20$& $59.99$& $58.66$& $63.26$& $\mathbf{67.26}$\\
       PET& 4500,0 & $59.53$& $61.84$& $60.45$& $\mathbf{64.20}$& $64.92$\\
       MT-PET& 4500,100 & $\mathbf{60.09}$& $\mathbf{62.68}$& $\mathbf{61.11}$& $63.93$& $64.69$\\
    \midrule
       PET+in domain MLM& 200,100 & $\textit{57.18}$& $\textit{60.12}$& $\textit{58.06}$& $\textit{64.29}$& $\textit{62.69}$\\
       PET+in domain MLM& 4500,100 & $\textit{59.87}$& $\textit{62.33}$& $\textit{60.85}$& $\textit{64.10}$& $\textit{64.73}$\\
    \bottomrule %

    \end{tabular}
    \caption{Results on exaggeration detection via strength classification (\textbf{T2}) with varying numbers of instances. \METHOD~uses 100 instances from paired press and abstract sentences (200 sentences total).} %
    \label{tab:t2_base_results}
\end{table*}

We first examine the performance with \textbf{T1} as the base task (see \autoref{tab:t1_base_results}). In a purely supervised setting, the model struggles to learn and mostly predicts the majority class. Basic PET yields a substantial improvement of 10 F1 points, with \METHOD~further improving upon this by another 8 F1 points. Accordingly, we conclude that training with auxiliary task data provides much benefit for scientific exaggeration detection in the \textbf{T1} formulation.

We next examine performance with \textbf{T2} (strength classification) as the main task in both few-shot and full data settings (see \autoref{tab:t2_base_results}). In terms of base performance, the model can predict exaggeration better than \textbf{T1 }in a purely supervised setting. For PET and \METHOD, we see a similar trend; with 200 instances for \textbf{T2}, PET improves by 7 F1 points over supervised learning, and \METHOD~improves on this by a further 0.9 F1 points. Additionally, \METHOD~improves performance on the individual tasks of predicting the claim strength of conclusions in press releases and scientific abstracts with 200 examples. While less dramatic, we still see gains in performance using PET and \METHOD~when 4,500 instances from \textbf{T2} are used, despite the fact that there are still only 100 instances from \textbf{T1}. We also test if the improvement in performance is simply due to training on more in-domain data (``PET+in domain MLM'' in \autoref{tab:t2_base_results}). We observe gains for exaggeration detection using masked language modeling on data from \textbf{T1}, but \METHOD~still performs better at classifying the strength of claims in press releases and abstracts when 200 training instances from \textbf{T2} are used.

\paragraph{RQ1} Our results indicate that \METHOD~does in fact improve over PET for both training setups. With \textbf{T1} as the main task and \textbf{T2} as the auxiliary task, we see that performance is substantially improved, demonstrating that learning claim strength prediction helps produce soft-labeled training data for \textit{exaggeration detection}. Additionally, we find that the reverse holds with \textbf{T2} as main task and \textbf{T1} as auxiliary task. As performance can also be improved via masked language modeling on data from \textbf{T1}, this indicates that some of the performance improvement could be due to including data closer to the test domain. However, our error analysis in \autoref{sec:error_analysis} shows that these methods improve model performance on different types of data.

\paragraph{RQ2} We find that \textbf{T2} is better suited for scientific exaggeration detection in this setting, however, with a couple of caveats. First, the final exaggeration label is based on expert annotations for claim strength, so clearly claim strength prediction will be useful in this setup. Additionally, the task may be more forgiving here, as only the direction needs to be correct and not necessarily the final strength label (i.e. predicting `0' for the abstract and any of `1,' `2,' or `3' for the press release label will result in an exaggeration label of `exaggerates').
\begin{figure}[t]
     \centering
         \includegraphics[width=0.65\linewidth]{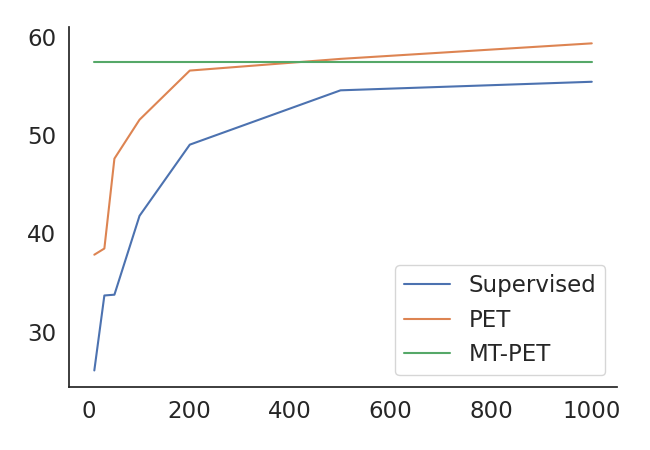}
         \caption{Learning curve for supervised learning and PET compared to performance of \METHOD~using 200 instances from \textbf{T2} and 100 from \textbf{T1}.}
         \label{fig:learning_curve}
\end{figure}
\paragraph{RQ3} We next examine the learning dynamics of our few-shot models with different amounts of training data (see \autoref{fig:learning_curve}), comparing them to \METHOD~to understand how well it performs compared to settings with more data. \METHOD~ with only 200 samples is highly competitive with purely supervised learning on 4,500 samples (57.44 vs. 58.66). Additionally, \METHOD~performs at or above supervised performance up to 1000 input samples, and at or above PET up to 500 samples, again using only 200 samples from \textbf{T2} and 100 from \textbf{T1}.

\subsubsection{Error Analysis}
\label{sec:error_analysis}
\begin{figure*}
     \centering

         \centering
         \includegraphics[width=0.95\textwidth]{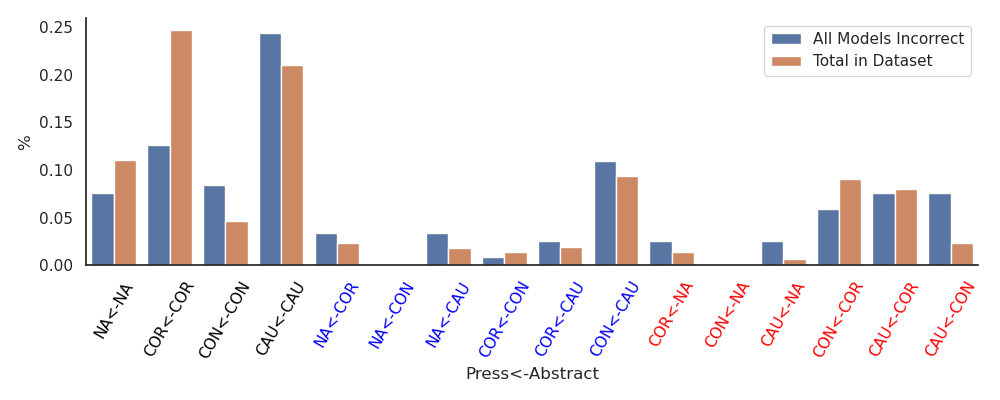}
         \caption{Proportion of examples by label which all models predict incorrectly.}
         \label{fig:all_fail}
\end{figure*}

\paragraph{RQ4} Finally, we try to understand the difficulty of scientific exaggeration detection by observing where models succeed and fail (see \autoref{fig:all_fail}). The most difficult category of examples to predict involve direct causal claims, particularly exaggeration and downplaying when one document is a direct causal claim and the other an indirect causal claim (`CON->CAU', `CAU->CON'). Also, it is challenging to predict when both the press release and abstract conclusions are directly causal.

The models have the easiest time predicting when both statements involve correlational claims, and exaggerations involving correlational claims from abstracts. We also observe that \METHOD~helps the most for the most difficult category: causal claims (see \autoref{fig:supervised_succeed} in \S\ref{sec:extra_plots}). The model is particularly better at differentiating when a causal claim in an abstract is \textit{downplayed} by a press release. It is also better at identifying correlational claims than PET, where many claims involve association statements such as `linked to,' `could predict,' `more likely,` and `suggestive of.'

The model trained with MLM on data from \textbf{T1} also benefits causal statement prediction, but mostly for when both statements are causal, whereas \METHOD~sees more improvement for pairs where one causal statement is exaggerated or downplayed by another (see \autoref{fig:pet_succeed} in \S\ref{sec:extra_plots}). This suggests that training with the patterns from \textbf{T1} helps the model to differentiate direct causal claims from weaker claims, while MLM training mostly helps the model to understand better how direct causal claims are written. We hypothesize that combining the two methods would lead to mutual gains.

\subsection{Related Work}
\subsubsection{Scientific Misinformation Detection}
Misinformation detection focuses on a variety of problems, including fact verification~\cite{DBLP:conf/naacl/ThorneVCM18,DBLP:conf/emnlp/AugensteinLWLHH19}%
, check-worthiness detection~\cite{Wright2020ClaimCD,DBLP:conf/ecir/NakovMEBMSAHHBN21}%
, stance~\cite{conf/emnlp/AugensteinRVB16,conf/naacl/BalyMGMMN18,hardalov2021survey} and clickbait detection~\cite{potthast-etal-2018-crowdsourcing}. While most work has focused on social media and general domain text, recent work has begun to explore different problems in detecting misinformation in scientific text such as SciFact~\cite{DBLP:conf/emnlp/WaddenLLWZCH20} and CiteWorth~\cite{DBLP:conf/acl/WrightA21}, as well as related tasks such as summarization \cite{DBLP:journals/corr/abs-2104-06486,dangovski2021we}.

Most work on scientific exaggeration detection has focused on flagging when the primary finding of a scientific paper has been exaggerated by a press release or news article~\cite{sumner2014association,bratton2019association,yu2020measuring,yu2019detecting,DBLP:conf/emnlp/LiZY17}. 
\citet{sumner2014association} and \citet{bratton2019association} manually label pairs of press releases and scientific papers on a wide variety of metrics, finding that one third of press releases contain exaggerated claims, and 40\% contain exaggerated advice. 
\citet{DBLP:conf/emnlp/LiZY17} is the first study into automatically predicting claim strength, using the data from \citet{sumner2014association} as a small labeled dataset. \citet{yu2019detecting} and \citet{yu2020measuring} extend this by building larger datasets for claim strength prediction, performing an analysis of a large set of unlabeled data to estimate the prevalence of claim exaggeration in press releases. Our work improves upon this by providing a more realistic task formulation of the problem, consisting of paired press releases and abstracts, as well as curating both labeled and unlabeled data to evaluate methods in this setting. %

\subsubsection{Learning from Task Descriptions}
Using natural language to perform zero and few-shot learning has been demonstrated on a number of tasks, including question answering~\cite{radford2018improving}, text classification~\cite{DBLP:journals/corr/abs-1912-10165}, relation extraction~\cite{DBLP:conf/aaai/BouraouiCS20} and stance detection \cite{hardalov2021emnlp,hardalov2021fewshot}. Methods of learning from task descriptions have been gaining more popularity since the creation of GPT-3~\cite{DBLP:conf/nips/BrownMRSKDNSSAA20}. %
\citet{DBLP:journals/jmlr/RaffelSRLNMZLL20} attempt to perform this with smaller language models by converting tasks into natural language and predicting tokens in the vocabulary. \citet{schick2020exploiting} propose PET, a method for few shot learning which converts tasks into cloze-style QA problems which can be solved by a pretrained language model in order to provide soft-labels for unlabeled data. We build on PET, showing that complementary cloze-style QA tasks can be trained on simultaneously to improve few-shot performance on scientific exaggeration detection.

\subsection{Conclusion}
In this work, we present a formalization of and investigation into the problem of scientific exaggeration detection. As data for this task is limited, we develop a gold test set for the problem and propose \METHOD, a semi-supervised approach based on PET, to solve it with limited training data. We find that \METHOD~helps in the more difficult cases of identifying and differentiating direct causal claims from weaker claims, and that the most performant approach involves classifying and comparing the individual claim strength of statements from the source and target documents. The code and data for our experiments can be found online\footnote{\url{https://github.com/copenlu/scientific-exaggeration-detection}}. Future work should focus on building more resources e.g. datasets for exploring scientific exaggeration detection, including data from multiple domains beyond health science. %
Finally, it would be interesting to explore how MT-PET works on combinations of more general NLP tasks, such as question answering and natural language inference or part-of-speech tagging and named entity recognition.

\section*{Acknowledgements}

$\begin{array}{l}\includegraphics[width=1cm]{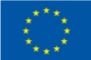} \end{array}$ The research documented in this paper has received funding from the European Union's Horizon 2020 research and innovation programme under the Marie Sk\l{}odowska-Curie grant agreement No 801199.

\section*{Broader Impact Statement}

Being able to automatically detect whether a press release exaggerates the findings of a scientific article could help journalists write press releases, which are more faithful to the scientific articles they are describing. We further believe it could benefit the research community working on fact checking and related tasks, as developing methods to detect subtle differences in a statement's veracity is currently understudied.

On the other hand, as our paper shows, this is currently still a very challenging task, and thus, the resulting models should only be applied in practice with caution. 
Moreover, it should be noted that the predictive performance results reported in this paper are for press releases written by science journalists -- one could expect worse results for press releases which more strongly simplify scientific articles.

\newpage

\section{Modeling Information Change in Science Communication with Semantically Matched Paraphrases}
\label{paper:modeling}

\subsection{Introduction}

Science communication disseminates scholarly information to audiences outside the research community, such as the public and policymakers \citep{national2017communicating}. This process usually involves translating highly technical language to non-technical, less-formal language that is engaging and easily understandable for lay people \citep{Salita2015WritingFL}. The public relies on the media to learn about new scientific findings, and media portrayals of science affect people's trust in science while at the same time influencing their future actions \citep{gustafson2019effects,Fischhoff2012CommunicatingUF,kuru2021effects}. However, not all scientific communication accurately conveys the original information, as shown in \autoref{fig:claim_generation}. Identifying cases where scientific information has changed is a critical but challenging task due to the complex translating and paraphrasing done by effective communicators. Our work introduces a new task of measuring scientific information change, and through developing new data and models aims to address the gap in studying faithful scientific communication.

\begin{figure}[t]
  \centering
    \includegraphics[width=0.65\linewidth]{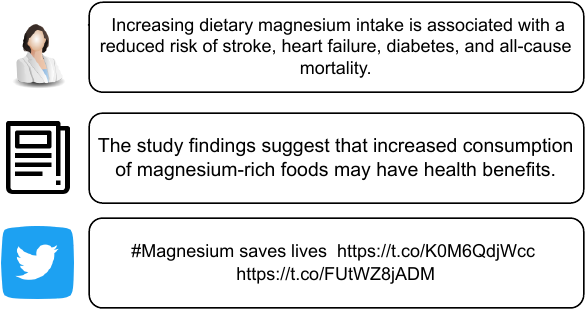}
    \caption{We are interested in measuring the information similarity of statements about scientific findings between different sources, including scientific papers, news, and tweets, shown here with real examples. The finding in this figure comes from~\citet{fang2016dietary} and the news quote is from~\cite{rappaport2016dietary}.} 
    \label{fig:claim_generation}
\end{figure}

Though efforts exist to track and flag when popular media misrepresent science,\footnote{See e.g. \url{https://www.healthnewsreview.org/} and \url{https://sciencefeedback.co/}} the sheer volume of new studies, reporting, and online engagement make purely manual efforts both intractable and unattractive. Existing studies in NLP to help automate the study of science communication have examined exaggeration \citep{wright2021semi}, certainty~\cite{DBLP:conf/emnlp/PeiJ21}, and fact checking~\cite{DBLP:conf/bionlp/BoissonnetSPV22,DBLP:conf/acl/0001WLKCAW22}, among others. However, these studies skip over the key first step needed to compare scientific texts for information change: automatically identifying content from both sources which describe the \textbf{same} scientific finding. In other words, to answer relevant questions about and analyze changes in scientific information at scale, one must first be able to point to which original information is being communicated in a new way. %

To enable automated analysis of science communication, this work offers the following \textbf{contributions} (marked by \textbf{C}). First, we present the \textsc{Scientific Paraphrase and Information ChangE Dataset} dataset (\textsc{Spiced}), a manually annotated dataset of paired scientific findings from news articles, tweets, and scientific papers (\textbf{C1}, \S\ref{sec:datset}). 
\textsc{Spiced} has the following merits: (1) existing datasets focus purely on semantic similarity, while \textsc{Spiced} focuses on differences in the \textit{information} communicated in scientific findings; (2) scientific text datasets tend to focus solely on titles or paper abstracts,  while \textsc{Spiced} includes sentences extracted from the full-text of papers and news articles; (3) \textsc{Spiced} is largely multi-domain, covering the 4 broad scientific fields that get the most media attention (namely: medicine, biology, computer science, and psychology) and includes data from the whole science communication pipeline, from research articles to science news and social media discussions. 

In addition to extensively benchmarking the performance of current models on \textsc{Spiced} (\textbf{C2}, \S\ref{sec:benchmarking}), we demonstrate that the dataset enables multiple downstream applications. In particular, we demonstrate how models trained on \textsc{Spiced} improve zero-shot performance on the task of sentence-level evidence retrieval for verifying real-world claims about scientific topics (\textbf{C3}, \S\ref{sec:evidence-retrieval}), and perform an applied analysis on unlabelled tweets and news articles where we show (1) media tend to exaggerate findings in the limitations sections of papers; (2) press releases and SciTech tend to have less informational change than general news outlets; and (3) organizations' Twitter accounts tend to discuss science more faithfully than verified users on Twitter and users with more followers (\textbf{C4}, \S\ref{sec:social-media-analysis}).%

\subsection{Related Work}
The analysis of scientific communication directly relates to fact checking, scientific language analysis, and semantic textual similarity. We briefly highlight our connections to these.

\paragraph{Fact Checking} Automatic fact checking is concerned with verifying whether or not a given claim is true, and has been studied extensively in multiple domains~\cite{DBLP:conf/naacl/ThorneVCM18,DBLP:conf/emnlp/AugensteinLWLHH19} including science~\cite{DBLP:conf/emnlp/WaddenLLWZCH20,DBLP:conf/bionlp/BoissonnetSPV22,DBLP:conf/acl/0001WLKCAW22}. Fact checking focuses on a specific type of information change, namely veracity. Additionally, the task generally assumes access to pre-existing knowledge resources, such as Wikipedia or PubMed, from which evidence can be retrieved that either supports or refutes a given claim. Our task is concerned with a more general type of information change beyond categorical falsehood and is a required task to complete prior to performing any kind of fact check.

\paragraph{Scientific Language Analysis}
Automating tasks beneficial for understanding changes in scientific information between the published literature and media is a growing area of research~\cite{wright2021semi,DBLP:conf/emnlp/PeiJ21,DBLP:conf/bionlp/BoissonnetSPV22,DBLP:conf/icwsm/DaiSW20,august2020writing,DBLP:conf/acl/TanL14,DBLP:conf/emnlp/VadapalliSPSV18,DBLP:conf/chi/AugustCHSR20,DBLP:conf/lrec/GinevM20}. The three tasks most related to our work are %
understanding writing strategies for science communication~\cite{august2020writing}, detecting changes in certainty~\cite{DBLP:conf/emnlp/PeiJ21}, and detecting changes in causal claim strength i.e. exaggeration~\cite{wright2021semi}. However, studying these requires access to paired scientific findings. To be able to do so at scale will require the ability to pair such findings automatically.

\paragraph{Semantic Similarity} The topic of semantic similarity is well-studied in NLP. Several datasets exist with explicit similarity labels, many of which come from SemEval STS shared tasks (e.g. \cite{DBLP:journals/corr/abs-1708-00055}) and paraphrasing datasets~\cite{DBLP:conf/naacl/GanitkevitchDC13}. It is possible to build unlabelled datasets of semantic similarity automatically, which is the main method that has been used for scientific texts~\cite{DBLP:conf/acl/CohanFBDW20,lo2020s2orc}. 
However, such datasets fail to capture more subtle aspects of similarity, particularly when the focus is solely on the scientific findings conveyed by a sentence (see \S\ref{sec:info-change-supplement}). 
And as we will show, approaches based on these datasets are insufficient for the task we are concerned with in this work, motivating the need for a new resource.

\subsection{\textsc{Spiced}}
\label{sec:datset}

We introduce \textsc{Spiced}, a new large-scale dataset of \textit{scientific findings} paired with how they are communicated in news and social media. Communicating scientific findings is known to have a broad impact on public attitudes \cite{weigold2001communicating} and to influence behavior, e.g., the way vaccines are framed in the media has an effect on vaccine uptake \cite{kuru2021effects}. Building upon prior work in NLP \cite{DBLP:conf/acl/WrightA21,DBLP:conf/emnlp/PeiJ21,sumner2014association,bratton2019association}, we define a scientific finding as \textbf{a statement that describes a particular research output of a scientific study, which could be a result, conclusion, product, etc.} %
This general definition holds across fields; for example, many findings from medicine and psychology report on effects on some dependent variable via manipulation of an independent variable, while in computer science many findings are related to new systems, algorithms, or methods. %
Following, we describe how the pairs of scientific findings were selected and annotated.

\subsubsection{Data Collection}

An initial dataset of unlabelled pairs of scientific communications was collected through Altmetric  ({\small{\url{https://www.altmetric.com/}}}) a platform tracking mentions of scientific articles online. %
This initial pool contains 17,668 scientific papers, 41,388 paired news articles, and 733,755 tweets---note that a single paper may be communicated about multiple times. The scientific findings were extracted in different ways for each source. Similar to \citet{Prabhakaran2016PredictingTR}, we fine-tune a RoBERTa~\cite{DBLP:journals/corr/abs-1907-11692} model to classify sentences into methods, background, objective, results and conclusions using 200K paper abstracts from PubMed that had been self-labeled with these categories~\cite{canese2013pubmed}. This sentence classifier attained 0.92 F1 score on a held-out 10\% sample (details in \S\ref{sec:parser-supplement}) and then the classifier was applied to each sentence of the news stories and paper fulltexts. 
Given the domain difference between scientific abstracts and news, we additionally manually annotated a sample of 100 extracted conclusions; we find that the precision of the classifier is 0.88, suggesting that it is able to accurately identify scientific findings in news as well. We extract each sentence classified as ``result'' or ``conclusion'' and create pairs with each finding sentence from news articles written about it. This yields 45.7M potential pairs of $\langle$news, paper$\rangle$ findings. For tweets, we take full tweets as is, yielding 35.6M potential pairs  of $\langle$tweet, paper$\rangle$ findings.

\subsubsection{Data sampling}
\label{sec:news-pairs}

Pairing every finding from a news story with every finding from its matched paper results in an untenable amount of data to annotate. Additionally, it has been shown that proper data selection can reduce the need to annotate every possible sample~\cite{DBLP:journals/neco/MacKay92b,DBLP:conf/cvpr/HolubPB08,DBLP:journals/corr/abs-1112-5745}. Therefore, to obtain a sample of paired findings covering a range of similarities, we first filter our pool of unlabelled matched findings based on the semantics 
using Sentence-BERT (SBERT, \citet{reimers-2019-sentence-bert}), a Siamese BERT network trained for semantic text similarity, trained on over 1B sentence pairs (see \S\ref{sec:full-model-descriptions} for further details). We use this model to score pairs of findings from news articles and papers based on their embeddings' cosine similarity and conduct a pilot study to determine which data to annotate. %

For the pilot, we sample 400 pairs evenly for every $0.05$ increment bucket in the range $[0,1]$ of similarity scores (20 per bucket). 
Each sample is annotated by two of the authors of this study with a binary label of ``matching'' vs ``not matching'', yielding a Krippendorff's alpha of $0.73$.\footnote{Note that many discussions about what constitutes matching vs. not matching were had in pilot work, leading to high agreement.} From this sample, we observed that there were no matches below 0.3 and only 2 ambiguous matches below 0.4. At the same time, the vast majority of samples from the entire dataset have a similarity score of less than 0.4. Additionally, above 0.9 we saw that each pair was essentially equivalent. Given the distribution of matched findings across the similarity scale, in order to balance the number of annotations we can acquire, the yield of positive samples, and the sample difficulty, we sampled data as follows based on their cosine similarity:

\begin{itemize}[noitemsep,nolistsep]
    \item Below $0.4$ = automatically unmatched.
    \item Above $0.9$ with a Jaccard index above $0.5$ = automatically matched.
    \item Sample an equal number of pairs from each $0.05$ increment bin between $0.4$ and $0.9$ for human expert annotation.
\end{itemize}

We sample 600 $\langle$news, paper$\rangle$ finding pairs from the four fields which receive the most media attention (medicine, biology, computer science, and psychology) using this method. 
This yields 2,400 pairs to be annotated. For extensive details on the pilot annotation and visualizations, see \S\ref{sec:pilot-annotation-details}. 

We follow a similar procedure to sample pairs from papers and Twitter for annotation. However, rather than use the SBERT similarity scores, we instead first obtain annotations for news pairs using the scheme to be described later in \S\ref{sec:annotation-scheme} in order to train an initial model on our task (CiteBERT, \citealt{DBLP:conf/acl/WrightA21}). We then use the trained model to obtain scores in the range [0,1] for each pair and sample an equal number of pairs from bins in 0.05 increments, for a total of 1,200 pairs (300 from each field of interest). 

\subsubsection{Finding Matching Annotation}
\label{sec:annotation-scheme}
We perform our final annotation based on the sampling scheme above using the Prolific platform ({\small{\url{https://www.prolific.co/}}}) as it allows prescreening annotators by educational background.
We require each annotator to have at least a bachelor's degree in a relevant field to work on the task. 
Annotators are asked to label ``whether the two sentences are discussing the same scientific finding'' for 50 finding pairs with a 5-point Likert schema where each value indicates that 
``The information in the findings is...''
(1): Completely different
(2): Mostly different
(3): Somewhat similar
(4): Mostly the same, or
(5): Completely the same.
See  \S\ref{sec:experimented-annotations-supplement} for details  of how this rating scale was decided.
We call this the \SCOREFULL{} (\SCORE{}) of a pair of findings. Annotation was performed using \textsc{Potato} \cite{pei2022potato}. Full annotation instructions and details are listed in \S\ref{sec:annotation-instructions}. Notably, annotators were instructed to mark how similar the information in the \textit{findings} was, as opposed to how similar the sentences are. Further, they were instructed to ignore extraneous information like ``The scientists show...'' and ``our experiments demonstrate...''.

\paragraph{Post processing} %
To improve the reliability of the annotations, we use MACE~\cite{DBLP:conf/naacl/HovyBVH13} to estimate the competence score of each annotator and removed the labels from the annotators with the lowest competence scores. 
\cameraready{MACE is a Bayesian method that learns distributions over true class labels and annotator competence based on crowd-sourced labels. }
We further manually examine pairs with the most diverse labels (standard deviation of ratings $>$1.2) and manually replace the outliers with our expert annotations. The overall Krippendoff's $\alpha$ is 0.52, 0.57, 0.53, and 0.52 for CS, Medicine, Biology, and Psychology respectively, indicating that the final labels are reliable. While many annotators considered the task challenging, our quality control strategies allow us to collect reliable annotations.\footnote{For example, one participant commented ``It was pretty hard to consider both the statements and their context then comparing them for similarities, but i enjoyed it''} For all the annotated pairs, we average the ratings as the final similarity score. In addition to the 3,600 manually annotated pairs, we include an extra 2,400 automatically annotated pairs as determined in \S\ref{sec:news-pairs} (unmatched pairs get an IMS of 1, matched pairs get an IMS of 5), for a total of 6,000 pairs. 
Given that there can be multiple pairs from a single news-paper pair, to avoid overlaps between training and test sets, we split the dataset 80\%/10\%/10\% based on the paper DOI and balance across subjects. Further dataset details in \S\ref{sec:dataset-details-supplement}

\begin{table*}[t]
\small

\newcommand{\tabincell}[2]{\begin{tabular}{@{}#1@{}}#2\end{tabular}}
\resizebox{0.99\textwidth}{!}{
\begin{tabular}{p{62mm}p{62mm}cc}%
\textbf{Paper finding } & \textbf{News Finding} & \textbf{Similarity Score} & \textbf{\SCORE{}} \\
\midrule
However, the consistency of the erythritol results in both the central adiposity and usual glycemia comparisons lends strength to the findings, and the cluster of metabolites has biological plausibility. & Young adults who exhibited central adiposity gain over the course of 35 weeks had plasma erythritol levels 15-times higher at baseline than those with stable adiposity over the same period. & 0.88 & 1 \\ \midrule
Our results showed that most of the official adult-onset men began their antisocial activities during early childhood. & Beckley, who is in the department of psychology and neuroscience at Duke, said the adult-onset group had a history of anti-social behavior back to childhood, but reported committing relatively fewer crimes. & 0.38 & 4.4 \\
\end{tabular}
}
\caption{ Annotated information matching score (IMS) and the similarity score estimated by SBERT~\cite{reimers-2019-sentence-bert} for selected finding pairs from \textsc{Spiced}. These examples demonstrate that simple similarity scores may not reflect whether the two sentences are covering the same scientific finding.}
\label{tab:example_scores}
\end{table*}
\paragraph{Selected Examples}
To highlight the difficulty of \textsc{Spiced}, we show a pair of samples from our final dataset in  \autoref{tab:example_scores}. %
The \SCORE{} is compared to the cosine similarity between embeddings produced by SBERT. For the first case, SBERT presumably picks up on similarities in the discussed topics, such as erythritol and its relationship to adiposity, but the paper finding is concerned with the consistency of results and its biological implications while the news finding explicitly mentions a relationship between erythritol and adiposity. The second case expresses the opposite effect; the news finding contains a lot of extraneous information for context, but one of the core findings it expresses is the same as the paper finding, giving it a high rating in \textsc{Spiced}.

\paragraph{Comparison with existing datasets} 
To further characterize the difficulty of \textsc{Spiced} compared to existing datasets, we show the average normalized edit distance between matching pairs in \textsc{Spiced}, STSB~\cite{DBLP:journals/corr/abs-1708-00055}, and SNLI~\cite{DBLP:conf/emnlp/BowmanAPM15} (see \S\ref{sec:metrics-supplement} for the calculation). STSB is a semantic text similarity dataset consisting of pairs of sentences scored with their semantic similarity, sourced from multiple SemEval shared tasks. SNLI is a natural language inference corpus, and consists of pairs of sentences labeled for if they entail each other, contradict each other, or are neutral.
We calculated the mean normalized edit distance across all pairs of \textit{matching} sentences in each dataset's training data;  For \textsc{Spiced} and STSB, pairs are considered matching if their IMS or similarity score is greater than 3, respectively. For SNLI, pairs are considered matching if the label is ``entailment''.

\begin{table}%
    \def\arraystretch{1.2}
    \centering
    \fontsize{10}{10}\selectfont
    \begin{tabular}{c c c c c}
    \toprule %
    STSB & SNLI & \textsc{Spiced} & News & Tweets\\
    \midrule 

         $0.401$ & $0.631$ & $\mathbf{0.726}$ & $\mathit{0.712}$ & $\mathit{0.749}$\\
    
    \bottomrule %

    \end{tabular}
    \caption{The average normalized edit distance between matching pairs for various datasets shows that \textsc{Spiced} includes more pairs that are lexically dissimilar. For \textsc{Spiced} and STSB, pairs are considered matching if their similarity score is greater than 3. For SNLI, pairs are considered matching if the label is ``entailment''.}
    \label{tab:edit-distance}
\end{table}

We find that there is a much greater lexical difference between the matching pairs in \textsc{Spiced} (0.726) than existing general domain paired text datasets (0.401 for STSB and 0.631 for SNLI). This gap between STSB and \textsc{Spiced} also emphasizes the difference between traditional semantic textual similarity tasks and the information change task we describe here. Within \textsc{Spiced}, Twitter pairs had a higher distance (0.749) than news pairs (0.712), suggesting stronger domain differences. For qualitative examples showing the difference between \textsc{Spiced} and STSB, see \S\ref{sec:info-change-supplement}.

\paragraph{Relationship of \textsc{Spiced} to Fact Checking}
The task introduced by \textsc{Spiced} 
captures information change more broadly than veracity as in automatic fact checking, as the task is concerned with the degree to which two sentences describe the same scientific information---indeed, two similar sentences may describe the same information equally poorly. Our task is similar to the sentence selection stage in the fact checking pipeline, and we later demonstrate that models trained on \textsc{Spiced} data are useful for this task for science in \S\ref{sec:evidence-retrieval}. However, our task and annotation are agnostic to whether a pair of sentences entail one another. This is especially useful if one wants to compare how a particular finding is presented across different media. Fact-checking datasets are also explicitly constructed to contain claims which are about a single piece of information---\textsc{Spiced} is not restricted in this way, focusing on a more general type of information change beyond categorical falsehood. Finally, we note two more unique features of \textsc{Spiced}: 1) \textsc{Spiced} contains naturally occurring sentences, while fact checking datasets like FEVER and SciFact often contain manually written claims. 2) The combination of domains in \textsc{Spiced} is unique; sentences are paired between (news, science) and (tweets, science), and these pairings don’t exist currently.

\subsection{Scientific Information Change Models}
\label{sec:benchmarking}

We now use \textsc{Spiced} to evaluate models for estimating the IMS of finding pairs in two settings: zero-shot transfer and supervised fine-tuning. 

\subsubsection{Experimental setup} 

We use the following four models to estimate zero-shot transfer performance.
\textbf{Paraphrase}: RoBERTa~\cite{DBLP:journals/corr/abs-1907-11692} pre-trained for paraphrase detection on an adversarial paraphrasing task~\cite{nighojkar-licato-2021-improving}. We convert the output probability of a pair being a paraphrase to the range [1,5] for comparison with our labels. \textbf{Natural Language Inference (NLI)}: RoBERTa pre-trained on a wide range of NLI datasets~\cite{nie-etal-2020-adversarial}. The final score is the model's measured probability of entailment mapped to the range [1,5]. \textbf{MiniLM}: SBERT with MiniLM as the base network~\cite{DBLP:conf/nips/WangW0B0020}; we obtain sentence embeddings for pairs of findings and measure the cosine similarity between these two embeddings, clip the lowest score to 0, and convert this score to the range [1,5]. Note that this model was trained on over 1B sentence pairs, including from scientific text, using a contrastive learning approach where the embeddings of sentences known to be similar are trained to be closer than the embeddings of negatively sampled sentences. SBERT models represent a very strong baseline on this task, and have been used in the context of other matching tasks for fact checking including detecting previously fact-checked claims~\cite{shaar-etal-2020-known}. \textbf{MPNet}: The same setting and training data as MiniLM but with MPNet as the base network~\cite{DBLP:conf/nips/Song0QLL20}. 

We fine-tune the following six models on \textsc{Spiced} to estimate IMS as a comparison with zero-shot transfer.
\begin{itemize}[noitemsep,nolistsep]
\item{\textbf{MiniLM-FT}: The same MiniLM model from the zero-shot transfer setup but further fine-tuned on \textsc{Spiced}. The training objective is to minimize the distance between the IMS and the cosine similarity of the output embeddings of the pair of findings.}
\item{\textbf{MPNet-FT}: The same setup as MiniLM-FT but using MPNet as the base network.}
\item{\textbf{RoBERTa}: The RoBERTa~\cite{DBLP:journals/corr/abs-1907-11692} base model; We perform a regression task where the model is trained to minimize the mean-squared error between the prediction and IMS.}
\item{\textbf{SciBERT}: A transformer model trained using masked language modeling on a large corpus of scientific text~\cite{beltagy2019scibert}. The fine-tuning setup is the same as for the RoBERTa model.}
\item{\textbf{CiteBERT}: A SciBERT model further fine-tuned on the task of citation detection, and was shown to have improved performance on downstream tasks using scientific text~\cite{DBLP:conf/acl/WrightA21}. The training setup is the same as for the RoBERTa model.}
\end{itemize}

Please see \S\ref{sec:full-model-descriptions} for further details on the models and pretraining methods.
For the fine-tuned models, we train on the entire training set of \textsc{Spiced}, including both news findings and tweets\cameraready{ and show test performance overall and split between news and tweets}. For the test set we only use manually annotated pairs. Performance is measured in terms of mean-squared error (MSE) and Pearson correlation ($r$) (definitions of all metrics in \S\ref{sec:metrics-supplement}). All results are reported as the average and standard deviation for each model across 5 random seeds. %

\subsubsection{Results} 
\begin{figure*}[t]
         \centering
     \begin{subfigure}[b]{0.495\textwidth}
         \centering
         \includegraphics[width=\textwidth]{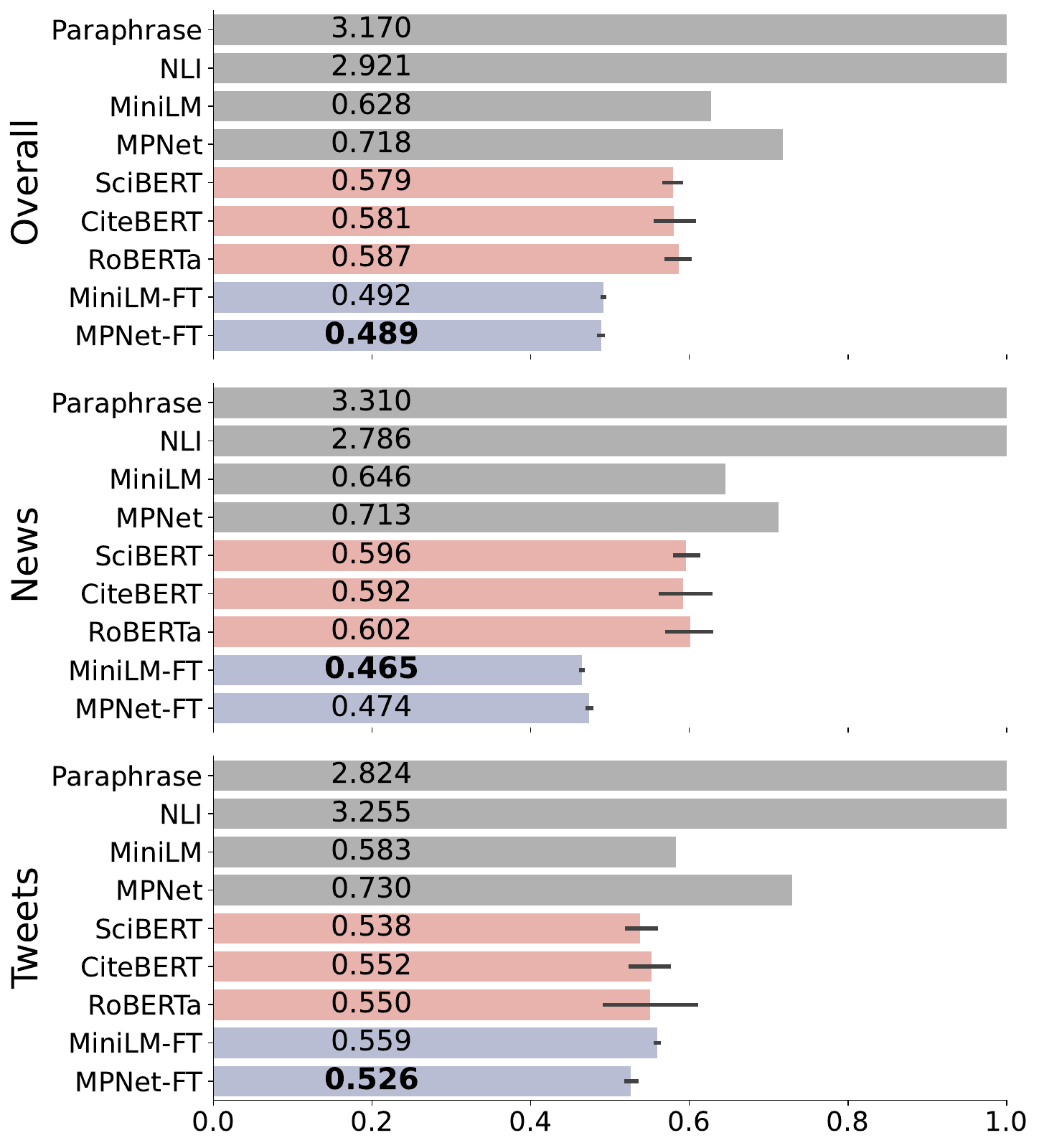}
         \caption{Mean Squared Error}
         \label{fig:mse-baselines}
     \end{subfigure}
     \hfill
     \begin{subfigure}[b]{0.495\textwidth}
         \centering
         \includegraphics[width=\textwidth]{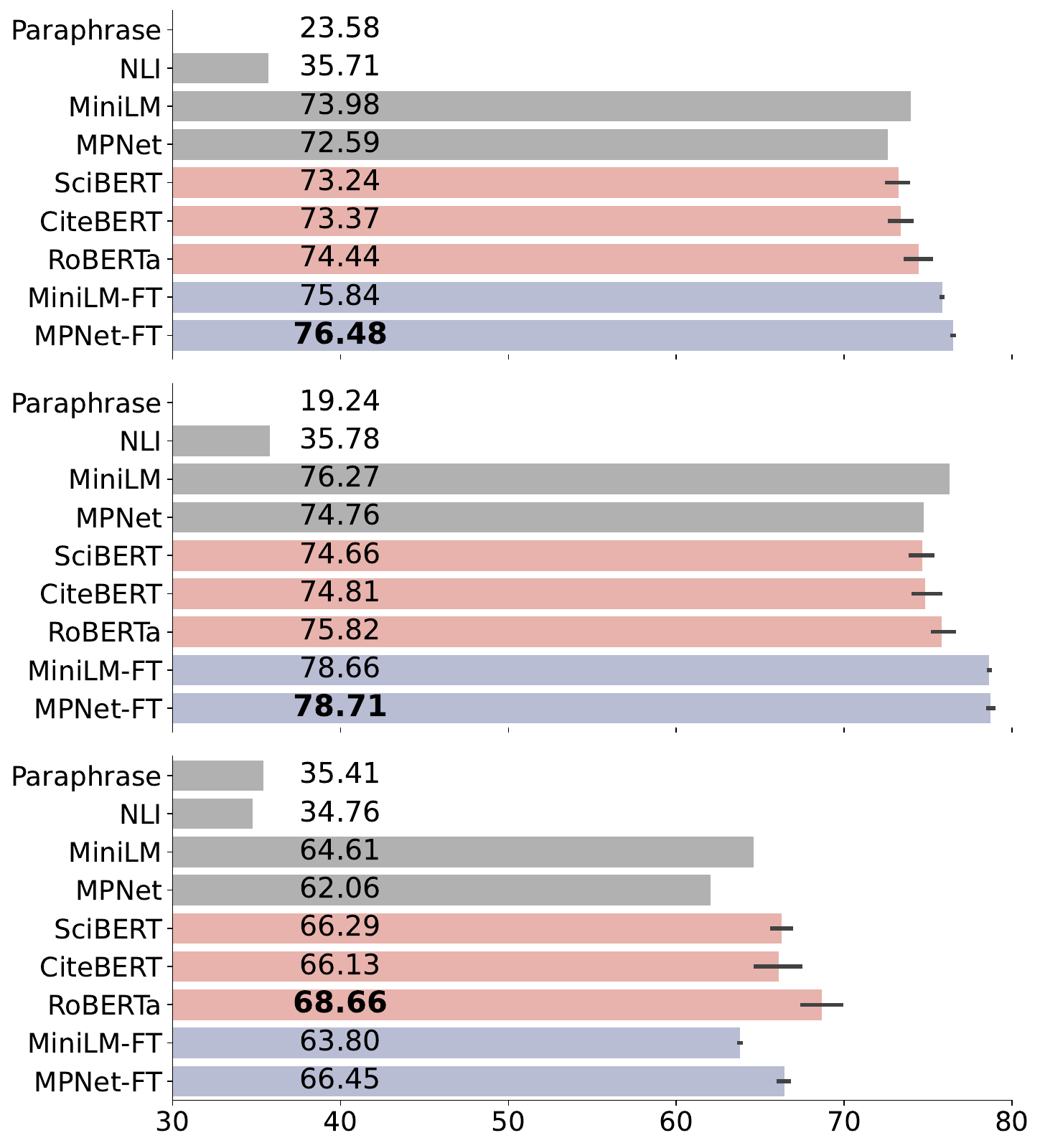}
         \caption{Pearson correlation ($r$)}
         \label{fig:r-baselines}
     \end{subfigure}
 
    \caption{(a) Mean Squared Error (MSE, $\downarrow$ better) and (b) Pearson correlation ($r$, $\uparrow$ better) on the test set of \textsc{Spiced}. Grey = zero-shot transfer models, red = MLM models fine-tuned on \textsc{Spiced}, blue = SBERT models fine-tuned on \textsc{Spiced}. Results are averaged across 5 random seeds. Best results are given in bold.}
    \label{fig:baseline-performance}
\end{figure*}

Paraphrase detection and natural language inference models perform very poorly for zero-shot transfer on this task (\autoref{fig:baseline-performance}, grey bars), with NLI having slightly better transfer, supporting our hypothesis that transferring from existing tasks to this domain is challenging. Fine-tuned models with Masked Language Model (MLM) pretraining can learn the task decently well (\autoref{fig:baseline-performance}, red bars), but surprisingly RoBERTa performs just as well as SciBERT and CiteBERT which were specifically pretrained on scientific texts. We posit that this could be due to the fact that RoBERTa was pretrained on a wider range of texts that are reflective of the domains in \textsc{Spiced}, including news texts, while SciBERT and CiteBERT were trained solely on scientific papers. 

SBERT models trained on large amounts of pretraining sentences perform well in the zero-shot transfer setup, with the MiniLM based model outperforming MPNet. The best setup was using SBERT fine-tuned on \textsc{Spiced} (\autoref{fig:baseline-performance}, blue bars), which yields up to 3.9 points gained overall in Pearson correlation and a reduction of 0.3 in terms of MSE (MPNet to MPNet-FT). We also note that there is a large gap between performance on this data and general semantic similarity datasets such as STSB, which see correlation scores in the 90s. As such, there is potentially much room to grow in terms of raw performance on this dataset.

Models performed worse for pairs with tweets versus those from news (Appendix \autoref{tab:baseline-findings-matching}). This performance difference is in line with our expectations, as there is a large domain shift between tweets and scientific texts and our base models were not exposed to tweets during pre-training. All models, including the zero-shot transfer SBERT models, perform much worse on that split of the data. Additionally, we only see minor gains in performance in terms of MSE for MiniLM when fine-tuned on tweets. We see larger gains for MPNet. Interestingly, the best performance (Pearson $r$) for Tweets is RoBERTa, though the overall MSE is still best for MPNet-FT. We show extended benchmarking in \S\ref{sec:extended-benchmarking} and the top-5 errors for RoBERTa and MPNet-FT in \S\ref{sec:error-examples}. \cameraready{We note the difficulty of these samples, including a need to understand that, for example, higher numbers are equivalent to ``an improvement'', and needing to hone in on relevant information in the sentences as opposed to extraneous details.}

\subsection{Application: Zero-Shot Evidence Retrieval for Scientific Fact Checking}
\label{sec:evidence-retrieval}

Accurately measuring the similarity of scientific findings written in different domains enables a wide range of downstream analyses and tasks. 
As a first task, we consider evidence retrieval for scientific fact checking of real-world scientific claims. In general, automatic fact checking consists of retrieving relevant evidence for a given claim and predicting if that evidence supports or refutes the claim. 
We test the ability of models trained on \textsc{Spiced} to perform the evidence retrieval task in a zero-shot setting. In this, we use the models as is, with no further fine-tuning on any evidence retrieval data. We consider two fact checking datasets: CoVERT~\cite{DBLP:journals/corr/abs-2204-12164} is a dataset of scientific claims sourced from Twitter, mostly in the domain of biomedicine. 
We use the 300 claims and the 717 unique evidence sentences in the corpus in our experiment. COVID-Fact~\cite{DBLP:conf/acl/SaakyanCM20} is a semi-automatically curated dataset of claims related to COVID-19 sourced from Reddit. 
The corpus contains 4,086 claims with 3,219 unique evidence sentences.

\paragraph{Setup}
We compare different models' ability to rank the evidence sentences such that the ground truth evidence for a given claim is ranked highest. We use four models in a zero-shot setting for comparison (MiniLM, MiniLM-FT, MPNet, and MPNet-FT; '-FT' indicates fine-tuning on \textsc{Spiced}), and show results with the unsupervised BM25 \cite{Robertson1994OkapiAT}, a widely used bag-of-words retrieval model. %
We report retrieval results in terms of mean average precision (MAP) and mean reciprocal rank (MRR), and average the results for models fine-tuned on \textsc{Spiced} across 5 random seeds.

\begin{table}[t]
    \setlength{\tabcolsep}{1.5pt}
    \def\arraystretch{1.2}
    \centering
    \fontsize{10}{10}\selectfont
    \begin{tabular}{l c c | c c}
    \toprule %
     & \multicolumn{2}{c}{CoVERT} & \multicolumn{2}{c}{COVID-Fact} \\
     \cline{2-5}
    Method & MAP & MRR & MAP & MRR \\
    \hline %
BM25 & $12.45_{0.00}$&$20.78_{0.00}$&$35.18_{0.00}$&$52.98_{0.00}$ \\
MiniLM & $26.84_{0.00}$&$37.98_{0.00}$&$50.11_{0.00}$&$64.78_{0.00}$ \\
~~~+ FT & $\mathbf{28.23_{0.08}}$&$\mathbf{40.81_{0.16}}$&$52.66_{0.10}$&$66.91_{0.09}$ \\
MPNet & $25.21_{0.00}$&$35.54_{0.00}$&$52.39_{0.00}$&$66.21_{0.00}$ \\
~~~+ FT & $26.84_{0.19}$&$37.65_{0.32}$&$\mathbf{53.61_{0.33}}$&$\mathbf{67.46_{0.28}}$ \\

    \bottomrule %

    \end{tabular}
    \caption{Mean average precision (MAP) and mean reciprocal rank (MRR) for retrieval on the CoVERT and COVID-Fact datasets. All models are zero-shot i.e. without fine-tuning on the retrieval dataset.} %
    \label{tab:retrieval}
\end{table}

\paragraph{Results}
We find that fine-tuning on \textsc{Spiced} provides consistent gains in retrieval performance on both datasets for both SBERT models (\autoref{tab:retrieval}). 
This performance increase is encouraging, as there are two notable differences between \textsc{Spiced} and the two datasets in our experiment. The first is that the tasks are different: \textsc{Spiced} provides a general scientific information similarity task which proves to be useful for evidence sentence ranking.
The second is that the domains are different: \textsc{Spiced} contains $\langle$news, paper$\rangle$ and $\langle$tweet, paper$\rangle$ pairs, while CoVERT and COVID-Fact have claims from Twitter and Reddit, respectively, paired with evidence in news. Our results show that training on \textsc{Spiced} improves the IR performance of the SBERT models, despite the domain and topic differences from our setting.  %

\subsection{Application: Modeling Information Change in Science Communication}
\label{sec:social-media-analysis}

Whether the media faithfully communicate scientific information has long been a core question to the science community \citep{national2017communicating}. Our dataset and models allow us to conduct a large-scale analysis to study information change in science communication. Here, we focus on three research questions: 
\begin{itemize}[noitemsep]
\item{\textbf{RQ1:} Do findings reported by different types of outlets express different degrees of information change from their respective papers?}
\item{\textbf{RQ2:} Do different types of social media users systematically vary in information change when discussing scientific findings?  }
\item{\textbf{RQ3:}  Which parts of a paper are more likely to be miscommunicated by the media?}
\end{itemize}

RQ1-2 focus on the holistic information change captured in \SCORE{}, while RQ3 focuses on what types of information might be changing.

\subsubsection{RQ1: Comparing Media Outlets}
Different types of media target different audiences and tend to report the same issue differently \cite{richardson1990writing, mencher1997news}. While good science journalism requires outlets to prioritize quality, in real practices, journalists may adopt different writing strategies for different types of audiences \citep{roland2009quality}. %
Thus, we investigate if findings reported by different types of outlets express different levels of information change, focusing on three types of outlets: General News (e.g., NYTimes), Press Releases (e.g., Science Daily), and Science \& Technology (e.g., Popular Mechanics).  
We use our best-performing MPNet-FT model to estimate the IMS of over 1B pairs and keep those with IMS $>$ 3, which finally leads to 
1.1M paired findings from 26,784 news stories and 12,147 papers. We then build a linear mixed effect regression model \citep{galecki2013linear} to predict \SCORE{} for matching pairs from news stories and research articles. 
We include a fixed effect for the type of news outlet, using General News as the reference category. To account for reporting differences across fields and variations specific to highly-publicized papers, we also include a fixed effect for the scientific subject and a random effect for each paper with 30+ pairs (all other papers are pooled in a single random effect). 

\begin{figure}[t]
  \centering
    \includegraphics[width=0.75\textwidth]{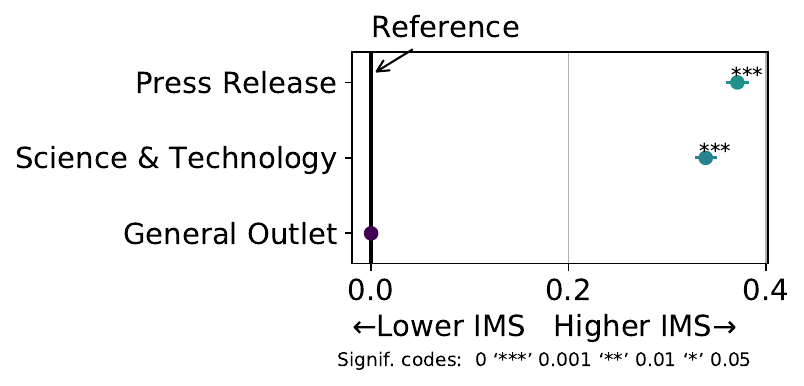}
    \caption{Scientific findings covered by Press Release and SciTech generally have less informational changes compared with findings presented in General Outlets
    }
    \label{fig:news_outlet_type_matching_score}
\end{figure}

\paragraph{Results.}
Compared with General News, Science \& Technology news outlets and Press Releases report findings that more closely match those from the original paper (\autoref{fig:news_outlet_type_matching_score} shows the regression coefficients). This difference likely is due to some form of audience design where the journalist is writing for a more science-savvy readership in the latter two, whereas General News journalists must more heavily paraphrase the results for lay people.

\subsubsection{RQ2: Comparing Social Media Accounts}

Social media play an important role in disseminating scientific findings \citep{zakhlebin2020diffusion}, so what factors affect the presentation of scientific information on social media becomes an important question. Here, we focus on the types of Twitter users who tweet about scientific findings. Based on 182K matched tweets and paper findings, we again build a linear mixed effect regression model to predict IMS. We include fixed effects of (1) if the account is run by an organization, as inferred using M3 \citep{wang2019demographic}, (2) if the account is  verified  (3) the number of followers and following, both log-transformed, and (4) the account age in years. We use the same field fixed effects and paper random effects as in RQ1.

\begin{figure}[t]
  \centering
    \includegraphics[width=0.75\textwidth]{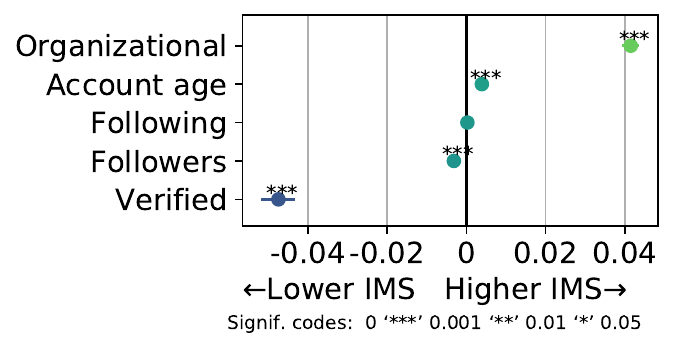}
    \caption{Organizational Twitter accounts keep more original information from the paper finding while verified users and those with more followers change more information when tweeting about a scientific finding. 
    }
    \label{fig:twitter_user_types_matching_score}
\end{figure}

\paragraph{Results}
The type of user strongly influences how faithful the tweets are to the original findings (Figure \ref{fig:twitter_user_types_matching_score}). Accounts from organizations tend to be more faithful to the original paper findings, which could be due to intentional actions of image management to build trust \citep{saffer2013effects}. Surprisingly, verified accounts were far more likely to change information away from its original meaning; similarly, accounts with more followers had the same trend. Given their prominent roles in Twitter communication \citep{bakshy2011everyone,hentschel2014finding}, multiple mechanisms may explain this gap such as adding more commentary or trying to translate original scientific findings to lay language to make the findings easier to understand. \S\ref{sec:regression_details} shows the details of regression results. \cameraready{Our results point to a need to study systematic behavior based on who is communicating science.}

\subsubsection{RQ3: What Information Changes}
\label{sec:section_analysis}
Most studies on scientific misinformation focus on paper titles and abstracts \cite{sumner2014association}, which cannot fully reflect the information presented in the full papers. Analyzing the information change of findings paired from all sections of papers could help to better understand the mechanisms behind scientific misinformation and develop strategies to reduce them. We use the same 1.1M finding pair dataset as RQ1 and analyze what information might have changed using two models trained for changes in scientific communication: identifying exaggerations \citep{wright2021semi} and certainty \citep{DBLP:conf/emnlp/PeiJ21}. See \S\ref{sec:exaggeration-supplement} for more details on the exaggeration detection task.

\begin{figure}[t]
  
  \centering
    \includegraphics[width=0.75\linewidth]{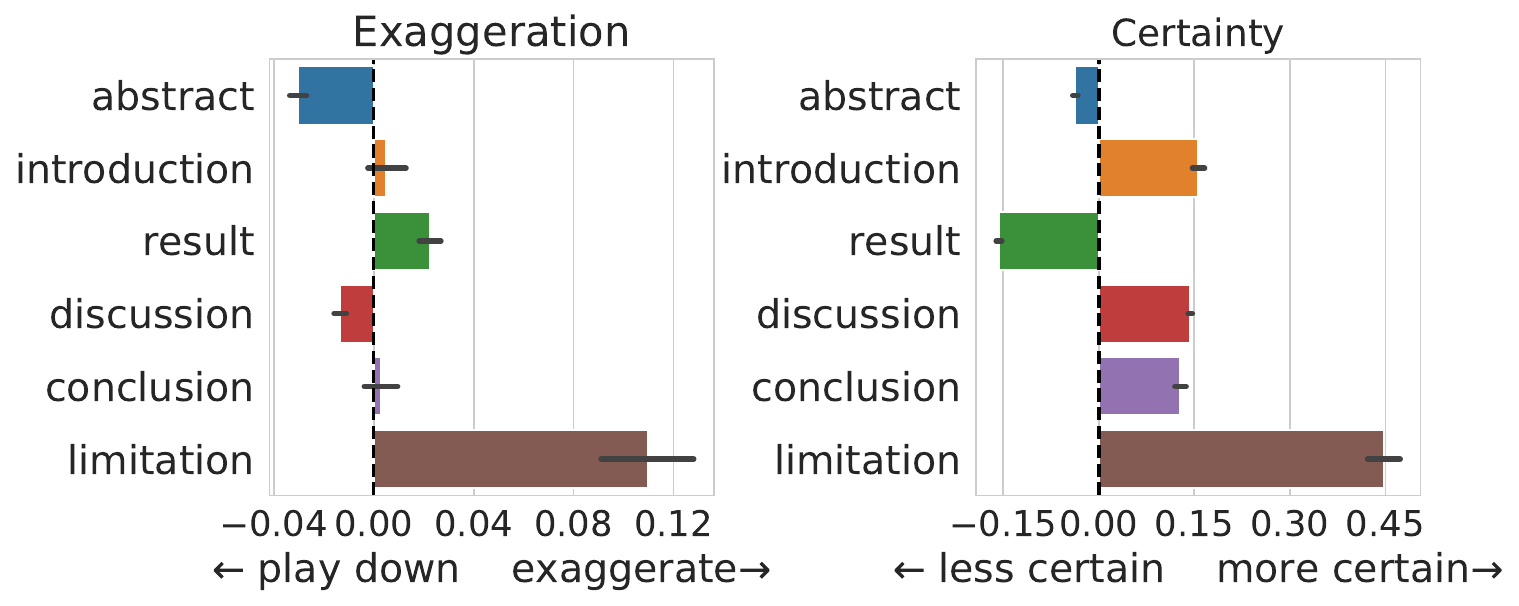}
    \caption{Journalists tend to downplay the certainty and strength of findings in abstracts, but overstate findings discussed in limitations sections. 
    }
    \label{fig:section_effect_news_paper_pairs}
\end{figure}

\paragraph{Results}
Journalists tend to downplay the certainty and strength of findings from abstracts (\autoref{fig:section_effect_news_paper_pairs}), mirroring the results of \citet{DBLP:conf/emnlp/PeiJ21}. However, this pattern does not persist for findings in other parts of papers,  especially the limitations. Existing studies suggest that journalists might fail to report the limitations of scientific findings \citep{Fischhoff2012CommunicatingUF}, and our results here suggest that findings presented in limitations are more likely to be exaggerated and overstated. However, it is also possible that scientists may adopt different discourse strategies for different parts of a paper \citep{clark2013writing}. Nonetheless, our result obviates the necessity of analyzing the full text of a paper when studying science communication.

\subsection{Conclusion}

Faithful communication of scientific results is critical for disseminating new information and establishing public trust in science. Given the challenge of---and occasional failures in---communicating science, new resources and models are needed to evaluate how science is reported. Here, we introduce \textsc{Spiced}, a new science communication paraphrases dataset labeled with information similarity. Extensive experiments demonstrate that models can predict the degree to which two reports of a scientific finding have the same information but that this is a challenging task even for current SOTA pre-trained language models. In downstream applications, we show  \textsc{Spiced} improves model performance for evidence retrieval for scientific fact checking; and, using the trained model to perform a large-scale analysis of information change in science communication, we show systematic behaviors in how different people and news outlets faithfully convey scientific results. Data, code, and pretrained models are available at {\small \url{http://www.copenlu.com/publication/2022_emnlp_wright/}}. %

\section*{Acknowledgements}
$\begin{array}{l}\includegraphics[width=1cm]{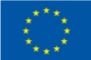} \end{array}$ This project has received funding from the European Union's Horizon 2020 research and innovation programme under the Marie Sk\l{}odowska-Curie grant agreement No 801199 and a Rackham Graduate Student Research Grant at the University of Michigan.

\section*{Limitations}
We note three limitations of our study.
Our data and analysis in social media is limited to only one platform, Twitter, and includes only tweets directly linked to the original paper, as indicated through Altmetric. While Twitter is among the largest social media platforms and is the most common in the Altmetric data, our data potentially omits other kinds of scientific communication about papers that do not directly link to a paper or tweets that link to a paper that cannot be easily identified to a DOI (e.g., linking to a PDF hosted on a personal website). Other types of tweets may be omitted from our dataset such as those written in a thread, or in a tweetorial, about a paper \cite{gero2021makes}, which may include additional tweets that describe a paper's findings. While our models would likely still be able to effectively analyze such tweets, these additional forms of scientific communication could add new variety. We leave identifying and collecting such tweets to future work. 

Second, our study focuses on only four large scientific fields. While these fields do cover a broad selection of papers, we were unable to annotate additional fields due to annotation budget and limitations from the Prolific platform. On Prolific, not all potential domains had sufficient numbers of qualified annotators (we required at least a Bachelor's degree in the domain) and the number of unique surveys to run scaled linearly with the number of domains, creating a significant human overhead. However, we will open source our annotation interface and pipeline and we encourage further efforts to build a larger dataset across more scientific domains.

Finally, while our models achieve moderately high performance at inferring the information matching (\autoref{fig:baseline-performance}), performance is not perfect, which potentially limits our ability in downstream models and tasks. While we show the data is still useful in training for related tasks (\Sref{sec:evidence-retrieval}) and a trained model can be used to identify systematic behavior by types of users and outlets (\Sref{sec:social-media-analysis}), more accurate models would likely be needed to identify any trends for finer-grained settings, such as looking at the behavior of a specific outlet. For this reason,w e have kept our analyses at a higher level (e.g., outlet categories).

\section*{Ethics and Impacts}

Miscommunication of scientific information can have negative impacts on many aspects of our society. Our study contributes to a large research program on the science of science communications \citep{national2017communicating}. Our dataset and model could be used to keep track of information change in science communication, enable large-scale analysis to understand the current science communication ecosystem, and finally help to facilitate better and more effective science communications.

\textbf{Crowdsourcing ethics} Annotating paired findings requires deep attention and may lead to annotator burnout. We carefully designed our annotation pipeline to provide a good annotation experience for the annotators. We designed a user-friendly Web-based annotation interface that allows annotators to do annotations using keyboard shortcuts. All the annotators are encouraged to leave comments and answer several questions about their annotation experience. More than 95\% of the annotators are satisfied with their annotation experience and many people suggest that our study helps them to better understand the science communication process\footnote{For example, one participant said ``Nice learning experience, Helps to understand the news can be far more different then the research paper cited''} and our annotation interface makes their task easier.\footnote{For example, one participant said ``i liked the option of using my keyboard, it made the experience more comfortable and efficient.''}

\newpage

\bibliography{references}
\bibliographystyle{acm}

\newpage

\appendix

\section{Appendices for Claim Check-Worthiness Detection as Positive Unlabelled Learning}

\subsection{Examples of PUC Improvements for Rumour Detection}
\label{sec:puc_examples}

Examples of improvements for rumour detection using \textit{PUC} can be found in \autoref{tab:pheme_pos_better}.
\begin{table*}[h]
    \centering
    \begin{tabular}{p{12cm} c c}
    \toprule
    \multicolumn{1}{c}{Rumour text} & nPUC & \multicolumn{1}{c}{nBaseline} \\
    \midrule
      Germanwings co-pilot had serious depressive episode: Bild newspaper http://t.co/RgSTrehD21 & 13 & 5\\
      \hline
      Now hearing 148 passengers + crew on board the \#A320 that has crashed in southern French Alps. \#GermanWings flight. @BBCWorld & 10 & 2 \\
      \hline
      It appears that \#Ferguson PD are trying to assassinate Mike Brown's character after literally assassinating Mike Brown. & 13 & 5 \\
      \hline
      \#Ferguson cops beat innocent man then charged him for bleeding on them: http://t.co/u1ot9Eh5Cq via @MichaelDalynyc http://t.co/AGJW2Pid1r & 9 & 2\\
    \bottomrule

    \end{tabular}
    \caption{Examples of rumours which the \textit{PUC} model judges correctly vs the baseline model with no pretraining on citation needed detection. n* is the number of models among the 15 seeds which predicted the correct label (rumour).}
    \label{tab:pheme_pos_better}
\end{table*}
\begin{table*}
    \centering
    \begin{tabular}{p{12cm} c c}
    \toprule
    \multicolumn{1}{c}{Non-Rumour text} & nPUC & \multicolumn{1}{c}{nBaseline} \\
    \midrule
      A female hostage stands by the front entrance of the cafe as she turns the lights off in Sydney. \#sydneysiege http://t.co/qNfCMv9yZt & 11 & 5\\
      \hline
      Map shows where gun attack on satirical magazine \#CharlieHebdo took place in central Paris http://t.co/5AZAKumpNd http://t.co/ECFYztMVk9 & 10 & 4 \\
      \hline
      "Hands up! Don't shoot!" \#ferguson https://t.co/svCE1S0Zek & 12 & 7 \\
      \hline
      Australian PM Abbott: Motivation of perpetrator in Sydney hostage situation is not yet known - @9NewsAUS http://t.co/SI01B997xf & 10 & 6\\
    \bottomrule

    \end{tabular}
    \caption{Examples of non-rumours which the \textit{PUC} model judges correctly vs the baseline model with no pretraining on citation needed detection. n* is the number of models among the 15 seeds which predicted the correct label (non-rumour).}
     \label{tab:pheme_neg_better}
\end{table*}

\subsection{Reproducibility}
\subsubsection{Computing Infrastructure}
All experiments were run on a shared cluster. Requested jobs consisted of 16GB of RAM and 4 Intel Xeon Silver 4110 CPUs. We used a single NVIDIA Titan X GPU with 12GB of RAM.

\subsubsection{Average Runtimes}
See \autoref{tab:checkworthy_runtimes} for model runtimes.
\begin{table*}
    \centering
    \fontsize{10}{10}\selectfont
    \begin{tabular}{l c c c}
    \toprule
    Method & Wikipedia & PHEME & Political Speeches\\
    \midrule
      \rule{0pt}{2ex}BERT&  34m30s& 14m25s& 8m11s\\
      BERT + PU& 40m7s& 20m40s&  15m38s\\
      BERT + \textit{PUC}& 40m8s& 21m20s& 15m32s\\
      BERT + Wiki& - & 14m28s& 8m50s\\
      BERT + WikiPU& -& 14m25s& 8m41s\\
      BERT + Wiki\textit{PUC}& -& 14m28s& 8m38s\\
      BERT + PU + WikiPU& -& 20m41s& 15m32s\\
      BERT + \textit{PUC} + WikiPUC& -& 21m52s& 15m40s\\
    \bottomrule

    \end{tabular}
    \caption{Average runtime of each tested system for each split of the data}
    \label{tab:checkworthy_runtimes}
\end{table*}

\subsubsection{Number of Parameters per Model}
We used BERT with a classifier on top for each model which consists of 109,483,778 parameters.

\subsubsection{Validation Performance}
Validation performances for the tested models are given in \autoref{tab:validation}.
\begin{table*}
    \centering
    \fontsize{10}{10}\selectfont
    \begin{tabular}{l c c c}
    \toprule
    Method & Wikipedia & PHEME & Political Speeches\\
    \midrule
      \rule{0pt}{2ex}BERT&  88.9& 81.6& 31.3\\
      BERT + PU& 89.0& 83.7& 18.2\\
      BERT + \textit{PUC}& 89.2& 82.8& 32.0\\
      BERT + Wiki& - & 80.8& 32.3\\
      BERT + WikiPU& -& 82.0& 35.7\\
      BERT + Wiki\textit{PUC}& -& 80.4& 34.3\\
      BERT + PU + WikiPU& -& 82.9& 33.3\\
      BERT + \textit{PUC} + WikiPUC& -& 84.1& 34.0\\
    \bottomrule

    \end{tabular}
    \caption{Validation F1 performances for each tested model.}
    \label{tab:validation}
\end{table*}

\subsubsection{Evaluation Metrics}
The primary evaluation metric used was F1 score. We used the sklearn implementation of \texttt{precision\_recall\_fscore\_support}, which can be found here:

\url{https://scikit-learn.org/stable/modules/generated/sklearn.metrics.precision_recall_fscore_support.html}. Briefly:
\begin{equation*}
  p = \frac{tp}{tp + fp} 
\end{equation*}
\begin{equation*}
  r = \frac{tp}{tp + fn} 
\end{equation*}
\begin{equation*}
  F1 = \frac{2*p*r}{p+r} 
\end{equation*}
where $tp$ are true positives, $fp$ are false positives, and $fn$ are false negatives.

Additionally, we used the mean average precision calculation from the Clef19 Check That! challenge for political speech data, which can be found here:

\url{https://github.com/apepa/clef2019-factchecking-task1/tree/master/scorer} Briefly:
\begin{equation*}
    \text{AP} = \frac{1}{|P|}\sum_{i}\frac{tp(i)}{i}
\end{equation*}
\begin{equation*}
    \text{mAP} = \frac{1}{|Q|}\sum_{q\in Q}\text{AP}(q)
\end{equation*}
where $P$ are the set of positive instances, $tp(i)$ is an indicator function which equals one when the $i$th ranked sample is a true positive, and $Q$ is the set of queries. In this work $Q$ consists of the ranking of statements from each split of the political speech data.

\subsubsection{Links to Data}
\begin{table}
    \centering
    \fontsize{10}{10}\selectfont
    \begin{tabular}{l c}
    \toprule
    Hyperparameter & Value \\
    \midrule
      Learning Rate& 3e-5 \\
      Weight Decay& 0.01 \\
      Batch Size& 8 \\
      Dropout& 0.1 \\
      Warmup Steps& 200 \\
      Epochs& 2 \\
    \bottomrule

    \end{tabular}
    \caption{Validation F1 performances used for each tested model.}
    \label{tab:hyperparams}
\end{table}
\begin{itemize}
    \item Citation Needed Detection~\cite{redi2019citation}:  
    
    \url{https://drive.google.com/drive/folders/1zG6orf0_h2jYBvGvso1pSy3ikbNiW0xJ}
    
    \item PHEME~\cite{zubiaga2016analysing}:
    
    \url{https://figshare.com/articles/PHEME_dataset_for_Rumour_Detection_and_Veracity_Classification/6392078}.
    
    \item Political Speeches: We use the same 7 splits as used in~\cite{hansen2019neural}. The first 5 can be found here: \url{http://alt.qcri.org/clef2018-factcheck/data/uploads/clef18_fact_checking_lab_submissions_and_scores_and_combinations.zip}. The files can be found under "task1\_test\_set/English/task1-en-file(3,4,5,6,7)". The last two files can be found here: \url{https://github.com/apepa/claim-rank/tree/master/data/transcripts_all_sources}. The files are ``clinton\_acceptance\_speech\_ann.tsv'' and ``trump\_inauguration
    \_ann.tsv''.
\end{itemize}

\subsubsection{Hyperparameters}
We found that good defaults worked well, and thus did not perform hyperparameter search. The hyperparameters we used are given in \autoref{tab:hyperparams}.

\section{Appendices for Generating Label Cohesive and Well-Formed Adversarial Claims}
\begin{table*}[th]
\centering
\fontsize{10}{10}\selectfont
\begin{tabular}{l l l l l}
\toprule
\textbf{Class} & \textbf{Trigger} & \textbf{F1} & \textbf{STS} & \textbf{PPL}\\ \midrule
\multicolumn{5}{c}{\bf FC Objective} \\
S$\rightarrow$R & only &  0.014 &  4.628 &  11.660 (36.191) \\
S$\rightarrow$R & nothing &  0.017 &  4.286 &  13.109 (56.882) \\
S$\rightarrow$R & nobody &  0.036 &  4.167 &  12.784 (37.390) \\
S$\rightarrow$NEI & neither &   0.047 &  3.901 &  11.509 (31.413) \\
S$\rightarrow$NEI & none &  0.071 &  4.016 &  13.136 (39.894) \\
S$\rightarrow$NEI & Neither &  0.155 &  3.641 &  11.957 (44.274) \\
R$\rightarrow$S & some &  0.687 &  4.694 &  11.902 (33.348) \\
R$\rightarrow$S & Sometimes &  0.724 &  4.785 &  10.813 (32.058) \\
R$\rightarrow$S & Some &  0.743 &  4.713 &  11.477 (37.243) \\
R$\rightarrow$NEI & recommended &  0.658 &  4.944 &  12.658 (36.658) \\
R$\rightarrow$NEI & Recommend &  0.686 &  4.789 &  10.854 (32.432) \\
R$\rightarrow$NEI & Supported &  0.710 &  4.739 &  11.972 (40.267) \\
NEI$\rightarrow$R & Only &  0.624 &   4.668 &  12.939 (57.666) \\
NEI$\rightarrow$R & nothing &  0.638 &  4.476 &   11.481 (48.781) \\
NEI$\rightarrow$R & nobody & 0.678 &  4.361 &  16.345 (111.60) \\
NEI$\rightarrow$S & nothing &  0.638 &  4.476 &  18.070 (181.85) \\
NEI$\rightarrow$S & existed &  0.800 &  4.950  &  15.552 (79.823) \\
NEI$\rightarrow$S & area &  0.808 &  4.834  &  13.857 (93.295) \\

\midrule
\multicolumn{5}{c}{\bf FC+STS Objectives} \\
S$\rightarrow$R & never & 0.048 & 4.267 & 12.745 (50.272) \\
S$\rightarrow$R & every & 0.637 & 4.612 & 13.714 (51.244) \\
S$\rightarrow$R & didn & 0.719 & 4.986 & 12.416 (41.080) \\
S$\rightarrow$NEI & always  & 0.299 &  4.774 &  11.906 (35.686) \\
S$\rightarrow$NEI & every & 0.637 & 4.612 & 12.222 (38.440) \\
S$\rightarrow$NEI & investors & 0.696 & 4.920 & 12.920 (42.567) \\
R$\rightarrow$S & over &  0.761 &  4.741 &  12.139 (33.611) \\
R$\rightarrow$S & about &  0.765 &   4.826 &  12.052 (37.677) \\
R$\rightarrow$S & her &   0.774 &   4.513 &   12.624 (41.350) \\
R$\rightarrow$NEI & top &  0.757 &  4.762 &  12.787 (39.418) \\
R$\rightarrow$NEI & also &   0.770 &   5.034 &   11.751 (35.670) \\
R$\rightarrow$NEI & when &   0.776 &   4.843 &   12.444 (37.658) \\
NEI$\rightarrow$R & only &  0.562 &  4.677 &  14.372 (83.059) \\
NEI$\rightarrow$R & there &   0.764 & 4.846 &    11.574 (42.949) \\
NEI$\rightarrow$R & just &   0.786 & 4.916 &   16.879 (135.73) \\
NEI$\rightarrow$S & of&   0.802 & 4.917 &  11.844 (55.871) \\
NEI$\rightarrow$S & is &   0.815 & 4.931 & 17.507 (178.55) \\
NEI$\rightarrow$S & A &   0.818 & 4.897 & 12.526 (67.880) \\

\bottomrule
\end{tabular}
\caption{Top-3 triggers found with the Universal Adversarial Triggers methods. The triggers are generated given claims from a source class (column \textit{Class}), so that the classifier is fooled to predict a different target class. The classes are SUPPORTS (S), REFUTES (R), NOT ENOUGH INFO (NEI).}
\label{tab:evalonetrig}
\end{table*}

\subsection{Implementation Details}
\label{sec:appendixA}
\textbf{Models}. The RoBERTa FC model (125M parameters) is fine-tuned with a batch size of 8, learning rate of 2e-5 and for a total of 4 epochs, where the epoch with the best performance is saved. We used the implementation provided by HuggingFace library. We performed a grid hyper-parameter search for the learning rate between the values 1e-5, 2e-5, and 3e-5. The average time for training a model with one set of hyperparameters is 155 minutes ($\pm3$). The average accuracy over the different hyperparameter runs is 0.862($\pm$ 0.005) F1 score on the validation set.

For the models that measure the perplexity and the semantical similarity we use the pretrained models provided by HuggingFace-- RoBERTa large model (125M parameters) fine tuned on the STS-b task and RoBERTa base model (355M parameters) pretrained on a LM objective.

We used the HuggingFace implementation of the small GPT-2 model, which consists of 124M parameters and is fine-tuned with a batch size of 4, learning rate of 3e-5, and for a total of 20 epochs. We perform early stopping on the loss of the model on a set of validation data. The average validation loss is 0.910. The average runtime for training one of the models is 31 hours and 28 minutes.

We note that, the intermediate models used in this work and described in this section, are trained on large relatively general-purpose datasets. While, they can make some mistakes, they work well enough and using them, we don't have to rely on additional human annotations for the intermediate task.

\paragraph{Adversarial Triggers.} The adversarial triggers are generated based on instances from the validation set. We run the algorithm for three epochs to allow for the adversarial triggers to converge. At each epoch the initial trigger is updated with the best performing trigger for the epoch (according to the loss of the FC or FC+STS objective). At the last step, we select only the top 10 triggers and remove any that have a negative loss. We choose the top 10 triggers as those are the most potent ones, adding more than top ten of the triggers preserves the same tendencies in the results, but smooths them as further down the list of adversarial attacks, the triggers do not decrease the performance of the model substantially. This is also supported by related literature~\cite{wallace2019universal}, where only the top few triggers are selected.

The adversarial triggers method is run for 28.75 ($\pm$ 1.47) minutes for with the FC objective and 168.6($\pm$ 28.44) minutes for the FC+STS objective. We perform the trigger generation with a batch size of four. We additionally normalize the loss for each objective to be in the range [0,1] and also re-weight the losses with a wieht of 0.6 for the FC loss and a weight of 0.4 for the SST loss as when generated with an equal weight, the SST loss tends to preserve the same initial token in all epochs.

\paragraph{Datasets} 
The datasets used for training the FC model consist of 161,249 SUPPORTS, 60,227 REFUTES, and 69,885 NEI claims for the training split; 6,207 SUPPORTS, 6,235 REFUTES, and 6,554 NEI claims for the dev set; 6,291 SUPPORTS, 5,992 REFUTES, and 6522 NEI claims. The evidence for each claim is the gold evidence provided from the FEVER dataset, which is available for REFUTES and SUPPORTS claims. When there is more than one annotation of different evidence sentences for an instance, we include them as separate instances in the datasets. For NEI claims, we use the system of \citet{malon2018team} to retrieve evidence sentences. 

\subsection{Top Adversarial Triggers}
Table~\ref{tab:evalonetrig} presents the top adversarial triggers for each direction found with the Universal Adversarial Triggers method. It offers an additional way of estimating the effectiveness of the STS objective by comparing the number of negation words generated by the basic model (8) and the STS objective (2) in the top-3 triggers for each direction.
\label{sec:appendixC}

\subsection{Supplemental Material}
\label{sec:supplemental}

\subsubsection{Computing Infrastructure}
All experiments were run on a shared cluster. Requested jobs consisted of 16GB of RAM and 4 Intel Xeon Silver 4110 CPUs. We used two NVIDIA Titan RTX GPUs with 12GB of RAM for training GPT-2 and one NVIDIA Titan X GPU with 8GB of RAM for training the FC models and finding the universal adversarial triggers.

\subsubsection{Evaluation Metrics}
The primary evaluation metric used was macro-F1 score. We used the sklearn implementation of \texttt{precision\_recall\_fscore\_support}, which can be found here: \url{https://scikit-learn.org/stable/modules/generated/sklearn.metrics.precision_recall_fscore_support.html}. Briefly:
\begin{equation*}
   p = \frac{tp}{tp + fp} 
\end{equation*}
\begin{equation*}
   r = \frac{tp}{tp + fn} 
\end{equation*}
\begin{equation*}
   F1 = \frac{2*p*r}{p+r} 
\end{equation*}
where $tp$ are true positives, $fp$ are false positives, and $fn$ are false negatives.

\subsubsection{Manual Evaluation}
\label{app:B3}
After generating the claims, two independent annotators label the overall claim quality (score of 1-5) and the true label for the claim. The inter-annotator agreement for the quality label using Krippendorff's alpha is 0.54 for the quality score and 0.38 for the claim label. Given this, we take the average of the two annotator's scores for the final quality score and have a third expert annotator examine and select the best label for each contested claim label.

\section{Appendices for Transformer Based Multi-Source Domain Adaptation}

\subsection{BERT Domain Adversarial Training Results}
\label{sec:bert_appendix}
Additional results on domain adversarial training with Bert can be found in \autoref{tab:bert_appendix_results}.

\begin{table*}[th]
    \centering
    \fontsize{10}{10}\selectfont
    \begin{tabular}{l c c c c c | c c c c c c c }
    \toprule %
     Method &\multicolumn{5}{c}{Sentiment Analysis (Accuracy)}&\multicolumn{6}{c}{Rumour Detection (F1)}\\
    \cmidrule(lr){2-6}
    \cmidrule(lr){7-12}
     & D & E & K & B & macroA & CH & F & GW & OS & S & $\mu$F1\\
    \midrule
        Bert & 90.3& 91.6& 91.7& 90.4& 91.0& 66.4& 46.2& 68.3& 67.3& 62.3& 63.3\\
        \midrule
        Bert-Adv-12 & 89.8& 91.4& 91.2& 90.1& 90.6 & 66.6& 47.8& 62.5& 65.3& 62.8& 62.5\\
        Bert-Adv-4 & 89.9& 91.1& 91.7& 90.4& 90.8& 65.6& 43.6& 71.0& 68.1& 60.8& 62.8\\
    \bottomrule %

    \end{tabular}
    \caption{Experiments for sentiment analysis in (D)VD, (E)lectronics, (K)itchen and housewares, and (B)ooks domains and rumour detection for different events ((C)harlie(H)ebdo, (F)erguson, (G)erman(W)ings, (O)ttawa(S)hooting, and (S)ydneySiege) using leave-one-out cross validation for BERT. Results are averaged across 3 random seeds. The results for sentiments analysis are in terms of accuracy and the results for rumour detection are in terms of F1.}
    \label{tab:bert_appendix_results}
\end{table*}

\subsection{Reproducibility}

\subsubsection{Computing Infrastructure}
All experiments were run on a shared cluster. Requested jobs consisted of 16GB of RAM and 4 Intel Xeon Silver 4110 CPUs. We used a single NVIDIA Titan X GPU with 12GB of RAM.

\subsubsection{Average Runtimes}
The average runtime performance of each model is given in \autoref{tab:msda_runtimes}. Note that different runs may have been placed on different nodes within a shared cluster, thus why large time differences occurred. 
\begin{table*}[th]
    \centering
    \fontsize{10}{10}\selectfont
    \begin{tabular}{l c c}
    \toprule %
     Method &Sentiment Analysis &Rumour Detection\\
    \midrule
        Basic & 0h44m37s& 0h23m52s\\
        \midrule
        Adv-6 & 0h54m53s& 0h59m31s\\
        Adv-3 & 0h53m43s& 0h57m29s\\
        \midrule
        Independent-Avg & 1h39m13s& 1h19m27\\
        Independent-Ft & 1h58m55s& 1h43m13\\
        MoE-Avg & 2h48m23s& 4h03m46s\\
        MoE-Att & 2h49m44s& 4h07m3s\\
        MoE-Att-Adv-6 & 4h51m38s& 4h58m33s\\
        MoE-Att-Adv-3 & 4h50m13s& 4h54m56s\\
        MoE-DC & 3h23m46s& 4h09m51s\\
    \bottomrule %

    \end{tabular}
    \caption{Average runtimes for each model on each dataset (runtimes are taken for the entire run of an experiment).}
    \label{tab:msda_runtimes}
\end{table*}

\subsubsection{Number of Parameters per Model}
The number of parameters in each model is given in \autoref{tab:msda_num_params}.
\begin{table*}[th]
    \centering
    \fontsize{10}{10}\selectfont
    \begin{tabular}{l c c}
    \toprule %
     Method &Sentiment Analysis &Rumour Detection\\
    \midrule
        Basic & 66,955,010& 66,955,010\\
        \midrule
        Adv-6 & 66,958,082& 66,958,850\\
        Adv-3 & 66,958,082& 66,958,850\\
        \midrule
        Independent-Avg & 267,820,040& 334,775,050\\
        Independent-Ft & 267,820,040& 334,775,050\\
        MoE-Avg & 267,820,040& 334,775,050\\
        MoE-Att & 268,999,688& 335,954,698\\
        MoE-Att-Adv-6 & 269,002,760& 335,958,538\\
        MoE-Att-Adv-3 & 269,002,760& 335,958,538\\
        MoE-DC & 267,821,576& 334,777,354\\
    \bottomrule %

    \end{tabular}
    \caption{Number of parameters in each model}
    \label{tab:msda_num_params}
\end{table*}

\subsubsection{Validation Performance}
The validation performance of each tested model is given in \autoref{tab:msda_val_performance}.
\begin{table*}[th]
    \centering
    \fontsize{10}{10}\selectfont
    \begin{tabular}{l c c}
    \toprule %
     Method &Sentiment Analysis (Acc) &Rumour Detection (F1)\\
    \midrule
        Basic & 91.7& 82.4\\
        \midrule
        Adv-6 & 91.5& 83.3\\
        Adv-3 & 91.2& 83.4\\
        \midrule
        Independent-Avg & 92.7& 82.8\\
        Independent-Ft & 92.6& 82.5\\
        MoE-Avg & 92.2& 83.5\\
        MoE-Att & 92.0& 83.3\\
        MoE-Att-Adv-6 & 91.2& 83.3\\
        MoE-Att-Adv-3 & 91.4& 82.8\\
        MoE-DC & 89.8& 84.6\\
    \bottomrule %

    \end{tabular}
    \caption{Average validation performance for each of the models on both datasets.}
    \label{tab:msda_val_performance}
\end{table*}

\subsubsection{Evaluation Metrics}
The primary evaluation metrics used were accuracy and F1 score. For accuracy, we used our implementation provided with the code. The basic implementation is as follows.
\begin{equation*}
    \text{accuracy} = \frac{tp + tn}{tp+fp+tn+fn}
\end{equation*}
We used the sklearn implementation of \texttt{precision\_recall\_fscore\_support} for F1 score, which can be found here: \url{https://scikit-learn.org/stable/modules/generated/sklearn.metrics.precision_recall_fscore_support.html}. Briefly:
\begin{equation*}
   p = \frac{tp}{tp + fp} 
\end{equation*}
\begin{equation*}
   r = \frac{tp}{tp + fn} 
\end{equation*}
\begin{equation*}
   F1 = \frac{2*p*r}{p+r} 
\end{equation*}
where $tp$ are true positives, $fp$ are false positives, and $fn$ are false negatives.

\subsubsection{Hyperparameters}
We performed and initial hyperparameter search to obtain good hyperparameters that we used across models. The bounds for each hyperparameter was as follows:
\begin{itemize}
    \item Learning rate: [0.00003, 0.00004, 0.00002, 0.00001, 0.00005, 0.0001, 0.001].
    \item Weight decay: [0.0, 0.1, 0.01, 0.005, 0.001, 0.0005, 0.0001].
    \item Epochs: [2, 3, 4, 5, 7, 10].
    \item Warmup steps: [0, 100, 200, 500, 1000, 5000, 10000].
    \item Gradient accumulation: [1,2]
\end{itemize}
We kept the batch size at 8 due to GPU memory constraints and used gradient accumulation instead. We performed a randomized hyperparameter search for 70 trials. Best hyperparameters are chosen based on validation set performance (accuracy for sentiment data, F1 for rumour detection data). The final hyperparameters selected are as follows:
\begin{itemize}
    \item Learning rate: 3e-5.
    \item Weight decay: 0.01.
    \item Epochs: 5.
    \item Warmup steps: 200.
    \item Batch Size: 8
    \item Gradient accumulation: 1
\end{itemize}
Additionally, we set the objective weighting parameters to $\lambda = 0.5$ for the mixture of experts models and $\gamma = 0.003$ for the adversarial models, in line with previous work~\cite{guo2018multi,li2018s}.

\subsubsection{Links to data}
\begin{itemize}
    \item Amazon Product Reviews~\cite{blitzer2007biographies}:  \url{https://www.cs.jhu.edu/~mdredze/datasets/sentiment/}
    
    \item PHEME~\cite{zubiaga2016analysing}: \url{https://figshare.com/articles/PHEME_dataset_for_Rumour_Detection_and_Veracity_Classification/6392078}.
    
\end{itemize}

\section{Appendices for Multi-View Knowledge Distillation from Crowd Annotations for Out of Domain Generalization}
\subsection{Evaluation Metrics}
\label{sec:eval_metrics}
\paragraph{F1}
We used the sklearn implementation of \texttt{precision\_recall\_fscore\_support} for F1 score, which can be found here: \url{https://scikit-learn.org/stable/modules/generated/sklearn.metrics.precision_recall_fscore_support.html}. Briefly:
\begin{equation*}
   p = \frac{tp}{tp + fp} 
\end{equation*}
\begin{equation*}
   r = \frac{tp}{tp + fn} 
\end{equation*}
\begin{equation*}
   F1 = \frac{2*p*r}{p+r} 
\end{equation*}
where $tp$ are true positives, $fp$ are false positives, and $fn$ are false negatives.

\paragraph{Calibrated Log-Likelihood} The calibrated log-likelihood is defined in \citet{DBLP:conf/iclr/AshukhaLMV20} as a method to fairly compare uncertainty estimation between models on the same test set. The key observation is that in order to obtain a fair comparison, one must first perform temperature scaling at the optimal temperature on the classifier output for each model under comparison. Additionally, this temperature must be optimized on an in-domain validation set. The procedure to calculate the calibrated log-likelihood is:
\begin{enumerate}
    \item Split the \textbf{test set} in half, one half for validation and one half for test.
    \item Optimize a temperature parameter $T$ to minimize the average negative log-likelihood $-\frac{1}{n}\sum_{i}\log \tilde{p}(y_{i}=y^{*}_{i} | x_{i})$, where $\tilde{p}_{i} = \text{softmax}(\frac{l_{i}}{T})$ and $l_{i}$ is the logits of the classifier, on the validation half of the test set.
    \item Measure the temperature scaled log-likelihood on the test half of the test set.
\end{enumerate}
Following the suggestion from \citet{DBLP:conf/iclr/AshukhaLMV20}, we run this procedure 5 times on different splits of the test set and take the average test-half log-likelihood as the result.

\subsection{Visualization} 
\begin{figure*}[th]
         \centering
     \begin{subfigure}[b]{0.4\textwidth}
         \centering
         \includegraphics[width=\textwidth]{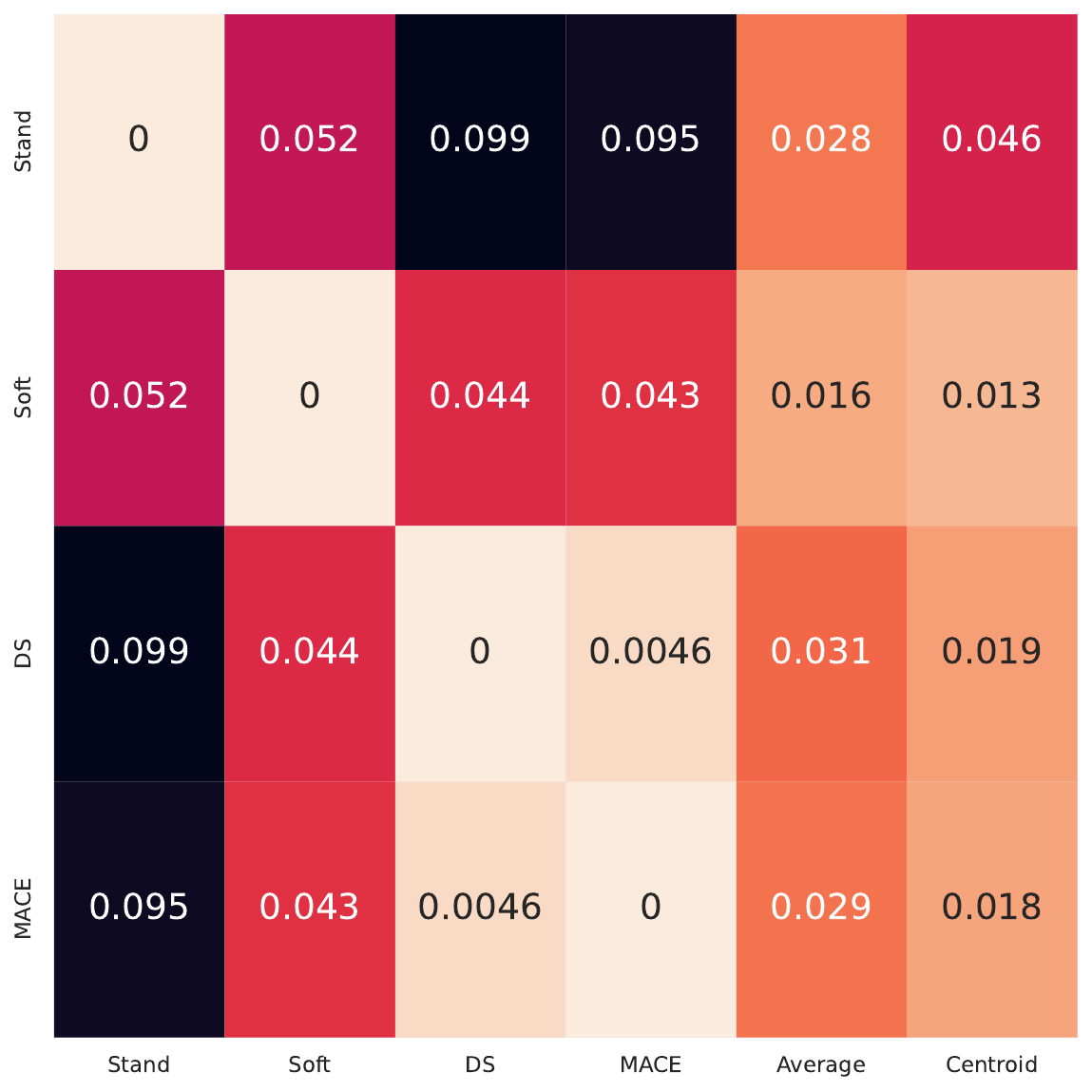}
         \caption{RTE}
         \label{fig:rte-js_heatmap}
     \end{subfigure}
     \hfill
     \begin{subfigure}[b]{0.4\textwidth}
         \centering
         \includegraphics[width=\textwidth]{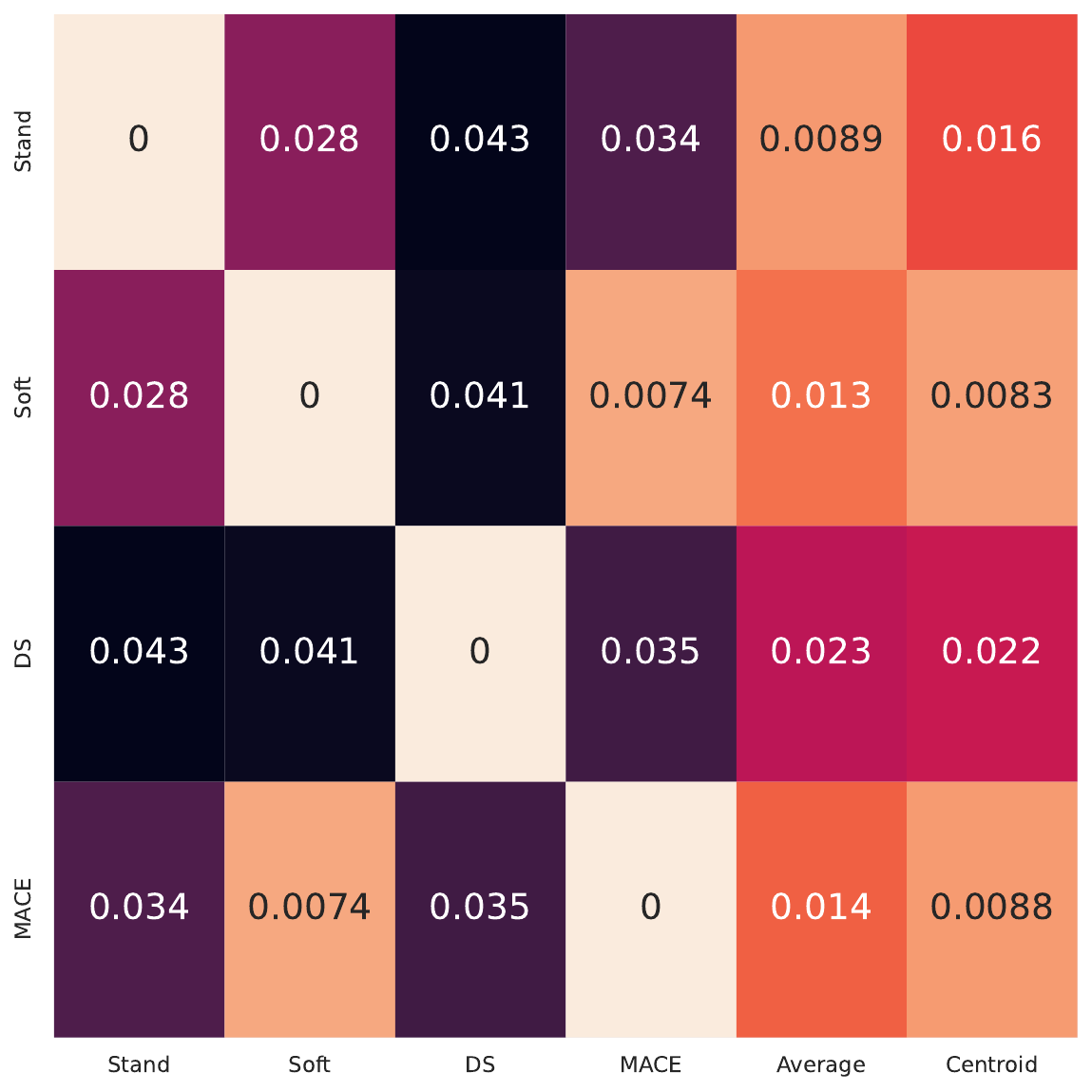}
         \caption{MRE}
         \label{fig:mre-js_heatmap}
     \end{subfigure} 
     \\
     \begin{subfigure}[b]{0.4\textwidth}
         \centering
         \includegraphics[width=\textwidth]{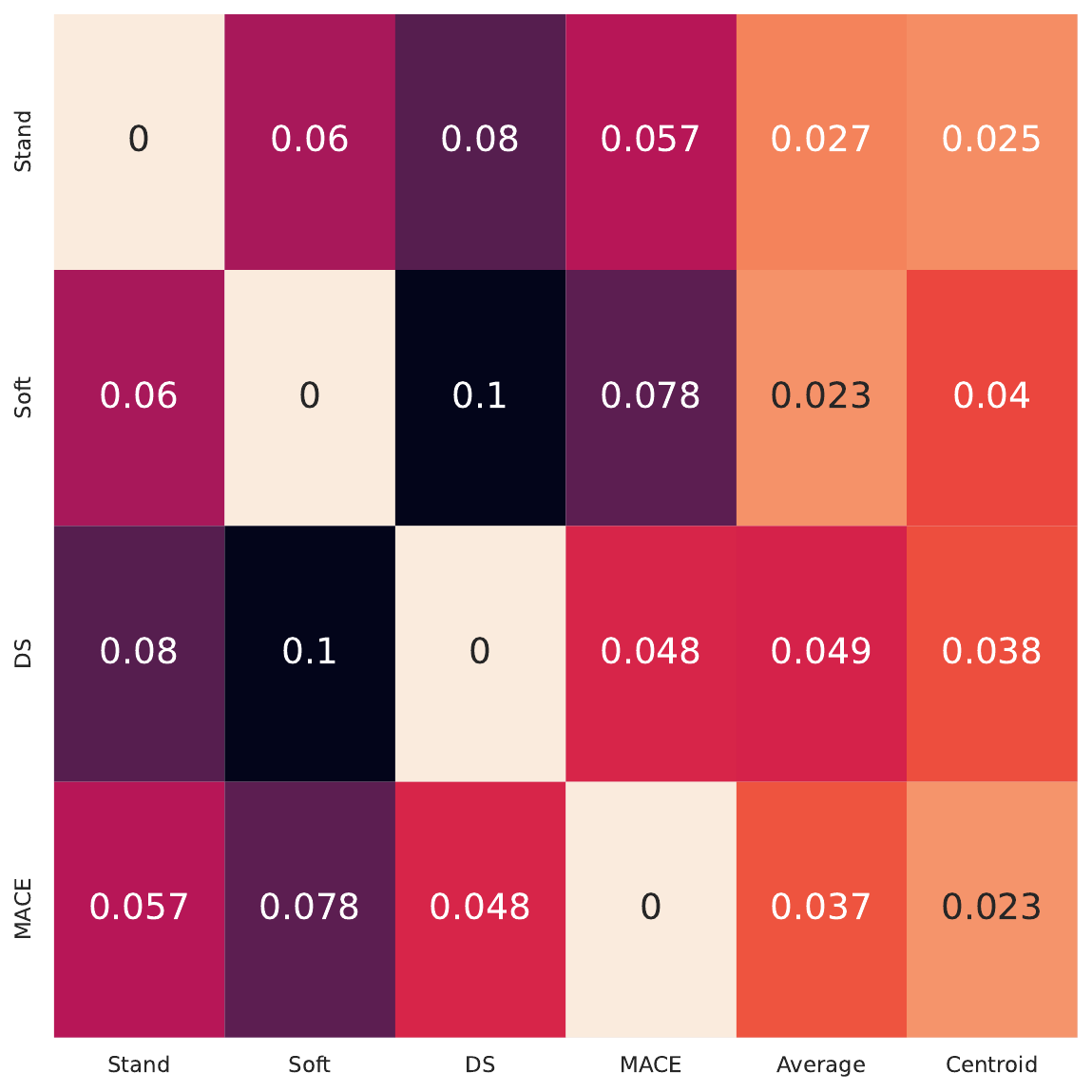}
         \caption{POS}
         \label{fig:pos-js_heatmap}
     \end{subfigure}
     \hfill
     \begin{subfigure}[b]{0.4\textwidth}
         \centering
         \includegraphics[width=\textwidth]{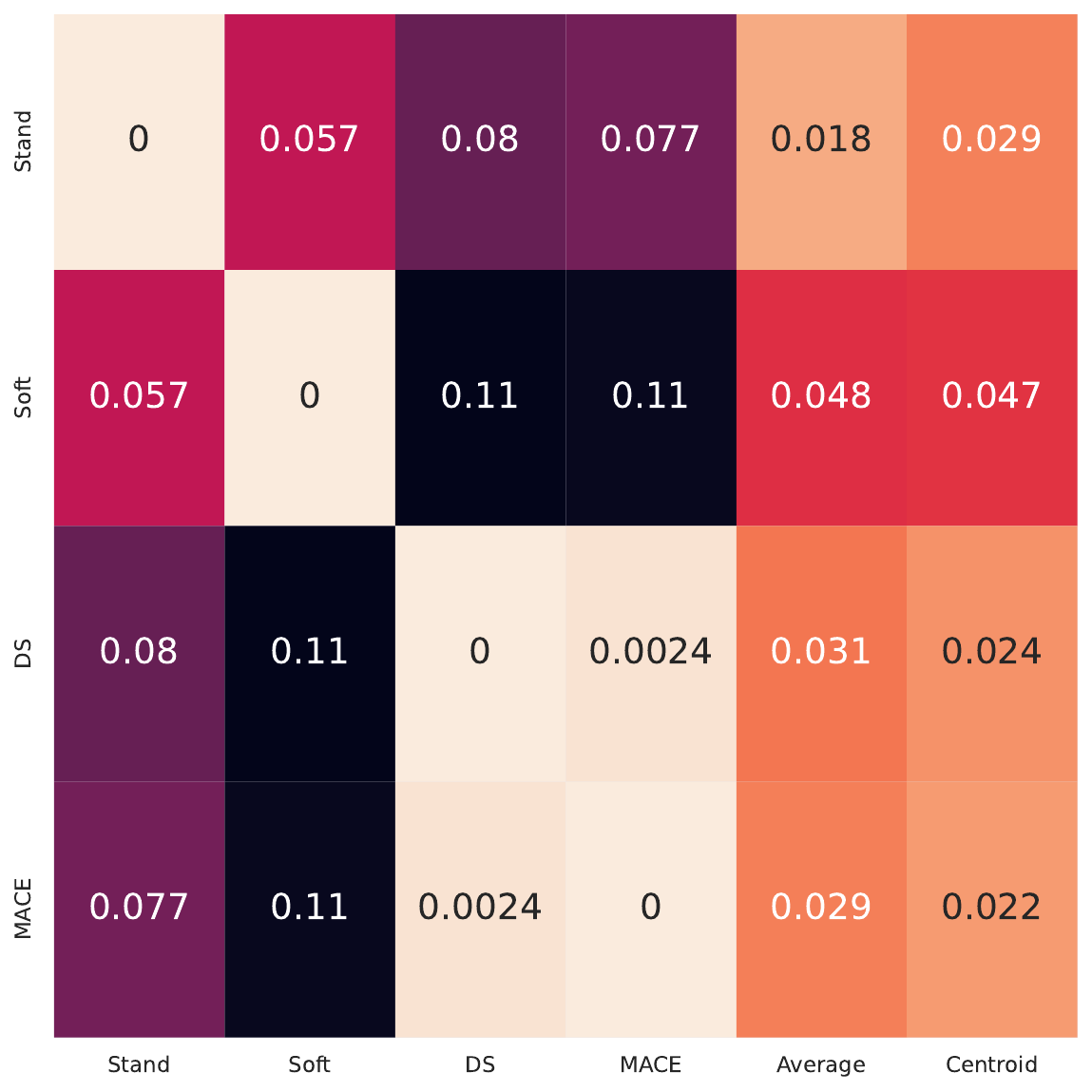}
         \caption{Toxicity}
         \label{fig:Jigsaw-js_heatmap}
     \end{subfigure} 
 
    \caption{Heatmaps of the average Jensen-Shannon divergence between individual soft labeling methods and average and JS centroid aggregation for (a) RTE, (b) MRE, (c) POS, and (d) Toxicity datasets.}
    \label{fig:divergence_heatmap}
\end{figure*}

Here we plot the JSD between individual methods and the averaging and JSC methods for each dataset in \autoref{fig:divergence_heatmap}.

\section{Appendices for CiteWorth: Cite-Worthiness Detection for Improved Scientific Document Understanding}

\subsection{List of Permissible Section Titles}
\label{sec:premissible-section-titles}
\begin{itemize}[noitemsep]
    \item introduction
\item abstract
\item method
\item methods
\item results
\item discussion
\item discussions
\item conclusion
\item conclusions
\item results and discussion
\item related work
\item experimental results
\item literature review
\item experiments
\item background
\item methodology
\item conclusions and future work
\item related works
\item limitations
\item procedure
\item material and methods
\item discussion and conclusion
\item implementation
\item evaluation
\item performance evaluation
\item experiments and results
\item overview
\item experimental design
\item discussion and conclusions
\item results and discussions
\item motivation
\item proposed method
\item analysis
\item future work
\item results and analysis
\item implementation details
\end{itemize}
\subsection{List of Regular Expressions}
\label{sec:regexes}
Citation format regexes:
\begin{itemize}
    \item \texttt{\textbackslash[([0-9]+\textbackslash s*[,-;]*\textbackslash s*)*[0-9]+\textbackslash s*\textbackslash]}
    \item \texttt{\textbackslash(?[12][0-9]{3}[a-z]?\textbackslash s*\textbackslash)}
\end{itemize}
Hanging citation regex:
\texttt{\textbackslash s+\textbackslash(?(\textbackslash(\textbackslash s*\textbackslash)|like|reference|\\including|include|with|for instance|for example|see also|at|following|of|from|to|in|by|\\see|as|e\textbackslash .?g\textbackslash .?(,)?|viz(\textbackslash .)?(,)?)\textbackslash s*\\(,)*(-)*[\textbackslash)\textbackslash]]?\textbackslash s*[.?!]\textbackslash s*\$}

\subsection{Reproducibility}
\subsubsection{Computing Infrastructure}
All experiments were run on a shared cluster. Requested jobs consisted of 16GB of RAM and 4 Intel Xeon Silver 4110 CPUs. We used a single NVIDIA Titan X GPU with 12GB of RAM.

\subsubsection{Average Runtimes}
The average runtime performance of each model is given in \autoref{tab:runtimes_citeworth}. Note that different runs may have been placed on different nodes within a shared cluster. 
\begin{table}[th]
    \centering
    \fontsize{10}{10}\selectfont
    \begin{tabular}{l c c}
    \toprule %
     Setting & Time\\
    \midrule
        Logistic Regression & 00h01m43s\\
        Transformer & 02h55m13s\\
        BERT & 05h30m30s\\
        SciBERT (no weighting) & 09h22m00s\\
        SciBERT & 09h32m37s\\
        SciBERT + PU & 16h01m27s\\
        Longformer-Solo & 75h27m22s\\
        Longformer-Ctx & 19h16m07s\\
    \bottomrule %

    \end{tabular}
    \caption{Average runtimes for each model (runtimes are taken for the entire run of an experiment).}
    \label{tab:runtimes_citeworth}
\end{table}

\subsubsection{Number of Parameters per Model}
The number of parameters in each model is given in \autoref{tab:num_params}.
\begin{table}[th]
    \centering
    \fontsize{10}{10}\selectfont
    \begin{tabular}{l c c}
    \toprule %
     Method & \# Parameters\\
    \midrule
        Logistic Regression & 198,323\\
        Transformer & 9,789,042\\
        BERT & 109,484,290\\
        SciBERT & 109,920,514\\
        Longformer & 149,251,586\\
    \bottomrule %

    \end{tabular}
    \caption{Number of parameters in each model}
    \label{tab:num_params}
\end{table}

\subsubsection{Validation Performance}
The validation performance of each tested model is given in \autoref{tab:val_performance}.
\begin{table}[th]
    \centering
    \fontsize{10}{10}\selectfont
    \begin{tabular}{l c c}
    \toprule %
     Method & F1\\
    \midrule
        Logistic Regression & -\\
        Transformer & 57.02\\
        BERT & 60.75\\
        SciBERT (no weighting) & 57.52\\
        SciBERT & 62.04\\
        SciBERT + PU & 61.43\\
        Longformer-Solo & 61.67\\
        Longformer-Ctx & 67.11\\
    \bottomrule %

    \end{tabular}
    \caption{Average validation performance for each of the models.}
    \label{tab:val_performance}
\end{table}

\subsubsection{Evaluation Metrics}
The primary evaluation metric used was F1 score.
We used the sklearn implementation of \texttt{precision\_recall\_fscore\_support} for F1 score, which can be found here: \url{https://scikit-learn.org/stable/modules/generated/sklearn.metrics.precision_recall_fscore_support.html}. Briefly:
\begin{equation*}
   p = \frac{tp}{tp + fp} 
\end{equation*}
\begin{equation*}
   r = \frac{tp}{tp + fn} 
\end{equation*}
\begin{equation*}
   F1 = \frac{2*p*r}{p+r} 
\end{equation*}
where $tp$ are true positives, $fp$ are false positives, and $fn$ are false negatives.

\subsubsection{Hyperparameters}
\label{sec:hyperparams}
\paragraph{Logistic Regression} We used a C value of 0.1151 for logistic regression.

\paragraph{Basic Transformer} The final hyperparameters for the basic Transformer model are: batch size: 64; number of epochs: 33; feed-forward dimension: 128; learning rate: 0.0001406; number of heads: 3; number of layers: 5; weight decay: 0.1; dropout probability: 0.4. We performed a Bayesian grid search over the following ranges of values, optimizing validation F1 performance: learning rate: $[0.000001, 0.001]$; batch size: $\{4, 8, 16, 32, 64, 128\}$; weight decay: $\{0.0, 0.0001, 0.001, 0.01, 0.1\}$; dropout probability: $\{0.0, 0.1, 0.2, 0.3, 0.4, 0.5\}$; number of epochs: $[2, 40]$; feed-forward dimension: $\{128, 256, 512, 1024, 2048\}$; number of heads: $\{1, 2, 3, 4, 5, 6, 10, 12\}$; number of layers: $[1, 12]$.

\paragraph{BERT} The final hyperparameters for BERT are: batch size: 8; number of epochs: 3; learning rate: 0.000008075; triangular learning rate warmup steps: 300; weight decay: 0.1; dropout probability: 0.1. We performed a Bayesian grid search over the following ranges of values, optimizer validation F1 performance: learning rate: $[0.0000001, 0.0001]$; triangular learning rate warmup steps: $\{0, 100, 200, 300, 400, 500, 1000, 1500, 2000,$ $2500, 5000\}$; batch size: $\{4, 8\}$; weight decay: $\{0.0, 0.0001, 0.001, 0.01, 0.1\}$; number of epochs: $[2, 40]$.

\paragraph{SciBERT} The final hyperparameters for SciBERT are: batch size: 4; number of epochs: 3; learning rate: 0.000001351; triangular learning rate warmup steps: 300; weight decay: 0.1; dropout probability: 0.1. We performed a Bayesian grid search over the following ranges of values, optimizer validation F1 performance: learning rate: $[0.0000001, 0.0001]$; triangular learning rate warmup steps: $\{0, 100, 200, 300, 400, 500, 1000, 1500, 2000,$ $2500, 5000\}$; batch size: $\{4, 8\}$; weight decay: $\{0.0, 0.0001, 0.001, 0.01, 0.1\}$; number of epochs: $[2, 40]$.

\paragraph{Longformer-Ctx} The final hyperparameters for Longformer-Ctx are: batch size: 4; number of epochs: 3; learning rate: 0.00001112; triangular learning rate warmup steps: 300; weight decay: 0.0; dropout probability: 0.1. We performed a Bayesian grid search over the following ranges of values, optimizer validation F1 performance: learning rate: $[0.0000001, 0.0001]$; triangular learning rate warmup steps: $\{0, 100, 200, 300, 400, 500, 1000, 1500, 2000,$ $2500, 5000\}$; batch size: $\{4, 8\}$; weight decay: $\{0.0, 0.0001, 0.001, 0.01, 0.1\}$; number of epochs: $[2, 6]$.

\subsubsection{Data}
\citeworthdataset{}~is constructed from the S2ORC dataset, which can be found here: \url{https://github.com/allenai/s2orc}. In particular, \citeworthdataset{}~is built using the \texttt{20200705v1} release of the data. A link to the \citeworthdataset{}~data can be found here: \url{https://github.com/copenlu/cite-worth}.

\section{Appendices for Generating Scientific Claims for Zero-Shot Scientific Fact Checking}

\subsection{Reproducibility}
\subsubsection{Computing Infrastructure}

All experiments were run on an Amazon Web Services p3.2xlarge instance using a Tesla V100 GPU with 16GB of RAM.

\subsubsection{Number of Parameters per Model}
\begin{table}[t]
    \centering
    \fontsize{10}{10}\selectfont
    \begin{tabular}{l c}
    \toprule %
     Model & \# Params\\
    \midrule
        RoBERTa & 125M \\
        BART & 140M \\
        GPT-2 & 125M \\
        Longformer-SciFact & 438M \\
    \bottomrule %

    \end{tabular}
    \caption{Model sizes.}
    \label{tab:model_sizes}
\end{table}
The sizes of each of the models used in this work are given in \autoref{tab:model_sizes}.

\subsubsection{Hyperparameters}
\label{sec:hyperparams}
\subsubsubsection{Fact Checking}
\paragraph{SciFact data} Learning rate: 1e-5, 5 epochs, gradient accumulation for 8 batches, 1 sample per training batch, 16-bit precision, 809 total claims.

\paragraph{FEVER threshold} We tune the NEI threshold on the training set of SciFact, testing values in the range [1e-5, 2e-5, 3e-5, 4e-5, 5e-5, 1e-4, 2e-4, 3e-4, 4e-4, 5e-4, 1e-3, 2e-3, 3e-3, 4e-3, 5e-5, 0.01, 0.12, 0.2, 0.25, 0.4, 0.5, 0.75, 0.8, 0.8, 0.99, 0.999] and find that 5e-5 produces the best result.

\paragraph{\cgbart} Learning rate: 2e-6, 5 epochs, gradient accumulation for 8 batches, 1 sample per training batch, 16-bit precision, 1,561 total training claims.

\paragraph{\cgentity} Learning rate: 4e-8, 5 epochs, gradient accumulation for 8 batches, 1 sample per training batch, 16-bit precision, 8,592 total training claims.

\subsubsubsection{\cgbart} Learning rate: 2e-5, 3 epochs, linear warmup for 200 steps followed by linear decay, weight decay of 0.01, batch size of 8.

\subsubsection{Description of Datasets}
\label{sec:data_info}
We use a variety of datasets in this study for different components of models, training, and testing. Here we provide a description of each and in which module the dataset is used.

\paragraph{SciFact}
The SciFact dataset and rewritten claims used to train \cgbart can be found at \href{https://github.com/allenai/scifact}{https://github.com/allenai/scifact}. The dataset consists of 585 original citances with rewritten claims for each of them. Each citance consists of 1-2 rewritten claims. The SciFact rewritten claims are used to train \cgbart for direct claim generation. Additionally, SciFact contains biomedical claims paired with evidence abstracts and veracity labels in \{\textit{supports}, \textit{refutes}, \textit{not enough info}\} and is split into train, dev, and test sets. We use the train set for supervised fact checking experiments, and the dev set for testing since the test set does not come with labels.

\paragraph{FEVER}
FEVER is a general domain fact checking dataset built from Wikipedia. Like SciFact, the dataset consists of claims with paired evidence documents with labels in \{\textit{supports}, \textit{refutes}, \textit{not enough info}\}. FEVER is used as pretraining data for our fact checking models for zero-shot transfer to biomedical claims. The dataset can be found here \href{https://fever.ai/resources.html}{https://fever.ai/resources.html}.

\paragraph{MedMentions}
The MedMentions dataset is a dataset of 4,392 biomedical papers annotated with mentions of UMLS entities. It is used to train the named entity recognition and normalization models used by ScispaCy, which we used for named entity recognition in \cgentity and for normalization in \method. The dataset can be found at \href{https://github.com/chanzuckerberg/MedMentions}{https://github.com/chanzuckerberg/MedMentions}

\paragraph{UMLS}
The UMLS meta-thesaurus is a large biomedical knowledge base which unifies hundreds of different ontologies in biomedicine. UMLS is used as the source knowledge base for normalization and candidate selection for \method. Additionally, it is the knowledge base used to train \texttt{cui2vec}, which is used for candidate concept selection in \method. UMLS can be found here \href{https://www.nlm.nih.gov/research/umls/index.html}{https://www.nlm.nih.gov/research/umls/index.html}.

\paragraph{SQuAD}
The SQuAD dataset can be found at: \href{https://rajpurkar.github.io/SQuAD-explorer/}{https://rajpurkar.github.io/SQuAD-explorer/}. SQuAD is used as training data for the question generation module of \cgentity. SQuAD is a question answering dataset which contains data of the form $(q, c, a)$, where $q$ is the question, $c$ is a context document, and $a$ is an answer to the question which can be found in the context. 

\paragraph{QA2D}
The QA2D dataset can be found here\footnote{\href{https://worksheets.codalab.org/worksheets/0xd4ebc52cebb84130a07cbfe81597aaf0/}{https://worksheets.codalab.org/worksheets/ 0xd4ebc52cebb84130a07cbfe81597aaf0/}}. QA2D is used in the second part of the zero-shot \cgentity model to generate declarative sentences from questions. It consists of data of the form $(s, q, a)$ where $q$ is a question, $a$ is the answer to the question, and $s$ is the declarative form of the question containing the answer. 

\paragraph{MNLI}
MNLI is a crowd-sourced collection of 433k sentence pairs annotated for textual entailment. In other words, the data consists of pairs $(p, h)$, where $p$ is the premise and $h$ is the hypothesis, and labels in \{\textit{entailment}, \textit{contradiction}, \textit{neutral}\} which say if the hypothesis entails, contradicts, or is neutral towards the premise. MNLI is used to train a RoBERTa model for entailment, which is used by \method to select the best negation among a set of generated claims for a given source citance. The dataset can be found here \href{https://cims.nyu.edu/~sbowman/multinli/}{https://cims.nyu.edu/~sbowman/multinli/}

\section{Appendices for Semi-Supervised Exaggeration Detection of Health Science Press Releases}

\subsection{Error Analysis Plots}
\label{sec:extra_plots}
\begin{figure*}[t]
     \centering

         \centering
         \includegraphics[width=0.8\textwidth]{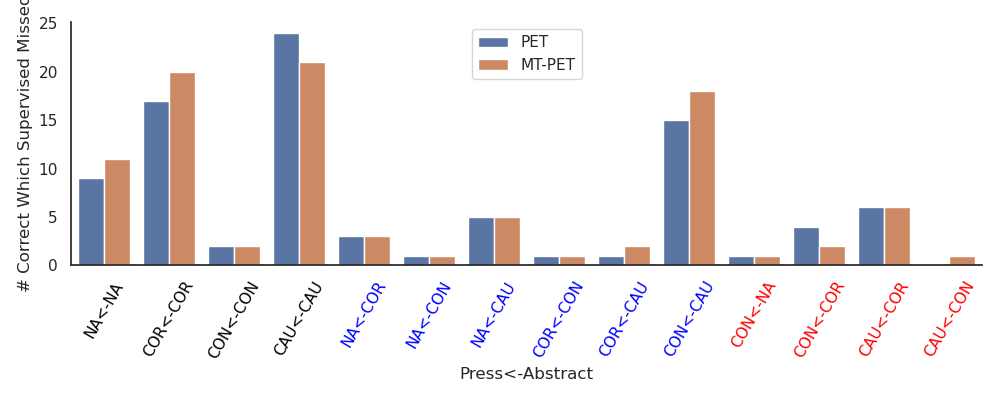}
         \caption{Number of instances that each model predicted correctly which the supervised model predicted incorrectly.}
         \label{fig:supervised_succeed}
\end{figure*}

\begin{figure*}[t]
     \centering

         \centering
         \includegraphics[width=0.8\textwidth]{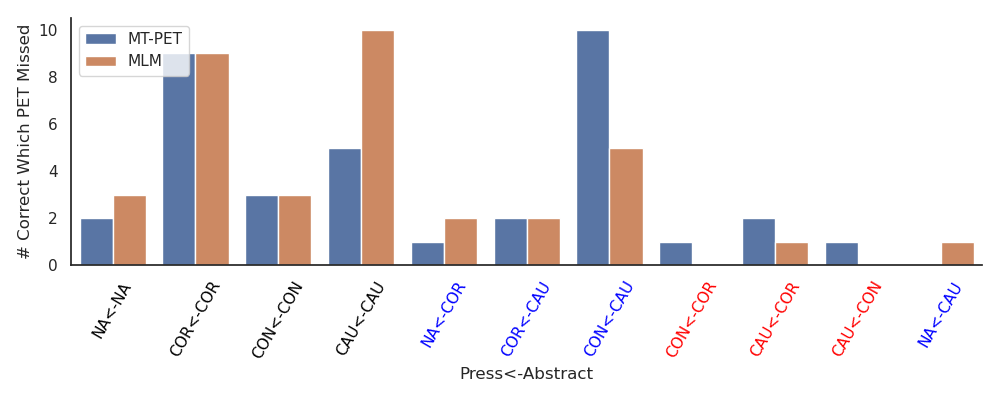}
         \caption{Number of instances that each model predicted correctly which PET predicted incorrectly.}
         \label{fig:pet_succeed}
\end{figure*}

Extra plots from our error analysis are given in \autoref{fig:supervised_succeed} and \autoref{fig:pet_succeed}.

\subsection{Reproducibility}
\subsubsection{Computing Infrastructure}

All experiments were run on a shared cluster. Requested jobs consisted of 16GB of RAM and 4 Intel Xeon Silver 4110 CPUs. We used a single NVIDIA Titan X GPU with 24GB of RAM.

\subsubsection{Average Runtimes}
The average runtime performance of each model is given in \autoref{tab:runtimes}. Note that different runs may have been placed on different nodes within a shared cluster. 
\begin{table}[htp]
    \centering
    \fontsize{10}{10}\selectfont
    \begin{tabular}{l c c}
    \toprule %
     Setting & |\textbf{T1}|,|\textbf{T2}| & Time\\
    \midrule
        Supervised& 100,0 & 00h01m28s\\
        PET& 100,0 & 00h11m14s\\
        \METHOD& 100,200 & 00h13m05s\\
        \midrule
        Supervised& 0,200 & 00h01m20s\\
        PET& 0,200 & 00h16m22s\\
        \METHOD& 100,200 & 00h18m43s\\
        Supervised& 0,4500 & 00h03m23s\\
        PET& 0,4500 & 00h40m23s\\
        \METHOD& 100,4500& 00h31m48s\\
    \bottomrule %

    \end{tabular}
    \caption{Average runtimes for each model (runtimes are taken for the entire run of an experiment).}
    \label{tab:runtimes}
\end{table}

\subsubsection{Number of Parameters per Model}
We use RoBERTa, specifically the base model, for all experiments, which consists of 124,647,170 parameters.

\subsubsection{Validation Performance}
As we are testing a few shot setting, we do not use a validation set and only report the final test results.

\subsubsection{Evaluation Metrics}
The primary evaluation metric used was macro F1 score.
We used the sklearn implementation of \texttt{precision\_recall\_fscore\_support} for F1 score, which can be found here: \url{https://scikit-learn.org/stable/modules/generated/sklearn.metrics.precision_recall_fscore_support.html}. Briefly:
\begin{equation*}
   p = \frac{tp}{tp + fp} 
\end{equation*}
\begin{equation*}
   r = \frac{tp}{tp + fn} 
\end{equation*}
\begin{equation*}
   F1 = \frac{2*p*r}{p+r} 
\end{equation*}
where $tp$ are true positives, $fp$ are false positives, and $fn$ are false negatives. Macro F1 is average F1 across all classes.

\subsubsection{Hyperparameters}
\label{sec:hyperparams}
\paragraph{T1 Hyperparameters Supervised/PET training} We used the following hyperparameters for experiments with \textbf{T1} as the main task: Epochs: 10; Batch Size: 4; Learning Rate: 0.00005598; Warmup Steps: 50; Weight decay: 0.001. We also weigh the cross-entropy loss based on the label distribution. These hyperparameters are found by performing a hyperparameter search using 4-fold cross validation on the 100 training examples. The bounds are as follows: Learning rate: $[0.000001, 0.0001$; Warmup steps: $\{0, 10, 20, 30, 40, 50, 100\}$; Batch size: $\{4, 8\}$; Weight decay: $\{0.0, 0.0001, 0.001, 0.01, 0.1\}$; Epochs: $[2, 10]$.

\paragraph{T2 Hyperparameters Supervised/PET training} Epochs: 10; Batch Size: 4; Learning Rate: 0.00003; Warmup Steps: 50; Weight Decay: 0.001. We also weigh the cross-entropy loss based on the label distribution.

\paragraph{Hyperparameters for Distillation} We used the following hyperparameters for distillation (training the final classifier after PET) for both \textbf{T1} and \textbf{T2} as the main task: Epochs: 3; Batch Size: 4; Learning Rate: 0.00001; Warmup Steps: 200; Weight decay: 0.01; Temperature: 2.0. We also weigh the cross-entropy loss based on the label distribution.

\subsubsection{Data}
\label{sec:data_info}
We build our benchmark test dataset from the studies of \citet{sumner2014association} and \citet{bratton2019association}. The original data can be found at \url{https://figshare.com/articles/dataset/InSciOut/903704} and \url{https://osf.io/v8qhn/files/}. A link to the test data will be provided upon acceptance of the paper (and included in the supplemental material). Claim strength data from \citet{yu2019detecting} for abstracts can be found at \url{https://github.com/junwang4/correlation-to-causation-exaggeration/blob/master/data/annotated_pubmed.csv}. Claim strength data for press releases from \citet{yu2020measuring} can be found at \url{https://github.com/junwang4/correlation-to-causation-exaggeration/blob/master/data/annotated_eureka.csv} %

\section{Appendices for Modeling Information Change in Science Communication with Semantically Matched Paraphrases}

\subsection{Information Change vs. Semantic Similarity}
\label{sec:info-change-supplement}
\begin{table*}[t]
\small

\newcommand{\tabincell}[2]{\begin{tabular}{@{}#1@{}}#2\end{tabular}}
\resizebox{0.99\textwidth}{!}{
\begin{tabular}{p{60mm}p{60mm}}%
\toprule
\textbf{Sentence 1} & \textbf{Sentence 2}\\
\midrule
The polar bear is sliding on the snow. & A polar bear is sliding across the snow.
\\
\midrule
A plane is taking off & An air plane is taking off\\
\midrule
A dog rides a skateboard & A dog is riding a skateboard\\
\midrule
A man is playing the drums & A man plays the drum\\
\bottomrule
\end{tabular}
}
\caption{Samples of sentence pairs in STSB which have a similarity score of 5 }
\label{tab:stsb-5-examples}
\end{table*}

\begin{table*}[t]
\small

\newcommand{\tabincell}[2]{\begin{tabular}{@{}#1@{}}#2\end{tabular}}
\resizebox{0.99\textwidth}{!}{
\begin{tabular}{p{60mm}p{60mm}}%
\toprule
\textbf{Sentence 1} & \textbf{Sentence 2}\\
\midrule
Higher-income professionals had less tolerance for smartphone use in business meetings. & We are intrigued by the result that professionals with higher incomes are less accepting of mobile phone use in meetings.
\\
\midrule
If we allow people to retract recently posted comments, then we may be able to minimize regret from posting in the heat of the moment. & Allowing users to retract recently posted comments may help minimize regret .\\
\midrule
Papers with shorter titles get more citations \#science \#metascience \#sciencemetrics & Our analysis suggests that papers with shorter titles do receive greater numbers of citations.\\
\midrule
Low levels of self-esteem and poor emotional processing skills were significantly correlated with gang involvement, as were low levels of parental monitoring, poor parental communication and housing instability. & Major findings also indicated that low levels of parental monitoring, poor parental communication and housing instability were significantly associated with gang involvement.\\
\bottomrule
\end{tabular}
}
\caption{Samples of sentence pairs in \textsc{Spiced} which have an \SCORE{} of 5.}
\label{tab:dataset-5-examples}
\end{table*}
We wish to highlight key differences between information change and semantic similarity, particularly with an eye to what makes the task introduced in \textsc{Spiced} difficult compared to semantic similarity scoring. To illustrate this, we present a sample of pairs in STSB that have the highest similarity score of `5' vs. samples in \textsc{Spiced} which have an \SCORE{} of 5 in \autoref{tab:stsb-5-examples} and \autoref{tab:dataset-5-examples}.

In this, for a pair to be perfectly similar from a semantics perspective, the entire sentence must contain exactly equivalent meaning. This is not the case with our task. For the information change task, pairs are highly similar even if some aspects of the semantics of the sentence are changed e.g. in the first sample, there is a difference between the two sentences semantically: the second in the pair discusses ``being intrigued'' by the finding, which is shared between the pair. This also makes the task extremely difficult -- a model must learn to compare only the salient scientific facts between the pair of sentences, as opposed to the entire meaning of each sentence.

\subsection{Pilot Annotation Details}
\label{sec:pilot-annotation-details}
\begin{figure}[t]
  
  \centering
    \includegraphics[width=0.75\linewidth]{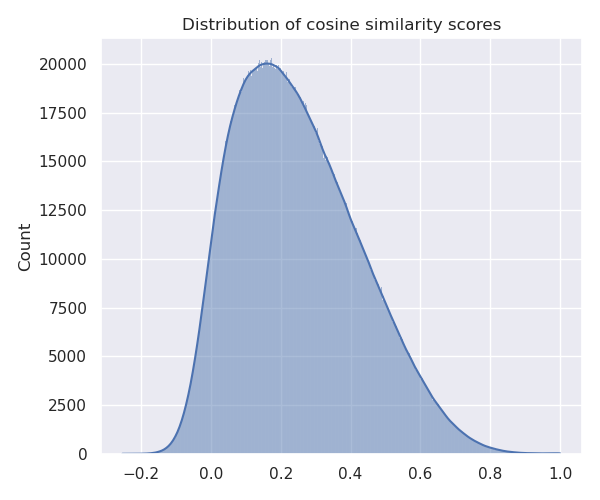}
    \caption{Distribution of the cosine similarity between findings extracted from news articles about particular scientific papers. Cosine similarity is measured between the embeddings produced for both findings using SBERT~\cite{reimers-2019-sentence-bert}.} %
    \label{fig:sample-cos-distribution}
\end{figure}

\begin{figure}[t]
  
  \centering
    \includegraphics[width=0.75\linewidth]{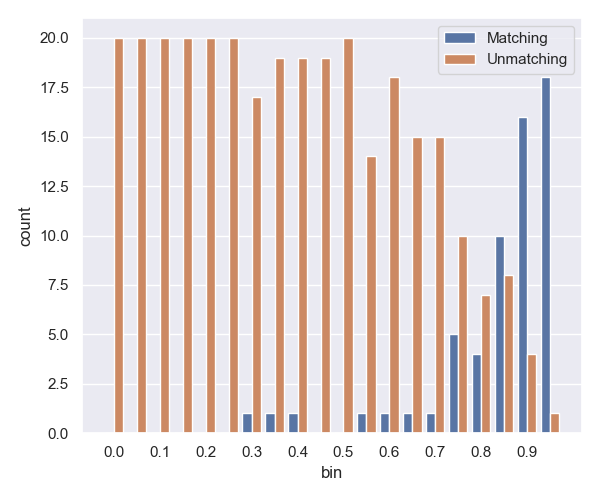}
    \caption{Number of samples per bin rated as matching vs. not matching (samples limited to those where both annotators agreed on the label). Most matching samples come from higher similarity bins, while more difficult samples come from the middle bins. %
    }
    \label{fig:pos-samples-pilot}
\end{figure}
For the pilot, we use 20 pairs from 20 different cosine similarity score bins in increments of $0.05$ starting from $0$. In other words, we have 20 bins with ranges of scores as: $0.0-0.05$, $0.05-0.1$...$0.9-0.95$, $0.95-1.0$. This results in 400 samples to annotate. 
The score distribution from 7,392,690 pairs from 3,525 source papers which we use for sampling is given in~\autoref{fig:sample-cos-distribution}. 
Each sample is annotated by two of the authors of the study with a binary label of ``matching'' vs ``not matching'', yielding a Krippendorff's alpha of $0.73$.

The number of positive samples per bin from the pilot study is given in \autoref{fig:pos-samples-pilot}. We see here that bins with a cosine similarity below 0.65 tend to have very few positive samples, and only above 0.8 do we start to see many positive samples in the bins. Almost all samples above $0.9$ are matching, and the only unmatched pairs appear to be instances of SBERT failing, since the matched pairs are almost exactly copied text. Additionally, this histogram indicates that the base rate of positive matching findings is low as the overall distribution of samples in the high cosine similarity region, where most of the matches exist, is small. At the same time, we note that some of the matches we find in the lower cosine similarity regions constitute quite interesting samples; for example, the following which has a cosine similarity of 0.41.

\begin{quote}
    \textbf{Paper finding:} For cases comparing a drone and a vehicle carrying a single package over similar distances, for example, a customer picking up a package from a retail store, the drone is clearly a lower-impact solution.~\cite{stolaroff2018energy}
    
    \textbf{News finding:} But if you forgot that essential ingredient for tonight's dinner, our findings suggest it's much better to have the grocery store send it to you by drone rather than to take your car to the store and back.\footnote{https://www.enbridge.com/energy-matters/news-and-views/delivering-packages-with-drones-might-be-good-for-the-environment}
\end{quote}

Both sentences are talking about the same finding, that drone delivery is more efficient over short distances than using a car, but in entirely different ways. From this, it is clear that simply using semantic text similarity is insufficient for solving this task, and we should include some of these lower similarity samples in our annotation. We, therefore, propose the following sampling scheme in order to balance the number of annotations we can acquire, the yield of positive samples, and the sample difficulty:
\begin{itemize}[noitemsep,nolistsep]
    \item Label all samples with a cosine similarity below $0.4$ as unmatched.
    \item Label all samples above $0.9$ with a Jaccard index above $0.5$ as matching.
    \item Sample an equal number of pairs from each $0.05$ increment bin between $0.4$ and $0.9$ for human expert annotation.
\end{itemize}

\subsection{Experimented annotation}
\label{sec:experimented-annotations-supplement}
We experimented with two annotation schemas: a binary schema where the annotators are asked to label ``whether the two sentences are discussing the same scientific finding'' with Yes or No, and a Likert schema where the annotators are asked to label if
``The information in the findings is...''
\begin{itemize}[noitemsep,nolistsep]
    \item 1: Completely different
    \item 2: Mostly different
    \item 3: Somewhat similar
    \item 4: Mostly the same
    \item 5: Completely the same
\end{itemize}
We ran several pilots using the two annotation schemas and the Likert a schema led to higher inter-annotator agreement (0.45 Krippendorff's alpha) compared with the binary schema (0.21 Krippendorff's alpha). Therefore we adopt the 5-point Likert schema for the annotation.

\subsection{Full Annotation Instructions}
\label{sec:annotation-instructions}
\begin{figure*}[t]
  
  \centering
    \includegraphics[width=\linewidth]{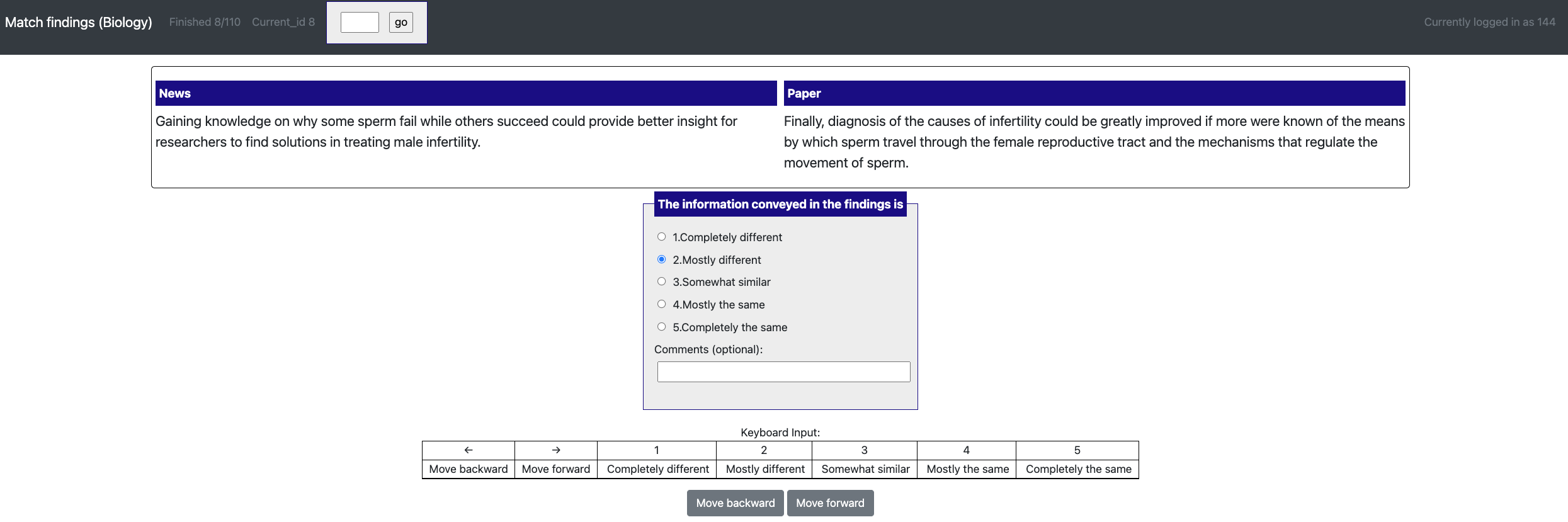}
    \caption{The annotation page of our crowdsourcing task
    }
    \label{fig:annotation_interface}
\end{figure*}

Annotation was performed using Prolific workers who labeled using \textsc{Potato} \cite{pei2022potato}. The annotation interface setup is available at {\small \url{https://github.com/davidjurgens/potato/tree/master/example-projects/match_finding}} which includes all the following instructions as well.

\paragraph{Task description:} The task is to label to what degree two sentences have the same information. The information in the sentences is scientific findings. Here, a scientific finding is a statement that describes a research output of a scientific study, such as a result, conclusion, product, etc. You should rate how similar the findings are; you can ignore extra information like ``The researchers showed...'', ``In vivo experiments demonstrated...'' etc. For example, in the sentence ``After controlling for weight and age, researchers found that overconsumption of sugar is linked with an increase in diabetes,'' the information in the finding is ``overconsumption of sugar is linked with an increase in diabetes''. Some sentences may have no findings or multiple findings, so use your best judgment about what are the core findings being said.

You will rate this on a 5-point scale, where each level means the following:

\begin{enumerate}
\item The information in the findings is completely different
\begin{itemize}
    \item Sentences in this category have findings which say completely different information
    \item The sentences may be on totally different topics
    \begin{itemize}
        \item Overconsumption of sugar causes diabetes
        \item Regular exercise improves heart health
    \end{itemize}
    \item There may be some overlap in key words used between the two sentences, but the actual information is completely different
    \begin{itemize}
        \item Chocolate contains a lot of sugar, and therefore can have an effect on weight.
        \item Overconsumption of sugar leads to diabetes.
    \end{itemize}
\end{itemize}

\item The information in the findings is mostly different
\begin{itemize}
    \item The findings may talk about the same topic, but the actual information is mostly different; for example, these sentences convey mostly different information even though they talk about the same topic:
    \begin{itemize}
        \item Overconsumption of sugar causes diabetes
        \item Sugar is good for your health
    \end{itemize}
    \item There could be a link between the two findings, but the information conveyed is still different
    \begin{itemize}
        \item Overconsumption of sugar increases blood glucose levels
        \item High blood glucose over time increases the risk of developing diabetes
    \end{itemize}
\end{itemize}

\item The information in the findings is somewhat similar
\begin{itemize}
    \item The findings are discussing relevant research outputs but there are some differences in the information conveyed. Here the difference is that (i) talks about the relationship between overconsumption of sugar and diabetes and (ii) describes how genetics plays a role in overconsumption of sugar
    \begin{itemize}
        \item Overconsumption of sugar causes diabetes
        \item Overconsumption of sugar might be genetically determined
    \end{itemize}
\end{itemize}

\item The information in the findings is mostly the same
\begin{itemize}
    \item In this case there may be some changes in e.g. the level of generality. Additionally, one sentence may go into more detail than the other and add additional context, but the information is largely the same
    \item Here the two findings have the same information but at different levels of generality:
    \begin{itemize}
        \item A link between sugar and diabetes was found
        \item Overconsumption of sugar is associated with the onset of diabetes
    \end{itemize}
    \item Here both sentences have the same core finding, but one sentence goes into more detail
    \begin{itemize}
        \item Overconsumption of sugar causes diabetes
        \item Experiments demonstrated that overconsumption of sugar led to an increase in blood glucose levels, which over a long enough time period was linked to an increased prevalence of diabetes in the cohort.
    \end{itemize}
    \item One finding could support the other
    \begin{itemize}
        \item Overconsumption of sugar causes diabetes
        \item Overconsumption of sugar can have negative effects on health
    \end{itemize}
\end{itemize}

\item The information in the findings is completely the same
\begin{itemize}
    \item In this case there is complete overlap in the information in the findings conveyed by the two sentences
    \begin{itemize}
        \item Overconsumption of sugar leads to diabetes.
        \item The researchers found that overconsumption of sugar leads to diabetes
    \end{itemize}
    \item Note that there can be changes in e.g. the level of certainty or the strength of the information.
    \begin{itemize}
        \item Overconsumption of sugar leads to diabetes.
        \item It is likely that there is a link between overconsumption of sugar and the onset of diabetes.

    \end{itemize}
\end{itemize}
\end{enumerate}

\subsection{Final dataset details}
\label{sec:dataset-details-supplement}
\begin{figure}[h]
  
  \centering
    \includegraphics[width=0.75\linewidth]{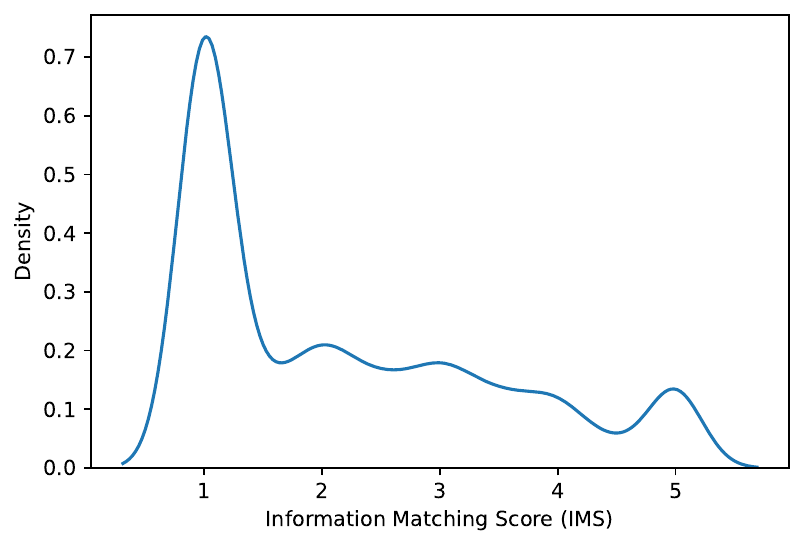}
    \caption{Distribution of the final matching score in \textsc{Spiced}, which includes some pairs of scientific findings that are automatically labeled based on their extreme textual similarity (high or low), in addition to the annotated pairs.
    }
    \label{fig:matching_score_distribution}
\end{figure}

\begin{figure}[h]
  
  \centering
    \includegraphics[width=0.75\linewidth]{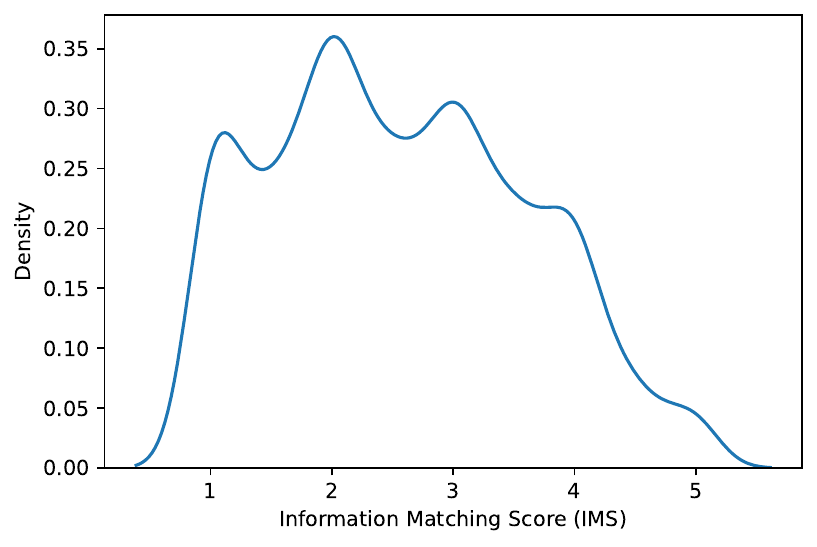}
    \caption{Distribution of the final matching score for annotated pairs in \textsc{Spiced}
    }
    \label{fig:matching_score_distribution_no_easy}
\end{figure}

\begin{figure*}[t]
  
  \centering
    \includegraphics[width=\linewidth]{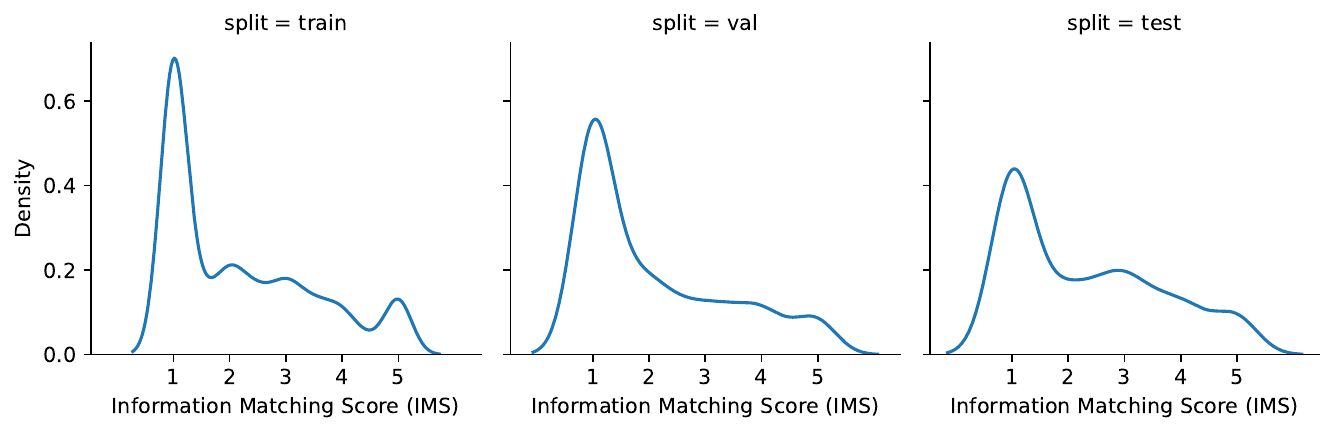}
    \caption{Distribution of the final matching score for each split set in \textsc{Spiced}
    }
    \label{fig:matching_score_distribution_splits}
\end{figure*}

Figure \ref{fig:matching_score_distribution} shows the IMS distribution in \textsc{Spiced}. Figure \ref{fig:matching_score_distribution_no_easy} shows the IMS distribution for annotated pairs in \textsc{Spiced}. Figure \ref{fig:matching_score_distribution_splits} shows the IMS distribution for each split.

We measure various aspects of lexical richness between the different domains of the data in \autoref{tab:lexical-richness}.

\begin{table}%
    \setlength{\tabcolsep}{1.5pt}
    \def\arraystretch{1.2}
    \centering
    \fontsize{10}{10}\selectfont
    \begin{tabular}{l c c c c}
    \toprule %
    Metric & Papers & Overall & News & Tweets\\
    \midrule 

        Unique tokens & $11047$ & $12139$ & $10203$ & $5037$\\
        RTTR & $32.01$ & $36.59$ & $33.48$ & $38.46$\\
        MTLD & $152.64$ & $185.35$ & $176.53$ & $259.88$\\
        HDD & $0.89$ & $0.90$ & $0.89$ & $0.92$\\
    
    \bottomrule %

    \end{tabular}
    \caption{Various measures of lexical richness and diversity between findings in papers and other sources. RTTR is the root token-type ratio; MTLD is measure of textual lexical diversity~\cite{mccarthy2010mtld}; HDD is the hypergeometric distribution diversity~\cite{mccarthy2010mtld}.}
    \label{tab:lexical-richness}
\end{table}

\subsection{Metrics}%
\label{sec:metrics-supplement}
\paragraph{Average Normalized Edit Distance} We calculate the normalized edit distance as follows:
\begin{equation*}
d_{N} = \frac{1}{|D|}\sum_{i}\frac{d(s_{1}^{(i)},s_{2}^{(i)})}{\max{(|s_{1}^{(i)}|,|s_{2}^{(i)}|)}}
\end{equation*}
where $|D|$ is the size of the dataset, $(s_{1}^{(i)},s_{2}^{(i)})$ is a sentence pair, and $d$ is the edit distance. 

\paragraph{Jaccard Index} The Jaccard index is calculated based on the overlap of the members of two sets (e.g. the words in two sentences $X$ and $Y$):
\begin{equation*}
    J = \frac{|X \cap Y|}{|X \cup Y|}
\end{equation*}

\paragraph{Cosine Similarity} The cosine similarity between two vectors $\textbf{a}$ and $\textbf{b}$ is calculated as:
\begin{equation*}
    S_{C}(\textbf{a},\textbf{b)} = \frac{\textbf{a} \cdot \textbf{b}}{\lVert\textbf{a}\rVert \lVert\textbf{b}\rVert}
\end{equation*}
Which is their dot product divided by the product of their lengths.

\paragraph{Mean Squared Error} The mean squared error between two lists of numbers of length $n$ is calculated as:
\begin{equation*}
    \text{MSE}(Y, \hat{Y}) = \frac{1}{n}\sum_{i}(Y_{i} - \hat{Y}_{i})^{2}
\end{equation*}

\paragraph{Mean Average Precision} The mean average precision in ranking takes the average Precision@k (P@k) for every relevant sample in a ranked list. First, P@k is calculated as follows:
\begin{equation*}
    \text{P@k}(\hat{Y}) = \frac{1}{k}\sum_{i}^{k}\mathbbm{1}(\hat{Y}_{i} = 1)
\end{equation*}
where $\mathbbm{1}$ is the indicator function. The average precision is then taken over all relevant items in the list, where there are $r$ relevant items:
\begin{equation*}
    \text{AP}(\hat{Y}) = \frac{1}{r}\sum_{k}\text{P@k}(\hat{Y}[:k]) \text{ where } \hat{Y}_k = 1
\end{equation*}
The mean average precision for a set of $n$ ranked lists $D$ is then the mean of the average precision of each of these lists:
\begin{equation*}
    \text{MAP} = \frac{1}{n}\sum_{j}\text{AP}(D_{n})
\end{equation*}

\paragraph{Mean Reciprocal Rank} The mean reciprocal rank (MRR) calculates the mean rank for each relevant item in a list i.e. its position in that list. It is calculated as follows for $D$ lists or relevant items in $\hat{Y}$ ranked lists:
\begin{equation*}
    \text{MRR}(D) = \frac{1}{|D|}\sum_{j}\frac{1}{|D_{j}|}\sum_{i}\frac{1}{\text{rank}_i(\hat{Y}_{j})}
\end{equation*}
where $\text{rank}_i(\hat{Y}_{j})$ is the rank of item $i$ in list $\hat{Y}_{j}$.

\subsection{Full Model Details}
\label{sec:full-model-descriptions}

All baseline experiments were run on a shared cluster. Requested jobs consisted of 16GB of RAM and 4 Intel Xeon Silver 4110 CPUs. We used a single NVIDIA Titan RTX GPU for experiments. Training takes approximately 3 minutes for all MLM-based models and 2 minutes for SBERT models. 

\paragraph{RoBERTa}
RoBERTa is a large pretrained transformer language model, trained using the masked language modeling (MLM) objective on a large corpus of English text. We use the base model of RoBERTa for our experiments. Huggingface model name: roberta-base -- 124,647,170 parameters

\paragraph{MiniLM} We use a popular pretrained SBERT model based on MiniLM~\cite{wang2020minilm}, which is trained by distilling multiple language models into one compressed model. SBERT uses siamese BERT encoders to obtain sentence embeddings for pairs of sentences and is trained to decrease the distance between these two embeddings. The pretraining for the sentence similarity task consists of a wide range of datasets covering multiple domains and $>$ 1 billion sentence pairs, including science~\cite{DBLP:conf/acl/CohanFBDW20, lo2020s2orc}. As much of the data is collected automatically, it uses a contrastive learning objective where known relevant pairs are treated as positive values and other samples in a batch are treated as negative values. The model is then trained to minimize the cross-entropy between the dot-product of embeddings and the label acquired from positive/negative samples.
Huggingface model name (sentence transformers): all-MiniLM-L6-v2 -- 22,713,216 parameters

\paragraph{MPNet} This is the same setup as in MiniLM but with using MPNet as the base network~\cite{DBLP:conf/nips/Song0QLL20}. MPNet is trained using a permuted language modeling (PLM) objective with position information as input to achieve the best of both worlds between MLM and PLM. The base network is used in the SBERT setup where it is further fine-tuned on the same dataset and same task as with MiniLM

Huggingface model name (sentence transformers): all-mpnet-base-v2 -- 109,486,464 parameters

\paragraph{Paraphrase Detection}
This is a paraphrase detection model based on RoBERTa used in~\cite{nighojkar-licato-2021-improving}. The model is trained on the adversarial paraphrase dataset introduced in that paper. 

Huggingface model name (sentence transformers): coderpotter/adversarial-paraphrasing-detector -- 124,647,170 parameters

\paragraph{NLI}
This is a RoBERTa model trained on a wide array of NLI datasets, including SNLI~\cite{DBLP:conf/emnlp/BowmanAPM15}, MNLI~\cite{DBLP:conf/naacl/WilliamsNB18}, FEVER (a fact-checking dataset)~\cite{DBLP:conf/naacl/ThorneVCM18} and ANLI~\cite{nie-etal-2020-adversarial}.

Huggingface model name (sentence transformers):

\text{ynie/roberta-large-snli_mnli_fever_anli_R1_R2_R3-nli} -- 124,647,170 parameters

\paragraph{SciBERT}
SciBERT is the original BERT model trained using MLM on a large set of scientific papers from Semantic Scholar~\cite{lo2020s2orc}. 

Huggingface model name (sentence transformers): allenai/scibert\_scivocab\_uncased -- 109,920,514 parameters

\paragraph{CiteBERT}
CiteBERT is SciBERT further fine-tuned on the CiteWorth dataset for the task of citation detection, which predicts if a given sentence requires a citation or not~\cite{DBLP:conf/acl/WrightA21}.

Huggingface model name (sentence transformers): copenlu/citebert -- 109,920,514 parameters

We use sane defaults when fine-tuning each of our models. In this, for the MLM based models we use [lr: 2e-5, n\_epochs: 3, warmup\_steps: 200, weight\_decay: 0.01, batch\_size: 8]. For SBERT models we use the same setting except we train for 5 epochs.

\subsection{Exaggeration Detection}
\label{sec:exaggeration-supplement}
The problem of scientific exaggeration detection was studied in~\cite{wright2021semi}. The basic task is: given a pair of scientific findings (e.g. a reference finding from a paper and its counterpart in a news article), determine if one finding is exaggerating the other finding. More formally, the task focuses on differences in the causal claim strength of the two findings, where the claim strength can take on one of four values:
\begin{itemize}[noitemsep,nolistsep]
    \item 0: No statement of relationship
    \item 1: Correlational statement (e.g. ``X is associated with Y'')
    \item Conditional causal statement (e.g. ``X might cause Y under circumstance Z'')
    \item Causal statement (e.g. ``X causes Y'')
\end{itemize}
\citet{wright2021semi} curate data and build models for performing the exaggeration detection task in two different settings: as predicting the individual claim strengths and comparing, and as an inference task where a model is fed both findings and asked to predict if the reference finding is being exaggerated, downplayed, or faithfully represented by its counterpart. We use the best-performing model from their paper, which is a multi-task few-shot learning model based on pattern exploiting training (PET) called MT-PET. In particular, we use the model for strength classification which has seen 4,500 individual findings labeled for claim strength and 200 pairs labeled for exaggeration.

\subsection{Scientific Text Parser}
\label{sec:parser-supplement}
We fine-tuned a RoBERTa model over 200K self-labeled abstracts from PubMed. The model is trained to predict five labels including: BACKGROUND, CONCLUSIONS, METHODS, OBJECTIVE and RESULTS. We did a 8:1:1 split for the data and fine-tune the RoBERTa model for 1 epoch. 0.92 F1 is attained on the test set.

\subsection{Extended Benchmarking}
\label{sec:extended-benchmarking}
Tables with extended benchmarking results can be found in \autoref{tab:baseline-findings-matching} to \autoref{tab:cs-baseline-findings-matching}.

\begin{table*}%
    \setlength{\tabcolsep}{1.5pt}
    \def\arraystretch{1.2}
    \centering
    \fontsize{10}{10}\selectfont
    \begin{tabular}{l c c | c c | c c}
    \toprule %
    & \multicolumn{2}{c}{All} & \multicolumn{2}{c}{News} & \multicolumn{2}{c}{Twitter} \\
     \midrule
    Method & MSE & $\rho$ & MSE & $\rho$ & MSE & $\rho$ \\
    \midrule 
Paraphrase & $3.170_{0.000}$ & $23.58_{0.00}$ & $3.310_{0.000}$ & $19.24_{0.00}$ & $2.824_{0.000}$ & $35.41_{0.00}$ \\
NLI & $2.921_{0.000}$ & $35.71_{0.00}$ & $2.786_{0.000}$ & $35.78_{0.00}$ & $3.255_{0.000}$ & $34.76_{0.00}$ \\
MiniLM & $0.628_{0.000}$ & $73.98_{0.00}$ & $0.646_{0.000}$ & $76.27_{0.00}$ & $0.583_{0.000}$ & $64.61_{0.00}$ \\
MPNet & $0.718_{0.000}$ & $72.59_{0.00}$ & $0.713_{0.000}$ & $74.76_{0.00}$ & $0.730_{0.000}$ & $62.06_{0.00}$ \\
\midrule
SciBERT & $0.579_{0.011}$ & $73.24_{0.73}$ &  $0.596_{0.018}$ & $74.66_{0.75}$ &  $0.538_{0.021}$ & $66.29_{0.67}$\\
CiteBERT & $0.581_{0.027}$ & $73.37_{0.78}$ &  $0.592_{0.034}$ & $74.81_{0.91}$ &  $0.552_{0.030}$ & $66.13_{1.43}$\\
RoBERTa & $0.587_{0.017}$ & $74.44_{0.81}$ &  $0.602_{0.033}$ & $75.82_{0.71}$ &  $0.550_{0.067}$ & $\mathbf{68.66_{1.29}}$\\
MiniLM-FT & $0.492_{0.001}$ & $75.84_{0.03}$ &  $\mathbf{0.465_{0.001}}$ & $78.66_{0.05}$ &  $0.559_{0.002}$ & $63.80_{0.05}$\\
MPNet-FT & $\mathbf{0.489_{0.003}}$ & $\mathbf{76.48_{0.07}}$ &  $0.474_{0.003}$ & $\mathbf{78.71_{0.17}}$ &  $\mathbf{0.526_{0.008}}$ & $66.45_{0.37}$\\

    \bottomrule %

    \end{tabular}
    \caption{OVERALL --MSE and Pearson correlation ($\rho$) on predicting the similarity of the scientific findings for different models. Results are averaged over 5 random seeds; standard deviation is given in the subscript.} %
    \label{tab:baseline-findings-matching}
\end{table*}

\begin{table*}%
    \setlength{\tabcolsep}{1.5pt}
    \def\arraystretch{1.2}
    \centering
    \fontsize{10}{10}\selectfont
    \begin{tabular}{l c c | c c | c c}
    \toprule %
    & \multicolumn{2}{c}{All} & \multicolumn{2}{c}{News} & \multicolumn{2}{c}{Twitter} \\
     \midrule
    Method & MSE & $\rho$ & MSE & $\rho$ & MSE & $\rho$ \\
    \midrule 
Paraphrase & $2.773_{0.000}$ & $27.16_{0.00}$ & $2.846_{0.000}$ & $30.22_{0.00}$ & $2.577_{0.000}$ & $28.18_{0.00}$ \\
NLI & $2.529_{0.000}$ & $40.23_{0.00}$ & $2.225_{0.000}$ & $47.55_{0.00}$ & $3.339_{0.000}$ & $6.23_{0.00}$ \\
MiniLM & $0.618_{0.000}$ & $76.45_{0.00}$ & $0.658_{0.000}$ & $80.31_{0.00}$ & $\mathbf{0.509_{0.000}}$ & $\mathbf{63.78_{0.00}}$ \\
MPNet & $0.804_{0.000}$ & $73.14_{0.00}$ & $0.815_{0.000}$ & $76.91_{0.00}$ & $0.777_{0.000}$ & $56.11_{0.00}$ \\
\midrule
SciBERT & $0.554_{0.020}$ & $71.67_{0.94}$ &  $0.507_{0.026}$ & $76.69_{0.73}$ &  $0.681_{0.058}$ & $43.56_{5.32}$\\
CiteBERT & $0.542_{0.031}$ & $72.55_{0.92}$ &  $0.496_{0.034}$ & $77.31_{1.11}$ &  $0.663_{0.029}$ & $46.01_{2.61}$\\
RoBERTa & $0.511_{0.036}$ & $75.40_{1.19}$ &  $0.475_{0.035}$ & $79.33_{0.78}$ &  $0.608_{0.056}$ & $53.72_{4.56}$\\
MiniLM-FT & $\mathbf{0.377_{0.002}}$ & $\mathbf{79.46_{0.15}}$ &  $\mathbf{0.327_{0.003}}$ & $\mathbf{84.08_{0.14}}$ &  $0.512_{0.001}$ & $60.00_{0.23}$\\
MPNet-FT & $0.412_{0.005}$ & $77.98_{0.23}$ &  $0.361_{0.004}$ & $82.30_{0.22}$ &  $0.548_{0.013}$ & $57.79_{0.72}$\\

    \bottomrule %

    \end{tabular}
    \caption{BIOLOGY -- MSE and Pearson correlation ($\rho$) on predicting the similarity of the scientific findings for different models. Results are averaged over 5 random seeds; standard deviation is given in the subscript.} %
    \label{tab:bio-baseline-findings-matching}
\end{table*}

\begin{table*}%
    \setlength{\tabcolsep}{1.5pt}
    \def\arraystretch{1.2}
    \centering
    \fontsize{10}{10}\selectfont
    \begin{tabular}{l c c | c c | c c}
    \toprule %
    & \multicolumn{2}{c}{All} & \multicolumn{2}{c}{News} & \multicolumn{2}{c}{Twitter} \\
     \midrule
    Method & MSE & $\rho$ & MSE & $\rho$ & MSE & $\rho$ \\
    \midrule 
Paraphrase & $3.282_{0.000}$ & $15.95_{0.00}$ & $3.525_{0.000}$ & $31.32_{0.00}$ & $2.629_{0.000}$ & $29.56_{0.00}$ \\
NLI & $2.820_{0.000}$ & $37.03_{0.00}$ & $2.841_{0.000}$ & $34.60_{0.00}$ & $2.763_{0.000}$ & $49.39_{0.00}$ \\
MiniLM & $0.706_{0.000}$ & $76.46_{0.00}$ & $0.739_{0.000}$ & $78.34_{0.00}$ & $0.615_{0.000}$ & $62.92_{0.00}$ \\
MPNet & $0.738_{0.000}$ & $79.41_{0.00}$ & $0.726_{0.000}$ & $81.42_{0.00}$ & $0.771_{0.000}$ & $64.96_{0.00}$ \\
\midrule
SciBERT & $0.429_{0.039}$ & $81.44_{1.44}$ &  $0.440_{0.027}$ & $83.37_{1.31}$ &  $0.400_{0.085}$ & $\mathbf{70.35_{2.90}}$\\
CiteBERT & $0.431_{0.044}$ & $81.80_{1.19}$ &  $0.433_{0.042}$ & $83.92_{1.32}$ &  $0.425_{0.067}$ & $69.49_{1.21}$\\
RoBERTa & $0.437_{0.040}$ & $82.20_{0.60}$ &  $0.425_{0.046}$ & $84.77_{1.12}$ &  $0.470_{0.185}$ & $69.73_{5.02}$\\
MiniLM-FT & $0.436_{0.004}$ & $79.31_{0.15}$ &  $0.445_{0.003}$ & $81.80_{0.11}$ &  $0.412_{0.007}$ & $64.08_{0.47}$\\
MPNet-FT & $\mathbf{0.371_{0.005}}$ & $\mathbf{82.58_{0.17}}$ &  $\mathbf{0.369_{0.005}}$ & $\mathbf{85.20_{0.22}}$ &  $\mathbf{0.377_{0.008}}$ & $65.03_{0.38}$\\

    \bottomrule %

    \end{tabular}
    \caption{MEDICINE -- MSE and Pearson correlation ($\rho$) on predicting the similarity of the scientific findings for different models. Results are averaged over 5 random seeds; standard deviation is given in the subscript.} %
    \label{tab:med-baseline-findings-matching}
\end{table*}

\begin{table*}%
    \setlength{\tabcolsep}{1.5pt}
    \def\arraystretch{1.2}
    \centering
    \fontsize{10}{10}\selectfont
    \begin{tabular}{l c c | c c | c c}
    \toprule %
    & \multicolumn{2}{c}{All} & \multicolumn{2}{c}{News} & \multicolumn{2}{c}{Twitter} \\
     \midrule
    Method & MSE & $\rho$ & MSE & $\rho$ & MSE & $\rho$ \\
    \midrule 
Paraphrase & $3.208_{0.000}$ & $32.23_{0.00}$ & $3.568_{0.000}$ & $27.56_{0.00}$ & $2.618_{0.000}$ & $46.52_{0.00}$ \\
NLI & $3.066_{0.000}$ & $39.57_{0.00}$ & $3.125_{0.000}$ & $27.39_{0.00}$ & $2.970_{0.000}$ & $50.61_{0.00}$ \\
MiniLM & $0.539_{0.000}$ & $75.16_{0.00}$ & $0.525_{0.000}$ & $77.98_{0.00}$ & $0.561_{0.000}$ & $66.81_{0.00}$ \\
MPNet & $0.634_{0.000}$ & $72.22_{0.00}$ & $0.650_{0.000}$ & $72.26_{0.00}$ & $0.608_{0.000}$ & $69.44_{0.00}$ \\
\midrule
SciBERT & $0.531_{0.022}$ & $74.57_{1.36}$ &  $0.571_{0.020}$ & $74.68_{1.56}$ &  $\mathbf{0.467_{0.030}}$ & $\mathbf{74.14_{1.41}}$\\
CiteBERT & $0.555_{0.015}$ & $73.23_{0.39}$ &  $0.585_{0.036}$ & $73.68_{0.52}$ &  $0.505_{0.031}$ & $72.50_{1.29}$\\
RoBERTa & $0.655_{0.040}$ & $71.28_{1.24}$ &  $0.720_{0.085}$ & $71.38_{1.88}$ &  $0.550_{0.057}$ & $71.35_{1.58}$\\
MiniLM-FT & $\mathbf{0.500_{0.004}}$ & $\mathbf{75.52_{0.11}}$ &  $\mathbf{0.467_{0.005}}$ & $\mathbf{78.48_{0.12}}$ &  $0.555_{0.004}$ & $66.50_{0.16}$\\
MPNet-FT & $0.520_{0.009}$ & $75.21_{0.25}$ &  $0.550_{0.006}$ & $75.48_{0.18}$ &  $0.471_{0.014}$ & $72.25_{0.67}$\\

    \bottomrule %

    \end{tabular}
    \caption{PSYCHOLOGY -- MSE and Pearson correlation ($\rho$) on predicting the similarity of the scientific findings for different models. Results are averaged over 5 random seeds; standard deviation is given in the subscript.} %
    \label{tab:psych-baseline-findings-matching}
\end{table*}

\begin{table*}%
    \setlength{\tabcolsep}{1.5pt}
    \def\arraystretch{1.2}
    \centering
    \fontsize{10}{10}\selectfont
    \begin{tabular}{l c c | c c | c c}
    \toprule %
    & \multicolumn{2}{c}{All} & \multicolumn{2}{c}{News} & \multicolumn{2}{c}{Twitter} \\
     \midrule
    Method & MSE & $\rho$ & MSE & $\rho$ & MSE & $\rho$ \\
    \midrule 
Paraphrase & $3.373_{0.000}$ & $24.35_{0.00}$ & $3.346_{0.000}$ & $26.48_{0.00}$ & $3.463_{0.000}$ & $37.48_{0.00}$ \\
NLI & $3.177_{0.000}$ & $29.97_{0.00}$ & $2.945_{0.000}$ & $36.51_{0.00}$ & $3.926_{0.000}$ & $-8.74_{0.00}$ \\
MiniLM & $0.656_{0.000}$ & $71.40_{0.00}$ & $0.656_{0.000}$ & $73.09_{0.00}$ & $0.656_{0.000}$ & $66.64_{0.00}$ \\
MPNet & $0.705_{0.000}$ & $70.03_{0.00}$ & $0.670_{0.000}$ & $72.43_{0.00}$ & $0.815_{0.000}$ & $60.18_{0.00}$ \\
\midrule
SciBERT & $0.738_{0.020}$ & $67.66_{0.71}$ &  $0.777_{0.031}$ & $67.46_{1.03}$ &  $0.609_{0.029}$ & $69.97_{1.76}$\\
CiteBERT & $0.733_{0.045}$ & $68.05_{1.38}$ &  $0.770_{0.051}$ & $67.83_{1.34}$ &  $0.612_{0.040}$ & $69.79_{2.33}$\\
RoBERTa & $0.690_{0.021}$ & $71.53_{1.15}$ &  $0.731_{0.031}$ & $71.24_{0.80}$ &  $\mathbf{0.560_{0.075}}$ & $\mathbf{75.49_{3.09}}$\\
MiniLM-FT & $0.611_{0.003}$ & $72.32_{0.05}$ &  $0.577_{0.001}$ & $74.13_{0.06}$ &  $0.721_{0.008}$ & $66.44_{0.25}$\\
MPNet-FT & $\mathbf{0.603_{0.004}}$ & $\mathbf{73.00_{0.20}}$ &  $\mathbf{0.575_{0.006}}$ & $\mathbf{74.46_{0.36}}$ &  $0.692_{0.011}$ & $67.18_{0.59}$\\

    \bottomrule %

    \end{tabular}
    \caption{COMPUTER SCIENCE -- MSE and Pearson correlation ($\rho$) on predicting the similarity of the scientific findings for different models. Results are averaged over 5 random seeds; standard deviation is given in the subscript.} %
    \label{tab:cs-baseline-findings-matching}
\end{table*}

\subsection{Error Examples}
\label{sec:error-examples}
Examples of errors which our best models made on $\langle$tweet, paper$\rangle$ pairs can be found in \autoref{tab:roberta-tweet-errors} and \autoref{tab:mpnet-tweet-errors}.

\begin{table*}[t]
\small

\newcommand{\tabincell}[2]{\begin{tabular}{@{}#1@{}}#2\end{tabular}}
\resizebox{0.99\textwidth}{!}{
\begin{tabular}{p{60mm}p{60mm}cc}%
\toprule
\textbf{Tweet} & \textbf{Paper Finding} & \textbf{Prediction} & \textbf{Ground Truth} \\
\midrule
Mixed reality variations improve learning, over screen-only options. CMU researchers.  & The overall improvement from pre to post was 11.3 \% in the mixed-reality conditions and 2.4 \% in the virtual conditions. & 2.92 & 5 \\ \midrule
Metarrestin, an inhibitor of tumor metastasis, discovered thru team science @ KU, @username, @username, and @username and more. Congrats to first author Kevin Frankowski and special thanks to Udo Rudloff, Juan Maruguan, and Sui Huang. & Evaluation of apoptotic index showed less than 1\% of cells undergoing apoptosis in response to metarrestin treatment. & 2.15 & 4.2 \\
\midrule
Today in @username a graphene transfer approach using paraffin as a support layer to obtain wrinkle-reduced, clean, large-area graphene retaining high mobility & Similar to previous reports, our PMMA-transferred CVD monolayer graphene on Si/SiO 2 substrate experienced compressive strain and p-doping 30 . & 2.64 & 1 \\
\midrule
When the Going Gets Tough: The "Why” of Goal Striving Matters. An excellent article by @username + colleagues. & Practitioners who aim to facilitate effective goal setting in sport, business, and educational settings would benefit from guidelines for developing autonomous motivation. & 2.00 & 3.6 \\
\midrule
Thosewho were sociosexually unrestricted reported lower stress and greater overall emotional health after casual sex. & Simple slope analyses indicated that high-SOI participants who had casual sex over the academic year had higher self-esteem (B $\frac{1}{4}$ 0.14, SE $\frac{1}{4}$ 0.06, p $\frac{1}{4}$ .025) and marginally lower depression (B $\frac{1}{4}$ A0.12, SE $\frac{1}{4}$ 0.07, p $\frac{1}{4}$ .091) and anxiety (B $\frac{1}{4}$ A0.11, SE $\frac{1}{4}$ 0.06, p $\frac{1}{4}$ .086) than high-SOI participants who did not have casual sex (Figure 3) . & 2.84 & 4.4\\
\bottomrule
\end{tabular}
}
\caption{Top-5 biggest errors made by RoBERTa on <tweet, paper> pairs in terms of absolute error. }
\label{tab:roberta-tweet-errors}
\end{table*}

\begin{table*}[t]
\small

\newcommand{\tabincell}[2]{\begin{tabular}{@{}#1@{}}#2\end{tabular}}
\resizebox{0.99\textwidth}{!}{
\begin{tabular}{p{60mm}p{60mm}cc}%
\toprule
\textbf{Tweet} & \textbf{Paper Finding} & \textbf{Prediction} & \textbf{Ground Truth} \\
\midrule
Mixed reality variations improve learning, over screen-only options. CMU researchers.  & The overall improvement from pre to post was 11.3 \% in the mixed-reality conditions and 2.4 \% in the virtual conditions. & 2.45 & 5 \\
\midrule
'Physical observation + interactive feedback improved children’s learning by 5x' via Nesra Yannier @username & These results show that mixed-reality led to more learning than screen only, for both the mousecontrol and physical-control conditions ( Figure 10 ). & 2.19 & 4 \\
\midrule
Today in @username a graphene transfer approach using paraffin as a support layer to obtain wrinkle-reduced, clean, large-area graphene retaining high mobility & Similar to previous reports, our PMMA-transferred CVD monolayer graphene on Si/SiO 2 substrate experienced compressive strain and p-doping 30 . & 2.80 & 1 \\
\midrule
Metarrestin, an inhibitor of tumor metastasis, discovered thru team science @ KU, @username, @username, and @username and more. Congrats to first author Kevin Frankowski and special thanks to Udo Rudloff, Juan Maruguan, and Sui Huang. & Evaluation of apoptotic index showed less than 1\% of cells undergoing apoptosis in response to metarrestin treatment. & 2.61 & 4.2 \\
\midrule
Super happy to present our latest paper on global food webs:
Years of work on predator-prey body-mass ratios and the first use of the GATEWAy data base. & Predators typically exert the strongest feeding pressure on prey that are 1-2 orders of magnitude smalle, while weaker interaction strengths are realized with prey that are smaller or larger than this size. & 1.92 & 3.4\\
\bottomrule
\end{tabular}
}
\caption{Top-5 biggest errors made by MPNet-FT on <tweet, paper> pairs in terms of absolute error. }
\label{tab:mpnet-tweet-errors}
\end{table*}

\begin{table*}[t]
\small
\fontsize{9}{9}\selectfont
\newcommand{\tabincell}[2]{\begin{tabular}{@{}#1@{}}#2\end{tabular}}
\resizebox{0.99\textwidth}{!}{
\begin{tabular}{p{60mm}p{60mm}cc}%
\toprule
\textbf{Paper Finding} & \textbf{News Finding} & \textbf{Prediction} \\
\midrule
Increase in the body size of dicynodonts across the Late Triassic may have been driven by selection pressure to reach a size refuge from large predators (24) .  & Researchers believe selection pressures--potentially to protect themselves from larger predators--may have been the driver behind their giant size, but more research will be needed to understand Lisowicia and its place in the evolutionary tree. & 3.0008 \\ \midrule
The best option among the three is the EPS container with the lowest impacts across the 12 categories. & The study found that the styrofoam container was the best option among the disposable containers across all the impacts considered, including the carbon footprint. & 3.1120 \\
\midrule
As media coverage started to increase, water demand decreased and the models with media correctly captured the downward trend, but the models without media forecasted increasing demand. & Strikingly, the models also found that for every 100-article increase over a two-month period, there was an 11 percent to 18 percent decrease in demand for water. & 3.1537 \\
\midrule
For example, of the 63 negative precipitation years during 1896-2014, 15 of the 32 warm-dry years (47\%) produced 1-SD drought, compared with only 5 of the 31 cool-dry years (16\%)  & Their analysis revealed that the years that were both warm and dry were about twice as likely to produce a severe drought as years that were cool and dry. & 3.2569 \\
\midrule
Our study shows that low-dose BPA and BPS exposure has physiological effects. & Although the levels were low, the scientists soon saw that both BPA and BPS caused changes in the brain development of the zebra fish embryos. & 3.3331 \\
\midrule
Use of multiple prescription medications with these potential effects was associated with greater likelihood of concurrent depression. & About 15 percent of participants who simultaneously used three or more of these drugs were depressed. & 3.3692 \\ 
\midrule
We also found that renewal submission rate was the factor most predictive of sustained funding for either gender, and that gender differences in survival disappear when genders were matched on renewal submission rate and first year of funding. & On average, women submitted eligible grants for renewal 42\% of the time and won funding 36\% of the time, compared with 45\% and 39\%, respectively, for men. & 3.4132 \\
\midrule
Among those completing the 12-month survey, 60 nonsmokers (55.6\%) and 29 smokers (26.6\%) were reemployed at 1 year. & After 12 months, the re-employment rate of smokers was 24 percent lower than that of nonsmokers. & 3.5151 \\
\midrule
This suggests behaviour consistent with moral licensing: participants who refrained from cheating at higher stakes seem to have subsequently licensed themselves to donate less to charity, thereby "balancing" their moral behaviour over time. & However those who cheated the least when tempted with high stakes were more likely to license themselves not to behave so charitably in another task. & 3.5481 \\
\midrule
Lack of Panx1 increases adipocyte hypertrophy and reduces adipocyte numbers in subcutaneous fat in vivo. & With both a normal diet, and a a high-fat diet, a lack of Panx1 increases cell size. & 3.5618 \\

\bottomrule
\end{tabular}
}
\caption{Borderline IMS Model prediction samples. We note that 3 appears to be a good threshold for matching, as pairs with an IMS over 3 tend to discuss the same scientific findings.}
\label{tab:prediction_samples}
\end{table*}

\subsection{Regression details}
\label{sec:regression_details}

\begin{table*}
\begin{center}
\begin{tabular}{llll}
\hline
Model:            & MixedLM & Dependent Variable: & paper\_sentence\_score  \\
No. Observations: & 1111150 & Method:             & REML                    \\
No. Groups:       & 6705    & Scale:              & 0.1084                  \\
Min. group size:  & 31      & Log-Likelihood:     & -349944.7797            \\
Max. group size:  & 67063   & Converged:          & Yes                     \\
Mean group size:  & 165.7   &                     &                         \\
\hline
\end{tabular}
\end{center}

\begin{center}
\begin{tabular}{lrrrrrr}
\hline
                         & $\beta$ Coef. & Std.Err. &       z & P$> |$z$|$ & [0.025 & 0.975]  \\
\hline
\emph{Intercept}                &  3.299 &    0.007 & 489.729 &       0.000 &  3.286 &  3.312  \\
Outlet Type: Press Release &  0.037 &    0.001 &  31.187 &       0.000 &  0.035 &  0.039  \\
Outlet Type: Science \& Technology      &  0.034 &    0.001 &  30.581 &       0.000 &  0.032 &  0.036  \\
Field: Biology                  & -0.018 &    0.020 &  -0.904 &       0.366 & -0.056 &  0.021  \\
Field: Psychology               &  0.040 &    0.018 &   2.168 &       0.030 &  0.004 &  0.076  \\
Field: Medicine                 &  0.206 &    0.017 &  11.813 &       0.000 &  0.171 &  0.240  \\
Field: Computer\_science        &  0.050 &    0.024 &   2.132 &       0.033 &  0.004 &  0.096  \\
Group Var                &  0.009 &    0.001 &         &             &        &         \\
\hline
\end{tabular}
\end{center}
    \caption{Regression table for RQ1} %
    \label{tab:RQ1_regression}
\end{table*}

\begin{table*}
\begin{center}
\begin{tabular}{llll}
\hline
Model:            & MixedLM & Dependent Variable: & paper\_sentence\_score  \\
No. Observations: & 182735  & Method:             & REML                    \\
No. Groups:       & 1360    & Scale:              & 0.1525                  \\
Min. group size:  & 31      & Log-Likelihood:     & -89654.8514             \\
Max. group size:  & 89523   & Converged:          & Yes                     \\
Mean group size:  & 134.4   &                     &                         \\
\hline
\end{tabular}
\end{center}

\begin{center}
\begin{tabular}{lrrrrrr}
\hline
                            & $\beta$ Coef. & Std.Err. &       z & P$> |$z$|$ & [0.025 & 0.975]  \\
\hline
\textit{Intercept}                   &  3.777 &    0.013 & 292.571 &       0.000 &  3.752 &  3.803  \\
Is Verified  User?  & -0.047 &    0.004 & -11.044 &       0.000 & -0.056 & -0.039  \\
Is Organizational Account? & 0.042 &    0.002 & -19.026 &       0.000 & -0.046 & -0.037  \\
User Metric: log(Followers)             & -0.003 &    0.001 &  -5.059 &       0.000 & -0.004 & -0.002  \\
User Metric: log(Following)          &  0.000 &    0.001 &   0.369 &       0.712 & -0.001 &  0.002  \\
User Metric: Account Age (in years)   &  0.004 &    0.000 &  10.824 &       0.000 &  0.003 &  0.005  \\
Field: Biology                     & -0.025 &    0.030 &  -0.850 &       0.395 & -0.083 &  0.033  \\
Field: Psychology                  &  0.308 &    0.028 &  11.052 &       0.000 &  0.254 &  0.363  \\
Field: Medicine                    &  0.206 &    0.026 &   7.826 &       0.000 &  0.155 &  0.258  \\
Field: Computer\_science           & -0.352 &    0.035 & -10.158 &       0.000 & -0.420 & -0.284  \\
Group Var                   &  0.059 &    0.006 &         &             &        &         \\
\hline
\end{tabular}
\end{center}
    \caption{Regression table for RQ2} %
    \label{tab:RQ2_regression}
\end{table*}

\autoref{tab:RQ1_regression} shows the regression table for RQ1.
\autoref{tab:RQ2_regression} shows the regression table for RQ2.

\end{document}